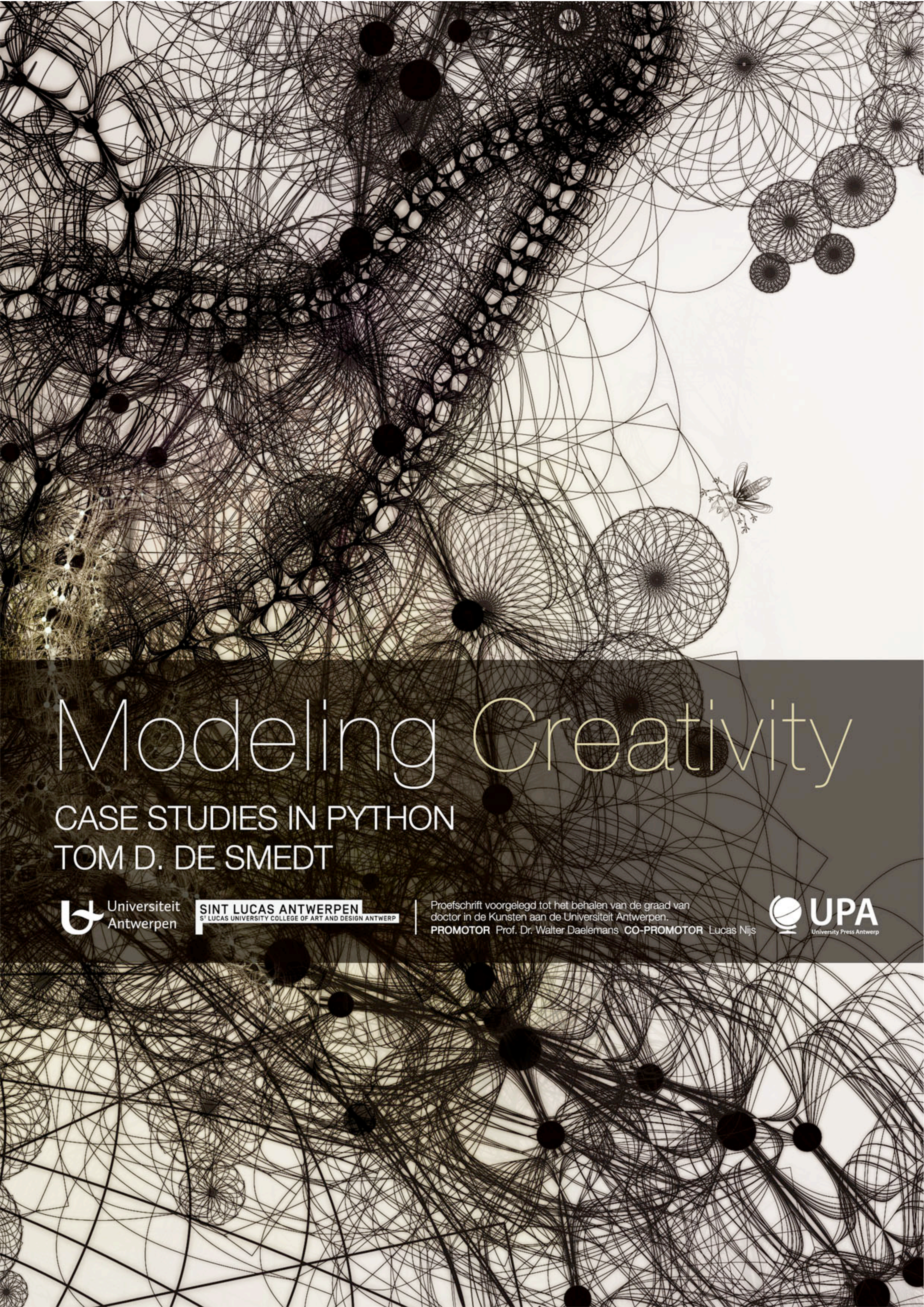

# Modeling Creativity

## CASE STUDIES IN PYTHON

TOM D. DE SMEDT

Faculteit Letteren en Wijsbegeerte
Departement Taalkunde

# Modeling Creativity

## CASE STUDIES IN PYTHON

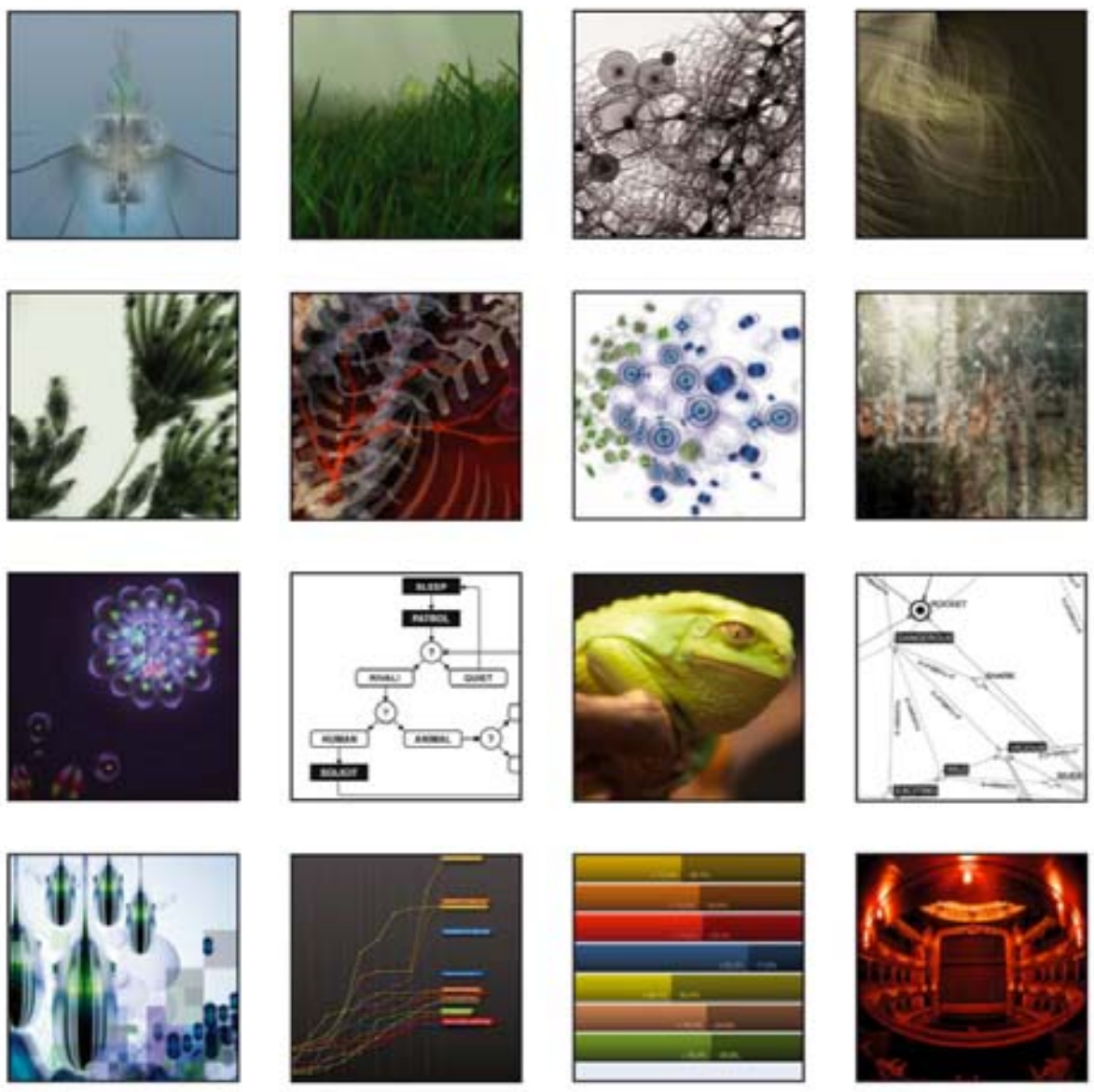

Computermodellen van Creativiteit:
Gevalstudies in de Python-programmeertaal

Proefschrift voorgelegd tot het behalen van
de graad van doctor in de Kunsten
aan de Universiteit Antwerpen
te verdedigen door

Tom De Smedt

PROMOTOR
Prof. dr. Walter Daelemans
CLiPS Computational Linguistics Group
Universiteit Antwerpen

CO-PROMOTOR
Lucas Nijs
EMRG Experimental Media Research Group
St. Lucas University College of Art & Design, Antwerpen

# Modeling Creativity

## CASE STUDIES IN PYTHON

Tom De Smedt

"Enough!" he said, hoarse with indignation. "How dare you waste a great talent on such drivel? Either give it decent poems to write or I call the whole thing off!"

"What, those aren't decent poems?" protested Klapaucius.

"Certainly not! I didn't build a machine to solve ridiculous crossword puzzles! That's hack work, not Great Art! Just give it a topic, any topic, as difficult as you like . . ."

Klapaucius thought, and thought some more. Finally he nodded and said:

"Very well. Let's have a poem, lyrical, pastoral, and expressed in the language of pure mathematics. Tensor algebra mainly, with a little topology and higher calculus, if need be. But with feeling, you understand, and in the cybernetic spirit."

"Love and tensor algebra? Have you taken leave of your senses?" Trurl began, but stopped, for his electronic bard was already declaiming:

Come, let us hasten to a higher plane,
Where dyads tread the fairy field of Venn,
Their indices bedecked from one to $n$,
Commingled in an endless Markov chain!

Come, every frustum longs to be a cone,
And every vector dreams of matrices.
Hark to the gentle gradient of the breeze:
It whispers of a more ergodic zone.

[...]

– Stanislaw Lem, The Cyberiad, Trurl's Electronic Bard (1974)

# Contents

# Preface

The topic of this work is to model creativity using computational approaches. Our aim is to construct computer models that exhibit creativity in an artistic context, that is, that are capable of generating or evaluating an artwork (visual or linguistic), an interesting new idea, a subjective opinion. The research was conducted in 2008–2012 at the Computational Linguistics Research Group (CLiPS, University of Antwerp) under the supervision of Prof. Walter Daelemans. Prior research was also conducted at the Experimental Media Research Group (EMRG, St. Lucas University College of Art & Design Antwerp) under the supervision of Lucas Nijs. Early research focuses on generative art. Related work is discussed in chapters 1, 2 and 3. More recent research focuses on computational creativity and computational linguistics. Related work is discussed in chapters 4, 5, 6 and 7.

Generative art, computational creativity and computational linguistics have a common ground in creativity. Specifically, they relate in some way to the question if and how machines can be creative and intelligent. In chapter 2, we provide a case study of a computer program that is the author of its own artworks. It will turn out that this program is not very creative because its actions are mostly random, whereas artistic creativity is the result of non-random thought processes and social consensus. After a summary of the theoretical basis of creativity in chapter 4, we direct our attention to a case study involving a program that generates creative ideas. It will turn out that the results are interesting, but also that it is difficult to evaluate whether or not they are novel and useful. In chapter 6, we discuss the PATTERN software package developed in the course of our research, which relies on more formal methods. Chapter 7 uses the software from chapter 6 to evaluate subjective opinion in text.

Intuitively, a model for (artistic) artificial creativity appears to involve the steps illustrated in figure 1. A "good idea", that is, a *novel* and *appropriate* solution to a given problem, is selected or combined from a pool of many possible ideas (chapters 4 and 5). A work (of art) that captures this idea is then created (chapters 2 and 3). The work is then evaluated by the author or a community of knowledgeable peers (chapters 6 and 7). Following a negative evaluation, the idea is subsequently adapted (chapter 1). We will test this conjecture in a number of case studies and ground it in the literature.

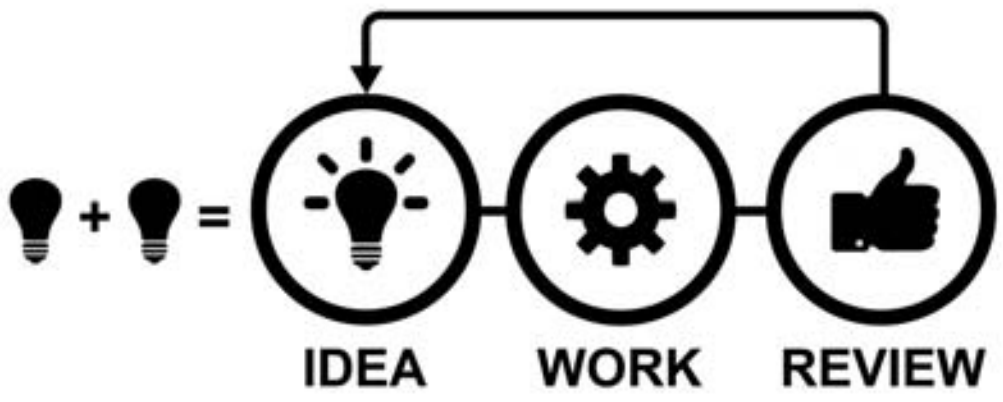

Figure 1. Illustrative example of a model for artistic creativity.

A brief overview of the different chapters:

1. Blind creativity

   In this chapter we outline the mechanisms of creativity in nature, which is essentially "blind". We discuss a computer model for evolution by natural selection, and a computer model for complexity in agent-based systems. We show how creativity does not arise *ex nihilo*, out of nothing, but rather as the result of a multitude of underlying, interacting processes. This is a key argument that allows us to further deconstruct the concept of creativity in following chapters.

2. Generative art

   In this chapter we give an overview of the history of generative art, a rule-based form of art influenced by complexity and self-organization in nature (chapter 1). We demonstrate our Python software packages for producing generative art, NODEBOX and NODEBOX FOR OPENGL, and provide a commentary on six case studies in NODEBOX. We argue how this kind of software offers creative leverage, since users are free to combine any kind of functionality they want in programming code. Finally, we discuss if and how art created by a machine can really be creative. This raises questions that are further addressed in chapter 4 and chapter 5.

3. Brain-computer interfaces

   In this chapter we discuss VALENCE, a generative art installation created with NODEBOX FOR OPENGL. It responds to brain activity using a wireless EEG headset. The aim of such direct brain-computer interfaces (BCI) is to augment the way in which humans communicate with machines. In our approach, we focus on alpha waves (relaxation) and the valence hypothesis (arousal). This kind of dynamic artwork would not be possible to realize using traditional software.

4. Human creativity

   In this chapter we provide a brief overview of the psychological study of human creativity, from the network of neurons in the brain that generates the conscious mind – and which produces the electrical activity recorded by VALENCE – to creative ideas, intuition and different styles of thought. In particular, we discuss a computational approach to creativity: Boden's concept search space. It is the theoretical foundation of our PERCEPTION case study in chapter 5.

5. Computational creativity

   In this chapter we will discuss PERCEPTION, a model for computational creativity. Instead of the visual output in chapters 2 and 3, PERCEPTION models creative thought by means of analogy. We use our PATTERN software package (chapter 6) to implement a concept search space as a semantic network of related concepts, and search heuristics to traverse the network. The chapter concludes with a discussion of the results and a long-lasting debate in the field of artificial intelligence.

6. Computational linguistics

   In this chapter we present MBSP FOR PYTHON, a memory-based shallow parser, and PATTERN, a Python web mining package. We briefly examine the statistical techniques that underpin the software, the vector space model. The aim of PATTERN is to provide a wide range of functionality that can be creatively combined to solve tasks in data mining and natural language processing. We provide short examples to demonstrate how this works, and an extensive case study in chapter 7.

7. Sentiment analysis

   In this chapter we discuss a sentiment lexicon for Dutch, and a derivate for English, that can be used to predict subjective opinion in text. It is built with and part of PATTERN. We examine the methods used, the results, and an interesting case study on political discourse in newspapers. We conclude with a discussion of how this approach can be used to evaluate art appreciation.

Our work is presented in a popular scientific tone, using a hands-on approach with case studies and examples in Python programming code. It is intended for an artistic audience, but researchers in the field of artificial intelligence may benefit as well from the comprehensive summary of the evolutionary principles in biology, the psychological study of creativity, and the statistical principles in natural language processing and machine learning. We hope that our work offers an engaging discourse on computational creativity.

## Acknowledgements

The research was funded by the Industrial Research Fund (IOF) of the University of Antwerp. Artwork for CITY IN A BOTTLE was funded by the Flemish Audiovisual Fund (VAF). Artwork for NANOPHYSICAL and VALENCE was funded by Imec (Interuniversity Microelectronics Centre).

Many thanks and honors go to my advisors: Prof. dr. Walter Daelemans at the Computational Linguistics Research Group (CLiPS) and Lucas Nijs at the Experimental Media Research Group (EMRG), without whose insight, patience and impatience this work would not exist. Many thanks go to my colleagues at CLiPS and EMRG for assistance and feedback. Many thanks to the members of the review committee: Prof. dr. Steven Gillis, dr. Tony Veale, dr. Jon McCormack and dr. Guy De Pauw. Many thanks also to Imke Debecker and Jo De Wachter at Imec. A special thanks goes to my family for support; to Tom Van Iersel, Ludivine Lechat, Giorgio Olivero and dr. Thomas Crombez for support, feedback and collaboration; and worldwide to the NodeBox, GitHub and MLOSS open source communities.

The subjectivity lexicon in chapter 7 was annotated by Nic De Houwer, Lore Douws, Aaricia Sobrino Fernández, Kimberly Ivens (Computational Models of Language Understanding master class), Tim De Cort (postgraduate student at St. Lucas University College of Art & Design), Tom De Smedt and Walter Daelemans.

# Part I

# NATURE

•

The necessary nasty screeching of
flirtatious fitness
quarrelsome and
contentious.

An entity exposition,
Restlessness

– FLOWEREWOLF, Nature (edited)

# 1 Blind creativity

One of the most compelling examples of creativity is no doubt life itself. These days, it is harder to spot nature and its diversity at work. Humans have been prolific in urbanizing the planet. But many of us will have pleasant childhood memories of going out to explore, to observe a bustling ant colony, to inspect the shapes and colors of flower petals, to wonder at the intricacy of a spider web. How can such purposeful intricacy exist, when nature is inherently without purpose?

## Spider webs

Webs are what make spiders spiders. But the spider's ancestors didn't venture onto land to begin making webs there and then, 400 million years ago. The skill was refined bit by bit. The oldest known proto-spider (Attercopus fimbriunguis) probably lined its burrow with silk to keep water out, not to catch prey. There were no flying insects around to catch in any case. Later, single silk threads would be used around the shelter, as trip-lines for a passerby snack. Gradually, over the course of 200 million years, traps became more elaborate, from small ground webs and bush webs to the orbicular webs we know today (Craig, 1997 and Vollrath & Selden, 2007).

Building a web is an efficient hunting strategy. Instead of rummaging about (which is hard work) the spider can wait for prey to wander in. The spider releases a sticky thread and waits for the wind to blow it away, hook it onto something. It cautiously traverses the thread and reinforces it with a second line. Then it extends the line with a downward Y-shape, forming the first three radials of the web (cf. wheel spokes). It constructs an encompassing frame to attach more radials to. Then, the spider will add a spiraling line from the inside out, using its own body to measure spacing between the lines. Most spiders have three types of spinnerets to produce silk. The first spiral is constructed with non-sticky silk. It allows the spider to easily move about during construction. It will progressively replace the non-sticky parts with smaller, sticky spirals, from the outside in. Building the web requires energy of course. After an unproductive day, the spider may eat part of its web to recycle the proteins it needs to make silk, and start over (Craig, 1987).

## Flower petals

Flowers primarily use pollination[1] to reproduce. They depend on other organisms (pollinators) to transport pollen grains from the flower stamens to receptive flower carpels, where fertile seeds are produced. As such, the flower's colorful petals and fragrance are not intended to gratify humans. They serve as cues to attract pollinators. Insects such as honeybees and butterflies are lured in with the promise of reward (nectar!) and in turn spread the pollen that stick to their legs when they move on to other flowers. Wind-pollinated plants on the other hand have scentless, inconspicuous flowers in shades of brown and green.

[1] Over 80% of crops cultivated for human consumption in Europe (35% worldwide) depend on insect pollinators, bees especially. Recent studies show a decline in pollinators, for example wild bee communities jeopardized by agricultural intensification (Klein et al., 2007, Gallai, Salles, Settele & Vaissière, 2009).

Once pollinated, some flowers change color, urging visitors to more rewarding neighbors. This "trick" benefits the flower. Fewer pollinator visits are wasted on siblings that can no longer receive or donate pollen (Delph & Lively, 1989). Some flowers will also wither their petals to decrease attraction once pollinated, since the pollinators will prefer flowers with more petals (van Doorn, 1997). But changing color is the more efficient strategy. This way, visitors can still recognize the plant from a distance and make their choice at closer range (Weiss, 1995). This scenario is not unlike checking the expiry dates in the supermarket.

## Ant colonies

Observing ants at work is fun. Leave a sugar cube in the vicinity of the nest and before long there will be a chain of worker ants going back and forth, hauling bits of sugar to the nest. Ants are social insects. They rely on pheromone scent for communication. When food is located, an ant will mark the trail with secreted pheromones as it heads back to the colony. Other ants can use their antennae to assess the direction and intensity of the scent. They will go over to the food source, grab a morsel, and reinforce the trail with more pheromones on the way back. Shorter paths will be traversed faster, reinforced more often, and hence attract even more ants until an optimal route is established. Once the food source is exhausted, no new scent markers are left, so the depleted path dissipates. The ants will abandon the path and begin to look for other nearby food sources.

Social behavior is key to the ant's success. The survival of the colony does not rely on a single individual, but on many ants cooperating. A single worker ant can specialize in the task at hand, for example moving stuff. As long as there are plenty of other workers in the chain, the ant does not need to know about the next step. It can simply go on moving stuff. Other ants in the chain will take care of the next step. If necessary, individual worker ants can switch to group behavior (e.g., a food object or an intruder is too big to handle) by means of recruitment and alarm scent signals (Hölldobler & Wilson, 1990).

#### SUPERORGANISM

Hölldobler & Wilson (2009) view the ant colony as a *superorganism* – an organism consisting of many organisms. Individual ants in the colony can be thought of as cells in an organism. Groups of ants represent different functions of the organism. For example, the worker ant caste resembles a circulatory system of food distribution (*trophallaxis*), the soldier ant caste parallels the immune system. Hofstadter (1979) uses a similar analogy: Aunt Hillary, a conscious ant hill, where individual ants are relatively simple ("as dumb as can be") but function like neurons in the brain by means of signal propagation. For example, a worker ant confronted with an intruder will release alarm signals (pheromones). These are picked up and propagated by more and more ants, until the signals either dissipate by removal of the threat or effectively put the entire colony/brain in **PANIC** mode.

## 1.1 Evolution by natural selection

Spider webs, flower petals and ant trails are just a few of the creative strategies found throughout nature. There are many more examples. How can such creative strategies exist, and why? How did spiders invent the orbicular web? Why do ants cooperate? This is explained by Darwin's theory of evolution (1859). A scientific theory is a well-researched framework of corroborating evidence grounded in empirical phenomena, by means of observation. This differs from the popular sense of the word "theory". The theory of evolution is a solid framework that explains how all life evolved from a universal common ancestor nearly four billion years ago. We provide a brief overview of the mechanisms, relevant to the topic of creativity. For more information, we recommend *The Selfish Gene* (Dawkins, 1976).

#### COMPETITION

The driving force behind evolution is competition: competition among plants for nutrients and light, among spiders that catch bees and bees that prefer not to be caught, among ant colonies occupying the same territory, among humans for mates and social status.

#### VARIATION + SELECTION

Evolution itself is the process by which organisms develop and diversify across successive generations. This happens by means of heredity. Traits are passed on from ancestors to their offspring (Fisher, 1930). Today, we know that such traits (e.g., ant foraging behavior) are controlled by genes, sequences of DNA. When two organisms of a species reproduce, each parent contributes half of the offspring's gene pool. Mutation, errors that occur during DNA replication, may cause changes in an individual's gene pool. Genetic recombination and mutation produces variation in the population; new, slightly different traits. Some traits will be beneficial to the organism's survival given its surrounding environment. The fittest individuals with the best traits can reproduce more rapidly, or simply live long enough to reproduce. Gradually, favorable traits will spread throughout the population. This is called natural selection. As such, evolution is a blind process: it is a complex, non-random process, but it has no predefined goal or design.

Spiders that build better webs have more food to eat. They will be fitter and survive longer. They will have more opportunities to reproduce (natural selection). When their genes are passed on, the offspring inherits the web-building traits of its predecessors (evolution). Due to genetic recombination and mutation, some of the offspring will mess up their webs. But others will refine the technique further (variation). When insects diversified 200 million years ago, spiders that took advantage of this new food source by building orb webs prospered. The spider did not "want" or "plan" to build a better web. Those that did simply did better. Likewise, floral color change or petal withering allows flowers to be pollinated faster. Consequently, such properties propagate. Ants cooperate because the behavior is mutually beneficial (Jackson & Ratnieks, 2007). Suppose one worker ant passes its chunk of sugar to another worker ant, which then takes off to keep the food for itself. The deceptive ant would thrive and the cooperative ant would starve. Natural selection will favor cooperation. All chunks of sugar eventually end up at the nest for all to enjoy, so there is no harm in cooperating. On the contrary, it enables the ants to process food sources that would otherwise be too large to handle individually.

## 1.2 EVOLUTION: genetic algorithm + competition

EVOLUTION (see also De Smedt, Lechat & Daelemans, 2011) is a computer model for evolution by natural selection. The project is an extension of Ludivine Lechat's thesis project GRAPHIC CELLULAR DOMESTICATION (GCD). Like many of Lechat's artworks[2], GCD consists of visual elements that can be combined into different shapes (typically lifeforms) using a set of rules. This kind of rule-based art is also called generative art, discussed in chapter 2. Lechat's work was done by hand. In EVOLUTION we expound on this by using an automated, computational approach. The game starts out with a set of insect creatures randomly designed from a pool of components: heads, legs, wings, tails, and so on. Different components have a behavioral impact. For example, bigger eyes and antennae allow a creature to employ better hunting strategies (ambush, intercept). Larger wings allow a creature to fly faster. A longer tail allows more efficient steering.

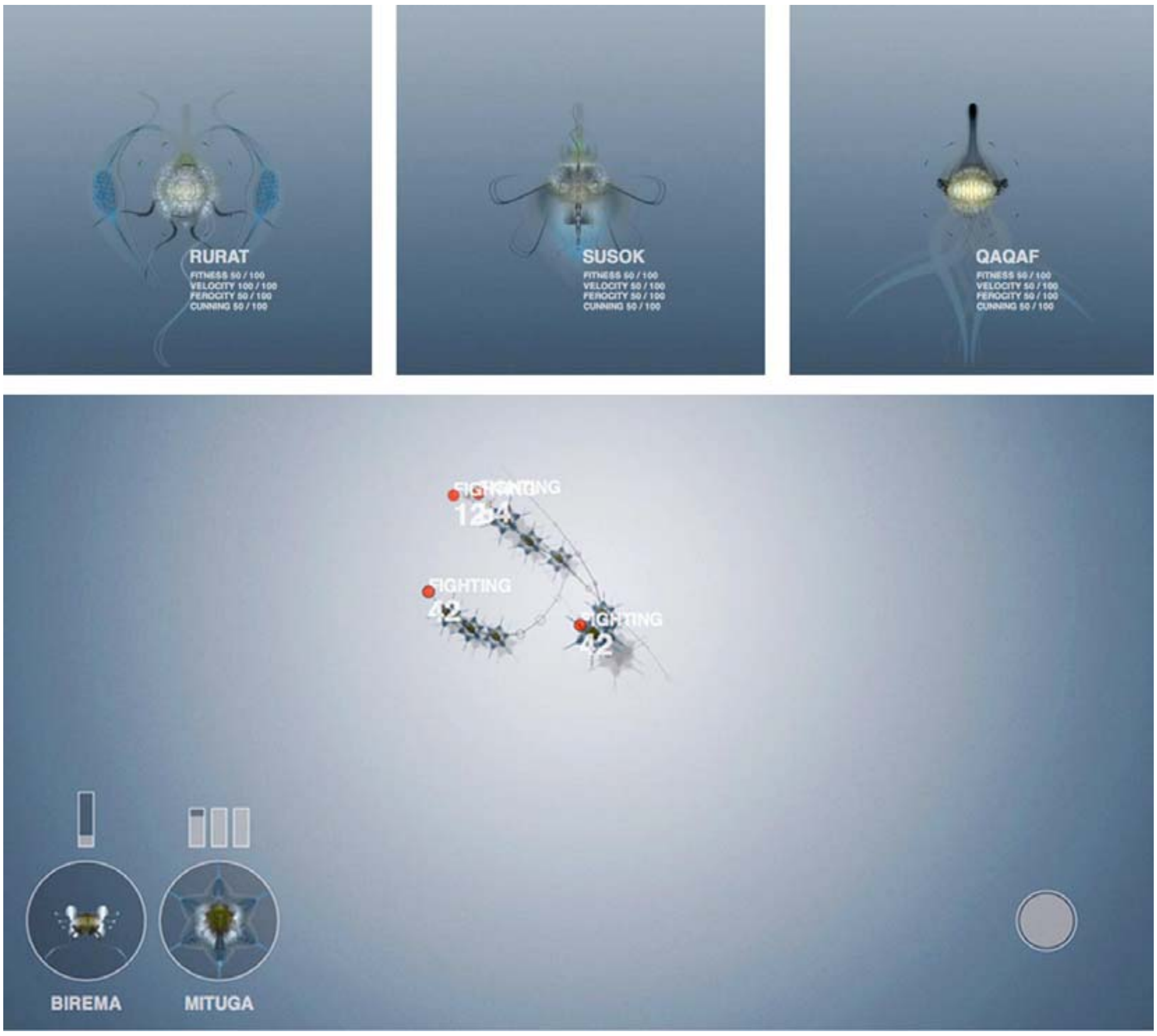

Figure 2. EVOLUTION. Three evolved creatures and an ongoing fight between two species.

[2] http://www.ludivinelechat.be/

**FITNESS → CONVERGENCE**

We then used a genetic algorithm to evolve better creatures. A genetic algorithm (Holland, 1992) is a search technique used in artificial intelligence (AI) computer programs, based on natural selection. A straightforward Python implementation is included in the appendix. Central to the technique is the *fitness function*. The fitness function calculates a score (e.g., `0.0`–`1.0`) for each candidate in a population. Given a population of candidates, the algorithm will then select, pair and recombine the fittest candidates to create a new population. With each consecutive generation the population's average fitness increases, slowly converging to an optimal solution. A known weakness of the technique is that the optimal solution can be either global or local. Local optima occur when the algorithm converges too fast. This happens when weaker candidates are discarded too early, since they can introduce useful variation if retained and combined with fitter candidates. A well-known solution is to use an amount of random mutation to maintain diversity and avoid local optima.

The GA's fitness function in EVOLUTION is the game itself: an interactive hunting ground where creatures are pitted against each other. Creatures that survive the competition are allowed to reproduce. Two creatures from the pool of survivors are randomly recombined into a new creature. For example, the "Rurat" and "Qaqaf" creatures shown in figure 2 might produce an offspring called "Quraf", that sports the large flippers of the first parent and the elegant tentacles of the second parent.

In our setup we used hierarchical fair competition (Hu & Goodman, 2002) instead of mutation to avoid local optima. This means that the game is hierarchically fair in the sense that even highly evolved creatures still face a steady supply of new genetic material (new random creatures to fight). Interestingly, the setup often produces an endless crash-and-bloom cycle of 1) creatures that are exceptional but flawed and 2) mediocre all-rounders. Random newcomers will eventually beat the current (mediocre) winner with an "exceptional trick" (e.g., very aggressive + very fast), but are in turn too unstable to survive over a longer period (e.g., unable to cope with cooperating adversaries). Their trick enters the gene pool but is dominated by generations of older DNA, leading to a slow overall evolution. Figure 2 shows three evolved creatures and a fight between two species.

In 2009, we expanded EVOLUTION into a procedural computer game called CITY IN A BOTTLE (Marinus, Lechat, Vets, De Bleser & De Smedt, 2009) that features more complex scenarios, such as interaction between plants and insects. Figures 3.1 and 3.2 show development screenshots.

Computer models that evolve virtual creatures are part of a field called artificial life. Other examples include Dawkins' BIOMORPH and Sims' EVOLVED VIRTUAL CREATURES. These focus on generating fitter creatures as a task in itself, whereas EVOLUTION focuses on competition between creatures. We will discuss artificial life in more detail shortly. First, we will look at evolutionary strategies such as competition and cooperation, and how a multitude of such strategies can give rise to complex systems in which creative behavior seemingly arises *ex nihilo*, out of nothing.

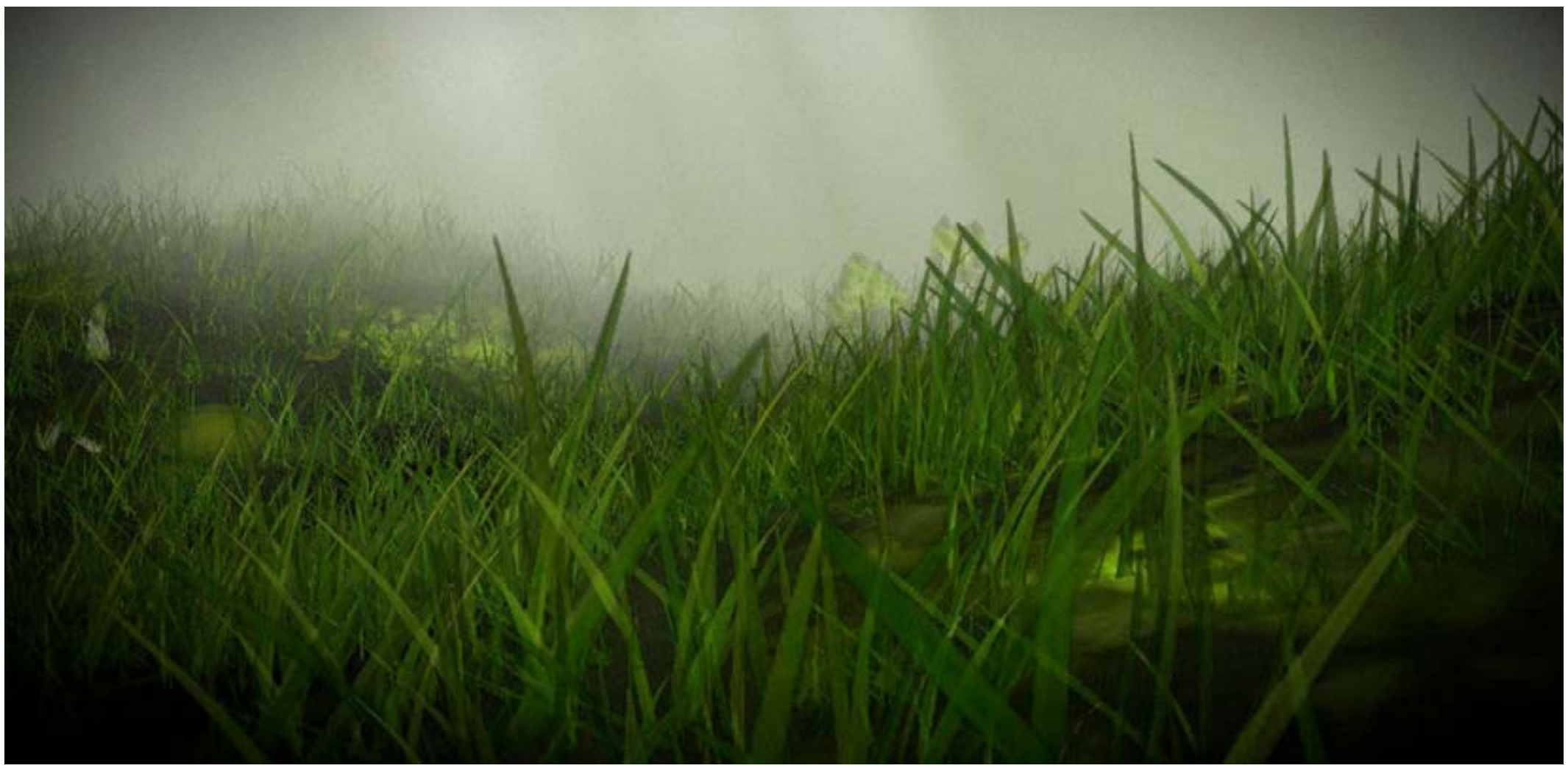

Figure 3.1. Screenshot of the procedural game world in CITY IN A BOTTLE.

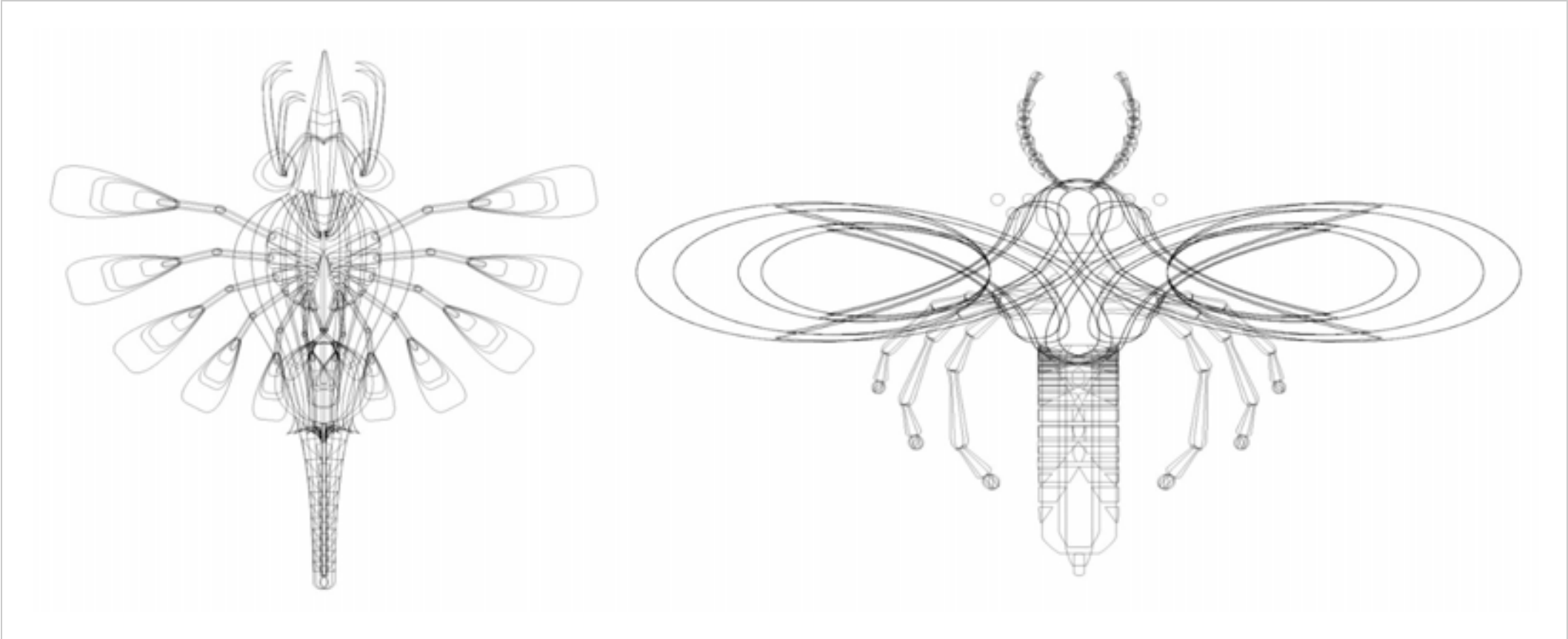

Figure 3.2. Screenshot of procedural creatures in CITY IN A BOTTLE.

## 1.3 Evolutionarily stable strategy (ESS) and cooperation

Evolution by natural selection is often understood as "the survival of the fittest". In this sense it seems to imply that only strong, selfish organisms survive. The idea that selfish behavior is the natural state is a deceptive simplification that can be exploited to promote for example greed and ethnic violence (Marsella, 2005). On the contrary, it is often rewarding to play a cooperative evolutionary strategy. Each strategy has its own costs and benefits. This concept was first formalized by Maynard Smith & Price (1973) using a branch of mathematics called game theory. A strategy is a pre-programmed (genetic) behavioral policy. For example, "attack opponent" or "attack opponent if provoked". An evolutionarily *stable* strategy (ESS) is a strategy adopted by most members of the population. Once an ESS is established it is very hard to replace it with another strategy. In other words, the best strategy for an individual depends on what the majority of the population is doing (Dawkins, 1976), as illustrated by the following example:

**NICE ↔ NASTY**

Imagine a population of ants where all ants are fighting each other. There is no benefit in sharing food, since none of the other ants will share. This kind of reciprocal "nasty" strategy is costly. Whenever an ant is hungry it may end up in a vicious fight over a food source. In fact, it is the worst possible strategy, since no party will yield; all ants need to eat. To clarify this we can construct a payoff matrix (table 1). Say that if an ant wins a conflict, it eats the food and scores `+10`. If it loses it dies and scores `-100`. All ants are equally nasty, so we can expect each ant to win half of the conflicts and lose half. On average, each ant scores `-45`. Now, suppose there is a goofy ant that doesn't fight, but instead attempts to share. If it doesn't work out it runs away. If it can share, it eats half of the food and scores `+5`. It it runs away, it gets `+0` for wasting time. On average it scores `+2.5`, which is better than `-45`. Opportunity arises for ants that cooperate or flee instead of starting a fight. As more of them appear in the population their individual scores will rise to `+5`. Both **NICE** and **NASTY** strategies can be evolutionarily stable. In a population where all ants are cooperating, a single **NASTY** always gets the food (since all others run away). Its individual score is `+10`, so it prospers. In the end there will be a stable ratio between a part of the population with a **NICE** ESS and a part with a **NASTY** ESS.

| PAYOFF MATRIX | NASTY | NICE |
|---|---|---|
| **NASTY** | `(V-C)/2, (V-C)/2` | `V, 0` |
| **NICE** | `0, V` | `V/2, V/2` |

Table 1. Payoff matrix for **NASTY** and **NICE** strategies, also known as Hawk-Dove.
Let the value of a food resource `V` = `+10` and the cost of an escalated fight `C` = `-100`.

Once an ESS is established, things calm down. This ties in with the proposition that evolutionary changes occur relatively fast, alternated with longer periods of relative evolutionary stability (Gould, 1972).

**RECIPROCAL ALTRUISM**

In social science, Axelrod (1984) shows that nice strategies work better in iterated games. In an iterated game there will be many successive encounters with the other player (Axelrod calls this "the shadow of the future"). In this case, playing **NASTY** is undesirable since it provokes the other player to retaliate and play **NASTY** too on the next encounter. The Cold War (1945–1991) is an exemplar of an endless cycle of nasty retaliation. Axelrod organized a computer tournament called Prisoner's Dilemma to which he invited game theory experts to submit programs. Each program consisted of a game strategy, for example "always play nasty" or "play nice with an occasional nasty sneak attack". The most successful program, Anatol Rapoport's TIT FOR TAT, was also the simplest, with four lines of programming code. TIT FOR TAT opens by playing **NICE**. In the next round it reciprocates the opponent's last move. Thereafter, it will always cooperate unless provoked. When provoked it retaliates in kind, but it is forgiving. In a game with enough TIT FOR TAT players, all-round cooperation emerges as the stable strategy.

Constructing a payoff matrix with the costs and benefits of each strategy explains how complex behavior such as cooperation can arise out of an interaction between underlying strategies. Creativity in nature is the result of evolution in a complex environment of interacting strategies.

## 1.4 Complex systems and emergent behavior

> The Company, with godlike modesty, shuns all publicity. Its agents, of course, are secret; the orders it constantly (perhaps continually) imparts are no different from those spread wholesale by impostors. Besides–who will boast of being a mere impostor? The drunken man who blurts out an absurd command, the sleeping man who suddenly awakes and turns and chokes to death the woman sleeping at his side–are they not, perhaps, implementing one of the Company's secret decisions? That silent functioning, like God's, inspires all manner of conjectures. One scurrilously suggests that the Company ceased to exist hundreds of years ago, and that the sacred disorder of our lives is purely hereditary, traditional; another believes that the Company is eternal, and teaches that it shall endure until the last night, when the last god shall annihilate the earth. Yet another declares that the Company is omnipotent, but affects only small things: the cry of a bird, the shades of rust and dust, the half dreams that come at dawn. Another, whispered by masked heresiarchs, says that *the Company has never existed, and never will.* Another, no less despicable, argues that it makes no difference whether one affirms or denies the reality of the shadowy corporation, because Babylon is nothing but an infinite game of chance.
>
> – Jorge Luis Borges, Fictions, The Lottery in Babylon (1962)

A rule or strategy is usually simple and straightforward. But a system in which multiple rules or strategies interact is complex to analyze. In the sense of Hölldobler & Wilson's superorganism, an ant colony as a whole exhibits complex behavior such as cooperative foraging or caterpillar herding, even though the individual parts (ants) are relatively simple. Other examples that demonstrate complex behavior include cells, nervous systems, social structures, human economies, the internet and the climate. Complexity (Waldrop, 1992) and self-organization (Kauffman, 1995) are studied in a wide range of domains, from biology, physics and chemistry to game theory, network theory, economics, politics and AI. The founding of the Santa Fe Institute (1984) has served as a significant milestone in the multidisciplinary study of systems.

**EMERGENCE**

Complex systems are said to operate "at the edge of chaos" but, not unlike "the survival of the fittest", this is rather vague and lends itself to embezzlement outside of its original context of cellular automata (Mitchell, Crutchfield & Hraber, 1994). We prefer to focus on the definition of emergence. According to Goldstein (1999), emergence is the arising of novel and coherent structures, patterns and properties during the process of self-organization in complex systems. In other words, emergent phenomena are unpredictable from the parts of the system. They manifest once the parts interact as a whole.

To illustrate this, we present an agent-based model that simulates cooperative foraging behavior. Agent-based modeling is an AI technique that describes a single autonomous agent (e.g., an ant) and its rules. Multiple agents can be put together to give insight in how they interact or how they evolve. A well-known example is BOIDS, a model for distributed flocking and swarming (Reynolds, 1987). Distributed flocking means that there is no leader in the flock. The fluid movement of the flock emerges from the interaction between the individual agents. Each boid (slang for "bird") observes its nearest neighbors to determine its own velocity and steering, using three rules:

| | |
|---|---|
| **SEPARATION** | Steer to avoid crowding neighbors. |
| **ALIGNMENT** | Steer towards the center of neighbors. |
| **COHESION** | Steer in the average direction of neighbors. |

By comparison, each ant in our foraging simulation has the following rules (figure 4):

| | |
|---|---|
| **ROAM** | Roam freely. Return home if nothing happens. |
| **TRACK** | Follow nearby trails to the food source, until the trail evaporates. |
| **HARVEST** | Collect nearby food. Mark the food source with pheromones. |
| **HOARD** | Return home with food. Mark the trail with pheromones. |

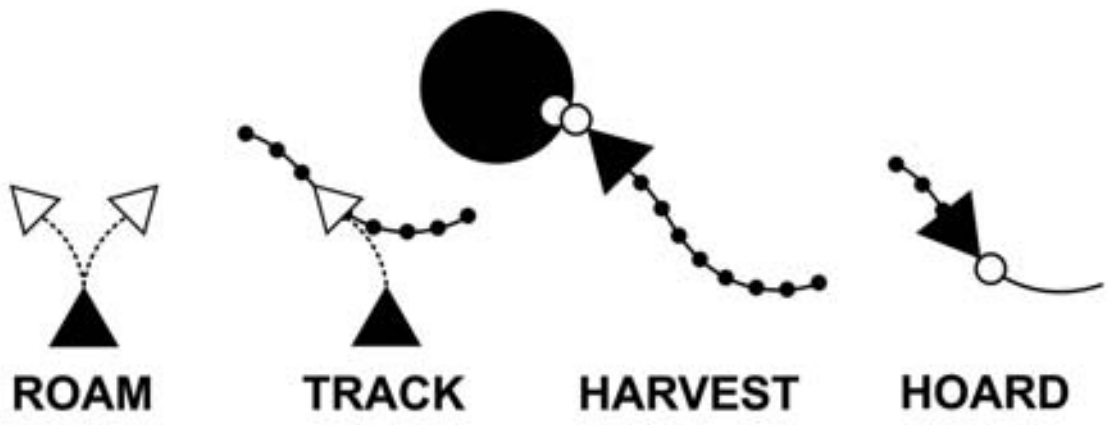

Figure 4. Distributed ant foraging behavior rules.

Collaborative behavior spontaneously emerges when multiple ants interact. When a given ant discovers a food source, it picks up some food and hauls it home while marking the trail with pheromones. Then other ants will discover the trail, head over to the food source, and so on. When observed as a whole, this self-organizing behavior is unexpected, since the individual agents are not programmed to collaborate. They adhere to four rules and that's it. The Python source code is given further below.

There are two different ways to consider the system. From a holistic stance, we can consider the whole. The whole exhibits emergent behavior (cooperation). It is more than the sum of its parts. From a reductionist stance, we can consider the individual parts. The four rules that define each part or agent are easily understood. It is not hard to predict what is going to happen, even without actually visualizing the simulation. The parts sum up to the whole. Note the `random()` in the program below, which yields a random number between `0.0` and `1.0`. It controls the ant's aimless **ROAM** behavior. It is used as an approximation of a subsystem, since "aimless roaming" is really the result of a complex of factors such as the terrain, the weather, scent, whether or not an ant's left-sided legs are slightly longer than its right-sided legs and so on. Because we do not know all the factors involved we take them as a given `random()`. In chapter 2, we will also rely on randomness to generate visual artworks, but we will abandon it in chapter 5, where our aim is to generate creative ideas in a deterministic, non-random way (i.e., from the parts).

Food sources are defined by a location in 2D space and an amount of available food. Larger food sources have a larger radius.

```
class Food:

    def __init__(self, x, y, amount=10):
        self.x = x
        self.y = y
        self.amount = amount
```

Pheromones are defined by a location in 2D space and a strength. Stronger pheromones have a larger radius. As time progresses, the strength of the scent evaporates.

```
class Pheromone:

    def __init__(self, x, y, strength=1.0):
        self.x = x
        self.y = y
        self.strength = strength

    def evaporate(self, m=0.99):
        self.strength *= self.strength > 0.01 and m or 0
```

Ants are defined by a location in 2D space, a direction (`v`) and a trail of `Pheromone` objects.
An ant may or may not be carrying a unit of food.

```
class Ant:

    def __init__(self, colony, x, y):
        self.colony = colony
        self.x = x
        self.y = y
        self.v = [0, 0]
        self.food = False
        self.trail = []
        self.roaming = random() * 100
```

The `roaming` attribute defines how many time cycles an ant will spend unsuccessfully looking for food before returning to the colony to start over.

We define a few helper functions: `steer()` directs an ant towards a pheromone, a food source, an ant or a colony; `pheromones()` yields all pheromones for a given colony.

```
def distance(v1, v2):
    return ((v1.x - v2.x) ** 2 + (v1.y - v2.y) ** 2) ** 0.5

def steer(ant, target):
    d = distance(ant, target) + 0.0001
    ant.v[0] = (target.x - ant.x) / d
    ant.v[1] = (target.y - ant.y) / d
    ant.roaming = 0

def pheromones(colony):
    for ant in colony.ants:
        for pheromone in ant.trail:
            yield pheromone
```

We can then define the rules: `roam()` randomly swerves the direction of an ant (or steers it to the colony if it is roaming too long), `track()` steers an ant to nearby pheromones, `harvest()` makes an ant pick up nearby food and start a trail of pheromones, `hoard()` steers an ant to the colony if it is carrying food, randomly leaving pheromones along the way.

```
from random import random

def roam(ant, m=0.3):
    ant.v[0] += m * (random() * 2 - 1)
    ant.v[1] += m * (random() * 2 - 1)
    ant.roaming += 1
    if ant.roaming > ant.colony.radius:
        steer(ant, ant.colony)
    if distance(ant, ant.colony) < 10:
        ant.roaming = 0

def track(ant):
    for pheromone in pheromones(ant.colony):
        if distance(ant, pheromone) < pheromone.strength * 30:
            if random() < pheromone.strength:
                steer(ant, pheromone)
            return

def harvest(ant):
    for food in ant.colony.foodsources:
        if distance(ant, food) < max(1, food.amount / 2):
            food.amount -= 1
            if food.amount <= 0:
                ant.colony.foodsources.remove(food)
            ant.trail = []
            ant.trail.append(Pheromone(food.x, food.y))
            ant.trail.append(Pheromone(ant.x, ant.y))
            ant.food = True

def hoard(ant, m=0.5):
    steer(ant, ant.colony)
    if random() < m:
        ant.trail.append(Pheromone(ant.x, ant.y))
    if distance(ant, ant.colony) < 5:
        ant.food = False
        ant.colony.food += 1
```

The `forage()` function simply bundles the rules and applies them:

```
def forage(ant, speed=1):
    if ant.food is False:
        roam(ant); track(ant); harvest(ant)
    else:
        hoard(ant)
    ant.v[0] = max(-speed, min(ant.v[0], speed))
    ant.v[1] = max(-speed, min(ant.v[1], speed))
    ant.x += ant.v[0]
    ant.y += ant.v[1]
```

A colony is defined as a collection of ants and available food sources in the vicinity of the colony's location in 2D space. When `Colony.update()` is called, pheromone strength evaporates while all ants continue foraging.

```
class Colony:

    def __init__(self, x, y, radius=200, size=30):
        self.x = x
        self.y = y
        self.radius = radius
        self.foodsources = []
        self.food = 0
        self.ants = [Ant(self, x, y) for i in range(size)]

    def update(self, speed=1):
        for ant in self.ants:
            for pheromone in ant.trail:
                pheromone.evaporate()
                if pheromone.strength == 0:
                    ant.trail.remove(pheromone)
            forage(ant, speed)
```

The system can be visualized in NODEBOX FOR OPENGL (see chapter 2, installation instructions in the appendix). The code below opens a 400 × 400 window displaying a colony of 30 ants and 10 random food sources. Each animation frame, a dot is drawn at the new position of each ant.

```
from nodebox.graphics import *

colony = Colony(200, 200, size=30)
colony.foodsources = [Food(random() * 400, random() * 400) for i in range(10)]

def draw(canvas):
    canvas.clear()
    for food in colony.foodsources:
        r = food.amount
        ellipse(food.x-r, food.y-r, r*2, r*2)
    for ant in colony.ants:
        ellipse(ant.x-1, ant.y-1, 2, 2)
    colony.update()

canvas.size = 400, 400
canvas.draw = draw
canvas.run()
```

A JavaScript implementation of the algorithm is provided as an example in PATTERN's online `canvas.js` editor[3] (discussed in chapter 6).

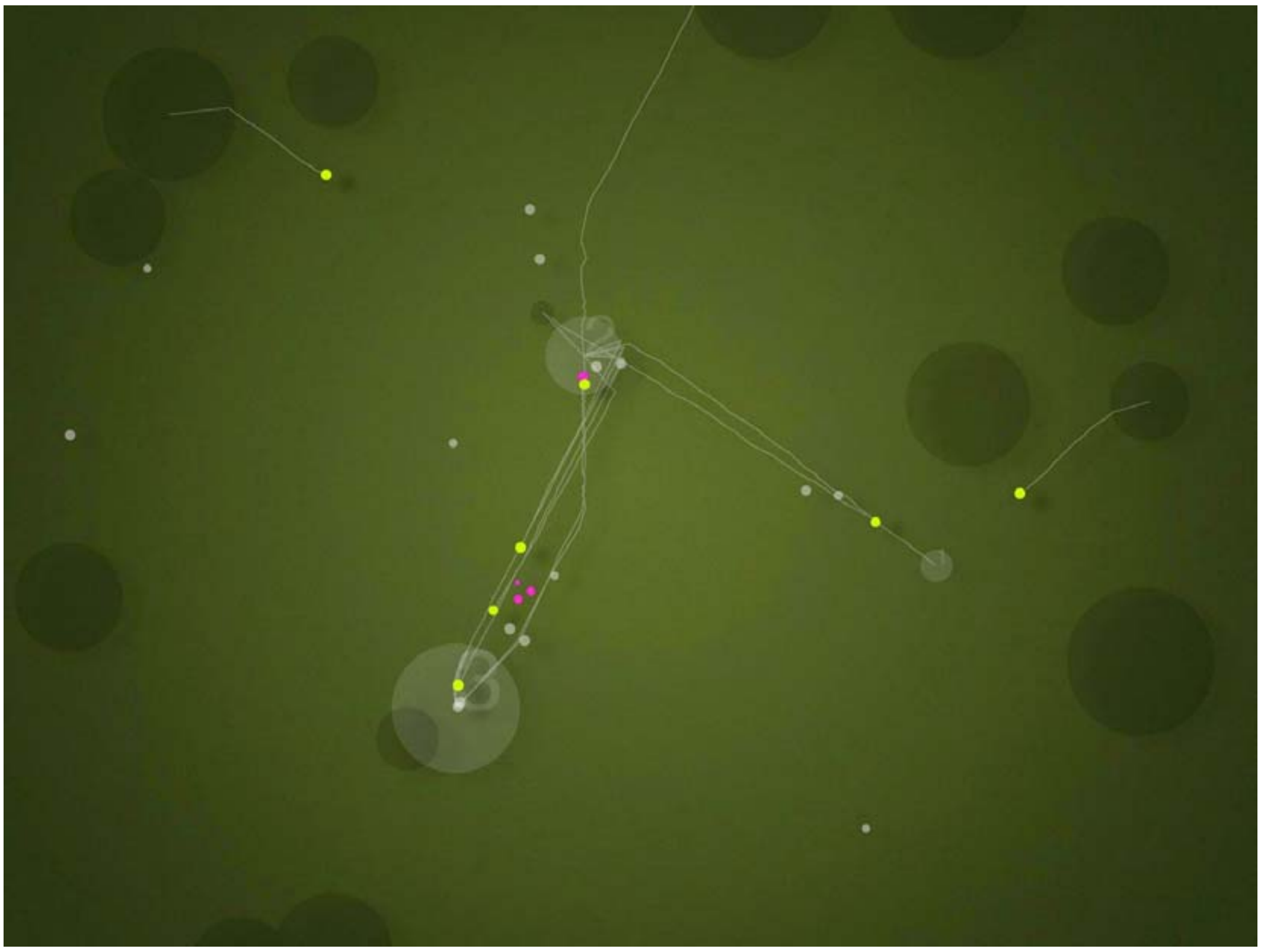

Figure 5. ANTAGONISM. Multiple colonies compete for food. Colonies 1, 2, 3 are marked in transparent grey. Ants carrying food are marked in green. Rivaling ants are marked in pink. Smaller dots are "injured" ants.

We can then define a 2D space with multiple ant colonies, where each colony's food hoard is also a food source for other colonies. A screenshot of the setup, called ANTAGONISM, is shown in figure 5. As food grows scarce, behavior than can be described as "plundering" and "warfare" arises. For example, colony 1 is located in a rich pocket of the virtual world, but while its ants are out collecting food it is consistently plundered by colony 2, which benefits from its central location. Colony 3 is located right on top of a food source, but it has few resources otherwise. Hence, it is involved in reciprocal parasitism with the nearby colony 2. The behavior seems to arise *ex nihilo* if the system is observed as a whole. But as we have seen, it is really the result of (complex) underlying strategies, in this case triggered by the setup of the artificial ecosystem (McCormack, 2007). This principle is fundamental to our stance on creativity in general. Creativity does not magically come out of nowhere. In nature, it is the result of evolution and competition and it is inherently blind. In humans, it is a conscious effort with the aim to solve a particular problem, rooted in the (complex) unconscious processes of the mind.

[3] http://www.clips.ua.ac.be/media/canvas/?example=class

## 1.5 Artificial life

The origin of artificial life systems such as EVOLUTION and ANTAGONISM can be traced back to von Neumann's hypothetical self-replicating machine in the 1940's, some years before Watson and Crick's discovery of DNA in 1953. In the 1960's, the first minicomputers appeared. Despite being called "mini" they were as big as a house. Because they remained unused at night (which was a waste of precious CPU cycles) scientists would use them for their own nightly entertainment. One of these pastimes was to implement Conway's GAME OF LIFE, a simplified version of the von Neumann replicator. GAME OF LIFE (Gardner, 1970) is a cellular automaton where each cell in an infinite grid is either **DEAD** or **ALIVE**. A set of rules defines how neighboring cells replicate or die out. For example: "each empty cell adjacent to exactly three neighbors becomes **ALIVE**". Different configurations of starting cells give rise to repeating patterns such as "guns" or "gliders" fired by guns. Combinations of patterns can be used for example to simulate a computer (i.e., it is Turing complete, Rennard, 2002) or a self-replicating machine (see Wade, 2010).

Another example of artificial life is Karl Sims' EVOLVED VIRTUAL CREATURES (Sims, 1994). In a virtual 3D world, creatures are evolved using a genetic algorithm. Each creature is composed of simple, connected blocks and selected for a specific task or behavior: swimming, hopping, following or competing with other creatures for the possession of a block. Another classic example is the VEHICLES thought experiment by Braitenberg (1984). In a Braitenberg vehicle, each of the wheels is connected to a light or heat sensor. Different connections make the vehicle behave differently and seemingly exhibit an emotional response (e.g., afraid of light). With the availability of inexpensive robotics kits such as Arduino[4] (2005) and Lego Mindstorms NXT[5] (2006), building Braitenberg vehicles has since become quite popular.

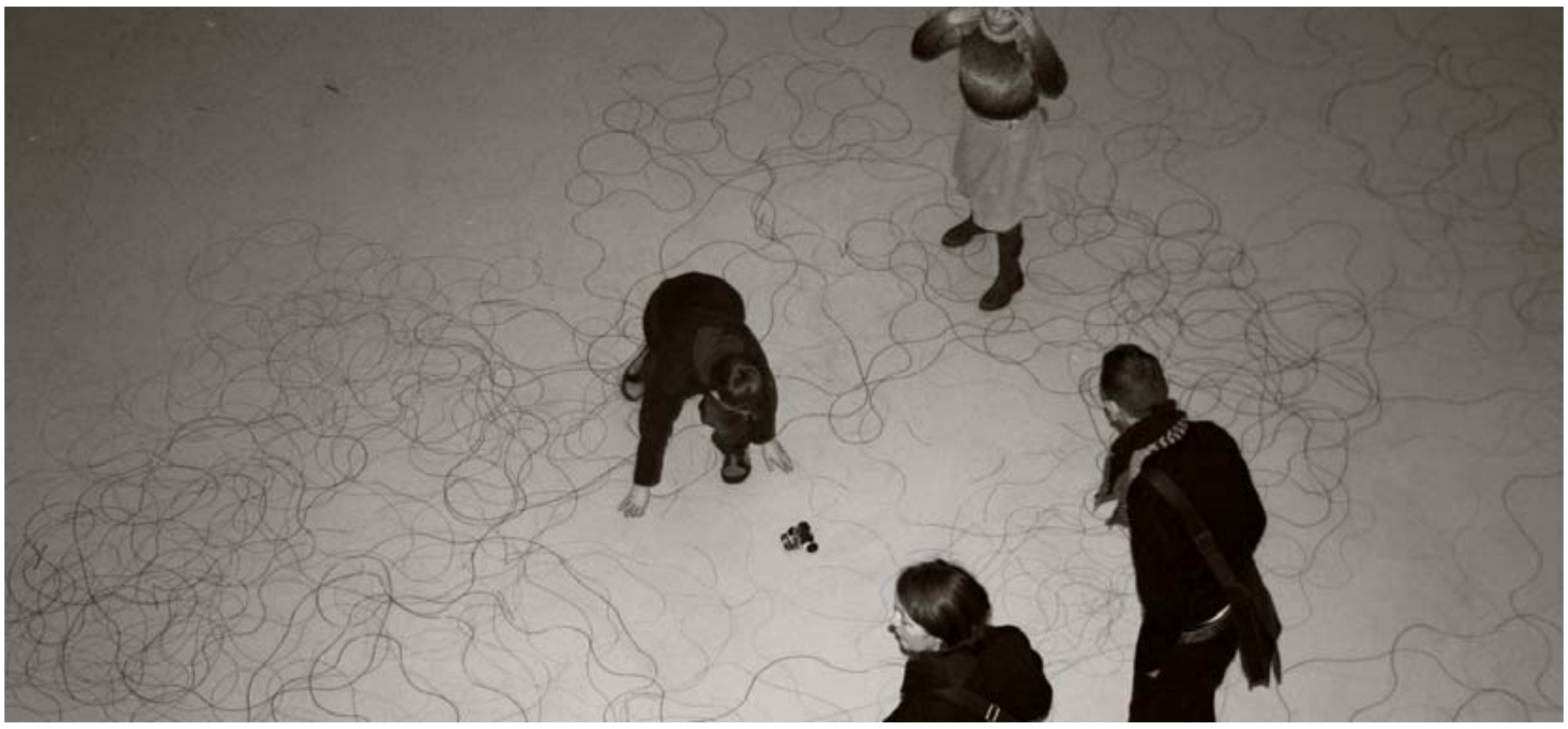

Figure 6. BLOTTER. Vehicles exhibited at the Museum of Modern Art, Antwerp, Belgium, 2005.
Photography © BLOTTER, Frederik De Bleser. Used with permission.

[4] http://www.arduino.cc/
[5] http://mindstorms.lego.com/

In 2005 we exhibited three vehicles at the Museum of Modern Art in Antwerp, Belgium. The project, called BLOTTER or "robot-plotter" (Menschaert, De Bleser & De Smedt, 2005), consisted of three vehicles that respond to sound by steering away from silent areas. This occasionally means running around in a circle chasing the noise produced by the vehicle's own wheel motors. Each of the vehicles had a waterproof marker attached to it, leaving an emergent pattern on the floor (figure 6). Our most intent audience was the museum's cleaning crew who thought it was the all-time most irritating art.

Interestingly, when a vehicle approached the feet of bystanders, some of them would stop to reprimand the robot or kneel down to pet it, not unlike they would a cat or a dog. Humans seem quick to attribute volition to machines or other organisms. Imagine you come home from work and the cat is waiting at the door to give you a warm greeting. But "warm greeting" is not in the cat's repertoire. It is exhibiting a trained behavior to promote feeding. You can reprimand the cat, in which case it will try again in a few minutes, or you can pet it and feed it, in which case you reinforce the behavior. When we say "the robot wants to inspect the noise" or "the cat wants to be fed" we are using "want" in a metaphorical sense. The robot does not have a feedback-processing loop to know what it wants. Whether the cat does is speculative, but we will attempt to answer it in chapter 4.

## 1.6 Discussion

In this chapter we have discussed how creative behavior can emerge as a result of evolution by natural selection. For example, in our EVOLUTION case study, virtual creatures adapt their behavior and gradually become more competitive, as weaker creatures are removed from both the tournament and the gene pool. This kind of creativity is essentially blind: there is no "want" in the repertoire. Proponents such as Dennett (2004) and Simonton (1999) argue that Darwinian creativity also applies to human creativity. The idea dates back to Campbell's BVSR model of "blind variation and selective retention" (Campbell, 1960). According to Simonton, since the incubation period between an individual's first attempt to solve a problem and the final solution is marked by all sorts of random input (everyday events, memories, associative thought), it can be considered a blind variation-selection process of the mind. Sternberg (1999) has criticized the universality of Darwinian creativity (i.e., in humans) as "pushing a good idea too far". He argues that Darwinian creativity is not-forward looking and not intentional, whereas creative humans are definitely intentional in their ways of selecting problems and solutions to solve them. This is discussed further in chapter 4.

We have discussed how creative behavior can emerge from complex systems that consist of a multitude of interacting strategies or rules. For example, in our ANTAGONISM case study, creative behavior such as cooperation arises when simple agents (ants) interact.

When multiple groups of agents (ant colonies) interact, even more complex behavior that can be described as plundering and warfare arises. A real-life example is the Cold War: an endless cycle of **NASTY** retaliation that sparked the "Space Race", thereby pioneering spaceflight, satellite technology and the internet. This shows how competition can be a creative catalyst.

EVOLUTION and ANTAGONISM were created in NODEBOX, a software application for generative art. Generative art is a form of art influenced by complexity and artificial life. This is discussed in more detail in chapter 2.

Central to our thesis is the idea that creativity does not magically arise *ex nihilo*, out of nothing. In nature it is grounded in the mechanisms of evolution and complexity. By comparison, human creativity is grounded in the unconscious processes of the mind, which is a complex system. It follows that we should be able to deconstruct the concept of creativity – open up the black boxes as we move deeper down – and then construct a computational model from the ground up. A case study along this line is presented in chapter 5. But in the next chapter we will first look at generative art, a rule-based form of art based on complexity in nature. We will present examples, and discuss what it means exactly to "move deeper down" in the context of artistic creativity.

# 2 Generative art

## What is generative art?

Generative art is a form of art inspired by nature and complexity (emergence in particular), and making use of techniques borrowed from AI and artificial life such as evolutionary algorithms and agent-based modeling. We have discussed evolution, complexity and artificial life in chapter 1.

In a generative artwork, individual features such as lines, shapes or colors interact autonomously using a system of rules, typically a computer program. An example called SELF-REPLICATION (De Smedt & Lechat, 2010) is shown in figure 7. It is also the cover image. The overlay figure shows the individual shapes; the building blocks that make up the composition. These are replicated over and over using a set of rules described in programming code. For example: "shape `X` rotates to create a circle", "shape `Y` advances to create a chain", "chains curl near circles" and "circles spawn adjacent circles". The last rule is an application of the Apollonius problem in geometry (see Coxeter, 1968).

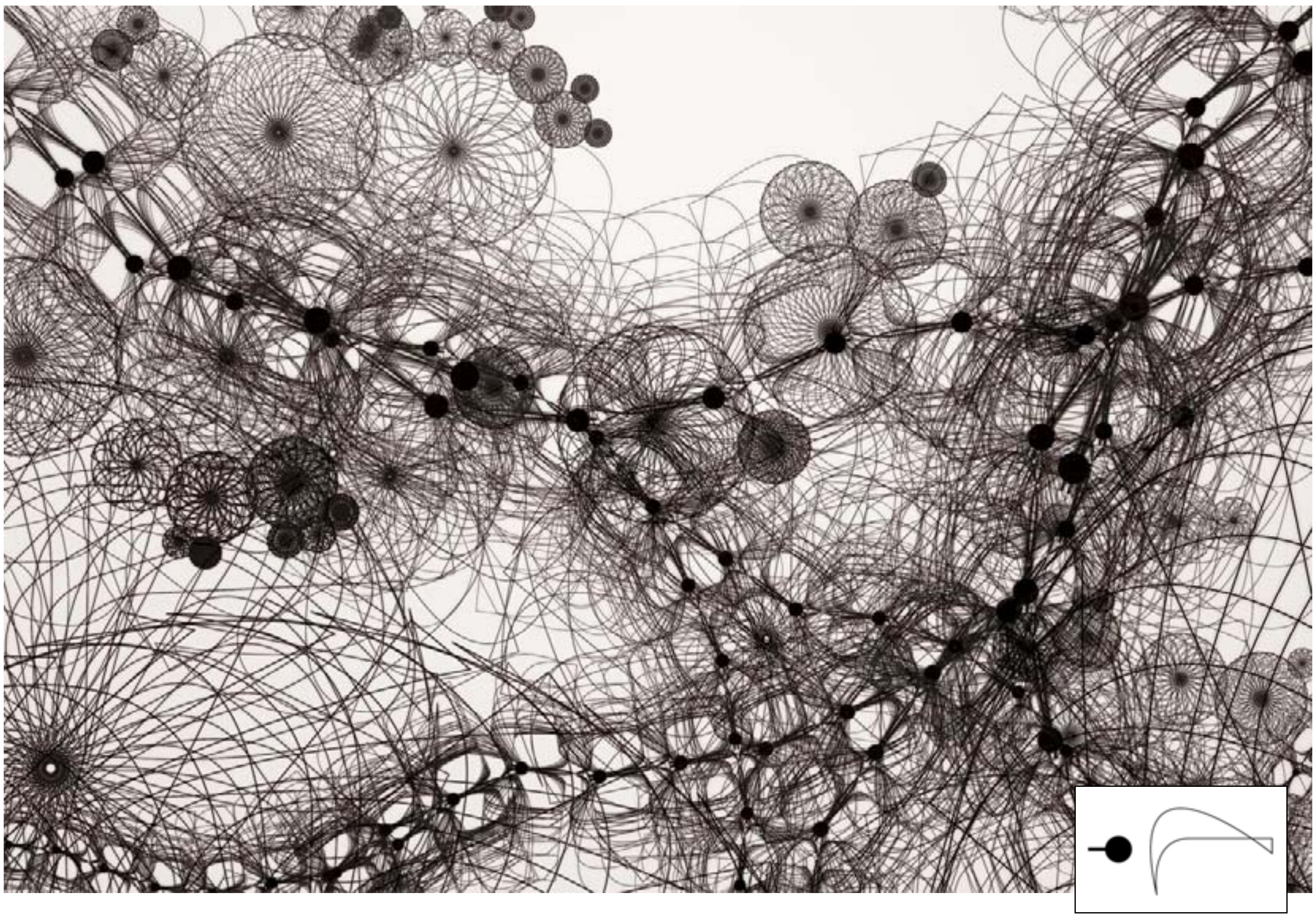

Figure 7. A variation of SELF-REPLICATION rendered in NODEBOX.
The overlay figure shows the individual building blocks.

Rule-based systems in generative art often involve an amount of randomness to promote unexpected variation. For example, in SELF-REPLICATION not all chains will curl near a circle. Some will ignore the rule and do nothing. Others might branch into new chains instead of curling. This way, each time the program is executed it produces a new artwork with different chains and circles. This unexpectedness, either as a result of randomness or of complexity, is referred to as self-organization or emergence (McCormack & Dorin, 2001).

#### PROCEDURAL

Generative art is related to the broader field of computer art. However, not all computer art is generative art (Boden, 2009). For example, in video games, web design or in Photoshop compositions the computer is used as a tool rather than as a generator. But there is at least one technique used in video games that qualifies as generative art: procedural modeling (Ebert, Musgrave, Peachey, Perlin & Worley, 2003). Procedural modeling involves rendering natural material such as marble, stone and wood with a mathematical algorithm instead of using photographs.

Generative art generally lacks a commercial or political motive. Many generative artists explore rule-based systems out of curiosity and for aesthetic or scientific pleasure, rather than as a means to express an opinion. In recent years, generative approaches are also being applied to more practical applications in the form of data visualization (Fry, 2008). For example, one approach to visualize a network of links between various web pages is to use the following rules: "web pages are represented as knots, links as lines" and "all knots repulse each other" but "lines act as springs that attract two knots". This is a specific kind of particle system (see further). Such kind of data-driven generative art is often characterized by a volatile, dynamic nature: the work in its current form may only exist at this particular time. The form changes as new data becomes available or as multiple authors contribute new rules (e.g., crowdsourcing).

Generative art represents a method to create art (i.e., rule-based) rather than a consistent artistic style (Galanter, 2003). Style and medium can vary greatly. Some generative artists use computer graphics while others build music composition systems, swarm robots, installations with growing plants, and so on. This diversity does not mean that generative art has a lack of interest in form as such. On the contrary, Galanter (2012) remarks:

> Generative art, and especially generative art that harnesses what we are learning from complexity science, is a unique opportunity to rehabilitate formalism in art. It presents form as anything but arbitrary. It presents beauty as the result of an understandable universe neutral to human social construction in a fair and unbiased way.

Galanter's description signals how generative art ties in with the scientific field, by taking an unbiased stance towards style.

## 2.1 Brief history

The term "generative art" has gained popularity over the last decade, with principal instigators such as the Processing[6] open source community and Marius Watz' Generator.x[7] forum. To date, the most widely quoted definition of generative art is again offered by Galanter (2003):

> Generative art refers to any art practice where the artist uses a system, such as a set of natural language rules, a computer program, a machine, or other procedural invention, which is set into motion with some degree of autonomy contributing to or resulting in a completed work of art.

In later work Galanter has further emphasized the aspect of autonomy. To qualify as generative art, he notes, a rule-based system must be well-defined and self-contained enough to operate autonomously (Galanter, 2008). This excludes a handmade drawing. But it includes ancient tiling patterns, where individual tiles are placed according to a symmetry-based algorithm, and for example Sol LeWitt's combinatorial sculptures and wall drawings. In fact, LeWitt's conceptual approach to art, where "the idea becomes a machine that makes the art" (LeWitt, 1967), was already well-established in the field of computer art some years before (Nake, 2010). Another example at the foundation of generative art is the graphic work of Escher (1960), exploring concepts such as self-reference (Prententoonstelling, 1956), recursion (Circle Limit III, 1959) and infinity (Waterfall, 1961). Since Escher's work is mostly hand drawn it is not generative in the true sense, but his notebooks contain resourceful rule systems which he used for tessellation or tiling. For example, the fish in Circle Limit III are based on the Poincaré disk model for hyperbolic geometry, introduced to Escher by Coxeter (1979).

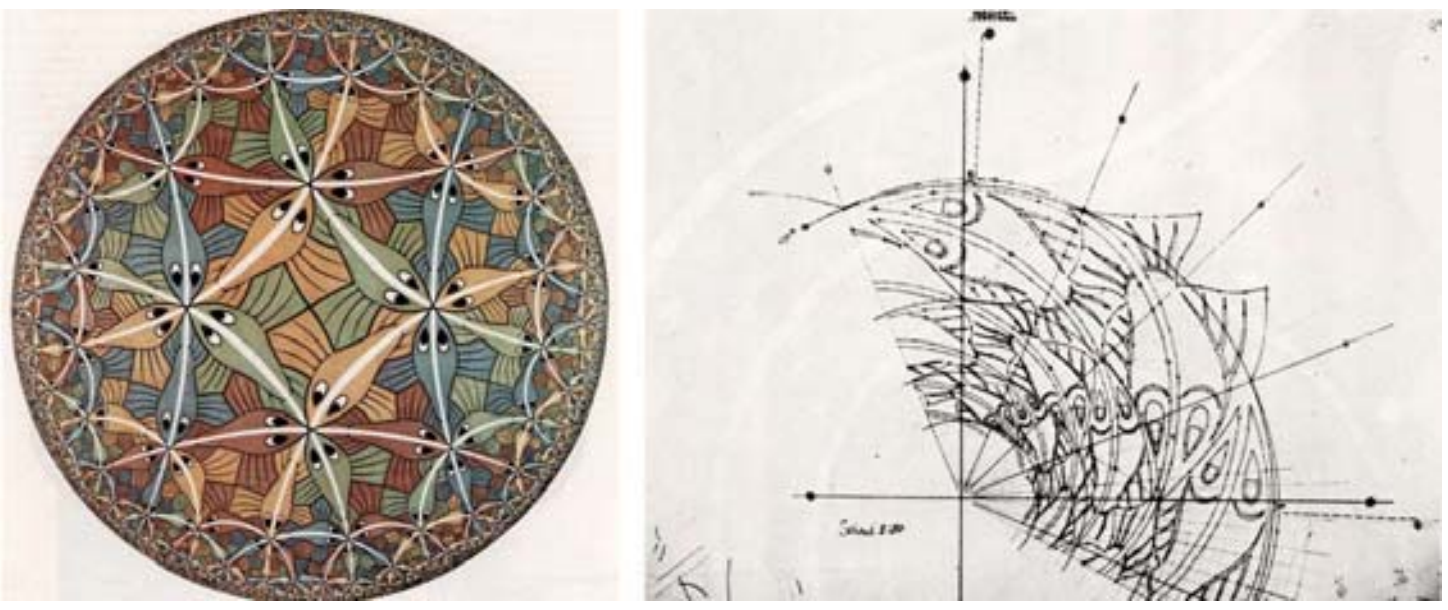

Figure 8. "Circle Limit" III and "Circle Limit III original sketch".

Watz (2005) agrees with Galanter that generative art can be found throughout history, but he attributes the term specifically to computer-based work created since the 1960's. This includes the Artificial Life systems discussed in chapter 1, for example Conway's GAME OF LIFE and BOIDS. Other milestones include the work on fractal geometry by Mandelbrot (1982), L-systems (Prusinkiewicz & Lindenmayer, 1990) and particle systems (Reeves, 1983).

[6] http://processing.org/
[7] http://www.generatorx.no/

#### FRACTALS

A fractal is a mathematical description of a self-similar (i.e., recursive) shape, consisting of smaller versions of itself. Examples include snowflakes, fern leaves and coastlines. In the 1970's Mandelbrot convincingly used computer graphics to illustrate and popularize the development of fractals. Today, fractals play an important role in procedural modeling of surface and terrain (Perlin, 1985). A variation of the fractal, the attractor, is also popular in generative art.

#### L-SYSTEMS

An L-system is a recursive rewriting system that can be used to model growth in plants. For example, algae growth can be described as `(A → AB), (B → A)` where `A` is rewritten to `AB`, then `ABA`, `ABAAB`, `ABAABABA`, and so on. Note that the length of the consecutive strings also satisfies the Fibonacci sequence (1, 2, 3, 5, 8, ...). L-systems were developed by Lindenmayer concurrent with Mandelbrot's computer-generated fractals. Today, they are widely used in computer games to render procedural plants[8], trees, or even entire cities (Parish & Müller, 2001).

#### PARTICLE SYSTEMS

A particle system is a technique where a multitude of particles interact using attractive and repulsive forces and collision. Particle systems are used in video games and film to model fuzzy phenomena such as fire, fluids, smoke, fog, dust and explosions. The term was coined in the 1980's by Bill Reeves while working at Lucasfilm on the special effects for the movie *Star Trek II: The Wrath of Khan*.

The 1970's and the 1980's started the home computer revolution. Video game consoles such as the Atari (1977) and the Commodore 64 (1982) appeared, along with personal computers such as the IBM PC (1981) and Apple's Macintosh (1984). In 1971, Lucasfilm was founded. For a time (1979–1986) it incorporated Pixar, well known for its computer animation films such as Toy Story and WALL-E. From the point of view of generative art, Pixar is important for its contribution to the RenderMan 3D graphics software (1989), which advanced the use of procedural techniques. RenderMan programs are written in the C programming language, developed by Dennis Ritchie at Bell Labs in 1972. Even though nowadays C is considered a "low-level" language it remains in wide use (Linux, Windows and Mac OS are based on C) and it has a close relationship with computer graphics because it is fast. Later, C++ (1983), Python (1991) and Java (1995) were developed to augment C with additional flexibility, for example object-oriented programming, garbage collection and simpler syntax, typically at the cost of reduced speed. Such "friendly" programming languages have helped to attract a more artistic-minded audience. In 1982, John Warnock founded Adobe Systems and created PostScript, a programming language for 2D page layout. Later on, Adobe produced computer graphics applications such as Photoshop and Illustrator, which continue to be popular with artists and designers.

[8] http://www.clips.ua.ac.be/media/canvas/?example=lsystem

By the 1990's the internet and consequently the World Wide Web were in full swing, with substantial impact on the way humans would process information. Evident examples include Google's search engine, Twitter's microblogging service, Facebook's social network, YouTube videos, Amazon.com and the iTunes Store. The control exerted by major corporations over available information (internet corporatization) has been subject to criticism (Rethemeyer, 2007, and Zimmer, 2008). By contrast, open source communities such as the Wikipedia online dictionary generally lack control structure or commercial interest. They play a significant role in the context of generative art.

#### OPEN SOURCE

Open source communities encourage collaborative development of free software. This means that anyone is allowed to edit the software. Bits and pieces are done by various people. To prevent abuse, software is typically released under a GNU General Public License (GPL). This strategy forces any derivatives of the software to become open source GPL as well. In other words, it is a viral license (Lerner & Tirole, 2002). In terms of artistic software, users are then encouraged to share artworks created with the software and to contribute bug reports. The developers benefit with a virtual showcase of art created using their tools. The artists benefit with free software. This process intensifies as the software grows: developers may move on to become artists, artists may become contributing developers. This reward in the form of increased personal proficiency is also called human capital (Hars & Ou, 2002).

A good example is the Processing community website, which combines the documentation of its open source software with a discussion forum and an online gallery. This collaborative approach has led to the renown of several generative artists, including Ben Fry, Casey Reas, Daniel Shiffman, Ira Greenberg, Jared Tarbell, Jer Thorp, Josh Nimoy, Karsten Schmidt, Marius Watz and Tom Carden.

Other well-known generative artists, not necessarily from the Processing community, include: Aaron Koblin (BICYCLE BUILT FOR TWO THOUSAND), Adrian Bowyer (REPRAP self-replicating 3D printer), ART+COM (KINETIC SCULPTURE), Bogdan Soban, Brendan Dawes, Celestino Soddu, David Rokeby, Frieder Nake, Golan Levin, Iannis Xenakis (generative avant-garde music), Jonah Brucker-Cohen, Jon McCormack (MORPHOGENESIS), Joshua Davis (pioneer of generative art in Flash), Ken Rinaldo, LAb[au], Mark Napier, Paul Bourke (geometry & fractals), Philip Worthington (SHADOW MONSTERS), Robert Hodgin (FLOCKING), Sergi Jordà (REACTABLE music instrument, together with Marcos Alonso, Martin Kaltenbrunner and Günter Geiger), Tim Blackwell (SWARM MUSIC), Tom Gerhardt (MUDTUB) and Tom Igoe (MAKING THINGS TALK).

Figure 9.1 shows TISSUE by Reas (2002), which models a synthetic neural system. Participants can then interact with the installation:

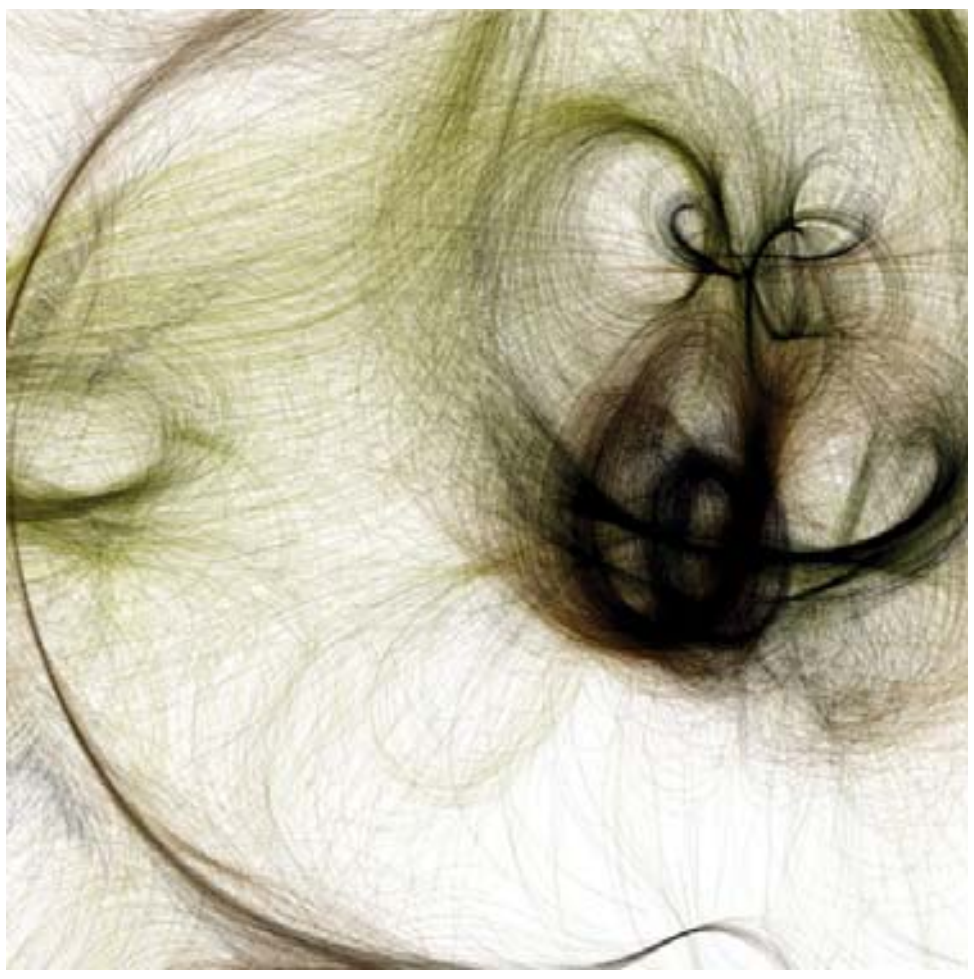

Figure 9.1. TISSUE (2002).
© Casey Reas. Used with permission.

Figure 9.3 shows SUPER HAPPY PARTICLES by Shiffman (2004), a live video installation where each pixel is a particle in a particle system:

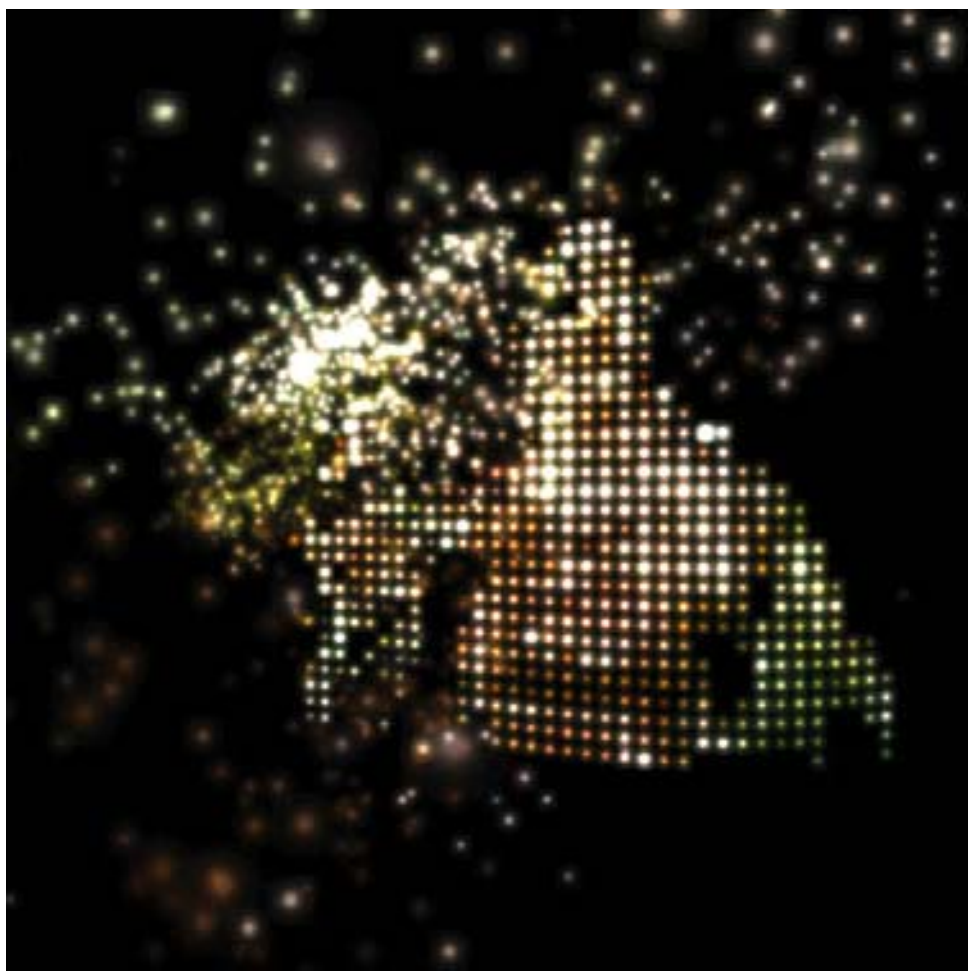

Figure 9.3. SUPER HAPPY PARTICLES (2004).
© Daniel Shiffman. Used with permission.

Figure 9.2 shows GENOME VALENCE by Fry (2002), a data visualization for DNA sequences retrieved from the BLAST search algorithm:

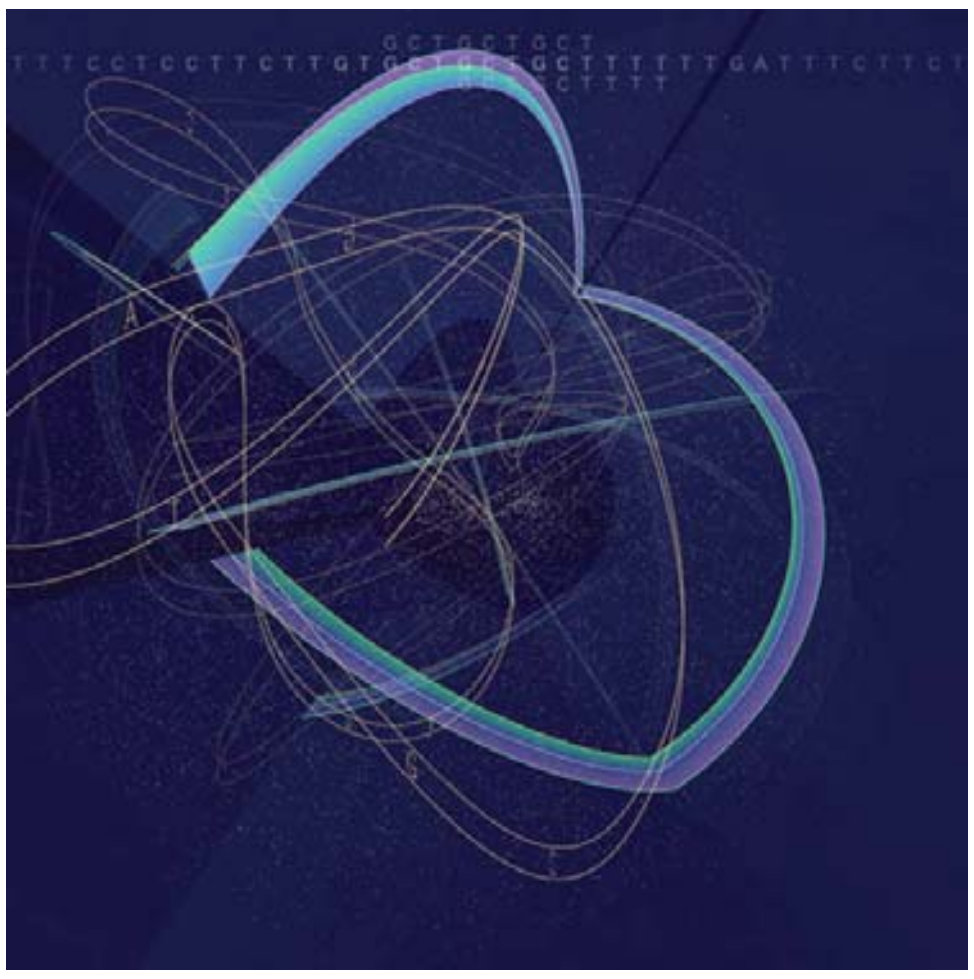

Figure 9.2. GENOME VALENCE (2002).
© Ben Fry. Used with permission.

Figure 9.4 shows METROPOP DENIM by Carden (2005), which combines attractors and particles with fashion photography:

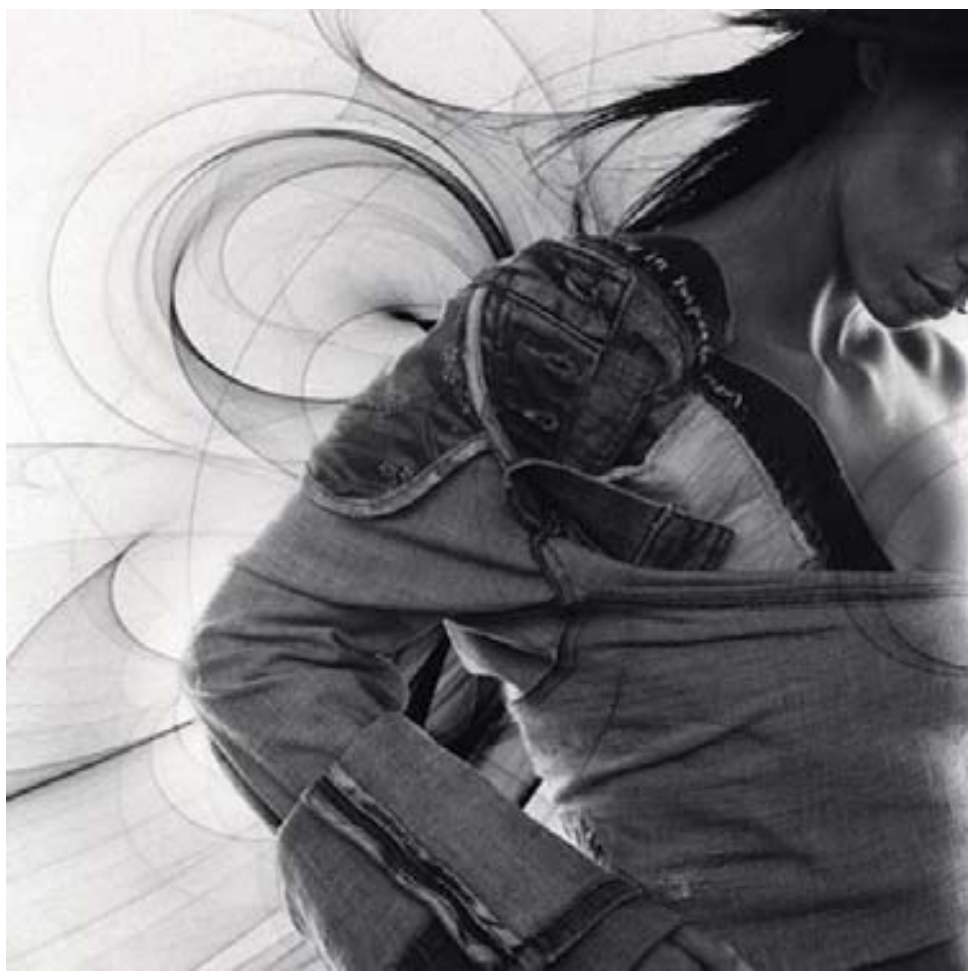

Figure 9.4. METROPOP DENIM (2005).
© Clayton Cubitt & Tom Carden. Used with permission.

Figure 9.5 displays an effect by Nimoy (2010) for the movie *Tron Legacy*, based on techniques such as particle systems and Perlin noise:

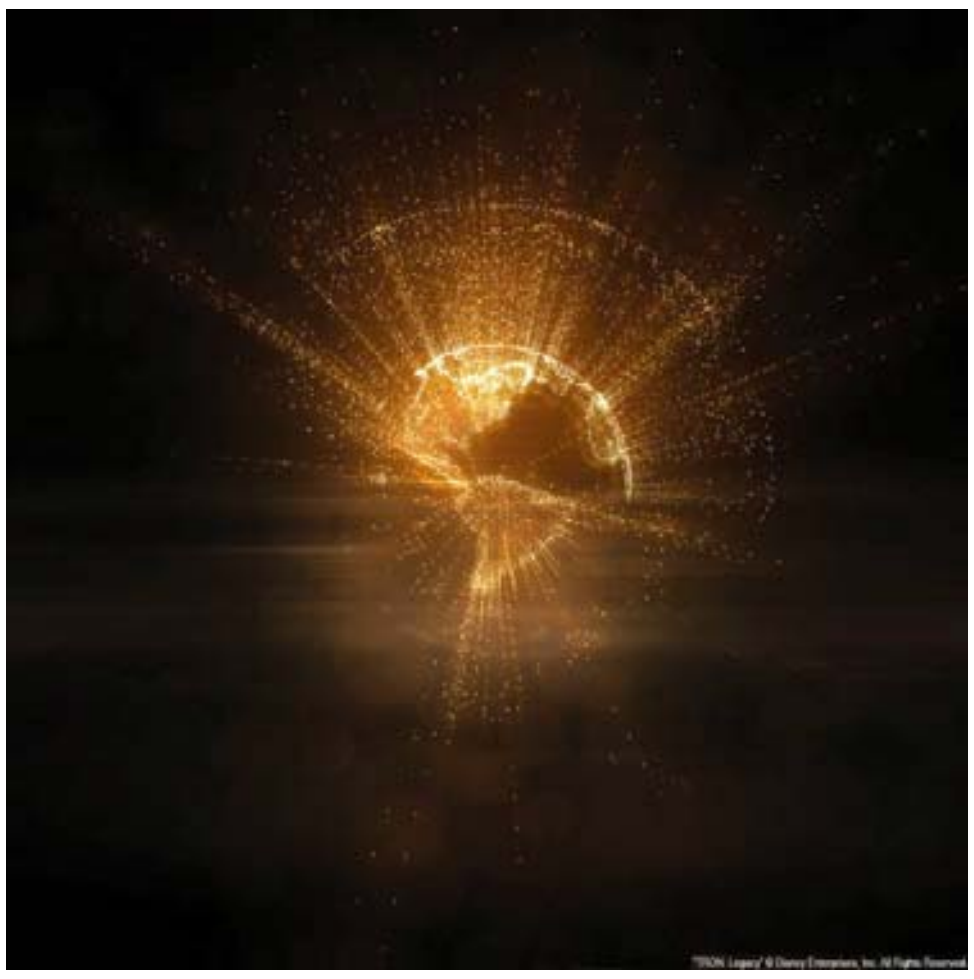

Figure 9.5. Tron Legacy special effect (2010).
© Josh Nimoy et al. Used with permission.
Disney Enterprises, Inc. All rights reserved.

Figure 9.6 displays 138 YEARS OF POPULAR SCIENCE by Thorp (2011), a data visualization of the Popular Science magazine archive:

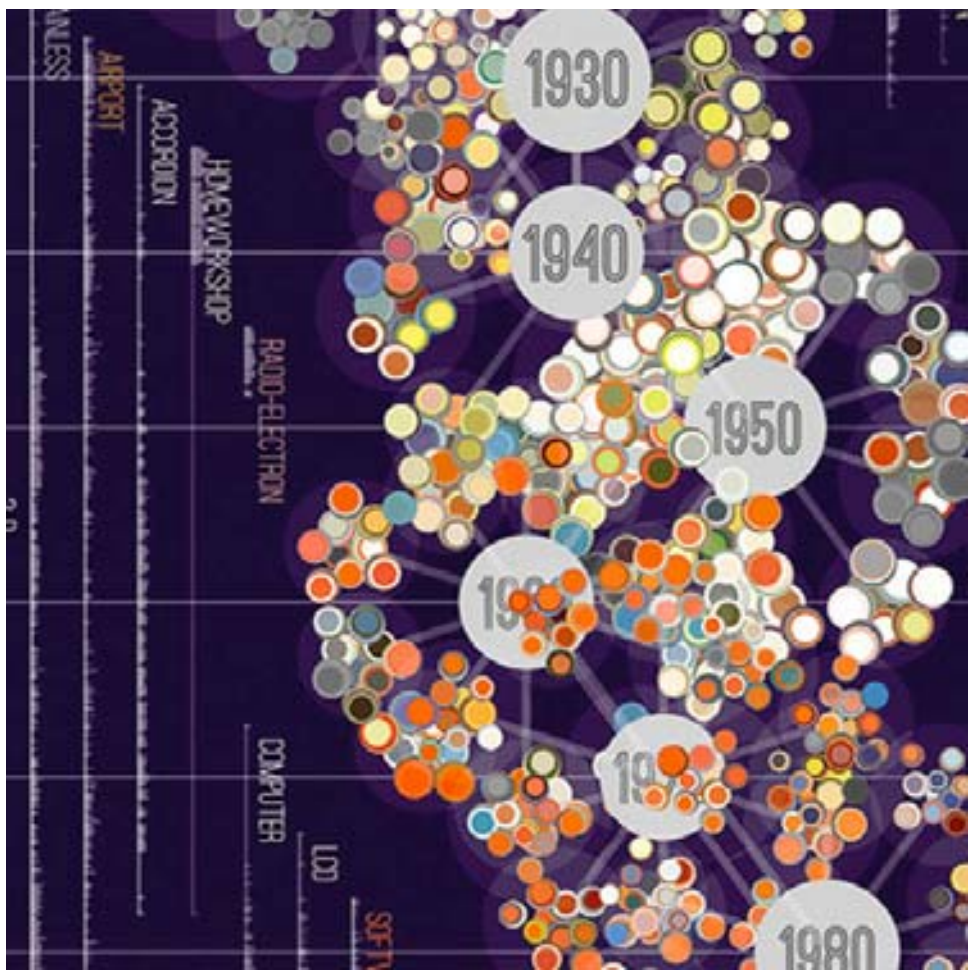

Figure 9.6. 138 YEARS OF POPULAR SCIENCE (2011).
© Jer Thorp & Mark Hansen. Used with permission.

Figure 9.7 is MORPHOGENESIS by McCormack (2011), an example of evolutionary art based on a genetic algorithm:

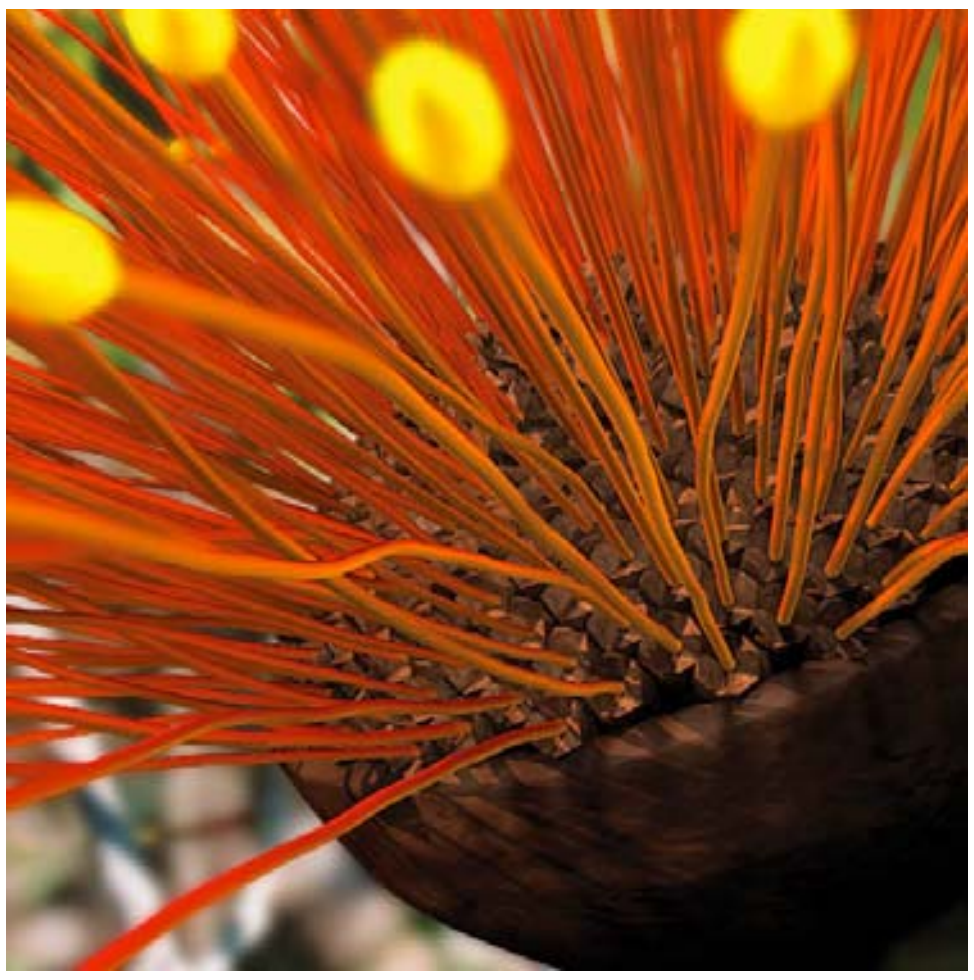

Figure 9.7. MORPHOGENESIS (2011).
© Jon McCormack. Used with permission.

## 2.2 Overview of software for generative art

Table 2 provides a 1990–2010 overview of programming languages and software applications that have been used to produce generative art. There may be others not to our knowledge. Over 70% of the projects listed is open source. Based on the steady introduction of new toolkits and derivatives we can say that rethinking artistic software, or at least adopting new toys, is a factor that contributes to the definition of generative art.

Of note is the work by Maeda (2001). While at MIT he created Design By Numbers (DBN), a software application that generates visual output based on simple programming commands such as `pen` and `line`. Students of Maeda later expanded his work into Processing (Reas & Fry, 2007), arguably the most well-known software application for generative art. In 2005 Reas and Fry received the prestigious Golden Nica award for their work. Processing is based on the Java 2D graphics library and influenced by technologies such as PostScript and OpenGL. This work has in turn led to the development of Wiring, a Processing spin-off for electronics, and Arduino, a microcontroller chip designed "to make things talk" in interdisciplinary projects (Igoe, 2007). It is interesting to note how derivatives of DBN are reminiscent of Maeda's initial goal – achieving simplicity in the digital age (Maeda, 2006). Processing describes itself as a "sketchbook" for programming, using combinations of simple commands such as `line()` and `rotate()`. Wiring is "thoughtfully created with designers and artists in mind"[9] (i.e., a non-technical audience) and openFrameworks is intended for "folks using computers for creative, artistic expression"[10]. DrawBot and its derivative NodeBox adopt the Python programming language, which arguably has simpler syntax than Java (no curly braces or semicolons).

The authors of DrawBot, Just van Rossum and Erik van Blokland, define the simplicity of a computational approach to art and design in terms of volume, complexity and variation:

**VOLUME** In large-volume assignments (e.g., a 1,000-page book), targeting the task as a whole is easier; we adopt a systematic approach instead of handling each feature (page) individually.

**COMPLEXITY** It is easier to adapt a layout system that controls color and composition across pages than to adjust each page by hand (e.g., "Can you make all titles bigger?").

**VARIATION** A layout system can be adapted up to the final moment, which leads to a form of digital sketching where many alternatives can be tried out in a non-destructive way.

The authors of NodeBox offer an additional metaphor. Traditionally, user interfaces in computer graphics applications have been based on real-world analogies, e.g., a pen for drawing, a stamp for copying and scissors for slicing. This model has creative limitations. First, the features can only be used as the software developers implemented them, creative recombination of tools is impossible when not foreseen.

[9] http://wiring.org.co/

[10] http://www.openframeworks.cc/about/

Second, there is not much room for abstraction: users will tend to think along the lines of what is possible with the built-in features (buttons, sliders, menus) and not about what they want (Cleveland, 2004). These limitations are mitigated when users can freely combine any kind of functionality in the form of programming code. For beginners, it is often helpful to explain this paradigm in terms of a cooking recipe:

**RECIPE** We can think of programming code as a recipe, a set of instructions that describes how to prepare or make something.

**INGREDIENTS** Any kind of functionality can freely be combined in programming code. Compared to a prepackaged dinner (i.e, buttons and sliders) you have to do a bit of work but you can do it any way you want, mixing any ingredients you like.

**COOKBOOK** Programming code is written down. It captures the creation process, as opposed to mouse gestures, and as such can be reused, adapted, shared and studied.

| # | YEAR | TOOLKIT | LANGUAGE | OPEN | DESCRIPTION |
|---|---|---|---|---|---|
| 1 | 1988 | DIRECTOR | `Lingo` | NO | Application for 2D multimedia CD-ROM's. |
| 2 | 1989 | RENDERMAN | `C` | NO | Application for 3D distributed rendering. |
| 3 | 1992 | OPENGL | - | YES | API for 3D computer graphics (cross-platform). |
| 4 | 1995 | DIRECTX | - | NO | API for 3D computer graphics (Windows, XBox) |
| 5 | 1996 | FLASH | `JavaScript` | NO | Application for 2D animation in web pages. |
| 6 | 1996 | SUPERCOLLIDER | `sclang` | YES | Programming language for audio synthesis. |
| 7 | 1996 | PURE DATA | dataflow | YES | Visual programming for computer music. |
| 8 | 1998 | VVVV | dataflow | NO | Visual programming for computer graphics. |
| 9 | 1998 | JAVA 2D | `Java` | NO | 2D graphics library for Java. |
| 10 | 1999 | OPENSCENEGRAPH | `C++` | YES | 3D graphics library for C++ and OpenGL. |
| 11 | 1999 | DBN | `DBN` | YES | 2D graphics application for teaching and sketching. |
| 12 | 2000 | PIL | `Python` | YES | 2D graphics library for Python. |
| 13 | 2001 | PROCESSING | `Java` | YES | 3D graphics application for teaching and sketching. |
| 14 | 2001 | SCRIPTOGRAPHER | `JavaScript` | YES | JavaScript plugin for Adobe Illustrator. |
| 15 | 2002 | QUARTZ | `Cocoa` | NO | 2D graphics library for Mac OS X. |
| 16 | 2002 | AGG | `C++` | YES | 2D rendering engine with anti-aliasing. |
| 17 | 2002 | DRAWBOT | `Python` | YES | 2D graphics application for teaching and sketching. |

| # | YEAR | TOOLKIT | LANGUAGE | OPEN | DESCRIPTION |
|---|---|---|---|---|---|
| 18 | 2003 | NODEBOX | `Python` | YES | 2D graphics application for teaching and sketching. |
| 19 | 2003 | WIRING | `Java` | YES | Application for electronics prototyping. |
| 20 | 2004 | QUARTZ COMPOSER | dataflow | NO | Visual programming language for Mac OS X. |
| 21 | 2004 | CONTEXT FREE | `CFDG` | YES | 2D graphics application based on formal grammar. |
| 22 | 2005 | ARDUINO | `Java` | YES | Microcontroller chip programmed in Wiring. |
| 23 | 2006 | OPENFRAMEWORKS | `C++` | YES | 3D graphics library for C++. |
| 24 | 2006 | <CANVAS> | `JavaScript` | NO | Scriptable graphics in a HTML <canvas> element. |
| 25 | 2007 | SHOEBOT | `Python` | YES | 2D graphics application for teaching and sketching. |
| 26 | 2007 | SILVERLIGHT | `.NET` | NO | Application for 2D web animation (cfr. Flash). |
| 27 | 2007 | PYGLET | `Python` | YES | 3D graphics library for Python. |
| 28 | 2008 | SHOES | `Ruby` | YES | 2D graphics application for teaching and sketching. |
| 29 | 2008 | RUBY-PROCESSING | `Ruby` | YES | Ruby-based Processing. |
| 30 | 2008 | FIELD | hybrid | YES | Application for digital art. |
| 31 | 2008 | RAPHAEL.JS | `JavaScript` | YES | 2D graphics library for JavaScript (SVG). |
| 32 | 2008 | PROCESSING.JS | `JavaScript` | YES | Web-based Processing using <canvas>. |
| 33 | 2009 | WEBGL | `JavaScript` | NO | Web-based OpenGL using <canvas>. |
| 34 | 2009 | NODEBOX 2 | dataflow | YES | NodeBox using a visual node-based interface. |
| 35 | 2010 | THREE.JS | `JavaScript` | YES | 3D graphics library for JavaScript. |
| 36 | 2011 | PAPER.JS | `JavaScript` | YES | Web-based Scriptographer using <canvas>. |
| 37 | 2011 | NOGL | `Python` | YES | NodeBox for OpenGL using Pyglet. |
| 38 | 2012 | D3.JS | `JavaScript` | YES | Data visualization library for JavaScript (SVG). |
| 39 | 2012 | CANVAS.JS | `JavaScript` | YES | Web-based NodeBox using <canvas>. |

Table 2. Overview of programming languages and software applications for generative art.

In recent years, effort appears to focus on JavaScript visualization toolkits such as WebGL and Processing.js. JavaScript is a programming language that is commonly used in web pages. As such, these visualization toolkits are directly available in the web browser, across different architectures and devices, without requiring installation. In chapter 6, we will briefly discuss our own JavaScript toolkit called canvas.js, which is part of the PATTERN software package.

## 2.3 NodeBox

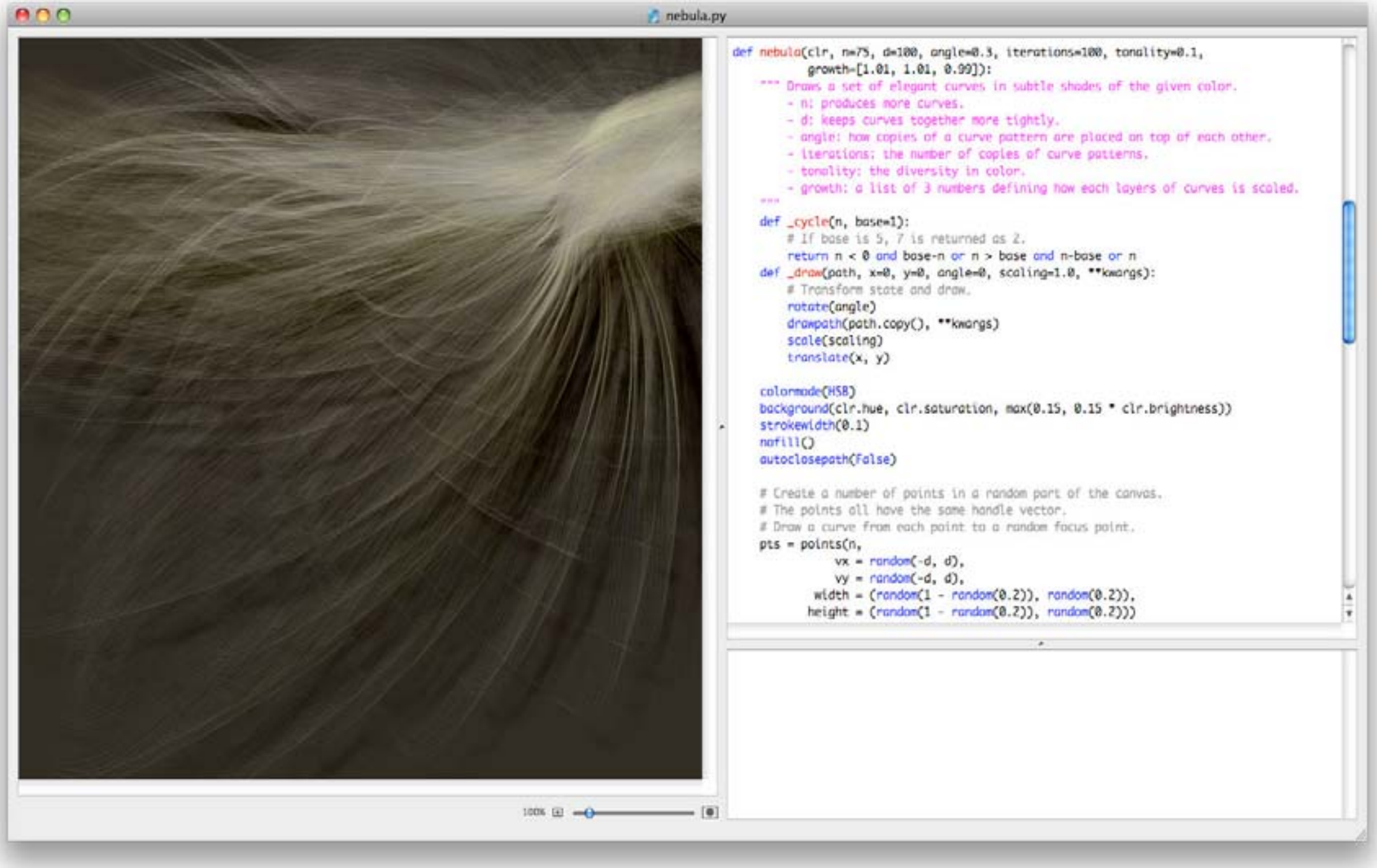

Figure 10. Screenshot of the NodeBox application.

NODEBOX (De Bleser, De Smedt & Nijs, 2003) is a free computer graphics application for Mac OS X that generates static, animated or interactive 2D visuals based on Python programming code. The visuals can be exported as PDF (vectors), PNG (pixels) or MOV (movie). The application implements a state machine with a scene graph built on Mac OS X's Quartz rendering engine. A state machine means that it has a current transformation state and a current color state, which are applied to all successive objects drawn to the canvas. A scene graph means that the objects are stored in a data structure so that they can be manipulated before or after they are drawn, as opposed to a direct drawing approach. Figure 10 shows a screenshot of the application.

Using Python code, users are free to combine any kind of functionality to produce visual output. Below is a simple example. It draws a random number of 10–100 squares on a 500 × 500 canvas, in random shades of transparent red, with random size, position and rotation.

```
size(500, 500)
for i in range(10, 100):
    x = random(WIDTH)
    y = random(HEIGHT)
    w = 50 + random(200)
    rotate(random(360))
    fill(random(), 0, 0, random(0.5)) # RGBA values 0.0-1.0
    rect(x - w/2, y - w/2, w, w)
```

NODEBOX has commands for shapes (e.g., `rect`), Bézier paths, images, text, transformation (e.g., `rotate`) and color (e.g., `fill`). The website[11] offers a reference of the command set, tutorials for beginners and advanced users, a collection of plug-in modules, a gallery and a discussion forum.

#### MODULAR

Over the course of two research projects (DESIGN AUTOMATION, GRAVITAL), we have expanded the application with a range of intermixable plug-in modules, for example for image compositing, color theory, layout systems, databases, web mining and natural language processing. A number of modules are inspired by nature and designed to produce generative art. For example, the `supershape` module implements the superformula (Gielis, 2003) and can be used to render many complex shapes found in nature. The `lsystem` module implements Lindenmayer's L-systems. The `noise` module implements Perlin's pseudo-random generator. Two modules provide examples of agent-based systems. The `ants` module can be used to model self-organizing ant colonies. The `boids` module implements Reynold's distributed model for flocking and swarming. The `graph` module combines graph theory with a force-based physics algorithm for network visualization.

Recent effort has produced two spin-off versions. NODEBOX 2 (De Bleser & Gabriëls, 2009) is a cross-platform implementation that focuses on node-based interfaces to alleviate the learning curve of programming code. NODEBOX FOR OPENGL (De Smedt, 2011) is a cross-platform implementation for simple games such as EVOLUTION and VALENCE, discussed in chapter 3.

## NodeBox 2 + 3

NODEBOX 2 (and more recently NODEBOX 3) is based on the Java 2D graphics library and uses a node-based user interface (Lee & Parks, 1995) where each operation is represented as a block or node that "does something". Each node functions as a generator that creates elements (e.g., `rectangle` node) or as a filter that modifies incoming elements (e.g., `rotate` node). A node has no fixed purpose. Or rather, it has many purposes that depend on the other nodes attached to it. In the user interface, nodes can be connected to form a graph. Creativity is encouraged by allowing users to combine and adapt nodes in various ways. The application includes the Bézier interpolation algorithms (De Smedt, F., personal communication, 2007) from the original version.

## NodeBox for OpenGL

NODEBOX FOR OPENGL (NOGL) is based on Pyglet (Holkner, 2008) and uses an updated version of the command set. For example, its `random()` function has an extra `bias` parameter that defines preference towards higher or lower numbers. Functionality from the plug-in modules has been bundled into the core of NOGL. This includes motion tweening, animation layers, hardware-accelerated pixel effects (using dynamic GLSL shaders), procedural images, geometry, Bézier path interpolation, tessellation, particle systems, graphs, flocking, supershapes, noise, GUI controls and audio. NOGL uses a direct drawing approach for performance.

[11] http://nodebox.net/code

## 2.4 NodeBox case studies

Following is a more detailed study of six generative artworks produced in NODEBOX, addressing topics such as the drawing syntax, recursion, and authorship in generative art.

### Nebula: random strands of curves

Figure 11 shows a number of variations of NEBULA (De Smedt, 2010), a NODEBOX script that generates strands of curves with subtle changes in color, shape and rotation. The source code is given below.

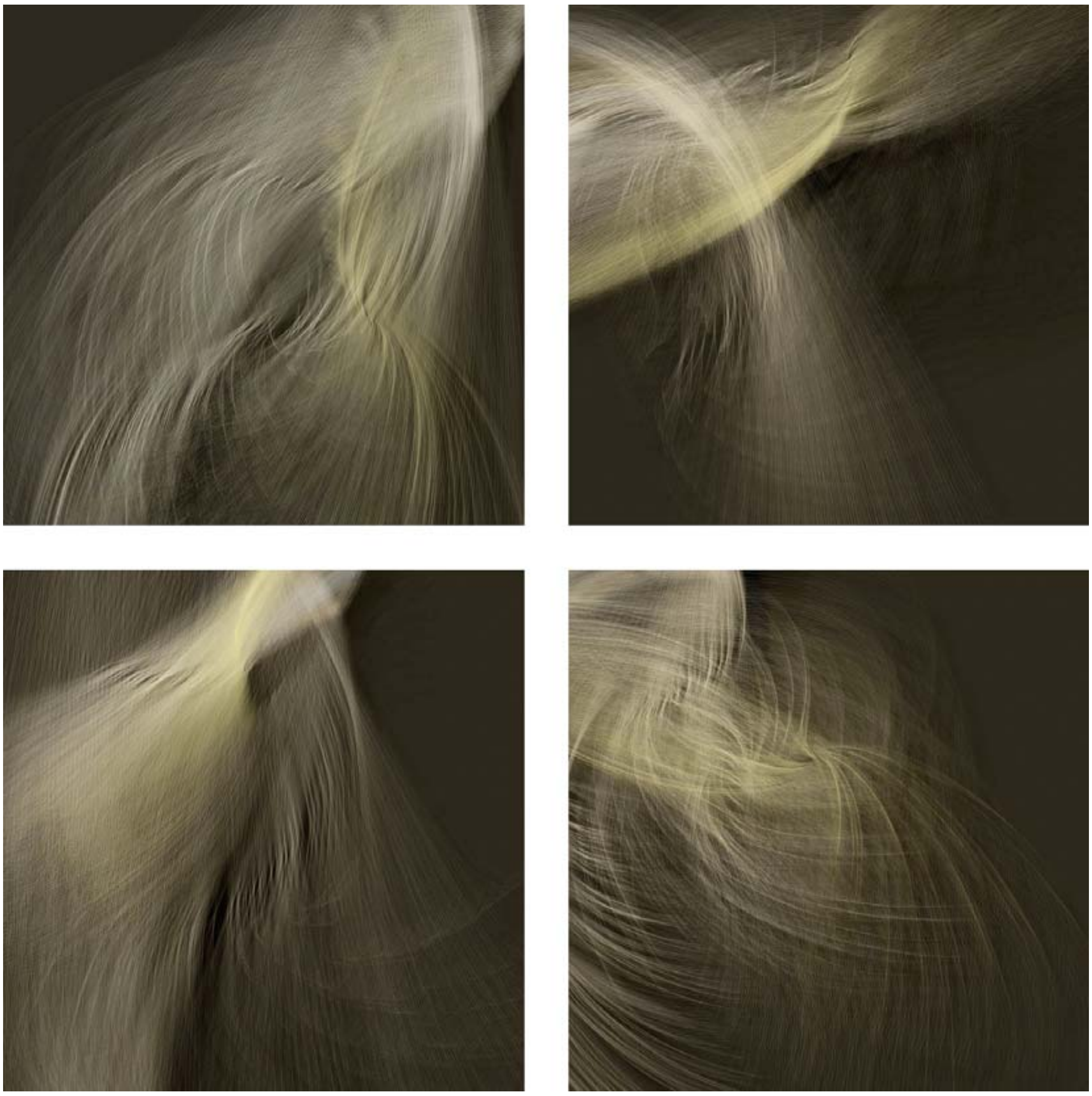

Figure 11. Four variations of NEBULA.

The first line of code imports the `nodebox.geo` module. It contains useful helper functions for 2D geometry, such as the `distance()` given two points or `coordinates()` given a point, distance and angle. It also has a `reflect()` function, which we will use to create smooth curves.

```
from nodebox import geo
```

See: http://nodebox.net/code/index.php/Math

The next few lines import the `colors` module. This is a plug-in module that can be downloaded from the website. It contains functionality for gradients, shadows and color harmony. In NEBULA, it is used to render a shadow for each curve, adding to the natural feel of the composition.

Note how the `ximport()` statement is wrapped in a `try...except` block so that the program doesn't crash for users that don't have the plug-in. The `ximport()` function is specific to NodeBox. It is used instead of Python's `import` to make the module aware of the current canvas.

```
try:
    colors = ximport('colors')
    colors.shadow(alpha=0.02, dx=30, dy=30)
except:
    pass
```

See: http://nodebox.net/code/index.php/Colors

The "nebula" is based on a bundle of curves that is copied over and over with slight variations. The curves that make up a bundle start at random positions and then merge into a single focus point. To generate a list of points randomly scattered across the canvas (or clamped to a relative extent of the canvas), we define the `scatter()` function. The function's `vx` and `vy` parameters set the direction (vector) of each point. All points have the same direction. So when we draw a curve from each point, all curves harmoniously bend in the same direction.

```
def scatter(n, vx, vy, extent=(0.0, 0.0, 1.0, 1.0)):
    points = []
    for i in range(n):
        x0, y0, x1, y1 = extent
        pt = PathElement()
        pt.x = WIDTH * x0 + random(WIDTH * (x1 - x0))
        pt.y = HEIGHT * y0 + random(HEIGHT * (y1 - y0))
        pt.ctrl1.x = pt.x + vx
        pt.ctrl1.y = pt.y + vy
        points.append(pt)
    return points
```

To generate the curves in a bundle we define the `bundle()` function. It returns a `BezierPath`, a group of straight lines and/or curves. It is more efficient to group multiple curves into one `BezierPath` than to draw each of them separately. The restriction is that all curves in a path will be drawn in the same color. Therefore we don't group different bundles in a single path; we want to vary their color.

```
def bundle(points, x, y, vx, vy, reflect=False, d=0):
    p = BezierPath()
    for pt in points:
        if reflect is False:
            vx0, vy0 = pt.ctrl1.x, pt.ctrl1.y
        else:
            vx0, vy0 = geo.reflect(pt.x, pt.y, pt.ctrl1.x, pt.ctrl1.y)
        p.moveto(pt.x, pt.y)
        p.curveto(vx0, vy0, vx, vy, x + random(-d, d), y + random(-d, d))
    return p
```

See: http://nodebox.net/code/index.php/Reference_|_BezierPath

NodeBox has an origin point state from which all `(x, y)` coordinates on the canvas originate. By default, this is the top left corner `(0, 0)`, but we can move the origin point around with the `translate()` function. Each time we draw a bundle, we want to move the origin a little bit, so that the next bundle is slightly displaced. This produces the thick strands.

```
def draw(path, x=0, y=0, angle=0, zoom=1.0, **kwargs):
    translate(x, y)
    rotate(angle)
    scale(zoom)
    drawpath(path.copy(), **kwargs)
```

Notice the `**kwargs` parameter. This is the Python idiom for optional parameters that can be passed to the function aside from those defined. This is useful since we can pass on unforeseen parameters to `drawpath()` inside the function, such as `fill` and `stroke` to colorize the path.

That is all we need to draw lots of bundles of curves. However, a simple `for`-loop with random bundles in random colors does not produce a very interesting composition. To make it more aesthetically pleasing requires further thought, fine-tuning and tweaking – the drawing code is a little more complicated. After some experimentation we ended up with three layers of bundles: a ground layer with darker variations of a single bundle, a middle layer with random, desaturated bundles, and a top layer with lighter variations of the last bundle we drew.

First we set the size of the canvas, along with the color and stroke width of the curves:

```
size(700, 700)
colormode(HSB); clr=color(0.1, 0.3, 1)
background(clr.h, clr.s, max(0.15, 0.15 * clr.brightness))
strokewidth(0.1)
nofill()
autoclosepath(False)
```

We create a list of points scattered across the canvas:

```
points = scatter(
         n = 75,
        vx = random(-100, 100),
        vy = random(-100, 100),
    extent = (random(), random(), random(), random())
)
```

We create a bundle of curves from the list of points:

```
x = random(WIDTH)
y = random(HEIGHT)
vx = x + random(-50, 200)
vy = y + random(-50, 200)
path = bundle(points, x, y, vx, vy, d=10)
```

We can then draw the ground layer. The `shift` variable determines the direction on the rainbow color wheel (e.g., shift red to orange or red to purple, but not both).

```
shift = choice((-1, 1))
for i in range(50):
    h = (clr.h + shift * random(0.1)) % 1
    s =  clr.s
    b =  clr.brightness * (0.4 + random(0.6)) # darker
    draw(path, x = 1.5,
               y = 0,
           angle = 0.3,
            zoom = 1.0,
          stroke = color(h,s,b, random(0.25)))
```

The middle layer, with variations of 2–3 random bundles:

```
for i in range(choice((2, 3))):
    reset() # reset origin
    points = scatter(
             n = 75,
            vx = random(-100, 100),
            vy = random(-100, 100),
        extent = (random(), random(), random(), random()))
    vx, vy = geo.reflect(x, y, vx, vy)
    path = bundle(points, x, y, vx, vy, reflect=True, d=10)

    for i in range(50):
        h = clr.h
        s = clr.s + random(-0.6) # desaturated
        b = random(0.2) + 0.8
        draw(path, x = random(),
                   y = random(),
               angle = 0.3,
                zoom = 1.01,
              stroke = color(h,s,b, random(0.25)))
```

Finally, the top layer, based on the last bundle drawn:

```
shift = choice((-1, 1))
for i in range(100):
    h = (clr.h + shift * random(0.1)) % 1
    s =  clr.s + random(-0.2)
    b = random(0.2) + 0.8
    draw(path, x = 1.5,
               y = 0,
           angle = 0.3,
            zoom = 0.99,
          stroke = color(h,s,b, random(0.25)))
```

Even though the result as a whole looks "profound", there is nothing remotely mystical about it. It is just a bunch of curves and lots of `random()` calls.

## Superfolia: a path filter

Figure 12 shows a number of variations of SUPERFOLIA (De Smedt & Lechat, 2010), a NODEBOX script that generates plant structures using a multitude of small, random curves drawn in varying shades of green.

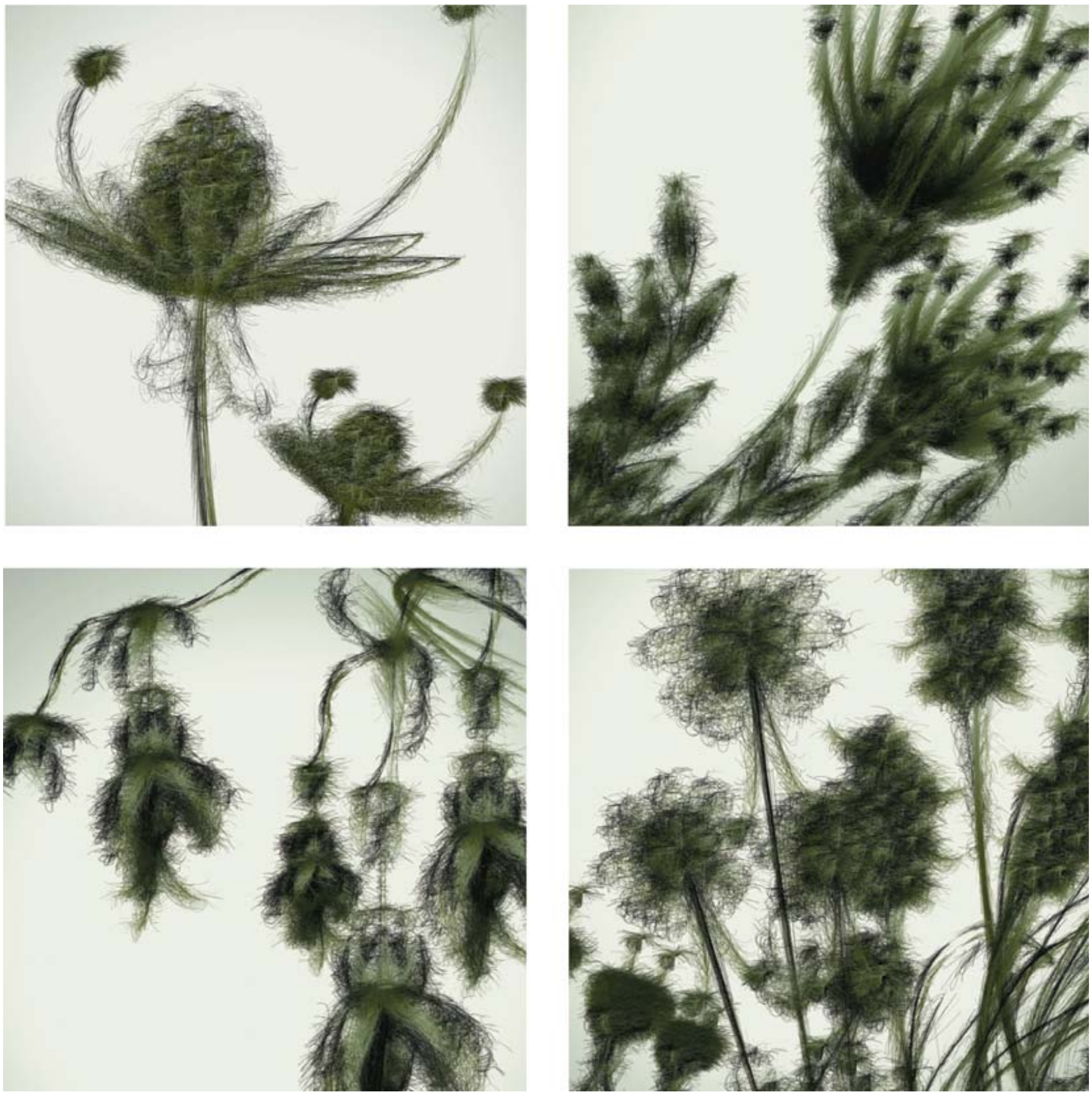

Figure 12. Four variations of SUPERFOLIA.

The skeleton of each plant was made in Adobe Illustrator and stored as an SVG file (Scalable Vector Graphics, an XML-based open standard for storing vector graphics). It is trivial to import it as a list of `BezierPaths` using the `svg` plug-in module:

```
svg = ximport('svg')
paths = svg.parse(open('plant1.svg').read())
```

Each `BezierPath` in the returned list of paths represents a feature in the plant skeleton (e.g., stem, leaves). We want to draw many random “hairs” in different shades along each path: darker hairs in the back and more colorful hairs on top. This adds a sense of depth to the composition. The hardest part in the source code is a linear interpolation function that calculates a gradient, given a list of colors. We have a general idea about the gradient (i.e., black → green) but we don’t know the amount of hairs per path beforehand – longer paths will have more hairs. The `gradient()` function transforms this general idea into a concrete list of colors: one `Color` for each hair given the length of a path.

```
def gradient(colors, steps=100):
    if len(colors) == 0:
        return []
    g = []
    for n in range(steps):
        x = float(n) / (steps-1 or 1) * (len(colors)-1)
        i = int(x)
        j = min(i + 1, (len(colors)-1))
        t = x - i
        g.append(Color(
            colors[i].r * (1-t) + colors[j].r * t, # R
            colors[i].g * (1-t) + colors[j].g * t, # G
            colors[i].b * (1-t) + colors[j].b * t, # B
            colors[i].a * (1-t) + colors[j].a * t  # A
        ))
    return g
```

NODEBOX is bundled with a fast path interpolation algorithm that can for example be used to calculate the length of a path, to split a path, or to find the point at `t` along a path, where `t=0.0` is the starting point of the path and `t=1.0` is the end point. We use it to draw the hairs on each path in the plant skeleton. This is called a path filter. In analogy to pixel filters for images, a path filter for vector art has a kernel function (in this case, “draw a hair”) that is executed for each point along the path. Note the `path.points()` in the code below, which yields a list of points interpolated along the path.

```
BLACK = color(0.00, 0.00, 0.00)
GREEN = color(0.16, 0.16, 0.05)
OLIVE = color(0.29, 0.27, 0.08)
BEIGE = color(0.57, 0.57, 0.44)

nofill()
transform(CORNER)
autoclosepath(False)

for path in paths:
    n = path.length > 900 and 100 or 50
    g = gradient([BLACK, GREEN, OLIVE, BEIGE], steps=n)
    x = None
    y = None
    for i, pt in enumerate(path.points(n)):
        clr = g[i]
        clr.alpha = 0.7 + random(0.3)
        if i > 0:
            p = BezierPath()
            p.moveto(x, y)
```

```
            for d1, d2 in ((20, 0), (10, 10)): # kernel
                p.curveto(
                    pt.ctrl1.x - random(d1),
                    pt.ctrl1.y,
                    pt.ctrl2.x,
                    pt.ctrl2.y + random(d1),
                    pt.x + random(-d2, d2),
                    pt.y + random(-d2, d2)
                )
            drawpath(p, stroke=clr, strokewidth=random(0.1, 0.3))
        x = pt.x
        y = pt.y
```

See: http://nodebox.net/code/index.php/Path_Mathematics

## Creature: recursive branching

Figure 13 shows CREATURE (De Smedt, Lechat & Cols, 2010), a NODEBOX script that generates a computational abstraction of animal morphology, developed in collaboration with the Department of Morphology at the University of Ghent.

Many animals are similar at the component level (eyes, joints, skeleton) but different as a whole. Some traits such as eyesight evolved from a common ancestor millions of years ago, but the species then diversified. The number of eyes (for example) differs across species. The majority of animals has eight eyes, or compound eyes (Nyffeler & Sunderland, 2003). Animal morphology is the result of careful selection, variation and retention. Taking a random selection of components (e.g., skeleton parts) and lumping them together to construct an animal is like expecting a whirlwind in a junkyard to construct an airplane. This is what CREATURE does. The result is somewhat alien, perhaps because the internal logic or purpose of the components is absent, that is, it is not the result of selection and adaptation (see chapter 1). Its nickname is ROADKILL.

The main part of the source code involves rendering an artistic interpretation of the cardiovascular system, using a recursive algorithm. The `root()` function draws a chain of lines. At each segment the direction changes randomly and there is a chance that `root()` is called, creating a new branch.

```
from math import sin, cos, radians

def root(x, y, angle=0, depth=5, alpha=1.0, decay=0.005):
    for i in range(depth * random(10, 20)):
        alpha -= i * decay
        if alpha > 0:
            angle += random(-60, 60)
            dx = x + cos(radians(angle)) * depth * 5
            dy = y + sin(radians(angle)) * depth * 5
            stroke(0, alpha)
            strokewidth(depth * 0.5)
            line(x, y, dx, dy)
            if random() > 0.75 and depth > 0:
                root(x, y, angle, depth-1, alpha) # recursion
            x = dx
            y = dy

root(50, 50)
```

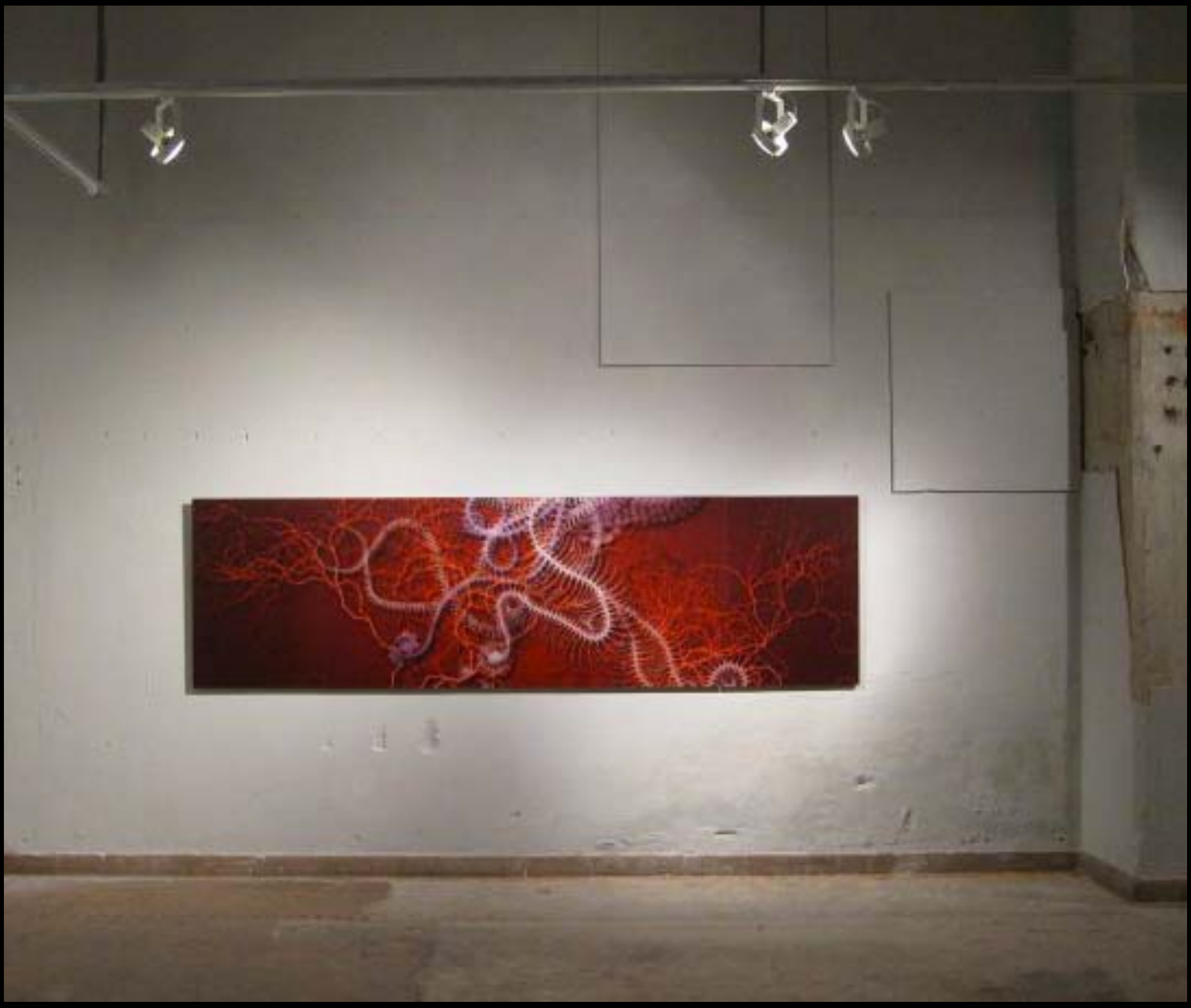

Figure 13.1. CREATURE at Creativity World Biennale, Oklahoma USA, 2010.

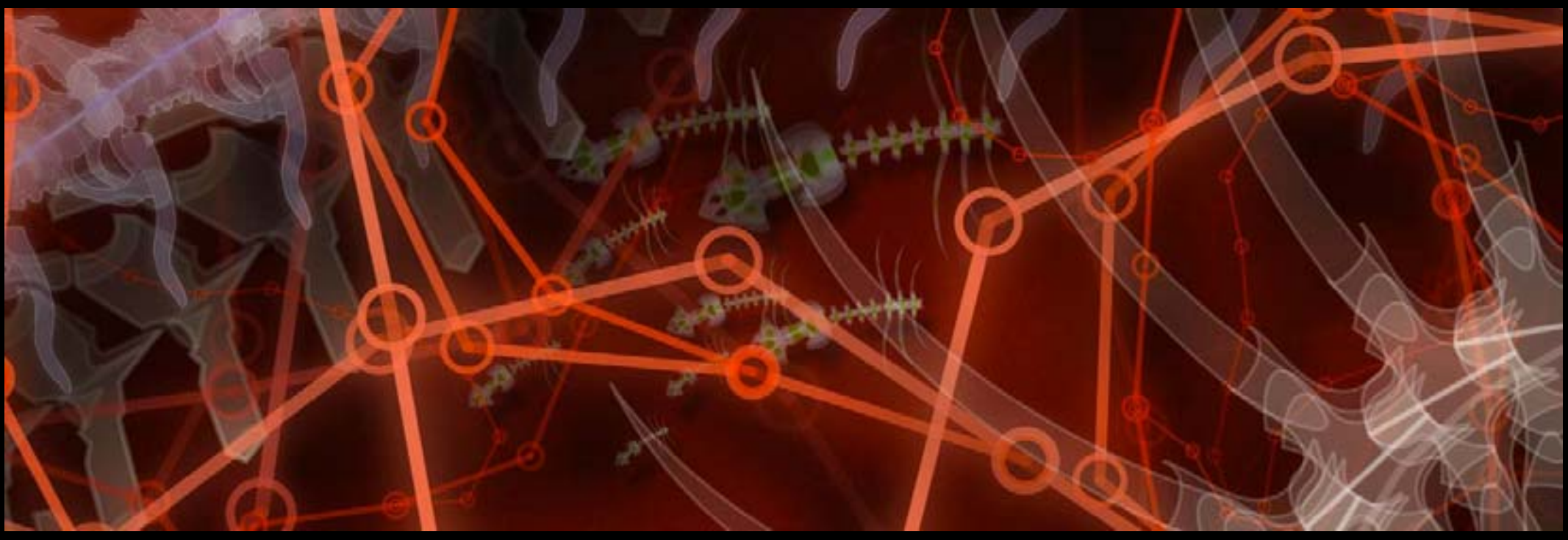

Figure 13.2. CREATURE detail.

## Nanophysical

Figures 14.1–14.11 show NANOPHYSICAL (Lechat & De Smedt, 2010), a visual metaphor that blends ideas from nanotechnology, cell biology and the brain. It is developed in collaboration with and permanently exhibited at Imec (Interuniversity Microelectronics Centre, Leuven). Some parts are handcrafted while others are generated with NODEBOX.

For example, figure 14.6 uses a particle system to visualize a dense group of cells that pack together. It is bundled as an example in NODEBOX FOR OPENGL and in PATTERN's online `canvas.js` editor[12]. Figure 14.7 uses a particle system to visualize a mesh of cells, where the connections between cells are springs.

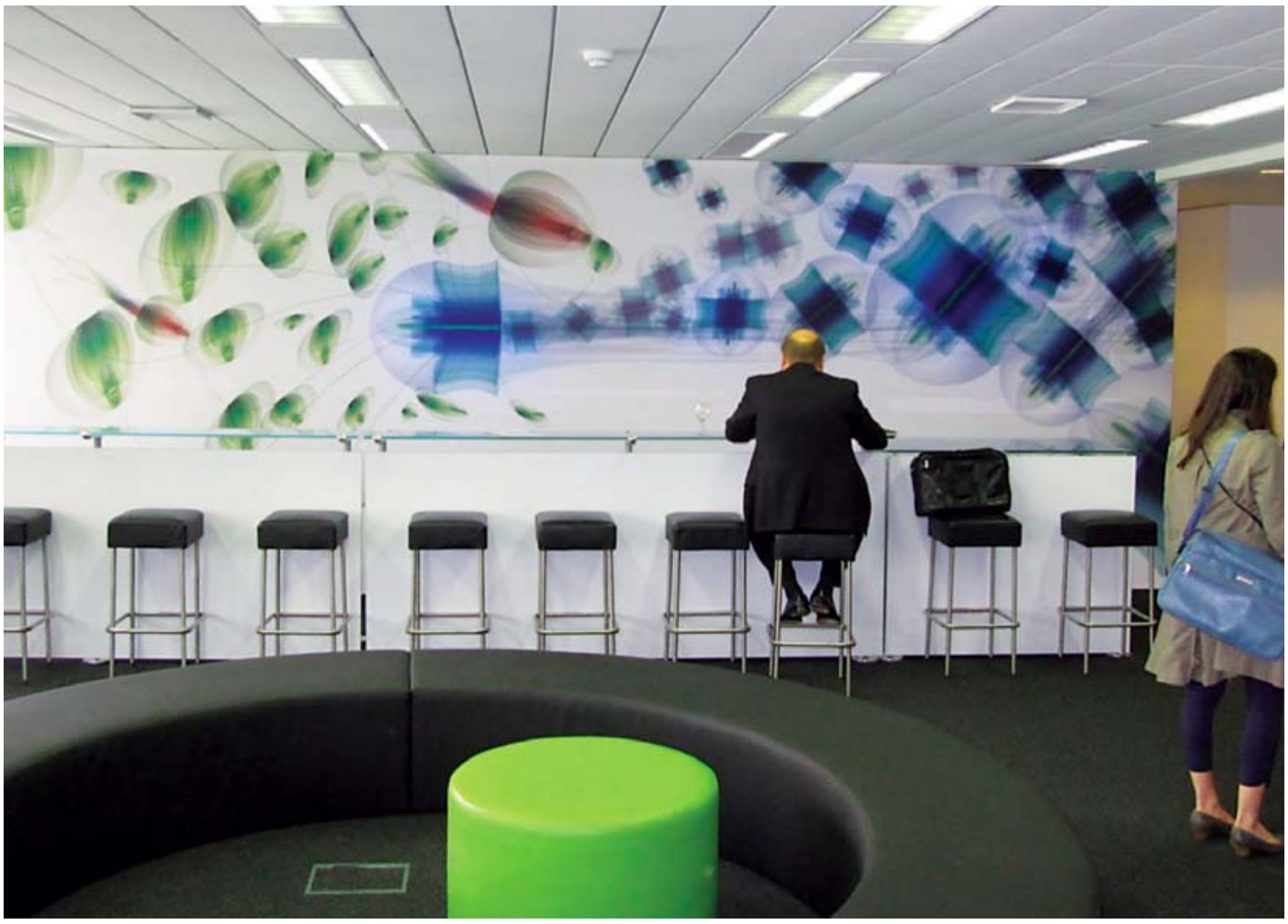

Figure 14.1. NANOPHYSICAL at Imec Leuven, permanent exhibition.

NANOPHYSICAL is produced on durable canvas. This is in contrast to the previously discussed works, which are produced as digital prints mounted on aluminum panels. This illustrates the often volatile nature of generative art: as the software used to generate the artwork ages, modern computer systems will be unable to reproduce it, while the original digital prints slowly fade due to sunlight exposure.

[12] http://www.clips.ua.ac.be/media/canvas/?example=pack

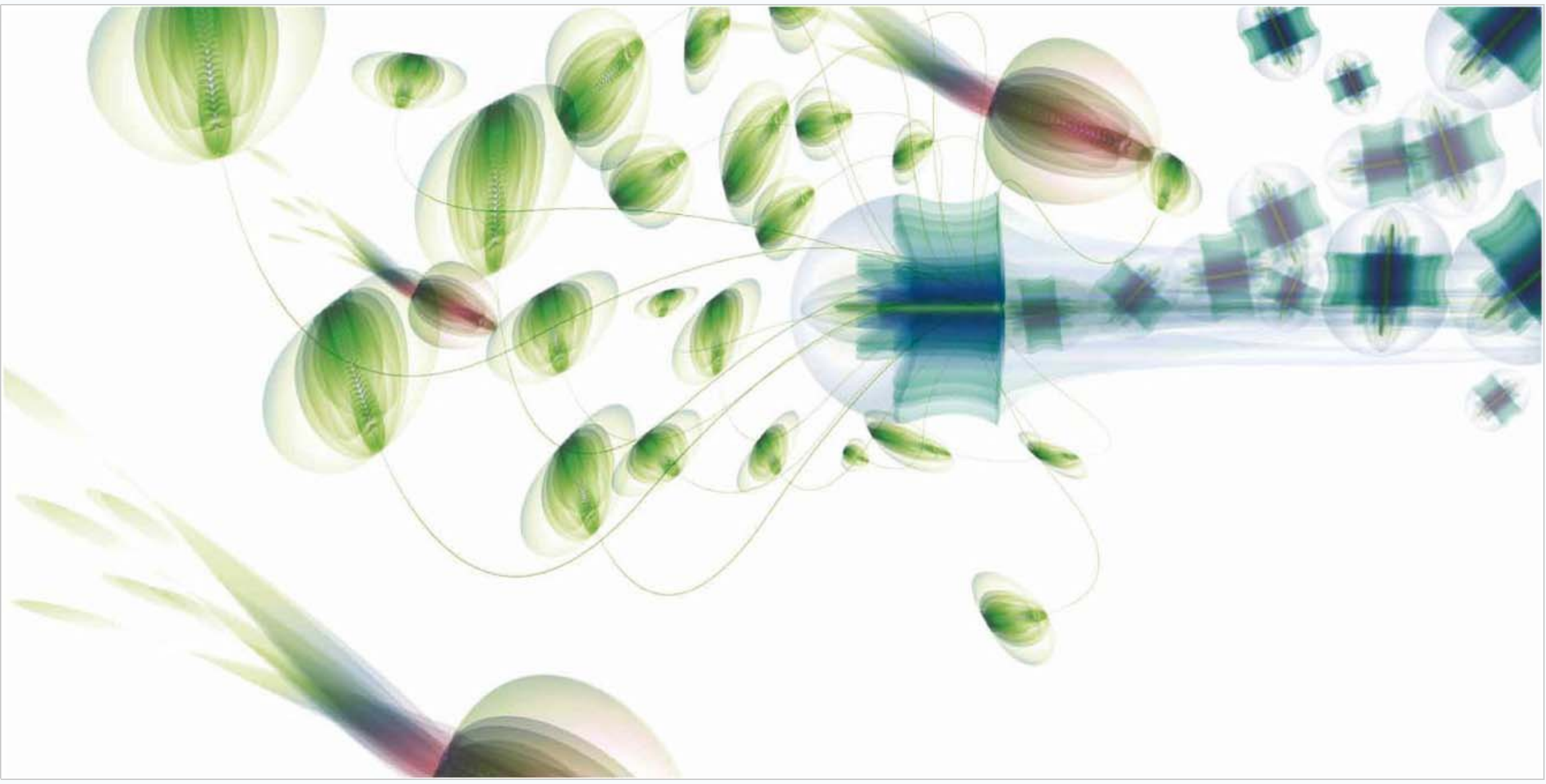

Figure 14.2. SOLAR PLANT. A solar cell converts sunlight energy into electricity. The blue elements represent solar cells, blended with the concept of parasitic symbiosis (green elements).

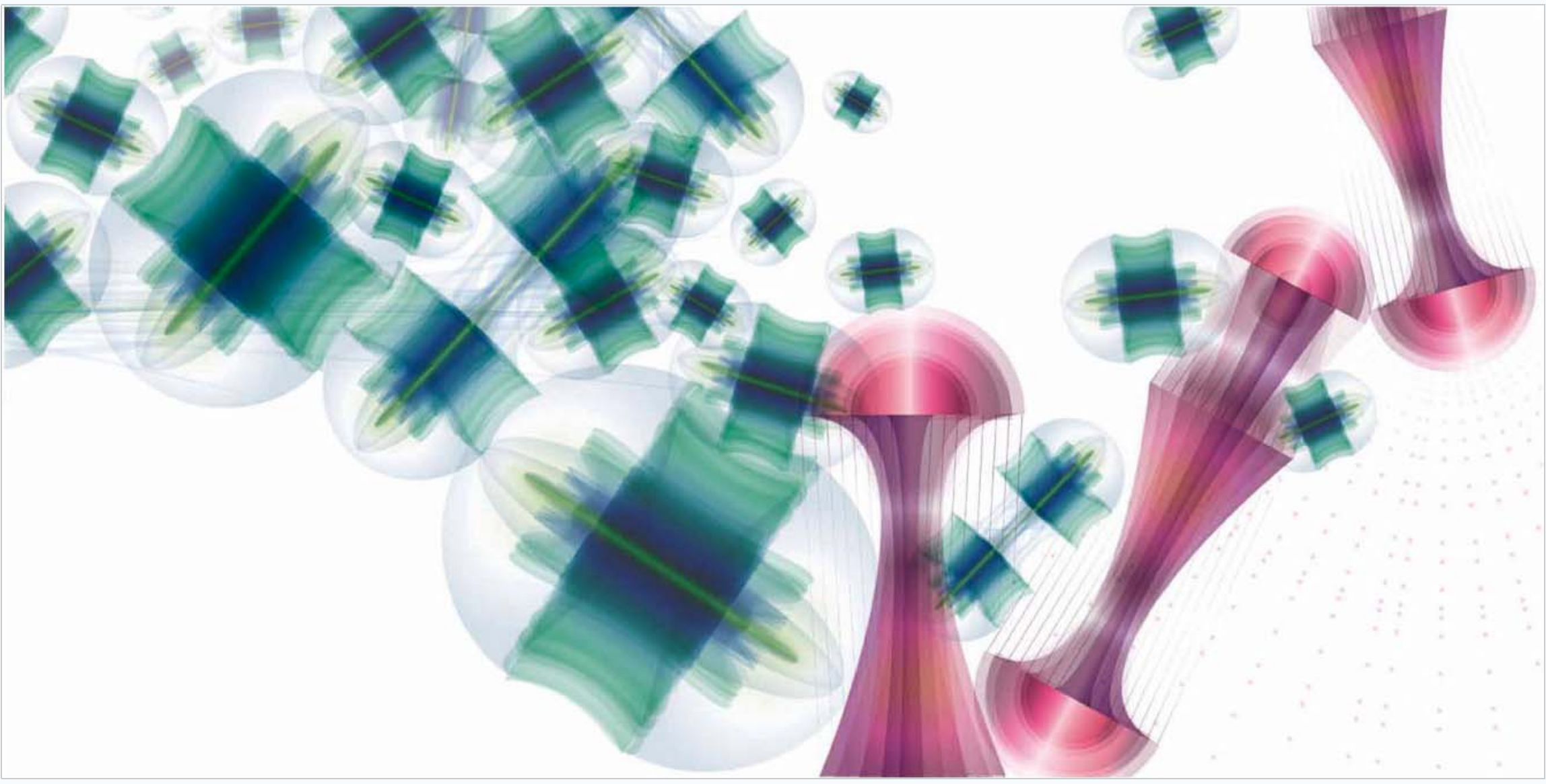

Figure 14.3. RESONANCE. The work spans a blank wall (for multimedia projections) using a wireless communication metaphor. Signals from 14.2 are depicted to be packed as wireless data and transmitted to 14.4.

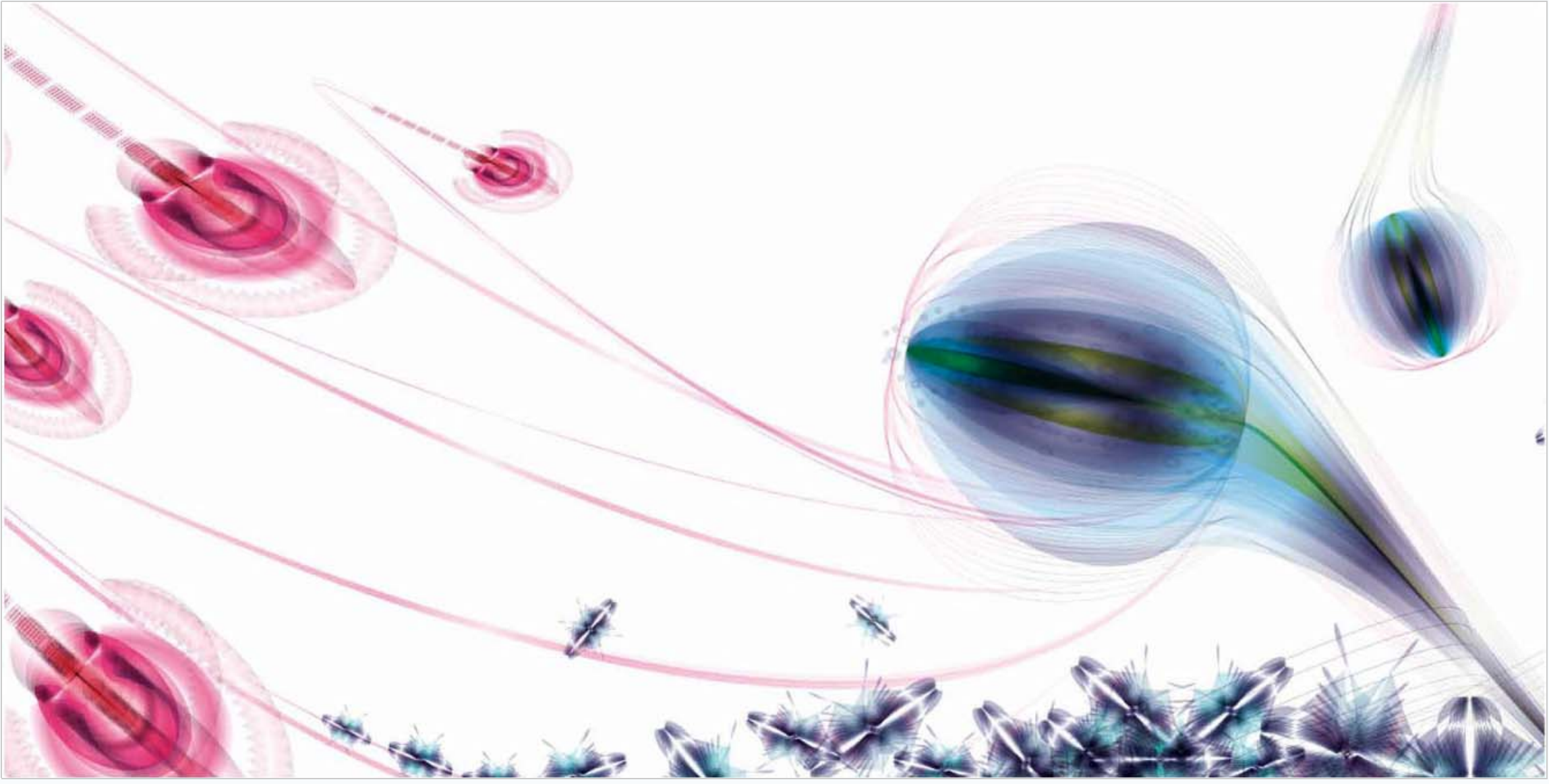

Figure 14.4. NEURON. In nerve cells, the bulbous end of a neuron receives chemical stimulation from the neuron's branched projections. This part represents a neuron passing on information to the stream in 14.5.

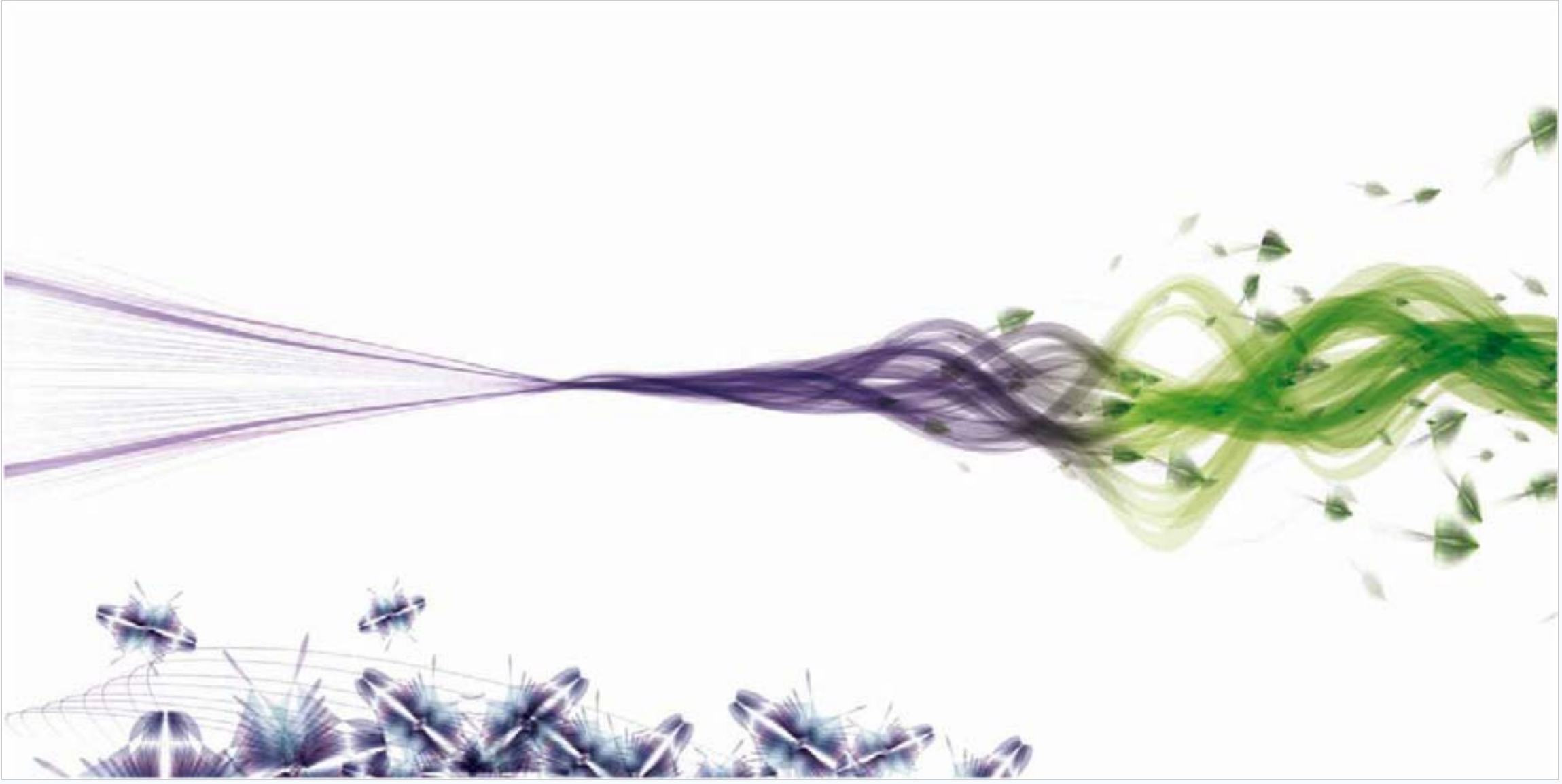

Figure 14.5. KINETIC STRING. A stream carries particles off to form a cell in 14.6, inspired by kinetic energy. The increase in pressure in the stream is visualized by a change in color.

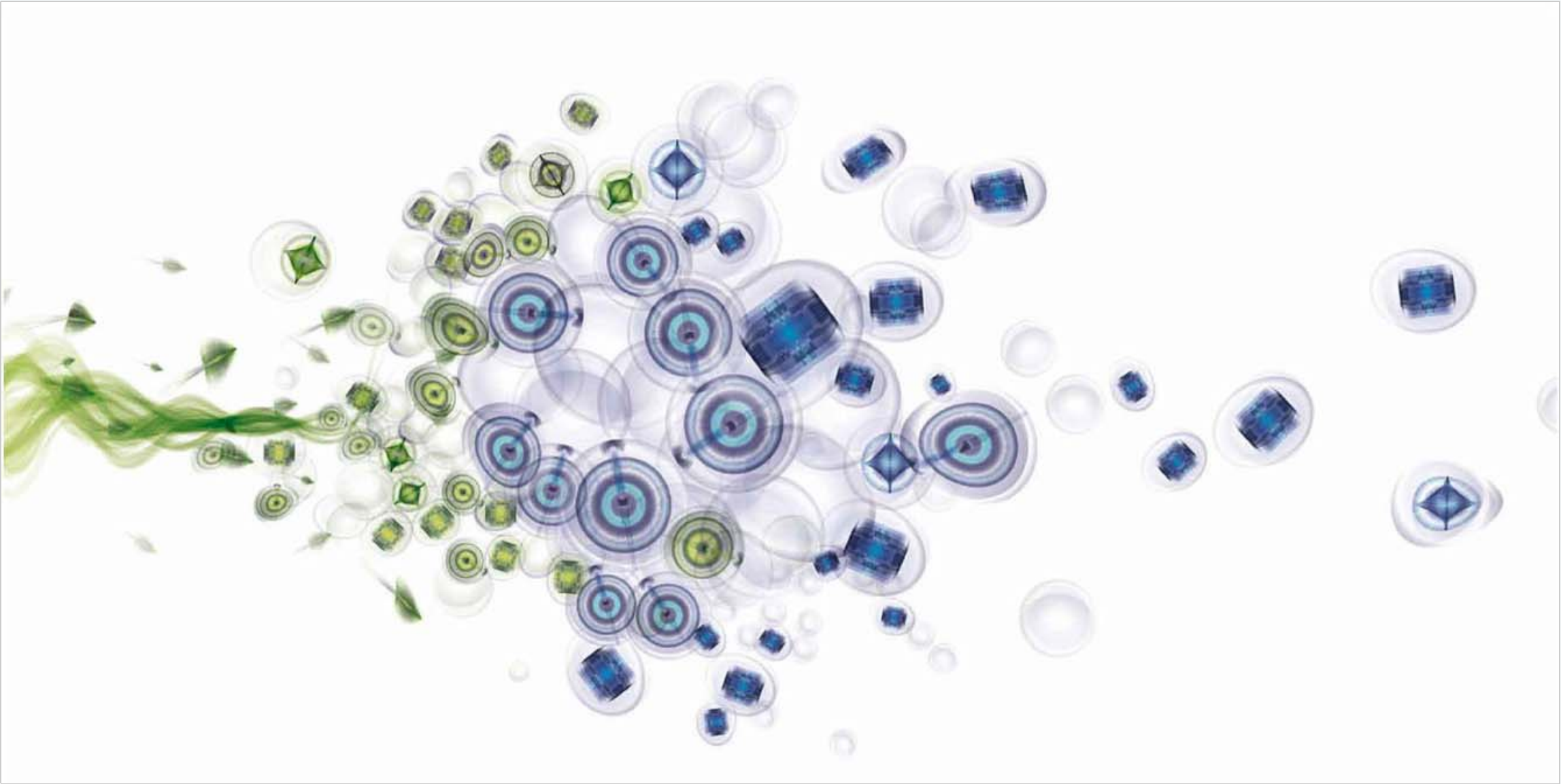

Figure 14.6. EXOCYTOSIS. Elements from the stream in 14.5 are organized into a cell. As the elements burst outwards, their capsule dissipates, leaving the content inside to form the tissue in 14.7.

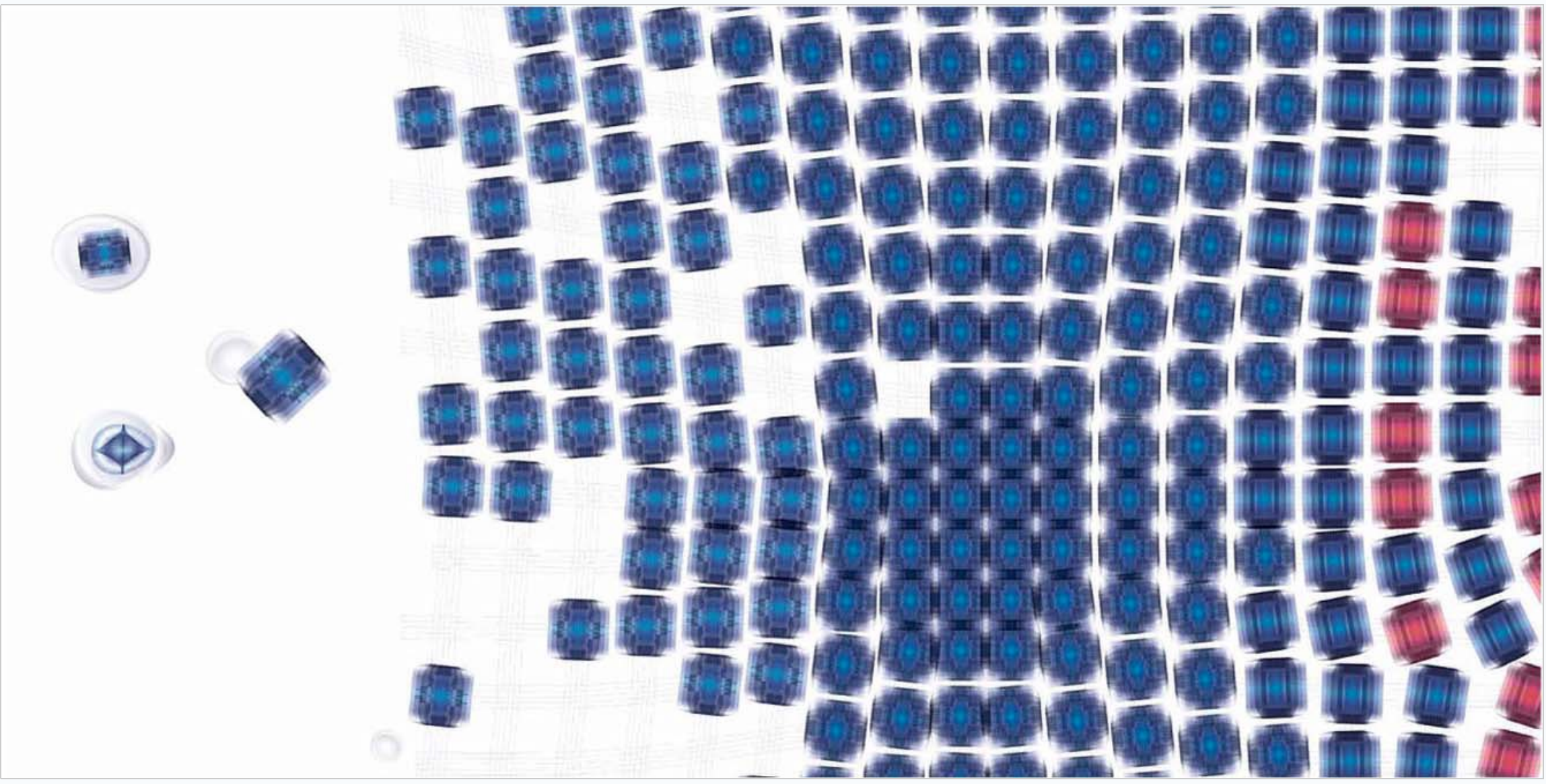

Figure 14.7. SELF-ASSEMBLY. The tissue is modeled by using a force-based algorithm in which the connections between elements are springs that pull at neighboring elements. The tissue functions as food for the cell in 14.8.

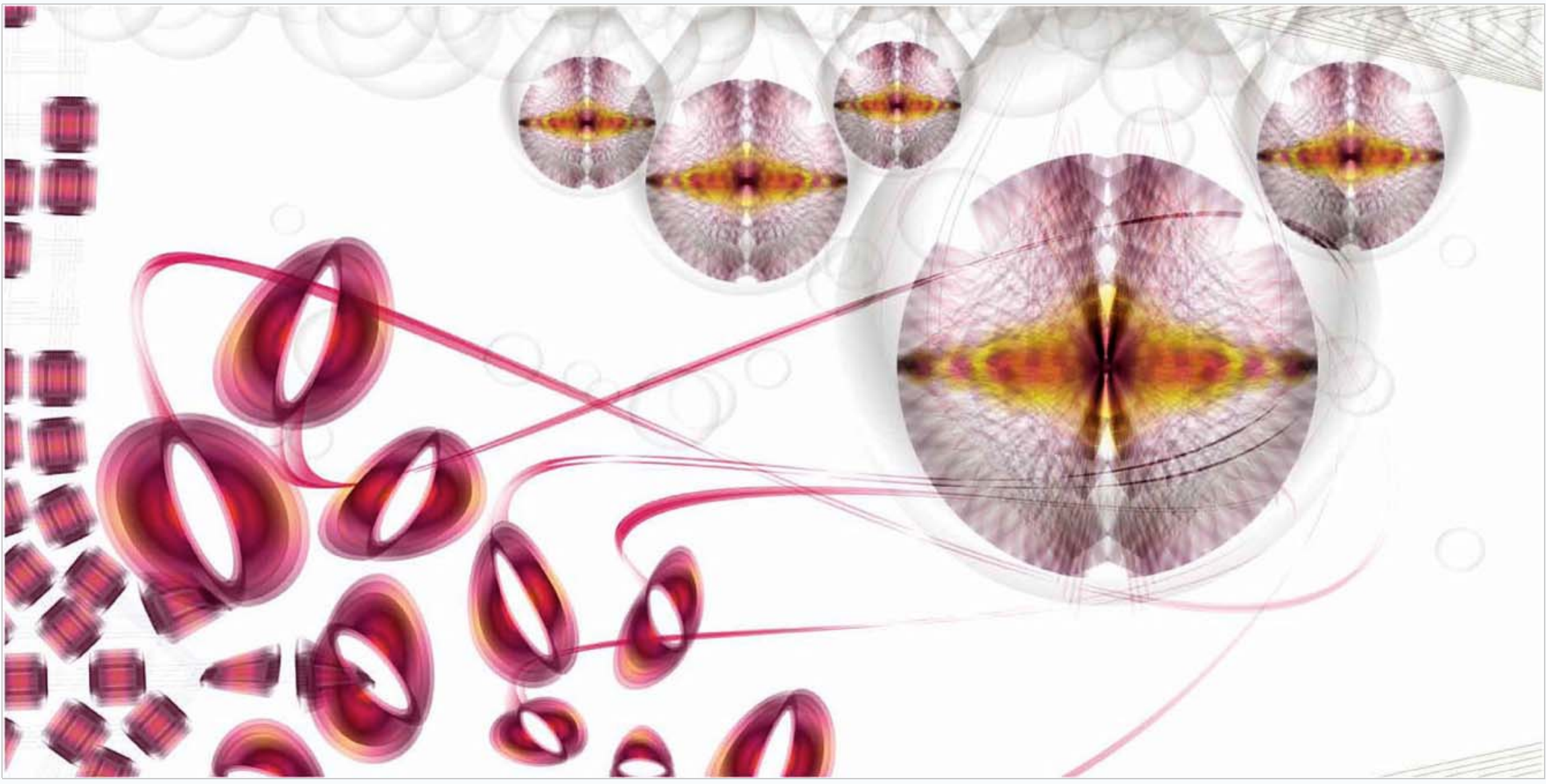

Figure 14.8. NUCLEID. The cocoon-like elements growing from the top draw are based on receptor cells, and the brain. The structure's fictional name is reminiscent of "nucleus", but with a slight astronomical touch.

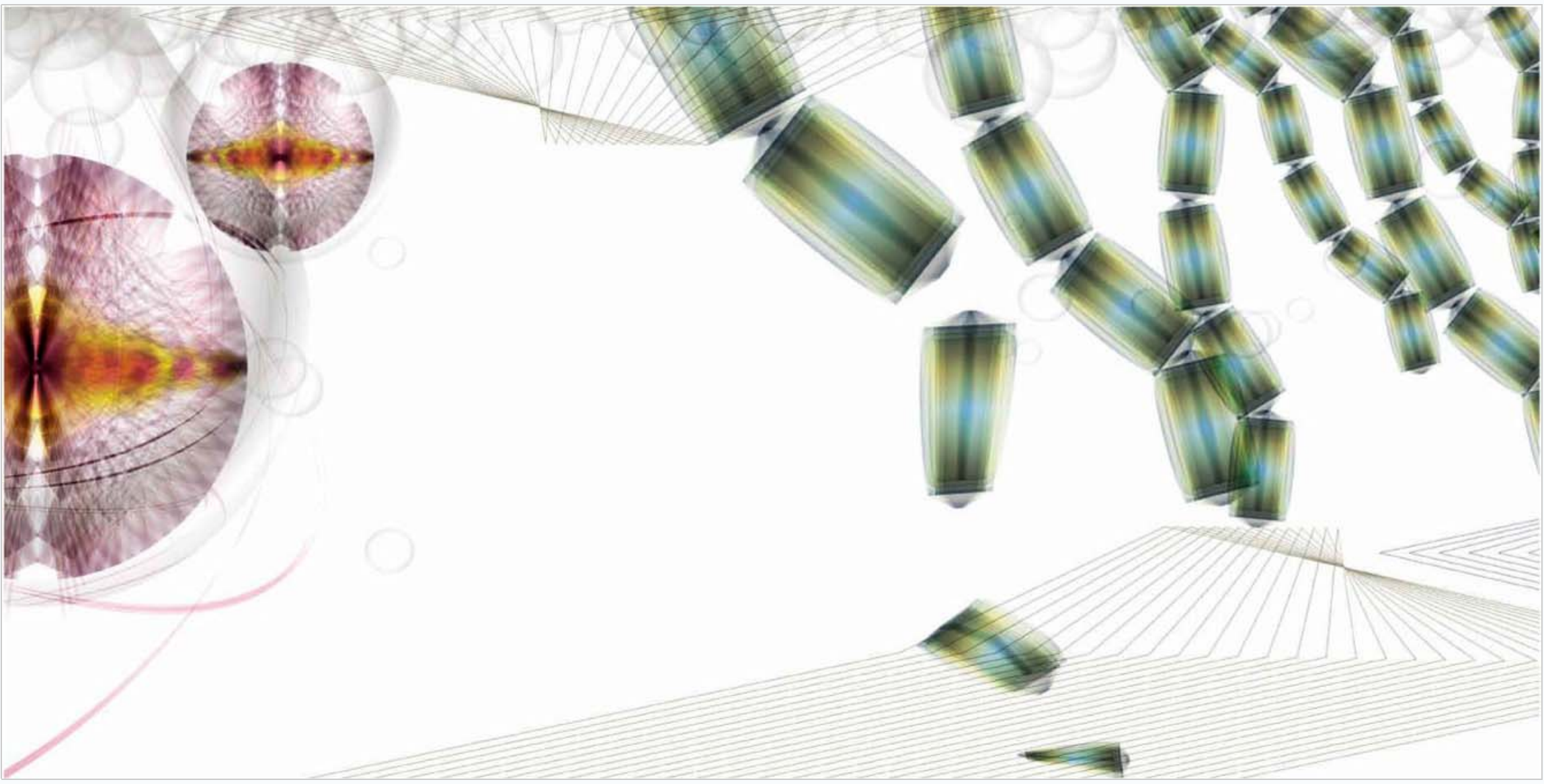

Figure 14.9. DNA. DNA contains the genetic instructions for building and maintaining an organism. The green strands represent DNA molecules passing information extracted from 14.10 to 14.8.

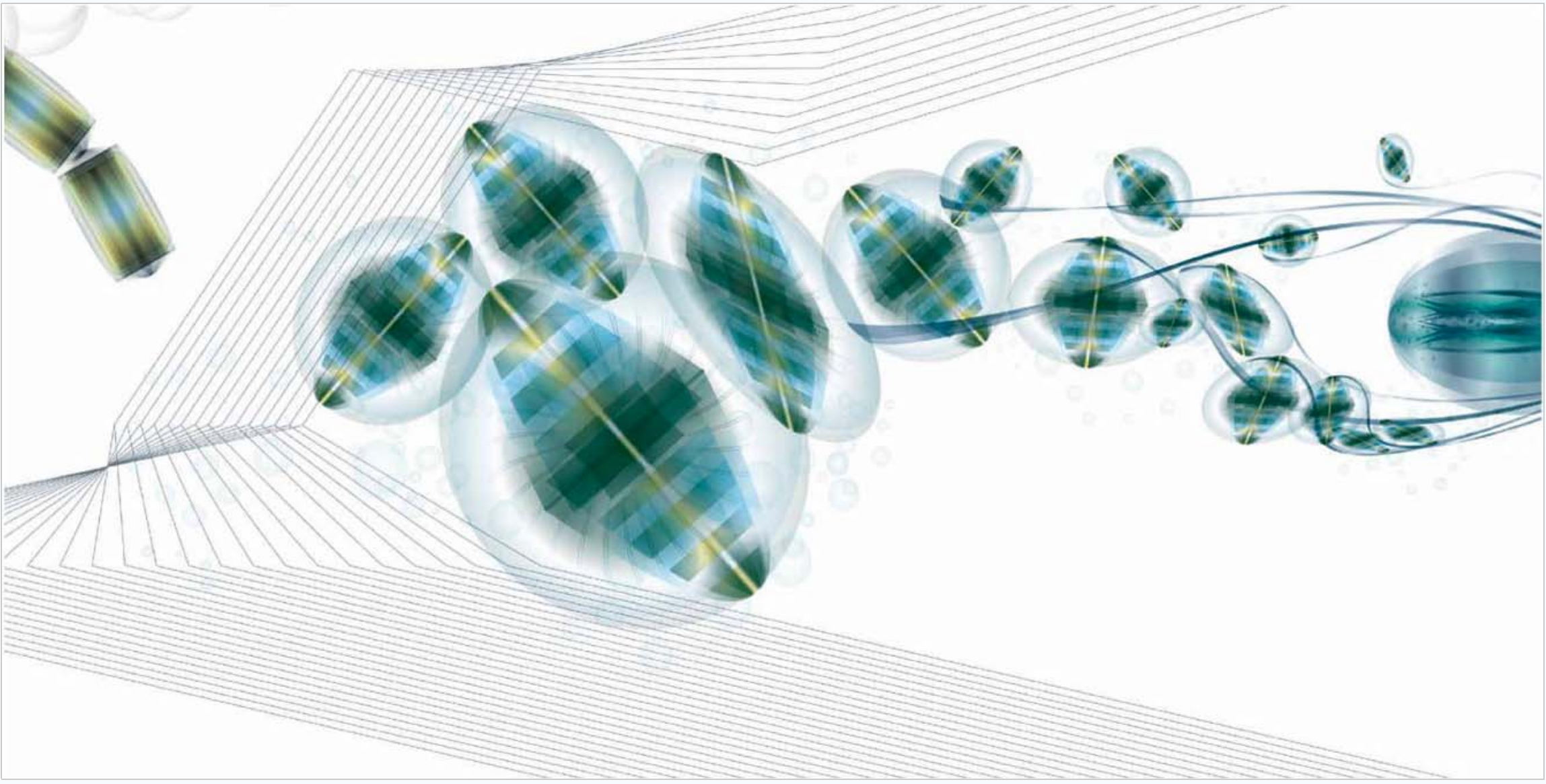

Figure 14.10. TRANSMITTER. A biochip is a miniaturized laboratory that performs biochemical reactions. This is represented with a chip-like wireframe that catches the signal emitted from the synapses in 14.11.

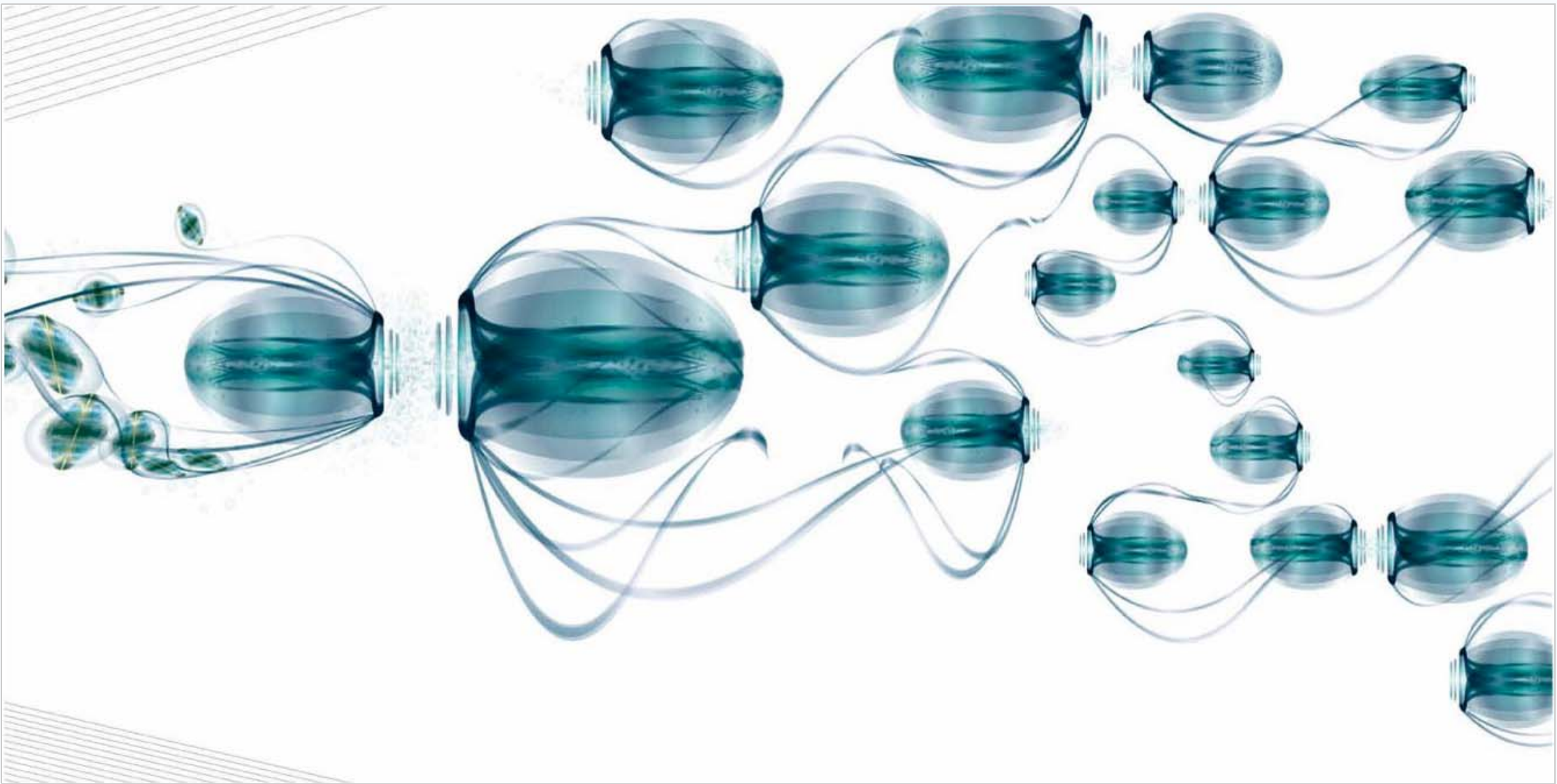

Figure 14.11. SYNAPSES. In the nervous system, synapses permit a neuron to pass signals to other cells. By analogy, the elements are designed to interconnect in a network and reinforce each other's signal.

A question that is sometimes raised is: “who is the author of a generative artwork?” Is it the person that creates the system, or the system itself? Clearly, in examples such as NEBULA, SUPERFOLIA, CREATURE and NANOPHYSICAL it is the person (or persons). The program acts as a generator, optimizing the production process of what would be tedious to achieve by hand. It follows orders, give or take a predefined amount of randomness. There is a creative advantage here in the sense that users are free to combine any kind of functionality in source code, instead of relying on predefined buttons and sliders. However, our concern is modeling creativity, so we are more interested in programs that are the author of their *own* creations. Instead of generating a product that is recognized as creative (i.e., art) such programs must involve an autonomous process that *is* creative (McCormack, 2005). We offer two additional case studies to see how this can be done: PRISM (De Smedt & De Bleser, 2005) and PERCOLATOR (De Smedt, 2005).

## Prism: color ranges + majority vote

PRISM matches colors to a given word. In the HSB color model, a color is represented using numerical values for hue, saturation and brightness, where hue corresponds to the actual spectral color (e.g., red, orange, yellow) having variations in saturation and brightness. PRISM is based on the idea that specific ranges of saturation and brightness can be expressed as adjectives in natural language. For example, *soft* colors are colorful but slightly subdued: lower in saturation and higher in brightness. The `soft()` function below takes a `Color` and returns a new `Color` in the same hue, but clamped to a softer tone:

```
def soft(clr):
    clr = clr.copy()
    clr.saturation = random(0.2, 0.3)
    clr.brightness = random(0.6, 0.9)
    return clr
```

**AUTHORSHIP ↔ CREATIVITY**

PRISM defines 11 ranges: light, dark, weak, bright, warm, cool, soft, hard, fresh, neutral and intense. Given a word (e.g., “ocean”) the algorithm then performs a series of Google search queries: “light blue ocean?”, “dark blue ocean?”, “soft orange ocean?” and so on. Each query yields the total count of pages known to Google containing the query term. These can be seen as a majority vote (for ocean, the winner is “deep blue”) and translated into HSB values. In short, PRISM independently matches colors to any given term. We must distinguish carefully between authorship and creativity. It may not be very creative to select colors using a majority vote but – however unimaginative – PRISM comes up with its own solution. The fact that the solution is obtained by parroting what was said online is a philosophical argument; we have known artists that did no better.

Following is a simplified implementation with two ranges, using NODEBOX and the PATTERN web mining package. It determines the color of the word "ocean", but any other query word `w` can also be given. For ocean, it yields dark blue colors.

```
def light(clr):
    clr = clr.copy()
    clr.saturation = random(0.3, 0.7)
    clr.brightness = random(0.9, 1.0)
    return clr

def dark(clr):
    clr = clr.copy()
    clr.saturation = random(0.6, 1.0)
    clr.brightness = random(0.2, 0.6)
    return clr

colormode(HSB)
colorwheel = {
       'red': color(0.00, 1.00, 1.00),
    'orange': color(0.10, 1.00, 1.00),
    'yellow': color(0.15, 1.00, 1.00),
     'green': color(0.35, 1.00, 1.00),
      'blue': color(0.65, 1.00, 1.00),
    'purple': color(0.85, 1.00, 1.00)
}

from pattern.web import Google

G = Google()
w = 'ocean'
vote = {}
for clr in colorwheel:
    for adj, rng in (('light', light), ('dark', dark)):
        q = G.search('"%s %s %s"' % (adj, clr, w))
        vote[(rng, clr, w)] = q.total

# Sort by total and select highest.
clr, rng = max((n, clr, rng) for (rng, clr, w), n in vote.items())[1:]
clr = colorwheel[clr]
clr = rng(clr)

rect(0, 0, WIDTH, HEIGHT, fill=clr)
```

## Percolator: online news → image composition

PERCOLATOR assembles image compositions based on recent news. First, it performs a web query to retrieve today's news updates. Using a technique from natural language processing called parsing (see chapter 6) it selects the most frequent nouns from the text. Sometimes it will translate these to related synonyms using the `wordnet` plug-in module (e.g., airplane → bomber). It then searches a database of cut-out images (i.e., with a transparent border) matching the keywords and assembles selected images into a composition, using random pixel filters such as blurring and blending. By analogy, we can say that PERCOLATOR is reading the newspapers for inspiration (search engine query), engages in associative brainstorming to interpret a topic of choice (synonyms), then rummages through piles of old magazines (image database), and finally cuts out pictures and collates them to create a visual.

The following example offers a simplified implementation using a combination of NODEBOX FOR OPENGL and PATTERN functionality. Figure 15 shows two generated visuals for a news update from the New York Times website on September 8, 2012:

> In Kabul, Suicide Bomber Strikes Near NATO Headquarters. A bomber blew himself up Saturday morning near Western embassies and the headquarters of NATO forces, killing at least four civilians, Afghan officials said.

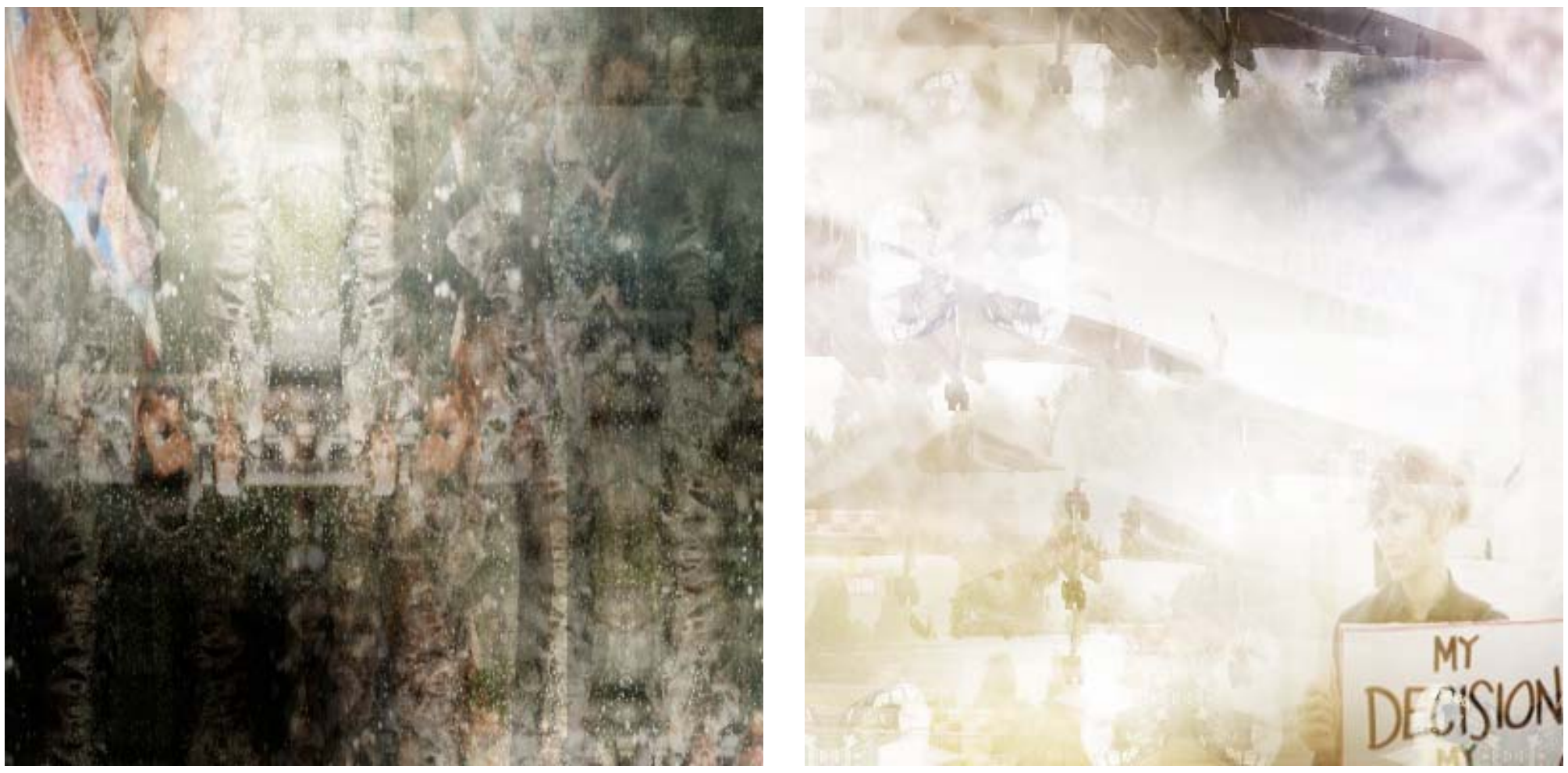

Figure 15. Two variations generated by PERCOLATOR for the NATO suicide bombing.

First, a random news update is mined from the New York Times' RSS feed:

```
from nodebox.graphics import *

from pattern.web import Newsfeed, plaintext
from pattern.web import Bing, IMAGE, URL
from pattern.en import tag
from pattern.en import wordnet

news = 'http://rss.nytimes.com/services/xml/rss/nyt/World.xml'
news = choice(Newsfeed().search(news))
news = plaintext(news.title + '. ' + news.description)
```

Then, singular nouns in the text are mapped to synonyms using `wordnet`, and corresponding images are mined from the Bing search engine:

```
images = []
for w, pos in tag(news):
    if pos == 'NN': # nouns
        try:
            w = wordnet.synsets(w)[0]
            w = choice([w] + w.hyponyms())
            w = choice(w.synonyms)
            img = choice(Bing().search(w, type=IMAGE))
            img = Image(None, data=URL(img.url).download())
            images.append(img)
        except:
            pass
```

The `cutout()` function uses a pixel filter to distort the image and to make its edges transparent:

```
def cutout(img, w, h, filters=(blur, invert, mirror)):
    img = render(lambda: img.draw(0, 0, w, h), w, h)
    img = choice(filters)(img)
    img = mask(img, invert(gradient(w, h, type='radial')))
    img = transparent(img, alpha=random(0.5, 1.0))
    return img
```

Finally, we cut out the images and stack them randomly using pixel blend modes. For example, the `multiply()` blend mode multiplies the pixel values of two layers into a new pixel color:

```
w, h = 400, 400
def draw(canvas):
    if canvas.frame == 1:
        layers = [cutout(img, w, h) for img in images]
        composition = solid(w, h, color(0))
        for i in range(75):
            composition = choice((screen, multiply))(
                img1 = composition,    # base layer
                img2 = choice(layers), # blend layer
                  dx = random(-w/2, w/2),
                  dy = random(-h/2, h/2))
        composition.draw()

canvas.size = w, h
canvas.draw = draw
canvas.run()
```

Is PERCOLATOR the author of the generated visuals? We think so. Is it also creative? Perhaps not, one objection is that the creative process is shallow. PERCOLATOR has a single topic of interest: news, and it is not picky about which news that is. It doesn't experience emotional responses. In fact, it has absolutely no idea what is going on, but news is news, so collate it PERCOLATOR will. Furthermore, it exercises only one artistic style: random combination of pixel filters, which it applies over and over. This is not so different from an undergraduate graphic design student. But students can learn, diversify and specialize whereas PERCOLATOR can not.

Mitigating these objections is not unfeasible. According to Csikszentmihalyi (1999), creativity is identified by peers (i.e., a knowledgeable audience, see chapter 4). Suppose we let PERCOLATOR publish its work on a website, so that visitors can vote what artworks they like best. We could then use a genetic algorithm that favors the steps used to produce visuals with a high rating. Suppose we integrate a semantic network of common sense such as CONCEPTNET (Liu & Singh, 2004) to improve the associative word mapping (airplane → **used-for** traveling, **found-in** sky, ...) We will explore this idea further in chapter 5. Suppose we introduce a reward system and make PERCOLATOR research the topics used in works with a high rating, and attempt to integrate these in new works. Suppose we set up an ensemble of competing systems. We could take it deeper and deeper, level after level, making it harder and harder to question the creative thought process.

## 2.5 Discussion

Twas brillig, and the slithy toves
Did gyre and gimble in the wabe:
All mimsy were the borogoves,
And the mome raths outgrabe.

[...]

– Lewis Carroll, Jabberwocky (1871)

In this chapter we discussed generative art: a rule-based form of art influenced by complexity and self-organization, using techniques from computer graphics, artificial life and artificial intelligence. The rules and constraints in a generative system (usually a computer program) are defined by a human artist. But the resulting artwork is allowed a degree of freedom within the constraints.

NODEBOX is an open source software application for producing generative art, where rules and constraints are written in Python programming code. Many other applications and toolkits exist. Many of them are open source and freely available. More recent toolkits focus at least in part on the Web and online data visualization. We use NODEBOX because Python syntax is flexible and straightforward. Curly braces (`{}`), semicolons (`;`) and type declarations are absent from the syntax so we can focus on the task at hand. NODEBOX has been in use in the community for many years. It is stable and provides plug-ins for all graphics tasks addressed by traditional software. Python has been claimed to be too slow for computer graphics, but we argue that this is alleviated with the advent of our NODEBOX FOR OPENGL package, which relies on fast, hardware-accelerated OpenGL drawing operations using PYGLET. Our PATTERN software package includes a JavaScript implementation of the NODEBOX FOR OPENGL API for online data visualization. Another advantage is the availability of several robust natural language processing packages in Python (e.g., NLTK, PATTERN). We will elaborate on the relation between creativity and language in further chapters.

There has been some debate as to whether or not generative art really is art. Two examples are widely discussed: AARON (Cohen, 1995), a program that produces paintings, and EMMY (Cope, 2001), a program that produces music. In his book *Virtual Music*, Cope invites the reader to a musical Turing test. Alan Turing, who is considered the father of computer science and artificial intelligence, proposed a test to determine if a machine could think. In its original form this test is a natural language conversation with a machine that would or would not be indistinguishable from a conversation with a human (Turing, 1950). In Cope's work, we are asked to distinguish between music scores produced by humans and computers. The task is by no means trivial. Cope's program EMMY (Experiments in Musical Intelligence) analyzes a database of famous composers such as Bach and Chopin[13] for recurring patterns. It uses these to generate new compositions. Classical works are chopped up and reassembled in what Cope calls "recombinant music". The assembly is not random however, it involves 20,000 lines of Lisp programming code with rules for local and global pattern positioning to promote coherence. EMMY does not simply plagiarize the original, it seems to capture the style and emotion of the original (Hofstadter, 2002). According to Hofstadter, the brain is a machine: the substrate of consciousness is irrelevant, it could emerge as easily out of electronic circuits as out of proteins and nucleic acids. This is in fact the central hypothesis of artificial intelligence, known as the physical symbol system hypothesis (Newell & Simon, 1976). But EMMY is not a model of a brain with a mind that experiences emotion. It is a computer program that generates music by manipulating rules and constraints. So how can a program generate emotional music without experiencing emotion?

[13] ftp://arts.ucsc.edu/pub/cope/chopin.mp3

It is not difficult to generate a random music score by syntax. Similarly, it would not be hard to write a computer program that generates poems like Lewis Carroll's Jabberwocky, which uses nonsense syntax. But how can semantics (i.e., meaning) be fabricated? A music score that expresses emotion, or a poem where words and rhymes make sense, tell a story? With EMMY the perceived emotional effect emerges from the pattern positioning rules. They copy small chunks of the original author's semantics in a way that is undetectable to the majority of the audience (say 80%). An even better version might fool 90% of the audience, the next version 95%, 98%, and so on. At some point, hard investments will have to be made to make it 1% (or less) better. This is also known as the law of diminishing returns. Hofstadter calls these improvements "nested circles of style". He notes that it is a depressing thought that the outer, superficial circle of style contributes to the most perceived effect (the 80%), whereas more intricate inner circles contribute less and less. Certainly, there is a difference between the brute force iterations of a music generator or a chess computer, and intuition, identity, personality, hope, fear, life and death – the complex mechanisms that influence human reasoning and creativity. But the difference is a gradual slope, not a vertical rock wall.

Our PERCOLATOR case study is somewhere at the base of this slope. Nevertheless, the rule-based approach of generative art in itself offers creative leverage. In chapter 3 we provide a case study in which we combine NODEBOX FOR OPENGL with a brainwave monitoring device to visualize relaxation in the human brain. It is not possible to produce this kind of art installation with traditional tools (or no technology at all).

Our PERCOLATOR case study is a preliminary computational approach for modeling creativity: a program that is the author of its own artworks. However, it is not very creative, because it only does random things. We have discussed how the approach can be improved by implementing a more involved creative thought process. After defining what exactly creativity is in chapter 4, we present such a non-random case study in chapter 5. Finally, a program such as PERCOLATOR should be able to adapt and evolve. If the audience says: "This is the worst thing ever", PERCOLATOR may need to re-evaluate its artistic style. We reflect on computational techniques to address this task in chapter 7. We will need the theory and software discussed in chapter 6.

# Part II

# THE MIND

•

That monotone monosyllable of the sublime,
That rotten reciprocal pronoun of the nominative,
Wherein spirit rapped
of
which
the whatever
is
U
N
A
W
A
R
E

Nonsense noddle wandered.
Wordless words fierce.

– FLOWEREWOLF, Mind (edited)

# 3 Brain-computer interfaces

All animals with a spine have a brain, as do most animals without a spine, with the exception of sponges, jellyfish and starfish. The brain is the center of the body's nervous system, which specializes in signal processing and coordination. Together with the spine the brain is made up of nerve cells (neurons) that transmit electrical and chemical signals. Some neurons respond to touch, sound or light, others to muscle contractions and glands. The brain has evolved differently in different species. The human brain can roughly be divided into three parts: the cerebrum, the cerebellum at the lower back of the head, and the brainstem that joins the brain to the spinal cord. The cerebrum controls voluntary movement of the body, sensory processing, speech, language and memory. The cerebellum is involved in fine-tuned motor control. While the human brain contributes to about 2% of the total body weight, it consumes about 20% of the energy, or twice as much as the heart (Elia, 1992). When the brain is busy it produces spontaneous electrical activity along the scalp, based on ionic current flows within the neurons of the brain (Niedermeyer & Lopes da Silva, 2004). The brain is always busy unless you are dead.

#### EEG

Electroencephalography (EEG) is a technique used to record the brain's electrical activity, mainly in the outermost sheet of the cerebrum called the cerebral cortex, by placing conducting metal electrodes along the scalp. EEG has a long history in clinical, biological and psychological research, dating back to the work of Caton (1875) and Berger (1924). For example, an EEG study by Hobson & McCarley (1977) showed that dreaming is an automatic neural mechanism (perhaps simply to keep the brain occupied) and not, as Freud (1913) suggested, a manifestation of our deepest desires and secrets.

## 3.1 EEG wireless headset

Researchers from Imec (Interuniversity Microelectronics Centre) and Holst Centre have developed a wireless, low-power EEG headset with the aim to improve patients' quality of life (Gyselinckx et al., 2005). This is called a brain-computer interface (BCI). In comparison to commercial devices such as EPOC[14], the IMEC/Holst prototype focuses on low power consumption and long-term use (Patki, Grundlehner, Nakada & Penders, 2011). For example, in the case of epilepsy, abnormal neuronal activity can be observed using EEG analysis (Fisher et al., 2005). Ambulatory monitoring allows epileptics patients to be tracked while continuing their daily activities.

#### ALPHA, BETA & DELTA WAVES

The headset consists of a microprocessor, a wireless communication module and dry electrodes located at specific positions on the head to capture EEG signals. Different signals are captured at different positions. For example, alpha waves are neural oscillations that primarily occur in the occipital lobe at the back of the brain, during wakeful relaxation with closed eyes. Other well-known oscillations include beta waves (waking consciousness) and delta waves (deep sleep).

[14] http://emotiv.com/

## 3.2 EEG affective visualization

VALENCE (De Smedt & Menschaert, 2012) is a generative art installation created in NODEBOX FOR OPENGL. It presents an interactive game based on brain activity recording. Traditionally, computer games respond to the player's interaction with a game controller such as a joystick or a gamepad. But there are other physiological (e.g., heart rate) and behavioral (e.g., gesture, posture, facial expression) indicators of the player's emotional state (Gilleade, Dix & Allanson, 2005). Monitoring such affective biofeedback (Bersak et al., 2001) can be useful to improve the gaming experience. With the introduction of immersive technology, the highly competitive games industry has changed substantially in a short amount of time (Sung, 2011). For example, with the Nintendo Wiimote controller, the body of the player replaces the traditional console. With the Xbox Kinect motion sensing system, the physical space around the player is transformed into a game world. Such devices have in turn been adopted by the scientific and artistic communities. For example, Gallo, De Pietro & Marra. (2008) use the Wiimote as a 3D user interface for physicians to investigate patients' anatomy. In the arts, Jordà, Geiger, Alonso & Kaltenbrunner (2007) used motion sensing to create the REACTABLE, a collaborative electronic music instrument.

In our work, we use the Imec/Holst headset to control a physics-based simulation. When the player wearing the headset relaxes, the virtual world responds by constructing an aesthetically pleasing composition. We focus on alpha waves (relaxation) and the valence hypothesis (arousal).

#### VALENCE

The valence hypothesis states that the right brain hemisphere is dominant for negative or unpleasant emotions, and that the left hemisphere is dominant for positive or pleasant emotions. Research in this area has been conducted by Penders, Grundlehner, Vullers & Gyselinckx (2009).

We monitor the readings of two electrodes to control the simulation, one left and one right on the back of the head. Since readings can be prone to fluctuation (e.g., a sudden high or low peak) we use a simple moving average (SMA) on the data. The SMA is calculated by taking the average of progressive subsets of the data, smoothing short-term fluctuations and highlighting long-term trends. The hardware setup is illustrated in figure 16.

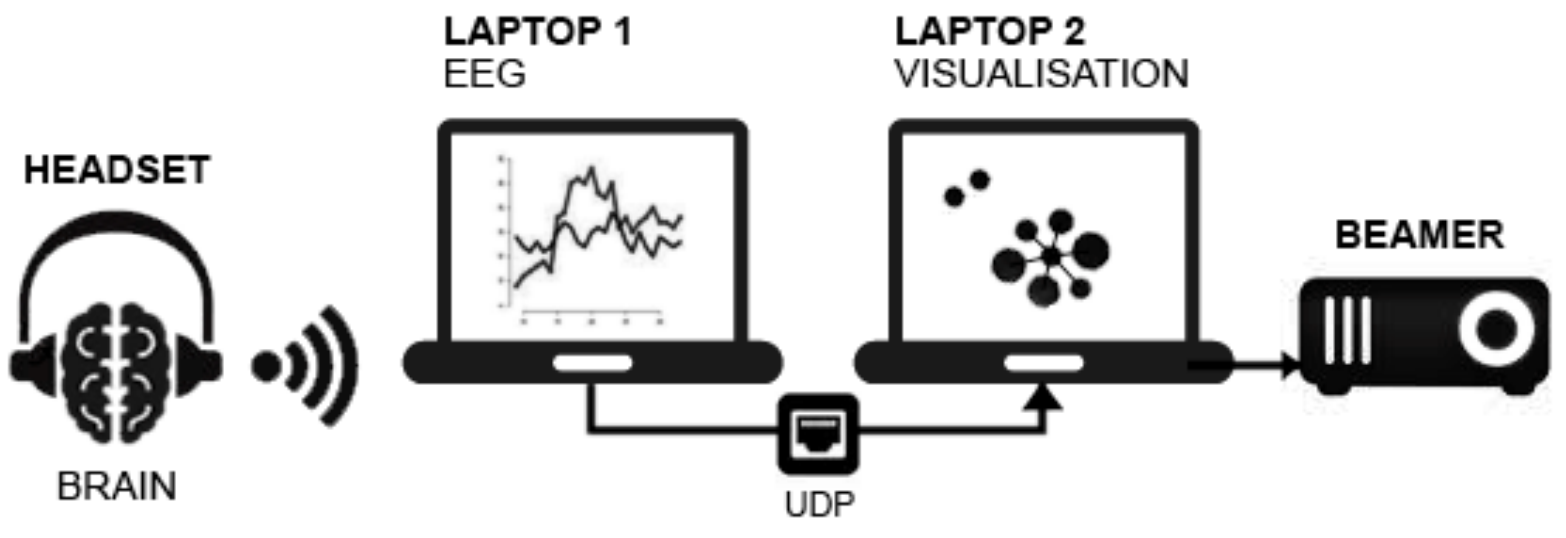

Figure 16. Hardware setup. Laptop 1 analyzes the EEG readings from the wireless headset and sends the SMA values to laptop 2, which renders the visualization using NODEBOX FOR OPENGL.

## Agent-based game environment

To create the virtual world we use NODEBOX FOR OPENGL. The following example animates a circle based on EEG monitoring of alpha waves. The full program is open source and available online on GitHub[15], but it requires the proprietary Imec/Holst headset and drivers to function properly of course.

```
from nodebox.graphics import *
from headset import Headset

# Create a connection to the headset controller.
headset = Headset('127.0.0.1' port=12000)

def draw(canvas):
    global headset
    headset.update(buffer=1024) # Read new EEG data.
    canvas.clear()
    x = canvas.width / 2
    y = canvas.height / 2
    r = headset.alpha[0].current * 100
    ellipse(x, y, r*2, r*2)

canvas.draw = draw
canvas.run()
```

The virtual world consists of graphical elements (or agents) controlled by a physics system of attractive and repulsive forces. The system controls 50–100 agents that roam freely, and one attractor agent. The attractor agent differs from other agents in that it has an invisible radius. This radius increases when the player's EEG alpha wave reading increases (i.e., he or she is relaxed). Agents within the radius are drawn in and clustered around the attractor using a circle packing algorithm.

Circle packing is a technique used to arrange elements that have a circular radius, so that no overlapping occurs (Collins & Stephenson, 2002). The technique is computationally fairly efficient, which is advantageous in our setup since polling the headset, processing the simulation and rendering effects all happen in real-time. As the player relaxes, more agents are attracted. The growing structure is simultaneously urged to the center of the screen.

The graphical elements used as agents are produced by Ludivine Lechat based on our previous work on NANOPHYSICAL (chapter 2). The basic elements have a neutral, oval shape and blue color. High EEG valence readings (normalized on a `-1.0` to `+1.0` axis) will lure more eccentric elements to the screen, such as pink "butterflies" or green "caterpillars". For the attractor we used a pixel filter that produces a rippling effect, based on work by Adrian Boeing[16].

[15] https://github.com/nodebox/valence
[16] http://adrianboeing.blogspot.com/2011/02/ripple-effect-in-webgl.html

Figure 17 shows different elements interacting in the game, where more colorful elements are used to represent emotional brain activity.

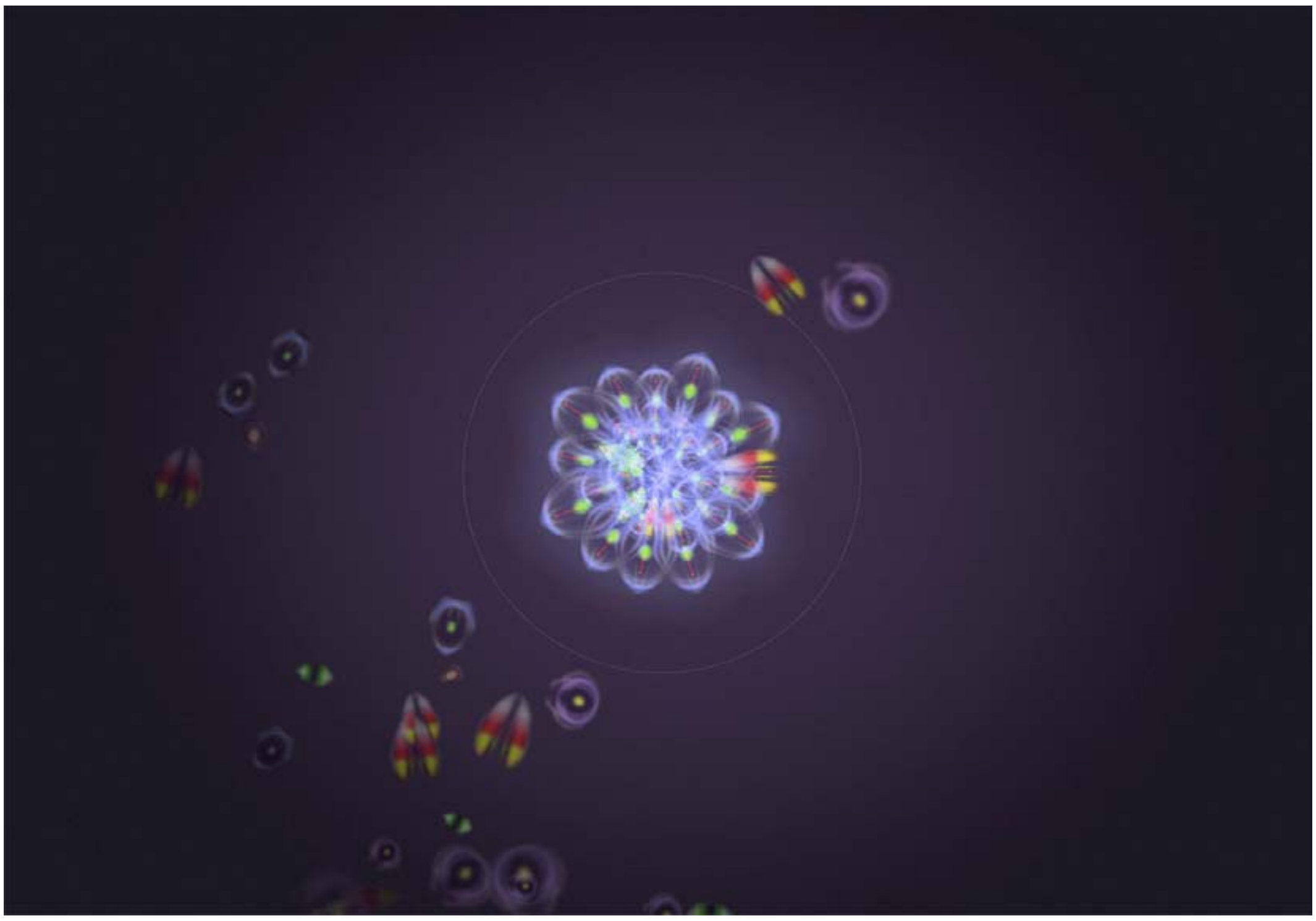

Figure 17. VALENCE animation screenshot.
The current state illustrates that the player is relaxed (alpha > SMA) and aroused (valence > `0.0`).

Similar work using EEG for audio has been done by Verle (2007) who used EEG input for VJ'ing. De Boeck (2009) created STAALHEMEL[17], an art installation that produces acoustic feedback based on the visitor's alpha and beta waves. Haill (2010) differentiates between left and right brain hemisphere to produce music. Sulzer (2012) also produces music based on EEG input. For a discussion of brain-computer interfaces in computer games, see Nijholt, Plass-Oude Bos & Reuderink (2009). Hjelm (2003) created BRAINBALL, a tabletop game where the ball moves away from the most relaxed player. We use a similar neurofeedback approach, that is, the player must reach a goal (relaxation and/or arousal) to control the visual output.

## Evaluation

VALENCE was exhibited during a two-day period at Creativity World Forum 2011. It consisted of a Chesterfield chair, two networked computers (one monitoring the EEG headset, one running the simulation), three display screens and two audio speakers. The headset had to be calibrated for 1–5 minutes to the player's head, accompanied by an explanation of what was going to happen.

[17] http://www.staalhemel.com/

We tested the setup with 25 players varying in age (20–60) and gender. For relaxation (alpha) results were good. Over 90% were able to relax and control the visual output in one or two attempts, with the duration of conscious relaxation varying from 2 to 10+ seconds. For arousal (valence) results were often poor: colorful elements tended to drift off and on the screen without apparent cause. One explanation is the distractive conditions of the setup. We used a camera to record players, while at the same time they were being watched by curious spectators, as illustrated in figure 18. A minority was unable to produce any changes in the game. An explanation is that some people find it difficult to relax in general. One highly effective player claimed proficiency with yoga and meditation.

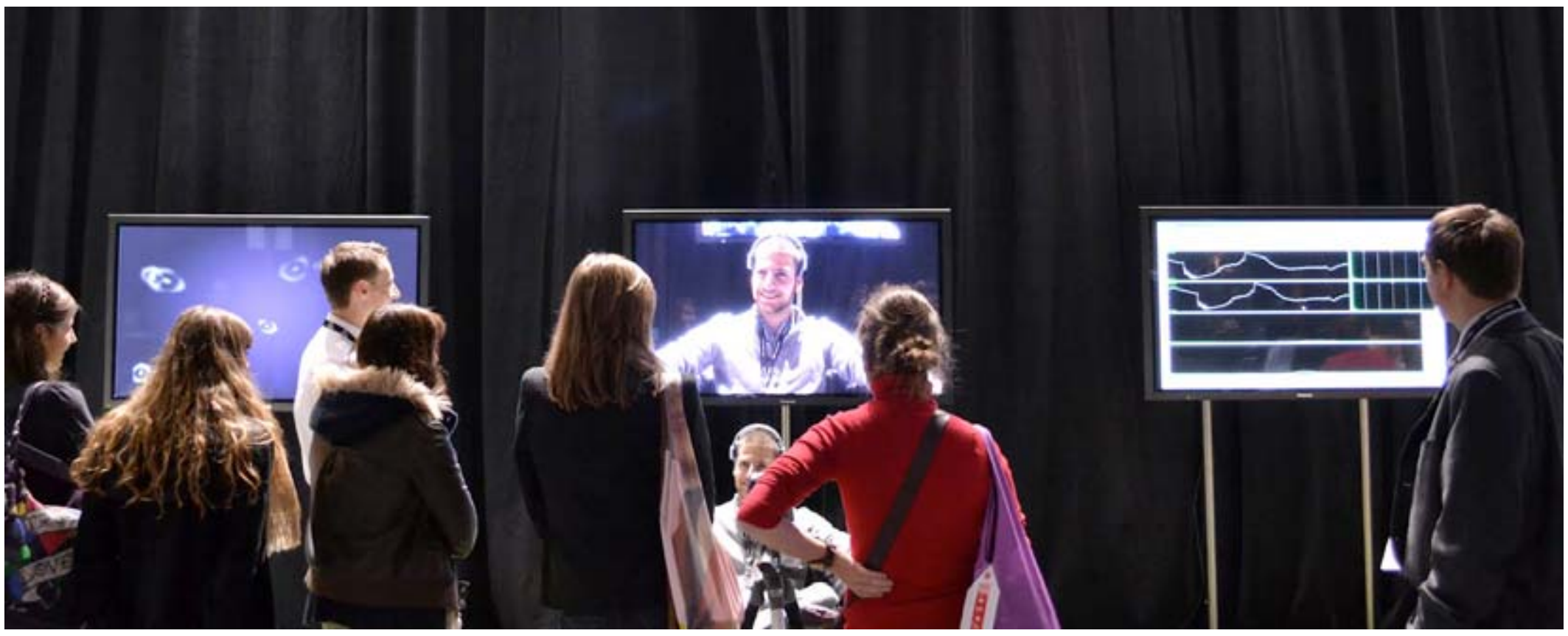

Figure 18. VALENCE at CWF 2011. Different screens display the visualization, the player, and the raw EEG data. Photography © Ludivine Lechat.

## 3.3 Discussion

In this chapter we gave a short overview of VALENCE, a generative art installation that responds to brain activity. Brain recording and imaging techniques are rapidly advancing what we know about the human brain. In VALENCE, we have used brain recording in an artistic context to control an agent-based simulation. The simulation responds when the player wearing the headset relaxes. However, this poses a dilemma since the player is consequently aroused. This can be observed as decreasing alpha waves in the EEG readings. Our setup may benefit from an adjustment of the game mechanics to better coincide with the player's relaxation state. Another problem pertains to the valence hypothesis. When the simulation responds, we noticed a correlation with arousal readings for some players. But since there is no emotionally connotative system (e.g., the installation is expected to evoke a pleasant or unpleasant sensation), our current setup is unable to verify the hypothesis. The setup may benefit from better validation methods, for example using ambient lighting and sound to evoke an expected sensation.

VALENCE is not a case study of how creativity works. Rather, it is an example of creativity itself. By creatively combining existing techniques from two different fields (agent-based modeling and brainwave recording) we can give rise to a new form of “brain-based” art that ties in with the scientific field. The setup would not be possible using only traditional software tools, or no prior technology at all. While in itself our setup is not entirely novel (i.e., H-creative as discussed in chapter 4), it introduces new elements, for example by focusing on the valence hypothesis. These elements may in turn lead to future work exploring new combinations.

This is in a nutshell how human innovation works. Comparable to evolution in nature, new ideas and new technologies are the result of building on previously acquired knowledge. This principle is well-established in the scientific field. A popular metaphor used to describe it is: “The dwarf sees farther than the giant, when he has the giant's shoulder to mount on.” (Coleridge, 1884). New ideas often start out as a vaguely defined whim. For example as in Einstein’s “What would it be like to ride a beam of light?” or, in the case of VALENCE, “How can we simulate The Force from the movie *Star Wars*?” We will discuss how novelty originates from the combination of existing ideas in more detail in the next chapter.

# 4 Human creativity

What is creativity? Is it a perfect dish in the friday night TV cooking show? The on-stage fashion cantrips of Lady Gaga? A smart presentation in the corporate meeting room? Retro eyeglasses and tattoo sleeves? With recent popular attention and its history in the divine, the definition of creativity has been obfuscated. To create a model of computational creativity that goes beyond random manipulation of rules, we need to examine more rigorously how human creativity works. Below we offer a short overview of the scientific literature, starting with what it means "to have an idea" and then discussing how creative ideas can arise.

## What is an idea?

An idea is a mental impression ranging from concrete to abstract. For example, a mental image of your housecat (concrete idea) or a generalized concept of all cats (abstract idea). Concepts are formed by means of inference (Mareschal, Quinn & Lea, 2010). If all domestic cats prowling your neighborhood have a tail and whiskers, it is plausible to assume that cats elsewhere will also have tails and whiskers. Some don't of course. The Manx cat for instance has a mutation that shortens the tail, but such exceptions are usually excluded from the conceptual model. Schematically, this model might look as follows: *cat* → tail + whiskers + fur + meow. Conceptual classification is flexible. Different models of a single concept can exist depending on the context: for example *valuable* → laptop & jewelry, as opposed to *valuable* + *burning house* → baby & cat.

#### MEME

An idea is a unit of thought. Like words (i.e., basic units of language), ideas can be combined (*cat* + *dog* → *pet*), related (*cat* ↔ *curiosity*) and shared. Ideas that spread from person to person within a culture are called memes (Dawkins, 1976). In analogy to genes, they adhere to the principles of competition, variation and selection. Interesting ideas are eagerly shared and flourish. They can also mutate, for better or for worse, in the process of writing or speech. This is called memetic drift. On the other hand, foolish and irrational ideas do not necessarily die out. They are often vehemently defended as part of a belief system (Sutherland, 1992). Ideas, both good and bad, can be crafted to become highly contagious, for example in viral marketing or religion (Lynch, 1996).

## Where do ideas come from?

Where do ideas come from? Sometimes we say that ideas come from the heart, but of course this is just a figure of speech used to accentuate earnesty. Ideas originate from conscious thought in the brain, the physical structure that generates the mind. In popular conviction, animals are frequently taken to be mindless whereas only humans are capable of conscious thought, since it is the only species with a developed sense of self (i.e., "me"). But we must differentiate carefully between self-awareness and consciousness.

#### SELF-AWARENESS

Can animals think? Do cats have ideas? The domestic cat may appear pensive before suddenly stirring to grab some lunch. It is hard to say if this is hunger instinct, conditioned behavior (owner-feeder at the doorstep!), boredom or genuine introspective thought. Possibly, the cat's brain is simply iterating a loop of selected strategies: "loiter?", "patrol?", "flee?", "fight?", "eat?", "mate?" Cats are quite complex machines but they lack a developed sense of self. The mirror test is a measure of self-awareness used to assess if an animal is capable of recognizing itself in a mirror (Gallup, 1970). According to Gallup (1982), self-awareness is the basis for the mind. It confers the ability to deal with social pressures more intricate than a fight-or-flight scenario, such as reciprocal altruism ("will I benefit by helping?"), deception ("will I benefit by cheating?") and grudging ("am I being cheated?"). Cats do not pass the mirror test. Most animals respond as if the image represents another animal. Presented with a mirror image, the cat will iterate its loop of strategies: "fight it?", "eat it?", "mate with it?" Since the mirror is not very responsive the cat loses interest. Occasionally, it will return to peek behind the mirror where the potential rival might be hiding. Humans pass the mirror test once they are about 18 months old (Gallup, Anderson & Shilito, 2002), along with the other great apes (bonobos, chimpanzees, orangutans, gorillas), elephants, dolphins, orcas and magpies. The magpie will scratch at its neck when it notices the colored marker in its neck in the mirror. When presented with a mirror image, we are aware of ourselves. We can try out party outfits or fret over the inevitable course of aging.

#### CONSCIOUSNESS

However, exactly what the mirror test demonstrates has also been debated. It is primarily a test of vision and some animals rely more on smell or hearing. Self-awareness is not the same as consciousness. To be conscious in the broadest sense, it is only necessary to be aware of the external world (Sutherland, 1989). The cat is *conscious of* the mirror's existence and some cat therein; only it does not understand that it is an image of itself. This does not reject that the cat is, for example, capable of preparing a voluntary attack on the rival behind the mirror ("fight it?"), which is a form of *implicit* self-awareness (Kriegel, 2003). It is likely that self-awareness emerges from higher-order consciousness. But in the broadest sense, consciousness is an analytical feedback processing loop.

The human brain is a self-organizing, parallel, distributed system. It consists of billions of neurons, each connected to thousands of other neurons in a complex network, ceaselessly transmitting information by exchanging electrical and chemical signals (Pakkenberg & Gundersen, 1997). By comparison, the cat brain has a few hundred million neurons (Roth & Dicke, 2005). The ant brain has a few hundred thousand. It follows that the human mind is more elaborate than the mind of the cat or the ant. Human consciousness, awareness of the self and the world around us, has more elbow room than cat consciousness, and cat consciousness more than ant consciousness. This does not mean that the housecat is incapable of taking pleasure in interacting with you. Even if the cat is only dimly aware of exactly who is being groomed, it still enjoys a fine back rub. We could think of consciousness as a grade (Hofstadter, 2007) from `0.0` to `1.0` instead of `0` *or* `1`. If adult humans score `0.8`, then elephants score `0.7`, cats `0.2` and ants `0.00000001`.

Processing the brain's neural signals to form a conscious experience takes time; about 300 milliseconds (Libet, Gleason, Wright & Pearl, 1983). This means that the signals meant to flick your hand may fire before realization dawns that you want to pet the cat. At a single point in time, there is no central command center that has all the information, no homunculus pulling levers (Dennett, 1991), "move hand, pet cat!" Consciousness is emergent (Dennett, 2003). After some grooming, the cat quite contently settles down on your lap. It drifts out of focus, but unconsciously you continue to scratch it behind the ears. You are vaguely aware of purring noises, yet the mind has turned to other thoughts. You may be contemplating what to eat, whether you'll wash the car or how to invade Poland. When the cat sinks its claws into your leg, it comes back into focus. You remember, sort of, scratching it behind the ears. Now that the cat is in onscious focus, whatever you were thinking of in turn recedes to the shadows of the mind.

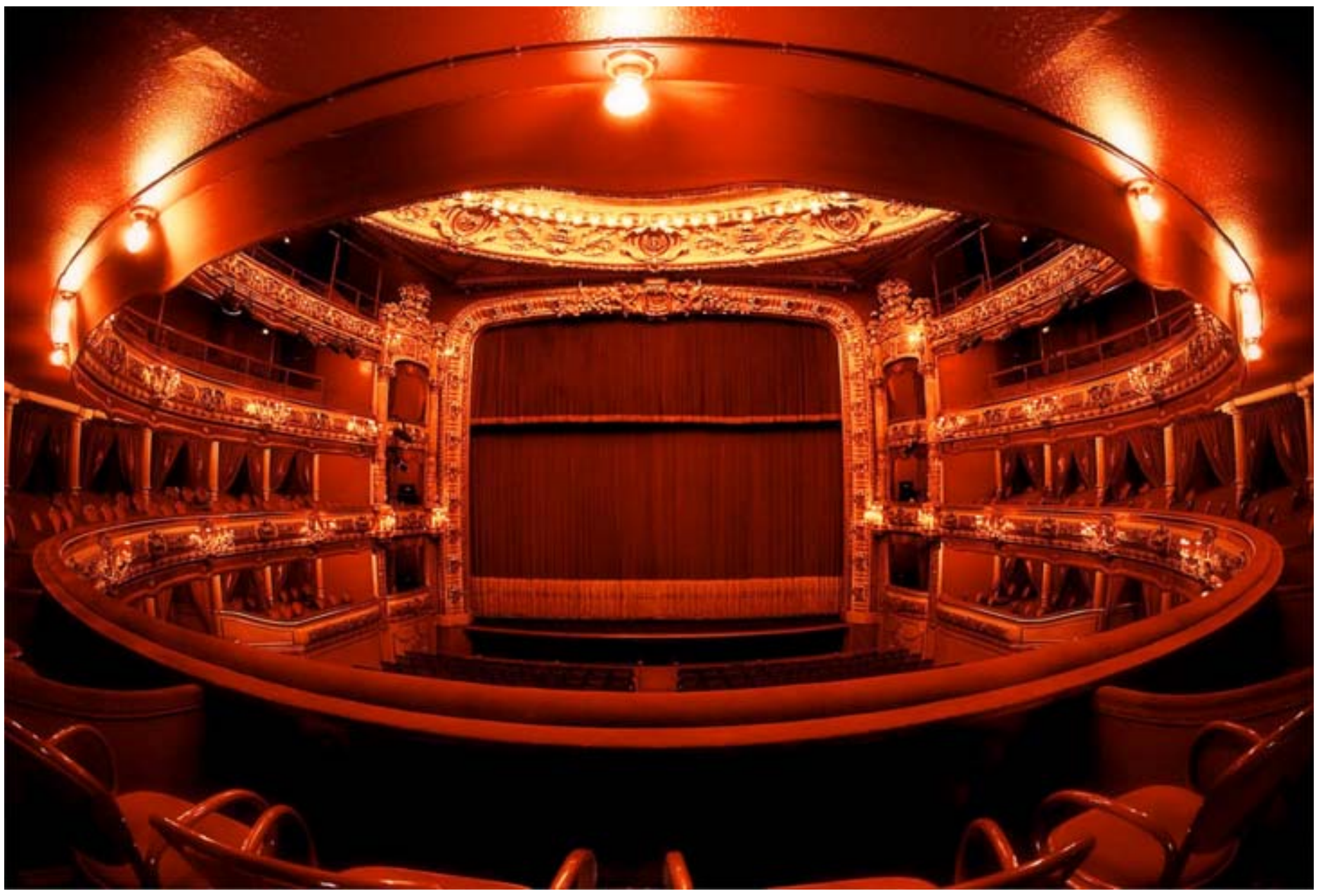

Figure 20. THEATRE IN RED. Photography © Fabrizio Argonauta. Used with permission.

A well-known metaphor compares consciousness to a theatre (figure 20) rather than a command center. Bottom-up instead of top-down. Whatever comes up on stage, under the spotlight, has the attention of the audience, watching from the dark. What comes up on stage has enough influence to affect what the mouth will say and what the hands will do. In the shadows, vaguely conscious events are waiting to take to the stage (claw!). Behind the scenes, expert unconscious processes set things in motion for the next act. Unfortunately the stage is rather small. It corresponds to the brain's working memory, the *global workspace* (Baars, 1997) with access to any single neuron in the brain (= the audience). Only "famous" events make it to the stage. Other matters are taken care of by (often conflicting) unconscious processes behind the scenes, for example driving the car while you are asking yourself if you forgot to buy cat food.

**MULTITASKING**

It is a popular misconception that the mind can engage in multiple conscious efforts at the same time. Imagine what it would be like to have two (or even three) consciousnesses. At best, some individuals are simply better at switching between tasks, but the context switch introduces a delay and errors (Meyer & Kieras, 1997). A good advice is not to mix work with social network status updates.

# 4.1 Creativity

In our work, we are interested in the origin of ideas that serve to tackle some kind of problem. Ideas that have a goal. Such ideas can offer bad solutions (e.g., climb in tree to grab cat) or good solutions (e.g., lure cat down with food). Successful solutions can be very creative. Unsuccessful solutions may appear to be creative as well, but they often also classify as "stupid", "absurd" or "inappropriate". This excludes them from the definition of creativity, as we will see.

Human beings have the ability to imagine themselves in the past, present and future, and to come up with intentional solutions to improve their condition. This involves having creative ideas in the first place, and then implementing their invention. Such inventions include the cooking fire, the fishing net, the wheel, language, democracy, zero, the printing press, the automobile, electricity, aspirin, theory of relativity, computers, human rights, even the questionable keyboard guitar. Clearly, not all ideas are equally great (theory of relativity ↔ keyboard guitar). But the point is that throughout history, all sorts of individuals have had interesting ideas: scientists, thinkers, artists, housewives, craftsmen, madmen. Creativity is intertwined with every aspect of our life, from luring a cat out of a tree to improvising a music instrument or inventing a combustion engine. It is not restricted to an elite. Like consciousness, we could grade the creativity of almost any idea on a scale from `0.0` to `1.0`.

Most often, creativity is combinatorial in nature. This holds that everything arises out of something, that creativity does not magically come into existence out of nothing (Perkins, 1988). A new idea emerges from the unfamiliar combination of existing ideas. To invent a spork (or foon) we'd need a spoon and a fork. To create a computer (1941), we'd need arithmetic (2000 BC), logic (19th century) and electricity (19th century) to power the circuits. To study electricity, Benjamin Franklin reportedly combined a kite with a dampened string attached to a metal key during a lightning storm. Humans have done well because we not only share genes but also ideas, outrunning natural selection. With each generation the pool of existing ideas to combine into new ideas, the "knowledge search space", expands. Knowledge and learning form the foundation of creativity (Weisberg, 1999). Imagine you are part of a tribe of cavemen some 40,000 years ago (Oakley, 1961). One day, Fred the Fire Maker is struck down by lightning, leaving the community without the knowledge of how to build a fire. Disaster! The skill could be lost for hundreds of years to come. No cooked meat, no hardened tools. It would have been worth the while if you had observed Fred closely prior to his untimely death. You might have agreed on a clever set of grunts to transfer the knowledge more easily. For example, "HRRRH" for "rub stick!" and "PHFFH" for "keep puffing!" Such rudimentary communication might give the tribe an evolutionary advantage (chapter 1) in passing down the secret of fire from generation to generation.

Sternberg (1999) defines creativity in terms of novelty and appropriateness:

> Creativity is the ability to produce work that is both novel (i.e., original, unexpected) and appropriate (i.e., useful, adaptive concerning task constraints).

Boden offers a similar definition:

> Creativity is the ability to come up with ideas or artefacts that are new, surprising, and valuable. "Ideas," here, includes concepts, poems, musical compositions, scientific theories, cooking recipes, choreography, jokes ... and so on, and on. "Artefacts" include paintings, sculpture, steam-engines, vacuum cleaners, pottery, origami, penny-whistles ... and you can name many more.

#### NOVELTY: BIG-C ↔ LITTLE-C

Sternberg and Boden both mention novelty. Novelty is commonly understood more specifically as adaptive novelty (Perkins, 1988), being able to cope with new problems (such as an ice age) using new strategies. Creative ideas are new ideas. But of course, there's new – and there's *new* (Boden, 2003). An idea may be original to the person who comes up with it, even though it is not necessarily new in a historical sense. When caveman Fred learned how to start a fire using friction, he must have felt quite surprised at first, terrified even, and gradually more and more delighted as he slowly discovered the advantages of a good fire. How clever of him. However, caveman Barney might have started doing likewise, far away and maybe even hundreds of years earlier. This does not make the idea any less valuable or creative from the perspective of Fred. Boden explains this as a distinction between P-creative ideas (personal) and H-creative ideas (historical). A similar distinction that is widely used is between little-c and Big-C creativity (Gardner, 1993, Kaufman & Beghetto, 2009). Little-c refers to everyday problem-solving and personal creative expression: luring the cat from a tree, making a better fishing net, rearranging the furniture. Big-C refers to eminent, great ideas, such as the theory of relativity or the combustion engine. Big-C is the stuff that fills history books.

#### APPROPRIATENESS

Sternberg and Boden both mention appropriateness, interestingly a synonym of fitness. Novelty alone does not account for a creative idea (Runco & Charles, 1992). For example, caveman Barney might have tried for a while jumping up and down as high as he could, stick held aloft, in an attempt to reach the sun and make the stick burn. Although his idea is quite original, it is not a very valuable solution for making fire. The idea would be more appropriate if Barney was a stand-up comedian. We can see how novelty and appropriateness relate in table 3 (Kilgour, 2007).

| NOVELTY | APPROPRIATENESS | |
|---|---|---|
| | LOW | HIGH |
| LOW | not creative = stupid | not creative = common |
| HIGH | not creative = absurd | **creative** |

Table 3. Novelty and appropriateness of creative ideas.

Because the term "creativity" is used in a diverse manner by scientists, artists and lay people, and because of its historical roots in the divine, scientific research has been scattered and obfuscated (Batey & Furnham, 2006). Unifying this research in a general theoretical framework of the mind is ongoing work. For a comprehensive overview, see *Handbook of Creativity* (Sternberg, 1999).

## 4.2 Inspiration

> A poet is a light and winged thing, and holy, and never able to compose until he has become inspired, and is beside himself and reason is no longer in him.
>
> – Plato, Ion, 534b–d

In most languages, personal opinions are expressed using adverbs and adjectives (Bruce & Wiebe, 1999, Benamara et al., 2007). For example, when we like a song we say that it is a "very good song" or a "great song". If it also makes us want to create music of our own, we say that it is "inspiring" or "thought-provoking". Inspiration refers to a sudden, unconscious onset of creative ideas, often in an artistic context such as literature, poetry or music. Because of the notion of inspiration, creativity has been shrouded in mystical beliefs. The realization that thought emerges from the brain is fairly recent. Historically, creative ideas were attributed to divine intervention (Sternberg & Lubart, 1999). The ancient Greeks believed that a creative person was an empty vessel filled with inspiration by a divine being (e.g., a Muse). The individual would then pour out the deity's directives in a frenzied fashion. For example, Archimedes was reportedly struck by inspiration while taking a bath, when he understood that the volume of water displaced must be equal to the submerged volume of his body, crying Eureka! Eureka! ("I have found it!") and ecstatically running into the streets of Syracuse naked. However, this account does not appear in his own writings. It is also plausible that Archimedes recognized the principle known in physics as buoyancy through careful study rather than frantic running about.

The inspirational view persisted throughout Christianity, where creativity was believed to be a gift of the Holy Spirit: "See, I have chosen Bezalel [...] and I have filled him with the Spirit of God, with wisdom, with understanding, with knowledge and with *all kinds of skills*" (Exodus 31:2–3). Since only God had the power to create something from nothing, any human creative act was necessarily an expression of God's work (Niu & Sternberg, 2006). During the 18th century Enlightenment – the dawn of reason, science, human rights – creativity was gradually attributed to the individual (Albert & Runco, 1999). Individuals such as Mozart were seen as exceptional or genius, gifted with a unique and innate talent. This talent could not be learned, taught or analyzed but it was not divine either (Boden, 2003). Only vague suggestions are offered as to how this would work. The best one could do was to give a genius room to work.

In the 19th century, Darwin's work on evolutionary adaptation and diversity came in sharp focus. Galton (1822–1911), who frequently corresponded with Darwin, took interest in the heritability of genius and consequently introduced the study of diversity to psychology, that is, diversity in terms of individual human differences. Albert & Runco (1999) see Galton's contribution as essential to psychological research since he established the statistical concept of correlation, and the use of questionnaires and surveys to collect quantifiable data. Instead of making assumptions, we started measuring and comparing things.

Today, talent or aptitude is understood as a predisposition towards certain interests. A readiness to learn and perform well in a particular situation or domain, involving a complex of factors such as heredity, environment, personality, intelligence, experience, knowledge and motivation (Eysenck, 1995, Snow et al., 2002). Each of these areas is studied in relation to creativity.

#### HEREDITY

There is no evidence that creativity runs in families (Martindale, 1999). However, as Martindale points out, several traits from which creativity emerges such as intelligence and antisocial behavior are heritable. For example, recent work by Comings et al. (2003) suggests that intelligence is a polygenic trait (i.e., depending on multiple genes). Other factors such as social environment that contribute to creativity, are not biological at all: human creative potential requires the contribution of both nature and nurture (Simonton, 2000).

#### ENVIRONMENT

Galton argued that eminence (Big-C creativity) relies on reputation: “the opinion of contemporaries, revised by posterity”. Most creators do not work in isolation from other creators, but inside a particular scientific, intellectual or artistic discipline (Simonton, 2000). It is this expert group, active in the same field and governed by the same rules and techniques, that recognizes and validates original contributions to the domain. This is called the systems model of creativity (Csikszentmihalyi, 1999). Furthermore, creativity is influenced by social context (war, revolt, heterogeneity) and family environment (birth order, parental loss, marginality, role models). Nurturing environments do not always foster creativity (Eisenstadt, 1978). Diverse and challenging experiences seem to be a requirement for nonconformity and perseverance (Simonton, 2000). Creative children tend to be more playful (Lieberman, 1965).

#### PERSONALITY

There is abundant evidence that creative individuals are disposed to be ambitious, independent, introverted, nonconformist, unconventional, more risk-taking, with wide interests and open to new experiences (Martindale, 1989, Feist, 1999, Simonton, 1999). Several studies correlate artistic creativity with a predisposition to mental illness; in particular obsessive–compulsive behavior, antisocial behavior, mood and alcohol-related disorders and manic depression (Nettle, 2006, and Burch, Pavelis, Hemsley & Corr, 2006). Artists in particular are also more sensitive.

#### INTELLIGENCE

It has been shown that creativity correlates positively with intelligence up to an IQ score of 120 (Guilford, 1967), but beyond that there is no significant correlation. This is called the threshold theory. The `< 120` threshold overlaps with little-c creativity. Since intelligence includes general problem-solving abilities, the smarter you are the better you will perform when solving everyday problems. But it does not explain Big-C creativity. If it takes an IQ of 130 to understand physics, then to be a great physicist with unusual and valuable ideas about physics, any additional increase in IQ is less important than other factors such as personality, experience and motivation (Simonton, 2004). In recent years, the IQ measurement has also been the subject of criticism. It is argued that IQ only measures a few aspects (i.e., logical, linguistic) of multiple intelligences (Gardner, 1999).

#### EXPERIENCE

A well-known study by Ericsson (1998) shows that it takes about 10,000 hours (3 hours daily = 9 years) of deliberate practice to become an expert musician, chess player or athlete. Deliberate practice implies doing it again and again to identify, evaluate and improve flaws. Boring? It will help if you are intrinsically motivated ("I can do this"), supported by devoted teachers ("you can do better") and having enthusiastic relatives to pay for instruments and courses ("you are doing well"). It is a popular myth that it is possible to chance upon some hidden talent and become famous overnight without doing anything. In the arts for example, entrance exams continue to err by preferring innate skill ("her drawings are wonderful") over eagerness to acquire new skill ("she's very motivated *but unfortunately* she can't draw...")

#### KNOWLEDGE

Modern theories about the mechanisms of creative thinking are grounded in, among other, psychology (e.g., Guilford, Mednick, Sternberg), Darwinian theory (e.g., Simonton), social perspectives (e.g., Amabile, Csikszentmihalyi) and cognitive science (e.g., Boden, Martindale). All of them share a common issue: knowledge. Many theories propose a tension between creativity and knowledge, where a minimum amount of domain-specific knowledge is required to be creative in a domain, but too much leads to stereotyped response. Others have argued a foundation view, where creativity and knowledge are positively correlated, with knowledge as the building blocks out of which new ideas are constructed (Weisberg, 1999). We will elaborate on the foundation view further on.

#### MOTIVATION

Motivation refers to a desire to do something, from getting up in the morning to feed the cat to staying up at night to study physics. Intrinsic motives include curiosity, interest and enjoyment in a particular task. Extrinsic motives include reward or recognition apart from the task itself. Amabile, Hill, Hennessey & Tighe (1994) show that intrinsic motivation correlates positively with creativity, whereas extrinsic motivation correlates negatively with creativity. An explanation for the negative correlation is that people try too hard when offered a reward (Sutherland, 1992). This kind of stress prevents flexibility of thinking. People will keep doing whatever is uppermost in their mind. This is also called the availability error. Amabile et al. suggest that a combination of intrinsic and extrinsic motivation may be favorable however. People with a mix of both tend to do well at complex problem-solving. "Passion" is often used to describe this attitude. An optimal state of intrinsic motivation is also called flow, in which one is totally immersed in an activity (Csikszentmihalyi, 1990).

Table 4 offers a brief summary of the factors that contribute to talent:

| | |
|---|---|
| **HEREDITY** | Have unconventional, intelligent parents (IQ >= 120). |
| **ENVIRONMENT** | Surround yourself with creative individuals with similar interests. Play. |
| **PERSONALITY** | Take time apart to think and investigate. Never take things for granted. |
| **INTELLIGENCE** | Pick a task your mental ability can barely handle. Then work extremely hard. |
| **EXPERIENCE** | Practice deliberately for 9 years. |
| **KNOWLEDGE** | Read, listen, observe, debate, fantasize, question. Keep asking silly questions. |
| **MOTIVATION** | Make a living out of the things you like doing (≠ what your dad likes doing). |

Table 4. Summary of the complex of factors that contribute to talent.

## 4.3 Intuition and insight

Intuition refers to an *a priori* understanding without the need for conscious thought: a hunch or hypothesis that we believe to be true. Many eminent creators have reported following a hunch of some sort (Runco & Sakamoto, 1999).

In Ancient Greece, Pythagoras and his students had a hunch about numbers and proportions (at least that is how the story goes, but let us indulge for a moment). The Pythagoreans' most beloved shape, the pentagram or five-pointed star, was a glimpse at the infinite. At the center of the star is a pentagon or five-sided polygon. Connecting the corners of the pentagon generates a new pentagram, and so on. The lines of each pentagram are governed by a beautiful proportion: the golden ratio, a division so that the ratio of the small part to the large part is the same as the ratio of the large part to the whole. It was perfect. It could be observed in nature over and over. Surely, if harmonic proportions governed numbers, shapes and nature alike, then did they not govern the construction of the entire universe?

But intuition is a systematic source of error in human judgment (Bowers, Regehr, Balthazard & Parker, 1990), misguided by cognitive heuristics such as representativeness and availability ("all cats have tails"). The golden ratio is an irrational number: $\phi = 1.618033987...$ It is not a proportion of nice, whole numbers. It is not 3/2, and not quite 5/3, and not entirely 8/5, and neither is it exactly 13/8. To the ratio-obsessed Pythagoreans, this wasn't perfect at all! We can only wonder about their disappointment. To top it off, the golden ratio does not appear "over and over" in nature. It appears *occasionally*, since it is the factor of a logarithmic spiral. Some natural phenomena such as shells, hurricanes or galaxies approximate logarithmic spirals (Livio, 2003).

Intuition is a poor method for prediction. However, it is an important faculty for discovering new ideas: an unonscious process that slowly converges into a conscious idea (Bowers et al., 1990). At first, a hunch is vague. An embryonically good idea that requires further development and testing (Forceville, C., personal communication, 2012). At the same time it feels promising already. But how can it feel promising if it is as yet undeveloped? In the mind, many (conflicting) unconscious processes are at work simultaneously. Some of these generate useless combinations (Boden, 2003), while others are gradually becoming more coherent in pattern, meaning and structure. These emerge as a promising hunch which can eventually develop into a flash of insight. In comic books, this moment of breakthrough is signaled with a light bulb over the character's head.

#### PRIMING

The transition from unconscious intuition → conscious insight is believed to involve a period of incubation, during which one does not consciously think about the task, but where the mind continues to work on it below the level of consciousness (Nickerson, 1999). This is different from multitasking, which really is a conscious (and costly) task-switch. The transition is long-term, automatic, unconscious. One of the mechanisms involved is called priming: increased sensitivity to particular stimuli as a result of recent experience such as playing, driving a car, reading a book, watching a movie or having a dream. If we offer the word "cat" and then ask you to name two pets, there is a good chance you will say "cat" and "dog". This is because "cat" is now saliently available in your mind, and "dog" is closely related to it. This is not so different from the hierarchical fair competition we discussed in chapter 1. A stream of primed stimuli to challenge, reinforce or hinder the formation of ideas. This implies that creativity can be enhanced by seeking positive stimuli (e.g., interesting books, movies, games) which appears to be supported by experimental studies (Mednick, 1964, Gruszka & Necka, 2002, Sassenberg & Moskowitz, 2005).

That said, we can now turn to the question how new ideas emerge out of unconscious processing.

## 4.4 A concept search space

Let us consider the mind as a search space: a roadmap of concepts with paths in between (e.g., highways, streets and trails), connecting the concepts that are related to each other. How many concepts are represented in the search space (i.e., how far the map spans) depends on knowledge and experience. How each concept relates to other concepts depends on education, environment and personality. If we think about a *cat*, we might recall related concepts such as *dog* and *mouse*. These are easier to reach from *cat* than *toaster*. If you once owned a cat, thinking about your cat could also recall memories of a toaster: *cat* → *my cat* → *6 a.m. meowing* → *breakfast* → *toast*, and so on. Because we cannot examine our own mind as a whole, neither can we "see" the search space as a whole. But we can wander along its paths, consciously or unconsciously. Each concept activates new associative paths to follow, by circumstance, similarity or mediation (Mednick, 1962, Balota & Lorch, 1986). Some paths may lead us to explore a part we didn't even know was there. Some paths may be too long, too old or too unreliable to investigate right now. Some will lead to a dead end or run in circles.

As more paths are traversed – as more related concepts get involved – the idea grows. Short paths to nearby concepts yield commonly available associations. For example: *cats chase mice*. Mundane ideas come to mind first, and then with effort more novel ideas (Osborn, 1963). For example: a *hungry cat makes a good alarm clock*. Distant or atypical relations yield Big-C creative ideas (Schilling, 2005) but come slower (Jung-Beeman et al., 2004). More relations in a larger search space enable more creative ideas.

#### THINKING STYLE

The search space can be traversed in different ways, using different thinking styles. A thinking style is a search strategy defined by rules and constraints that makes analogical pattern matching possible. Thinking styles primarily include unconscious, associative heuristics (shortcuts) such as intuition, rules of thumb, trial and error and common sense (Gilovich, Griffin & Kahneman, 2002). Because the search space is a simplified representation of reality, unconscious heuristics are subject to cognitive bias ("all cats have tails"), even though they may "feel right". A conscious thinking style can be constructed from a logical rule such as "find as many as possible" or "find the most unusual", combined with a constraint such as "pet" or "loud". For example, possible pets we can think of right now are cat, dog, fish, parrot and so on. Unusual and loud pets we can think of include rooster, tiger, dinosaur and, after giving it some more thought, a miniature Darth Vader. The last specimen is an example of imagination and conceptual blending, discussed further below. Thinking styles correspond to personality traits and environment (Kirton, 1976). Some people prefer to give unusual answers to tentative questions whenever possible. Others prefer logical answers to concrete problems. Mental agility (Hofstadter, 1995) is the ability to employ different thinking styles.

Style of thought was extensively studied first by Guilford (1956) and later by Torrance (1988). Guilford distinguishes two ends of a continuum: convergent and divergent thinking. Convergent thinking finds the single best solution. It can be used to solve practical problems such as: "How do you open a locked door if you don't have the key?" Divergent thinking freely explores many possible solutions. It can be used to solve open-ended problems such as: "What can you do with a brick?" Divergent thinking has been correlated to creative personality traits such as openness to new experiences and curiosity (McCrae, 1987). A well-known activity to promote divergent thought is brainstorming: rapidly generating ideas without immediately criticizing them.

Kahneman (2011) proposes a generalized dual-process model of thinking, with two styles called SYSTEM 1 and SYSTEM 2. System 1 is fast, automatic, unconscious and intuitive: "the secret author of many of the choices and judgments you make". System 2 corresponds to slower, conscious reasoning. It functions as a support system for more involved tasks. As Kahneman points out, System 1 is more present since consciousness is essentially lazy and distractible. The differences are outlined in table 5.

| SYSTEM 1 | SYSTEM 2 |
|---|---|
| fast | slow |
| unconscious | conscious |
| intuitive | analytic |
| associative | rule-based |
| affective | intentional |
| eager | lazy |

Table 5. Unconscious SYSTEM 1 vs. conscious SYSTEM 2 thinking styles.

Boden (2003) discerns three general styles of creative thought:

**COMBINATION**

Combinatorial creativity is the most common style. It pertains to a novel combination of existing ideas. Strategies include displacement (e.g., a dinosaur in your bed), comparison (e.g., toaster ↔ Darth Vader) and blending distinct ideas (e.g., a science fiction world with prehistoric animals).

**EXPLORATION**

Exploratory creativity can be seen as wandering aimlessly, finding out where paths lead to and how far the map stretches. This is manifested in activities such as sketching and daydreaming. Exploration uncovers concepts and relations we didn't realize were there. It generates new search strategies.

**TRANSFORMATION**

Contrary to a roadmap, the mental search space (i.e., the mind) can be changed. Transformational creativity changes the search space. It introduces new concepts, generates new relations or re-routes existing ones. Herein lies the origin of Big-C creativity; discoveries such as Einstein's theory of relativity that transform our minds' understanding of reality.

## 4.5 Imagination and metaphor

Imagination is a playful, analogical thinking style expressed in pretend play and daydreaming (Policastro & Gardner, 1999). It can be seen as a bridge between the conscious and unconscious. It is used to answer tentative questions such as Einstein's metaphorical "What would it be like to ride a beam of light?" There is considerable interest in analogy and metaphor in the field of computational creativity (see for example Veale, O'Donoghue & Keane, 2000, and Pereira, 2007). A metaphor is a figure of speech where a tangible concept is used to represent an abstract concept. For example, the mind as a *stage*, life as a *journey*, work as a *rat race*, or space-time as a *fabric*. Metaphors enable us to imagine how complex things work more easily.

**CONCEPTUAL BLENDING**

The theory of conceptual blending (Koestler, 1964, Lakoff & Johnson, 1980, Fauconnier & Turner, 2003) explores metaphor and imagination in further detail, particularly how concepts can be combined into new meaning. Fauconnier & Turner propose the idea of a blended search space: an *ad hoc* mental pocket where concepts are brought together and blended based on shared similarities.

An example is the "house" pretend game. Various dolls and Mr. Bear are seated around a small table. They are inert of course, and there is no real tea in the teapot. But the child playing the game can project what it knows from real life onto the scene. She herself is now the mom, Mr. Bear the dad, and the dolls are children that have to sit up straight and behave before being served invisible tea. Distinct concepts from the real world – family members and toys – are blended together in a PRETEND space. Dad's voice is mapped onto Mr. Bear. The dolls are reprimanded for their table manners in a way the girl would be in real life.

But events in the PRETEND space are allowed to run their own course; perhaps the dolls will rebel and fly off on their little ponies. Another stereotypical example is a PARODY blend space where the mannerisms of an army drill sergeant are mapped onto the angry boss during an after-work recount. Or a WISH blend space for example: a daydream in which you project the daily routine of a millionaire onto your own life.

Conceptual blending offers a framework to study creativity using computational techniques. We will elaborate on this in chapter 5.

## 4.6 Creative thinking test

Before concluding this chapter and moving on to discuss a computer model of creativity in the next chapter, we present the results of a creative thinking test. Here we are interested in the popular belief that artists are more creative than other people. Is creativity related to profession? Are artists more agile in employing thinking styles than other individuals? To answer this question, we subjected 187 participants varying in age, gender and profession to Guilford's Divergent Thinking test (DT) in an online survey. In a DT test, subjects are asked to come up with as many answers as possible to an open-ended problem solving task. In our setup they were asked to give 0–5 answers to the question: "How do you open a locked door if you don't have the key?" We did a second test with "What can you do with a brick?" with results comparable to those discussed here. Subjects were led to believe that they were participating in a personality test with a few additional questions, so they were unaware of the actual DT test. 124 participants indicated their profession as artistic, such as graphic design, fine arts or theatre. 45 indicated another profession. 19 indicated no profession. These were excluded from the experiment.

Creativity in the DT test is assessed as follows for each participant (Guilford, 1977):

| | |
|---|---|
| **ORIGINALITY** | Unusual answers across all answers score `1`, unique answers score `2`. |
| **FLUENCY** | The number of answers. |
| **FLEXIBILITY** | The number of different categories of answers. |
| **ELABORATION** | The amount of detail.<br>For example, "break door" = `0` and "break door with a crowbar" = `1`. |

To estimate flexibility, a categorization of all answers was performed manually by two annotators, with an inter-annotator agreement of `0.76` (Fleiss' kappa). The most common category in our survey is **VIOLENCE**, with instances such as "kick down the door" and "smash it open!" Other frequent categories include **DOORBELL**, **DOORMAT** and **WINDOW**.

In Guilford's DT test, exceptional creativity is represented by the top 1% of all answers. But we are more broadly interested in any answer that scores above average creativity. The lowest score in our setup is `0.0`. The highest score is `16.6`, with an average score of `6.2` and a standard deviation of `4.2` We then looked for outliers, that is, scores higher than average + standard deviation (> `10.4`).

Table 6 shows a breakdown of the results by profession and age.

| Artists | 01–18 | 19–24 | 25–35 | 36–50 | 51–69 |
|---|---|---|---|---|---|
| **CREATIVE** | 5 | 13 | 4 | 1 | 0 |
| **COMMON** | 10 | 62 | 14 | 9 | 6 |

| Other | 01–18 | 19–24 | 25–35 | 36–50 | 51–69 |
|---|---|---|---|---|---|
| **CREATIVE** | 0 | 1 | 6 | 3 | 0 |
| **COMMON** | 1 | 6 | 17 | 7 | 4 |

Table 6. Breakdown of creative vs. non-creative answers by profession and age.

Table 7 shows the results over all age categories:

| Answer | Artists | Other |
|---|---|---|
| **CREATIVE** | 23 | 10 |
| **COMMON** | 101 | 35 |

Table 7. Breakdown of creative vs. non-creative answers by profession (artistic–other).

One of the most creative answers was provided by a generative artist: "Get a bull from the farm nearby, wave a red flag in front of him and jump out of the way at the last given moment." Fisher's exact test for the 2 × 2 summary matrix yields `p=0.004`. This implies a significant correlation between profession and creativity, but not in favor of the artists, since their observed share of creative answers is 23% versus 29% for other professions. Interestingly, of the 45 non-artists, 9 were computational linguists and 8 were software engineers. If we exclude these two groups from the test then `p=0.054` or not significant. Excluding other non-artist groups such as people in accountancy or health care has no effect. But we must be careful to generalize from the results, given the small sample size. If anything, our results suggest a correlation between creativity and domain-specific knowledge (i.e., language) which seems to be backed up by the literature. Since the DT test is primarily a test of verbalizing thought (i.e., it doesn't measure how well you can paint) it follows that individuals proficient in language will perform better.

Realizing this, our further case studies will focus more on creativity in terms of knowledge representation and less on visual output. In chapter 2, we have discussed experiments such as PERCOLATOR that employ simple natural language processing algorithms to achieve visual results. Chapters 5, 6, 7 will focus more on knowledge representation and natural language processing, where visual output is sometimes used to clarify the data but no longer serves as our primary goal. We want to study what happens behind the scenes, how ideas are formed; or the opinions that people have about them.

## 4.7 Discussion

> The wonder takes place before the period of reflection, and (with the great mass of mankind) long before the individual is capable of directing his attention freely and consciously to the feeling, or even to its exciting causes. Surprise (the form and dress which the wonder of ignorance usually puts on) is worn away, if not precluded, by custom and familiarity.
>
> – Samuel Taylor Coleridge, Aids to Reflection (1825)

Creativity refers to the ability to think new ideas. Creative ideas are grounded in fast, unconscious processing such as intuition or imagination (Norris & Epstein, 2011) which is highly error-prone but allows us to "think things without thinking about them". Occasionally, this may feel surprising or promising. Some of these near-thoughts can emerge without warning as an interesting solution: a moment of insight. This usually happens while tackling everyday problems. This is called little-c creativity. However, Einstein did not think up the theory of relativity in a flash of insight while doing the dishes. Big-C creativity, eminent ideas that fill history books, develop gradually. They require interaction with slow, conscious processing (Takeuchi et al., 2011), turning a hunch into a rational hypothesis and mentally "toying" with it: analyzing it, immersing it back into the unconscious and analyzing it again (Koestler, 1964). This requires effort and motivation, because consciousness is lazy and tends to wander off (Kahneman, 2011). In the mind, consciousness in general is rare.

The mind, or memory, can be thought of as a search space of linked information (concepts). A private Wikipedia. Different thinking styles are used to retrieve, relate and combine different information. Flexibility to switch between styles of thought – from unconscious to conscious, from goal-oriented to open-ended, from combinatory to explorative and transformative – is key to creativity: an agile mind.

Interestingly, a concept search space can be modeled with AI techniques. In chapter 5 we will represent a concept search space using a so-called semantic network. We can then traverse the network using search algorithms or agent-based systems, for example. Such models can be used to study how creative ideas originate. This is a step forward over our visual case studies in chapter 2. In the arts, what constitutes art (be it a poem, a painting or a computer game) is generally taken to be grounded in an interesting or unusual thought process, over and before being aesthetically pleasing. An artistic problem (Gombrich, 1960) is concerned with how to depict for example space, movement or expression, to which there can be many possible (visual) solutions and interpretations. So in terms of our case studies in chapter 2, if we want a PERCOLATOR that is taken seriously, we need it to generate its own creative thought processes first, to back up the visual results it consequently produces. To accomplish this we will turn our attention to language. Language is more explicit in representing thought than visual methods (images may be packed with multiple or ambiguous meanings), and as such more useful in our computational approaches.

# 5 Computational creativity

Computational creativity is a multidisciplinary field that studies creativity using techniques from artificial intelligence, and drawing from psychology, cognitive science and the arts. Researchers in this field are interested in the theoretical concerns of creativity as well as developing computer models that exhibit, generate and evaluate creativity. A definition is offered by Wiggins (2006):

> The performance of tasks (by a computer) which, if performed by a human, would be deemed creative.

An early approach include Newell, Shaw & Simon's state space search (1962). Figure 21 shows an example of a state space. Newell, Shaw & Simon first proposed a multipartite AI definition of creative problem solving, where a solution must 1) be novel and useful, 2) reject previously accepted solutions, 3) result from motivation and persistence, and 4) clarify an originally vague problem (Cardoso, Veale & Wiggins, 2009). This can be represented in the search-in-a-state-space paradigm. For example, (2) could define a constraint to avoid pathways with high traffic, and (3) could define the extent of a depth-first search. Such early approaches represented problem-solving as a branching structure of IF A THEN B ELSE C statements (i.e., a decision tree). This is also called GOFAI (Good Old Fashioned AI).

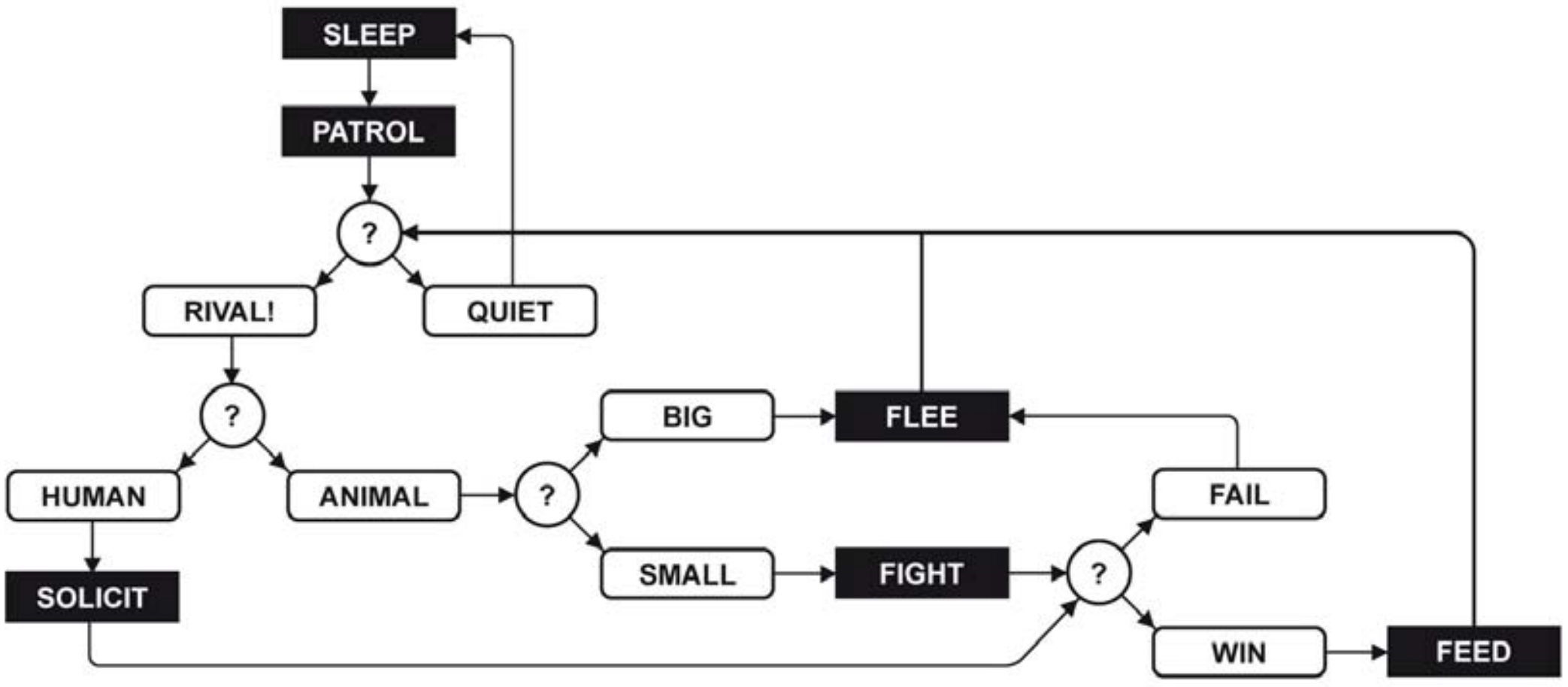

Figure 21. Example state space with a few possible states of a cat in black and the decision tree in white.

Current computational creativity research topics include problem solving, visual creativity, musical creativity and linguistic creativity. Linguistic creativity refers to computer models for language acquisition, language understanding, analogy, metaphor, humor, story generation and poetry. For example, in problem solving Saunders & Gero (2001) use an agent-based approach to model creative societies (Csikszentmihalyi's systems view). In visual creativity, Cohen's AARON and Colton's THE PAINTING FOOL (Colton, 2008) generate paintings in different styles and varying themes. In musical creativity, Cope's EMMY generates music scores based on pattern matching.

In linguistic creativity, Steels (1997) uses agent-based systems to evolve language in robots. COPYCAT (Hofstadter, 1995) is a model of analogy that solves problems such as "abc is to abd as xyz is to what?". SARDONICUS (Veale & Hao, 2007) acquires similes (e.g., "as proud as a peacock") from the Web and then uses these to generate metaphors, for example: brain = smart + box. SAPPER (Veale, O'Donoghue & Keane, 2000) uses a graph-based approach for conceptual blending (e.g., "surgeons are butchers"). RACTER (Chamberlain, 1984) generates poems by randomly filling in the blanks in a text template (e.g., **PERSON** loves **NOUN**).

## 5.1 FLOWEREWOLF: alliterative adjectives

*Tumultuous specular specter!*
*That announces its ungainly ubiquity*
*with R*
*A*
*P*
*P*
*I*
*N*
*G and the risen redaction of deep-sea dysaphia*

– FLOWEREWOLF, Soul

FLOWEREWOLF (De Smedt, 2005) generates poems by manipulating WordNet (Fellbaum, 1998) and by using alliteration. WordNet is a lexical database of English words grouped into synonym sets, with various semantic relations between synonym sets (e.g., hyponymy = **is-specific**). Part of the program is included in chapter 6. For a given noun (e.g., soul), it retrieves a random hyponym from WordNet (e.g., poltergeist) and its description ("a ghost that announces its presence with rapping"). The description serves as a template where each noun is mapped to a synonym (ghost → specter) and a random alliterative adjective is added (specter → specular specter). The fact that "specular" (= relating to the properties of a mirror) pertains well to "soul" is a coincidence. Another option may have been: "immaterial cavalier classifier".

The reason that the rhyme seems so fitting is because our mind will automatically imagine the story between the noun and the poem. This tendency to see meaningful relations (e.g., narrative) in meaningless noise is caused by the priming effect, where the mind is prepared to interpret stimuli according to an expected model (Shermer, 2008). With regard to poetry and the arts, the mind will attribute higher value and credibility to aesthetically pleasing form. This is called the Keats heuristic (McGlone & Tofighbakhsh, 1999). In other words, we *want* to find a meaningful poetry algorithm, because the words are pleasing and presented as a coherent whole, suggesting deliberate expertise. Similarly, it is plausible that we will want to find creativity in the work of artists which we know nothing about but which are highly regarded by others (i.e., famous).

In the remainder of this chapter, we focus on a case study that involves no randomness, although the interpretation of the output may also be subject to the aforementioned effects.

## 5.2 PERCEPTION: semantic network of common sense

Historically, there have been two fundamentally different paradigms in AI: the symbolic approach and the connectionist approach (Minsky, 1991). In the GOFAI symbolic approach, logical sets of rules such as `IF` A `THEN` B are employed for decision-making. In the connectionist approach, nonlinear dynamic systems such as neural networks are used. A neural network is an adaptive network that changes its structure during a learning phase. PERCEPTION (see also De Smedt, De Bleser, Van Asch, Nijs & Daelemans, 2013) uses the GOFAI approach with a connectionist technique (spreading activation) to model associative thinking. Concepts are related to each other in a semantic network (Quillian, 1968, Sowa, 1987). For example, a *doll* concept can be modeled as: *doll* **is-a** *toy*, *silent* **is-property-of** *doll*, *doll* **is-related-to** *girl*. A semantic network is a specific type of graph data structure. A graph is a network of nodes and edges (connections between nodes). Each node represents a concept (e.g., *doll*) and each edge has a type (e.g., **is-part-of**). Computational techniques from graph theory can be used to find important nodes (centrality) or paths between the nodes (shortest path).

PERCEPTION stores knowledge about what things look and feel like. The database has about 9,000 manually annotated relations of the types **is-a**, **is-part-of**, **is-opposite-of**, **is-property-of**, **is-related-to**, **is-same-as** and **is-effect-of** between 4,000 mundane concepts (*cat*, *doll*, *pasta, rain*, and so on). Relation types are distributed across 10 contexts. A portion is uncategorized. Table 8 shows a general breakdown. The data is available by means of an online visualizer[18] where new relations can be added to the semantic network. The visualizer shows the network using a force-directed algorithm (Hellesoy & Hoover, 2006). Following is a case study for comparing and shifting concepts in the semantic network, based on how the concepts look and feel. The analogies that PERCEPTION can generate are limited by the knowledge representation (i.e., 9,000 relations), since conceptual metaphor is a knowledge-hungry phenomenon (Veale & Hao, 2008). Further discussion will follow at the end of this chapter.

| | is-a | is-part-of | is-opposite-of | is-property-of | is-related-to | is-same-as |
|---|---|---|---|---|---|---|
| CULTURE | 573 | 491 | 25 | 1,060 | 1,060 | 74 |
| NATURE | 673 | 222 | 10 | 640 | 364 | 31 |
| PROPERTIES | 29 | 5 | 98 | 303 | 258 | 35 |
| PEOPLE | 340 | 44 | 3 | 80 | 91 | 13 |
| GEOGRAPHY | 233 | 292 | 0 | 36 | 11 | 3 |
| MEDIA | 109 | 75 | 2 | 148 | 143 | 5 |
| HISTORY | 33 | 35 | 1 | 32 | 86 | 9 |
| EMOTION | 28 | 18 | 3 | 66 | 72 | 3 |
| SOUND | 24 | 23 | 0 | 44 | 29 | 1 |
| SCIENCE | 12 | 26 | 1 | 33 | 38 | 4 |

Table 8. Distribution of relations by context. The PROPERTIES context is reserved for inter-adjective relations such as *slow* **is-property-of** *bored.* The **is-effect-of** relation is omitted; it contains very little data for now.

[18] http://www.nodebox.net/perception/?q=rocket

If a rocket is *fast* and *hot*, then what colors evoke speed and heat and might be suitable for a rocket toy? Since there is no right or wrong answer, this problem requires a divergent approach. A system such as PERCEPTION can help artists or designers discover a feasible solution to these so-called wicked problems – assignments in a social context with no clear solution (Rittel & Webber, 1973, Buchanan, 1992). If the toy company hires an artist to design a toy rocket, they could criticize her sketches in many ways (e.g., it looks a little menacing because too realistic). But probably not because the rocket design has bright red and yellow colors, which can be linked to heat, power, the sun, intensity, courage, and so on. The design choice of using red and yellow colors is a credible step towards a solution.

## Semantic network

The PATTERN package discussed in chapter 6 has a `Graph` object from which we can start:

```
from pattern.graph import Graph

g = Graph()
g.add_edge('doll', 'toy', type='is-a') # doll is-a toy
g.add_edge('silent', 'doll', type='is-property-of')
g.add_edge('doll', 'girl', type='is-related-to')

node = g['doll']
print node.id
print node.links
```

See: http://www.clips.ua.ac.be/pages/pattern-graph

The node's `id` is "doll". The node's `links` is a list of directly related nodes: `[Node(id='toy'), Node(id='silent'), Node(id='girl')]`. But we don't want to define all the relations by hand; the PERCEPTION database is included in PATTERN as a CSV-file that we can easily import:

```
from pattern.graph import Graph
from pattern.db import CSV

g = Graph()

data = 'pattern/graph/commonsense/commonsense.csv'
data = CSV.load(data)
for concept1, relation, concept2, context, weight in data:
    g.add_edge(
        concept1,
        concept2,
          type = relation,
        weight = min(int(weight) * 0.1, 1.0))
```

See: http://www.clips.ua.ac.be/pages/pattern-db#datasheet

In order to compare concepts, we will use a combination of graph theoretic methods: spreading activation, taxonomies, subgraph partitioning, betweenness centrality and shortest path finding.

## Concept halo

"What is a cat?" To answer this kind of question, we need to know how concepts relate to *cat* in the semantic network. Each concept is surrounded by related concepts that reinforce its associative meaning: *cat* → *purr*, *cat* → *mouse*, *furry* → *cat*, and so on. The commonsense halo (Hofstadter, 1985) is the concept itself, its directly related concepts, concepts related to those concepts, and so on, as deep as the representation requires. This is called spreading activation (Collins & Loftus, 1975, Norvig, 1987). When asked about mice, people can readily come up with closely related concepts such as cats and cheese, but it takes longer to come up with less obvious concepts (e.g., gene) because there are more intermediary steps in the semantic chain. Activation spreads out from the starting concept in a gradient of decreasing relatedness.

Spreading activation can be implemented simply using the `Node.flatten()` method:

```
def halo(node, depth=2):
    return node.flatten(depth)
```

The `depth` of the halo is an arbitrary choice. If it is too small, the concept will be underspecified. If it is too broad the halo will include irrelevant concepts. We use `depth=2`. In this case the halo is already rather broad: `halo(g['cat'])` yields over 350 concepts including *elegant*, *furry*, *tail*, *milk*, *sleep*, *explore*, but also *quantum theory* and, oddly, *french fries*. But this is beneficial: there will be more salient properties in the halo.

## Concept field

"What are creepy-looking animals?" To answer this kind of question we need to be able to compare each specific animal to all other animals. Some concepts belong to the same class or taxonomy. Cats, dogs, fish and squirrels are examples of animals – essentially all concepts with an implicit or explicit **is-a** relation to *animal* in the semantic network. Such a collection is also known as a semantic field (Brinton, 2000), a set of words that share a common semantic property, related to hyponymy but more loosely defined. So we need a `field()` function that returns a list with `['cat', 'dog', ...]` when given `'animal'` and sort it by creepiness.

Retrieving such a list is not hard to accomplish using a combination of `Graph` and `Node` methods. The `Node.flatten()` method returns a list containing the given node (`depth=0`), any nodes connected to it (`depth=1`), and so on. We can supply a filter function to restrict which edges are traversed. In our case, we only follow **is-a** relations. The `Graph.fringe()` method returns a list of nodes with a single edge (`depth=0`), any nodes connected to these nodes (`depth=1`), and so on. A combination of the two methods yields the "outer rim" of a node's taxonomy, which is a good approximation of our semantic field:

```
def field(node, depth=3, fringe=2):
    def traversable(node, edge):
        return edge.node2 == node and edge.type == 'is-a'
    g = node.graph.copy(nodes=node.flatten(depth, traversable))
    g = g.fringe(depth=fringe)
    g = [node.graph[n.id] for n in g if n != node]
    return g
```

## Concept properties

"How do cats and dogs compare?" For one, they both have a tail. Both of them are also furry, and they are both kept as pets. This is a featural approach to similarity (Tversky, 1977): comparing concepts by looking at the features or properties that they have in common. There are other measures of similarity, but in PERCEPTION we will use the feature approach because it is simple to implement. In the semantic network, some concepts (typically adjectives) are properties of other concepts. They describe what something looks or feels like: *romantic* **is-property-of** *France*, *fast* **is-property-of** *rocket*, *dark* **is-property-of** *evil*, *evil* **is-property-of** *Darth Vader*, and so on. If we can find the properties (e.g., romantic, fast, dark, evil) that define each concept we can construct an algorithm to compare them.

First, we store all the left-hand concepts that occur in **is-property-of** relations in a `PROPERTIES` dictionary. Dictionaries in Python are faster for lookup than lists:

```
PROPERTIES = [e.node1.id for e in g.edges if e.type == 'is-property-of']
PROPERTIES = dict.fromkeys(PROPERTIES, True)
```

We can then implement a `properties()` function that, given a concept, extracts its latent properties from the concept halo. Note how we sort the results by betweenness centrality (Brandes, 2001) using `Node.centrality` (= a value between `0.0` and `1.0`). This means that properties more central in the halo will appear first in the list:

```
cache = {} # Cache results for faster reuse.

def properties(node):
    if node.id in cache:
        return cache[node.id]
    g = node.graph.copy(nodes=halo(node))
    p = (n for n in g.nodes if n.id in PROPERTIES)
    p = reversed(sorted(p, key=lambda n: n.centrality))
    p = [node.graph[n.id] for n in p]
    cache[node.id] = p
    return p
```

In PERCEPTION, an animal that is semantically similar to a sword would be a hedgehog, since: *sharp* **is-property-of** *sword*, and also *sharp* **is-property-of** *spine* **is-part-of** *hedgehog*. The hedgehog's prickly spines are used as an associative bridge. This is a fine observation as it is, but what we may have wanted was a fairy-tale dragon. Whether or not a concept appears in a fairy tale is not expressed by **is-property-of** relations, so this will never happen. The system uses a script (Schank & Abelson, 1977) to explore the network. A script is a fixed sequence of commands for decision-making. In the `similarity()` function given below, the use of `properties()` is hard-coded.

This does not pertain very well to the agility of the mind, but it makes the algorithm easier to study. We can see how it would not be hard to adapt the code to incorporate (or evolve) other scripts besides our featural approach, by adapting the `properties()` function.

## Concept similarity

Similarity between two concepts is measured as the distance between their properties. The `similarity()` function retrieves the *k* most central properties in each concept's halo. It then measures the length of the shortest path (Dijkstra, 1959) connecting each two properties, preferring to traverse **is-property-of** relations over other relations. This yields a higher value for more similar concepts (lower value for distinct concepts):

```
def similarity(node1, node2, k=3):
    g = node1.graph
    h = lambda id1, id2: 1 - int(g.edge(id1, id2).type == 'is-property-of')
    w = 0.0
    for p1 in properties(node1)[:k]:
        for p2 in properties(node2)[:k]:
            p = g.shortest_path(p1, p2, heuristic=h)
            w += 1.0 / (p is None and 1e10 or len(p))
    return w / k
```

We can then use `similarity()` to implement a one-versus-all search:

```
def nearest_neighbors(node, candidates=[], k=3):
    w = lambda n: similarity(node, n, k)
    return sorted(candidates, key=w, reverse=True)
```

"What are creepy-looking animals?" For a given concept (e.g., creepy) and a list of candidates (e.g., animals), `nearest_neighbors()` yields the candidates with the highest similarity (the creepiest animals). In this particular example, it will suggest animals such as octopus, bat, owl, tick, spider, crow, ... No fluffy bunnies or frolicking ponies there!

```
print nearest_neighbors(g['creepy'], field(g['animal']))
```

Octopus has a direct *creepy* **is-property-of** *octopus* relation in PERCEPTION, so it is the obvious winner. However, things get more interesting when we look at the suggested bat. There is no *creepy* property for the *bat* concept. Instead, it is annotated with a *black* property and other relations to *cave*, *evil*, *night* and *radar*. It is associated to *creepy* by inference. More specifically, the system judges that the bat is a dark thing; and that dark is pretty creepy. Now where does dark come from? The direct relations of *bat* lead further to other concepts such as *Darth Vader* (via *black* and *evil*), *dark*, *dangerous*, *pessimistic*, *airplane*, *sky*, and so on. All of these concepts together make up the bat halo. Commonsense halo is explained in Chalmers, French and Hofstadter (1991), which argues the flexibility of human high-level perception – where objects and situations can be comprehended in many different ways depending on the context. The work includes an interesting quote by the late nineteenth century philosopher William James:

> There is no property ABSOLUTELY essential to one thing. The same property which figures as the essence of a thing on one occasion becomes a very inessential feature upon another. Now that I am writing, it is essential that I conceive my paper as a surface for inscription. [...] But if I wished to light a fire, and no other materials were by, the essential way of conceiving the paper would be as a combustible material. [...] The essence of a thing is that one of its properties which is so important for my interests that in comparison with it I may neglect the rest. [...] The properties which are important vary from man to man and from hour to hour. [...] Many objects of daily use—as paper,

> ink, butter, overcoat—have properties of such constant unwavering importance, and have such stereotyped names, that we end by believing that to conceive them in those ways is to conceive them in the only true way. Those are no truer ways of conceiving them than any others; there are only more frequently serviceable ways to us.

The bat halo is a flexible representation of the concept *bat*: something dangerous, flying in the sky, reminiscent of a character named Darth Vader, associated with pessimistic things, resembling an airplane, and so on. The halo includes many latent properties, none of which are very merry: *bad, black*, *brown*, *dark*, *deep*, *evil*, *negative*, *sad*, and so on. The most salient ones measured by betweenness (i.e., amount of traffic passing through) are *black*, *evil* and *dark*. They are a kind of conceptual glue that the system will use to reason about bats. For $k$=3, it will measure the distance of these three to *creepy* using Dijkstra's shortest path algorithm. The shorter paths are preferred, imitating the same cognitive bias in human intuition. The total "nearness" score is an indicator of the bat's creepiness. This approach will yield many possible solutions (i.e., divergent thinking) besides *bat*. As pointed out by Hofstadter, "making variations on a theme is really the crux of creativity". We can demonstrate the soundness of the approach by performing a search for happy animals instead of creepy animals. This will yield animals such as grasshopper, puppy, dog, chameleon, dolphin, kitten, and so on. These are certainly distinct from the creepy ones (we can argue about the chameleon).

The concept halo is important in another regard: computational tractability. It is intractable to calculate betweenness centrality for all concepts in large semantic networks as a whole: $O(nm)$ for $n$ concepts and $m$ relations. This would be the equivalent of "seeing" your mind as a whole.

## Brussels, the toad

As a thought experiment, suppose we want to create an advertising campaign to promote Brussels, the capital of the European Union. How can the system pick, for example, a striking image or photograph? We want something a little bit more thought-provoking than retrieving `brussels.jpg` from the Web. In the words of Veale, Feyaerts & Forceville: we want something that "compresses multiple meanings into a single form". Using the `nearest_neighbors()` function, we can shift the context of one concept to another concept as an exercise in combinatorial creativity. Unfortunately, PERCEPTION has no *Brussels* concept. But we can annotate the semantic network by mining the Web with the tools in PATTERN:

```
from pattern.web import Google, plaintext
from pattern.search import search

def learn(concept):
    q = 'I think %s is *' % concept
    p = []
    g = Google(language='en')
    for i in range(10):
        for result in g.search(q, start=i, cached=True):
            m = plaintext(result.description)
            m = search(q, m) # use * as wildcard
            if m:
                p.append(m[0][-1].string)
    return [w for w in p if w in PROPERTIES]
```

The `learn()` function returns a list of known properties for a given concept, mined from Google's search engine with an "`I think * is *`" query. This is adapted from a technique for finding similes (`as * as *`) described by Veale & Hao (2007). In this particular instance, the results are: *expensive*, *great* (2x), *green* and *proud*. We update the semantic network on the fly:

```
for p in learn('Brussels'):
    g.add_edge(p, 'Brussels', type='is-property-of')
```

Now we can do:

```
print nearest_neighbors(g['Brussels'], field(g['animal']))
```

The top three nearest neighbors yields *stag*, *frog* and *toad*, where *toad* is our personal favorite. Figure 22 shows the subnet of properties that connect *Brussels* to *toad*. The shortest paths that connect them include: *proud* → *confident* → *calm* and *great* → *pleasant* → *lazy* → *slow*. Figure 23 shows an artistic photograph of a toad.

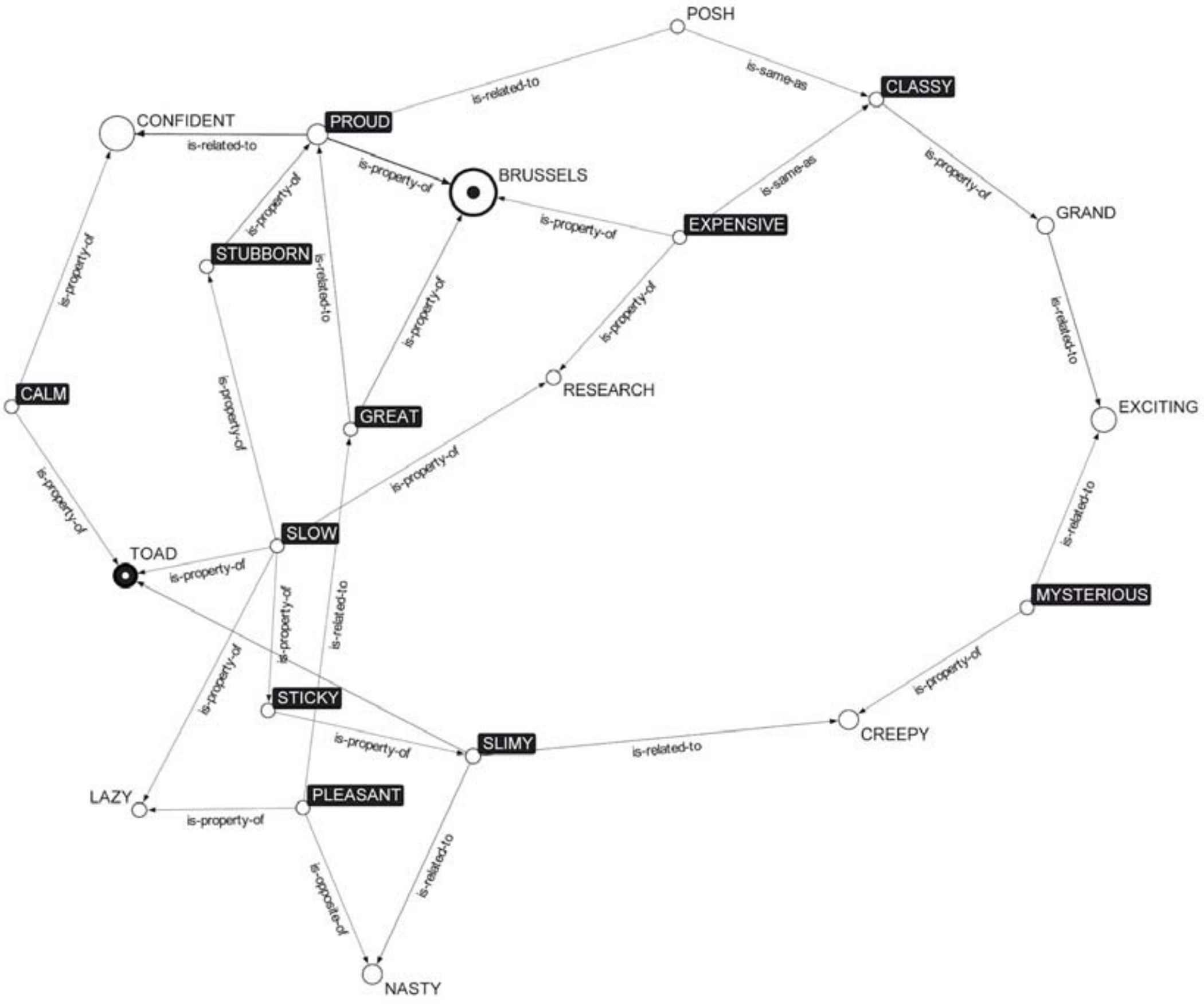

Figure 22. Subnet of properties that connect Brussels to a toad.
The shortest path is *Brussels* → *proud* → *confident* → *calm* → *toad*.

Let's try another one with comparisons to persons:

```
print nearest_neighbors('Brussels', field('person', 2, 1))
```

We tweaked the parameters of `field()` to exclude general concepts such as *brother* and *engineer*. The nearest neighbors then yield *Abraham Lincoln*, *Al Capone* and *Jerry Springer*. The path to *Abraham Lincoln* is short and strong: through *proud* as well as *great*. The shortest path to Al Capone is *expensive* → *drug* → *dangerous*. The shortest path to *Jerry Springer* switches tone: *proud* → *good* → *evil* → *creepy*. Let's avoid a lawsuit and stick with an image of a toad. Hopefully the EU committee funding our marketing campaign has a sense of humor.

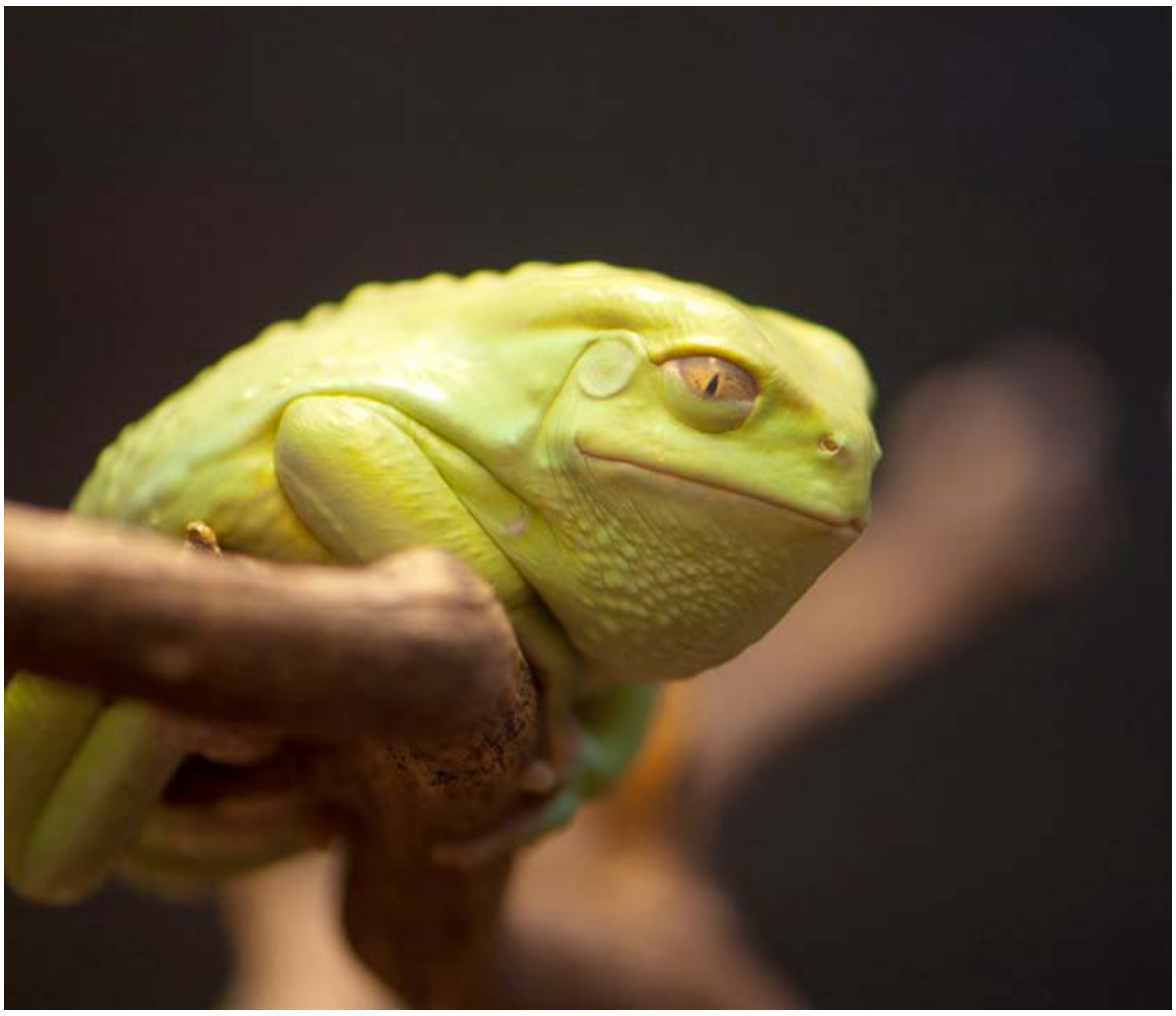

Figure 23. SMILING GREEN TOAD. Photography © Sarah Crites. Used with permission.

## Novelty assessment

As discussed in chapter 4, creative ideas are novel as well as appropriate. One approach to measure the novelty of PERCEPTION's suggestions is to count the number of web pages that mention each suggestion. It follows that suggestions that are less often mentioned are more novel.

Brussels + stag yields `25,000` results. Brussels + toad yields `26,000` and Brussels + frog yields `125,000`. By comparison, Brussels + waffles yields `1,700,000` and Brussels + sprouts yields `8,100,000`. Since frogs score about 5× more results than stags and toads, we should probably discard the analogy as too common. In the sense of Boden's distinction between H-creative and P-creative ideas, all solutions are P-creative since no solution scores `0` (i.e., historically novel). See Veale, Seco & Hayes (2004) on compound terms for further discussion.

```
from pattern.web import Google

def novelty(ideas=[]):
    candidates = [Google().search(idea) for idea in ideas]
    candidates = sorted(candidates, key=lambda results: results.total)
    candidates = [(results.query, results.total) for results in candidates]
    return candidates

print novelty(['Brussels stag', 'Brussels frog', 'Brussels toad'])
```

Assessing appropriateness is a much harder task. For example, in terms of a scientific innovation, appropriateness pertains to performance, production cost, scalability, undesired side effects, and so on. Entire fields such as process management, risk management and workflow optimization have been devoted to such tasks. In terms of an artistic idea, the solution is often formulated to solve a wicked problem for which there are many possible solutions. In this case, appropriateness is usually assessed by peers. Chapter 7 discusses a technique for evaluating the opinion of peers. PERCEPTION has no feedback loop to self-evaluate how and why it evaluated that Brussels has something in common with stags, frogs and toads.

## Biased search

As noted, it is not hard to adapt the `similarity()` function to simulate other approaches besides the featural approach. For example, we could introduce cognitive bias by diverting the paths between properties via a fixed set of concepts. A fixation. Some of the results will be irrational now but that is what we want of course. The following lines of code compute `p` in the `similarity()` function with a fixed deviation via the *sad* concept, perhaps not unlike a depression:

```
p  = g.shortest_path(p1, p2, heuristic=h)
p += g.shortest_path(p1, g['sad'], heuristic=h)
p += g.shortest_path(p2, g['sad'], heuristic=h)
```

A search for creepy animals then yields *grizzly*, *wolf*, *bat*, *raven* and *mole*. A search for happy animals yields *puppy*, *grasshopper*, *seal*, *reindeer*, *wolf*, and oddly, *meat*. The preferred colors for a rocket are *black* and *blue*. Some of the original results still bubble up, but with a gloomy tone it seems. For Brussels it yields *mockingbird*.

Whether or not this is how cognitive bias really works in the human mind is debatable; in fact it is debatable that this is how the mind works at all. But we cannot ignore that PERCEPTION is a creative step forward over the random juggling of pixel effects employed by the PERCOLATOR program discussed in chapter 2. By contrast, PERCEPTION uses a flexible and deterministic strategy based on comparing and blending ideas. The idea that Brussels is a toad or a stag is a little-c new idea. Finally, the bias approach discussed above is interesting for further experimentation – for example to model personality, which is a hallmark of human creativity as pointed out in chapter 4.

## Network visualization

To answer our initial problem, what colors does the system suggest for a rocket?

```
print nearest_neighbors(g['rocket'], field(g['color']))
```

It yields *yellow* and *red*: warm, energetic colors.

The functions in the `graph` module in PATTERN are identical to those in the `graph` module in NODEBOX FOR OPENGL. This way, we can combine the two implementations to visualize the network, using a force-directed algorithm to place the nodes on a 2D canvas. For example, figure 24 shows a visualization of the *rocket* concept halo:

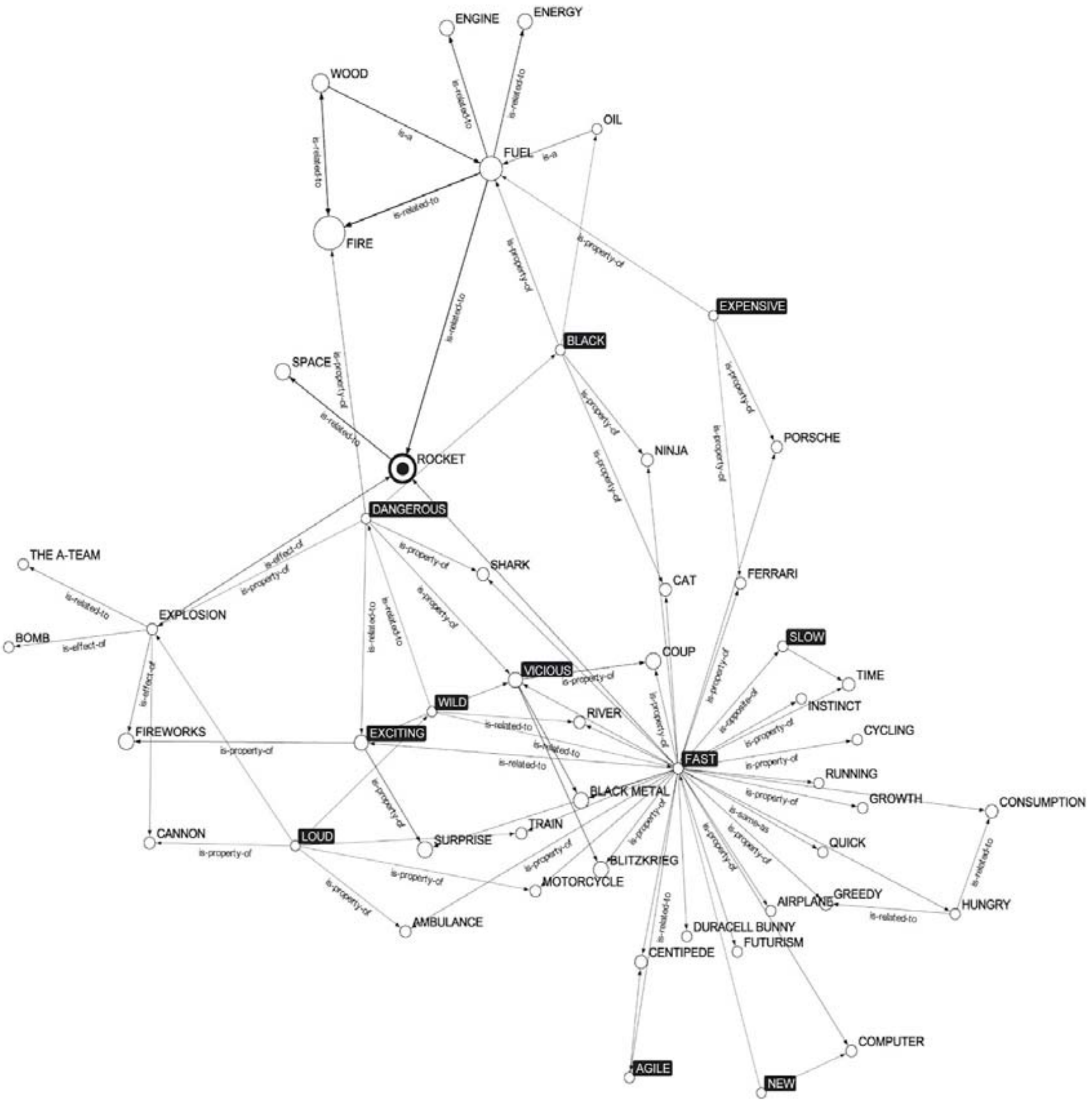

Figure 24. Visualization of the *rocket* halo in NODEBOX FOR OPENGL, with highlighted properties.

In Python code (simplified implementation):

```
from nodebox.graphics import *
from nodebox.graphics.physics import Graph as NodeBoxGraph

g1 = g.copy(nodes=halo(g['rocket']))
g2 = NodeBoxGraph()
for e in g1.edges:
    g2.add_edge(e.node1.id, e.node2.id)

def draw(canvas):
    canvas.clear()
    translate(canvas.width / 2, canvas.height / 2)
    g2.update()
    g2.draw()

canvas.draw = draw
canvas.run()
```

## 5.3 Discussion

PERCEPTION is a computer model of Boden's concept search space. We have used it to model a heuristic thinking style, based on concept properties to compare similar concepts. This can be seen as a combinatorial approach to creativity where two distinct concepts (e.g., Brussels and a toad) are blended together. Computer models can help us understand how creativity works. However, this does not necessarily mean that the mind *is* a desktop computer or vice versa. The debate is old and elaborate. In the 1960's, AI pioneer Simon claimed that "machines will be capable, within twenty years, of doing any work a man can do." Minsky, co-founder of the MIT AI Lab, agreed that "within a generation [...] the problem of creating artificial intelligence will substantially be solved." There was a general optimism towards strong AI, real machine consciousness and understanding, as opposed to weak AI, the use of computer models to *simulate* understanding. Minsky worked as a consultant for the HAL 9000 computer[19] in Kubrick & Clarke's movie *2001: A Space Odyssey*. It shows what the early pioneers actually believed what 2001 would be like.

At that time, the prevalent approach to AI was symbolic. But we can see how adding more `IF`'s to a program does not make a machine intelligent. First, the machine is clueless about what it is processing. The Chinese Room argument (Searle, 1999) offers an engaging thought experiment:

> Imagine a native English speaker who knows no Chinese locked in a room full of boxes of Chinese symbols (a data base) together with a book of instructions for manipulating the symbols (the program). Imagine that people outside the room send in other Chinese symbols which, unknown to the person in the room, are questions in Chinese (the input). And imagine that by following the instructions in the program the man in the room is able to pass out Chinese symbols which are correct answers to the questions (the output). The program enables the person in the room to pass the Turing Test for understanding Chinese but he does not understand a word of Chinese.

Second, a program is always one `IF` short: an inconsistency overlooked, a minor detail it doesn't handle yet. For a discussion of incompleteness, see Gödel's theorem, the Lucas-Penrose argument and the ensuing debate (Gillies, 1996). Understanding implies that a machine would be aware of its shortcomings: evaluate, and then learn and adapt its programs by itself.

[19] http://mitpress.mit.edu/e-books/Hal/chap2/two3.html

When PERCEPTION yields a clever response, is it inherently clever or does it only appear to be so? The idea of a logo of a stag as "the proud flagship to represent the EU capital" is certainly promising, but this description is our own interpretation (priming effect) and beyond the limitations of PERCEPTION. Searle's thought experiment in this regard has become so widely debated that some have claimed that cognitive science should be redefined as "the ongoing research program of showing Searle's Chinese Room argument to be false" (Harnad, 2001). Hofstadter has criticized Searle's argument as a "religious diatribe against AI masquerading as a serious scientific argument". There is more debate about the human mind as an information processing system, for example in *How The Mind Works* (Pinker, 1999), its rebuttal entitled *The Mind Doesn't Work That Way* (Fodor, 2000), and consequently Pinker's riposte *So How Does The Mind Work?* (Pinker, 2005).

We argue that PERCEPTION can be as clever as you are, given that you had the same limited amount of knowledge to work with, a dull personality, and only one thinking style. There is no magic property of creativity that would suggest otherwise. However, unlike you, our model does not have a conscious feedback loop to reflect on Big-C creative ideas, nor the flexibility to adopt different thinking styles and neither is it intrinsically motivated to do so (cfr. chapter 4.2 on motivation). These shortcomings pertain to an inability to evaluate and adapt. Knowledge, in the form of new concepts and relations in the semantic network, must be supplied by human annotators, either by hand or using a `learn()` function as discussed in the case study. We can refine the `learn()` function into an unsupervised, endless learning mechanism. If "`I think Brussels is *`" yields *great*, the search could then continue with "`I think * is great`", and so on. We can also refine the search heuristics to incorporate more search paths between concepts. In both cases the question of evaluation is raised. What is useful new knowledge and what is not? What are interesting new search paths, and which are not? So far we have focussed on generating (visuals + ideas). We will now direct our attention to learning and evaluating.

Computer algorithms for learning and evaluating are a well-researched discipline. Modern AI approaches have generally turned away from pursuing strong AI, in favor of statistical methods for practical learning and evaluation tasks in for example data mining and natural language processing (Russell & Norvig, 2003). In chapter 6, we discuss these topics in the context of computational linguistics. This is interesting, since it continues along the line of what we started with PERCEPTION: language as a model for representing and learning ideas, linguistic analogy as a hallmark of creativity. The advantage of statistical methods is that they are verifiable, whereas the heuristic approach in PERCEPTION can be subject to interpretation bias. Chapter 6 will first devote some time to explaining the statistical approaches and will then focus on two software packages for data mining, natural language processing and learning: MBSP FOR PYTHON and PATTERN. These packages have been designed with chapter 4 in mind. Each package consists of submodules that can be creatively combined to solve many different tasks, either among themselves or chained together, e.g., MBSP + PATTERN for robust natural language processing, or PATTERN + NODEBOX FOR OPENGL for network visualization.

# Part III

# LANGUAGE

•

The phonetician whispered
expressions of affection
honeyed words
into her ear:

"i n c o n s e q u e n t i a l i t h o g r a p h e m e"

– FLOWEREWOLF, Language (edited)

# 6 Computational linguistics

Natural language is not unlike playing with toy blocks. In Lego for example, there is a finite set of different blocks to combine. In language, there is a finite alphabet of characters. Characters (or phonemes in spoken language) can be combined into morphemes, and morphemes make up words, the basic units that carry meaning. Words make up sentences that express a statement, a question, a command or a suggestion. Just as one toy block may fit on some blocks but not on others, words in a sentence can be described by a set of rules called a grammar (Chomsky, 1957). Sentences make up a text used to communicate an idea. Arguably, language is one of our most creative spontaneous inventions. It is recombinable, flexible and adaptive. Numerous variations have evolved and still are evolving: English, German and Dutch (Germanic languages), Spanish, Portuguese and French (Romance languages), Arabic, Chinese, Hebrew, Hindi, Russian, Swahili, and so on. Some languages gradually die out as speakers shift to another language, or as the language splits into daughter languages (e.g., Latin → Romance languages). Many derivatives have been developed in the form of formal languages, programming languages and constructed languages (e.g., Esperanto).

In the human brain, language is processed in different areas, most notably Wernicke's area and Broca's area (Geschwind, 1970) in the cerebral cortex. Language is a window to the mind (Pinker, 2007). Since each individual's mind is different, language (metaphor in particular) underpins our ability to communicate and empathize with others, or to deceive them. A brain imaging study by Bedny, Pascual-Leone & Saxe (2009) on blind adults shows that the way in which humans reason about other humans does not rely on visual observation. Instead, research by Pyers & Senghas (2009) shows that the capacity to understand the beliefs or feelings of others relies on language over and above social experience.

Computational linguistics involves the theoretical study of language as well as the development of computer algorithms to process language, and ultimately to extract meaning and understanding. Figure 25 shows SCANNING–PARSING–UNDERSTANDING (Lechat & De Smedt, 2010), a visual metaphor that represents some chosen aspects of computational linguistics. Part of the artwork was generated in NODEBOX. It illustrates the process of extracting information from raw data (e.g., finding words in a string of characters), identifying word types (e.g., nouns, verbs, adjectives) and deriving meaning from the relations between words. Derived meaning can take different forms. For example, it might reveal what the author of a text is saying about a certain product (e.g., in an online review), or it might reveal psychological information about the author himself, based on his or her language use.

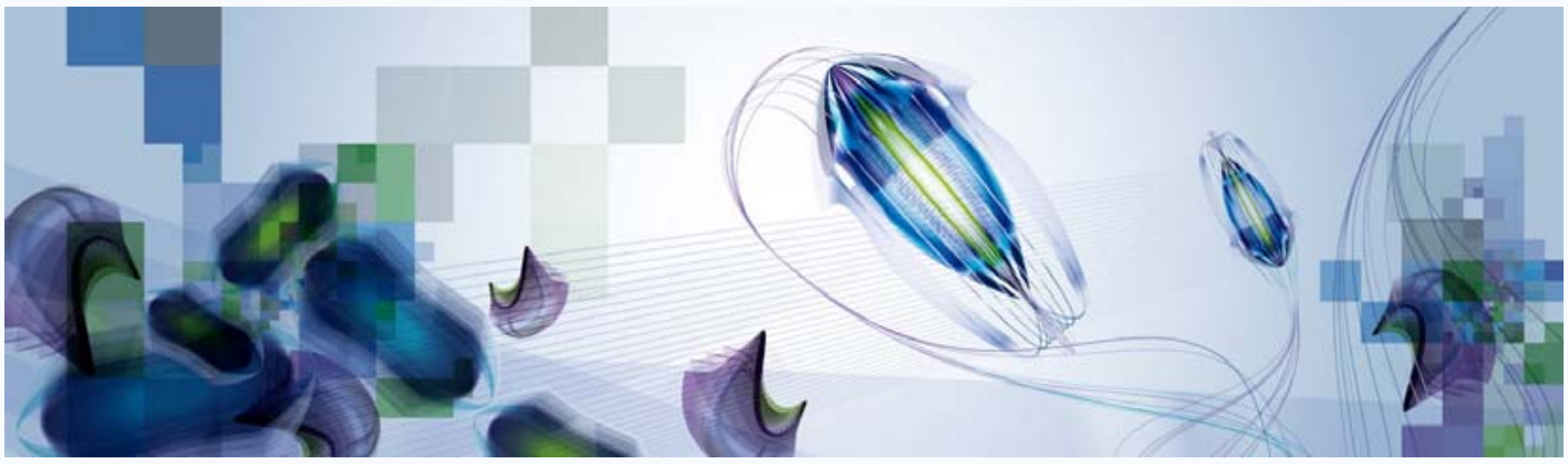

Figure 25.1: SCANNING: raw data is represented as a jumble of cell-like elements. The flowers in the center represent an algorithm that is scanning the data, identifying the words and passing them on one by one.

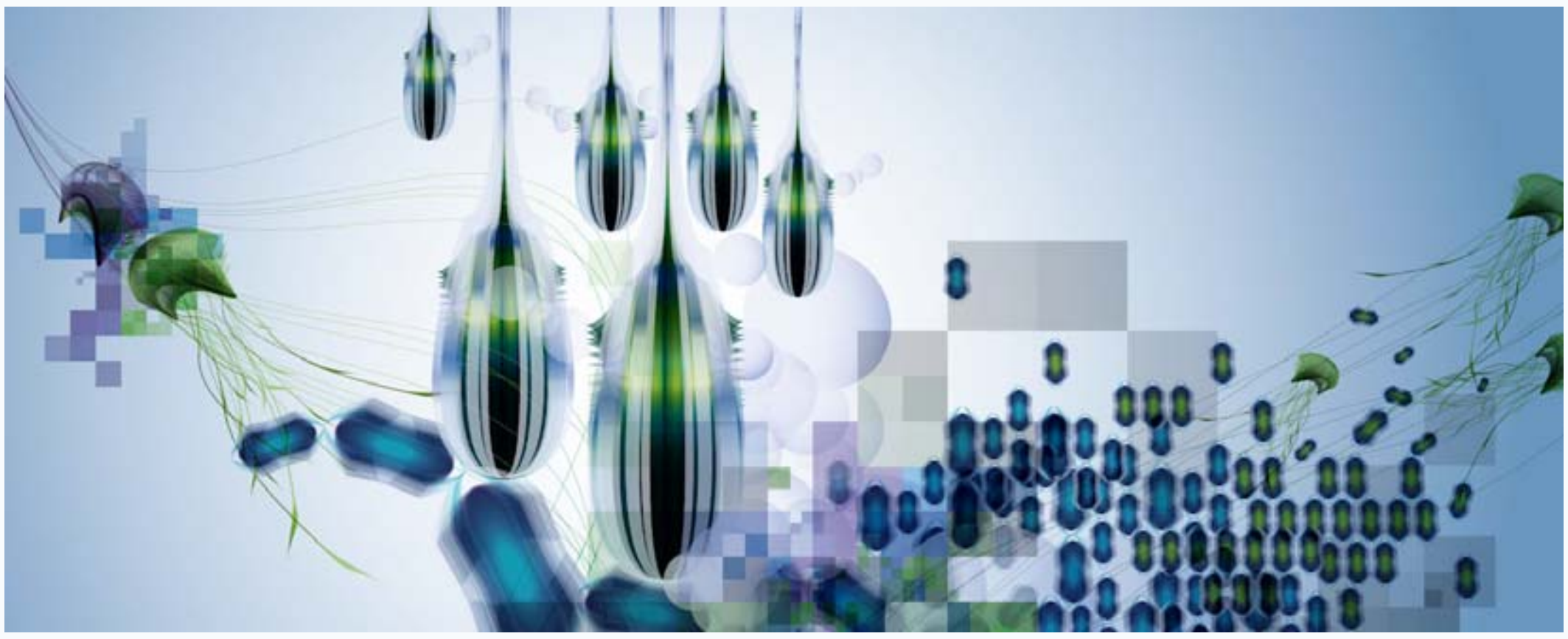

Figure 25.2: PARSING: the central cluster of shapes represents a parser. It appears to grind down on the cells, after which they are organized in a loose grid and assigned different colors, representing different word types.

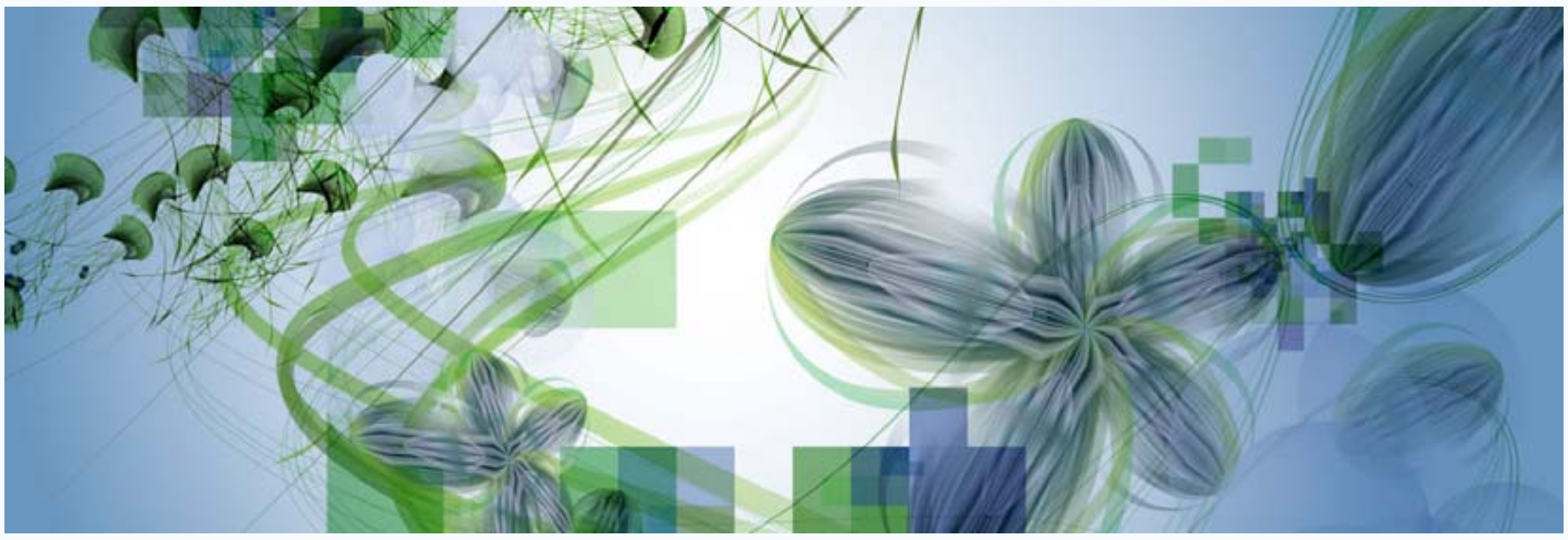

Figure 25.3: UNDERSTANDING: the cells become more complex in form as meaning is mined from them.

#### AMBIGUITY

Words can express multiple concepts depending on the context in which they are used. For example, “dash” can mean “crash”, “crush”, “rush”, “panache” or “–” depending on how it is used. This kind of ambiguity does not seem very efficient, but in fact it is: it makes a language more compact and allows us to communicate faster (Piantadosi, Tily & Gibson, 2012). On the other hand, different words can also express almost the exact same concept. For example, the verbs “rush”, “hurry”, “scurry”, “run” and “gallop” are all synonymous. This allows us to communicate more subtly, accurately or expressively.

#### PARTS OF SPEECH

Words fall into different categories. Nouns correspond to persons, objects or concepts. They can function as the subject of a sentence or the object of a verb or preposition. Verbs convey actions and events. In many languages, verbs inflect to encode tense or gender (e.g., dash → dashed). Adjectives qualify nouns. They correspond to properties such as size, shape, color, age, quantity and quality (e.g., dashing). Adverbs qualify adjectives and verbs (e.g., very dashing). Pronouns substitute previously mentioned or implicit nouns (e.g., it, he, his). Prepositions indicate a relation with the preceding or successive words (e.g., of, to, in). Conjunctions join two words, phrases or sentences (e.g., and, or). Nouns, verbs, adjectives, adverbs, pronouns, prepositions and conjunctions are the basic word classes or parts of speech in English, together with determiners (e.g., the) and interjections (e.g., uh). Word classes such as nouns, verbs and adjectives are open classes. They are constantly updated with new words borrowed from other languages or coined to describe a new idea (e.g., googling). Word classes such as pronouns, prepositions, conjunctions and determiners are closed. They are also known as function words and commonly used to establish grammatical structure in a sentence. In “the slithy toves gyred and gimbled in the wabe” it is not difficult to derive some meaning even if the nouns, adjectives and verbs are nonsensical. Word suffixes offer valuable clues: `-y` for adjectives, `-ed` for verbs. But nonsensical function words make the sentence incomprehensible: “thove mighty dove geared gard trembled tine thove waves” carries no meaning.

#### PART-OF-SPEECH TAGS

In natural language processing, the parts of speech are usually denoted with a short tag[20]: **NN** for nouns, **VB** for verbs, **JJ** for adjectives, **RB** for adverbs, **PR** for pronouns, **IN** for prepositions, **CC** for conjunctions and **DT** for determiners (Marcus, Santorini & Marcinkiewicz, 1993). For example:

| The | stuffed | bear | sported | a | dashing | cravat | with | polka | dots |
|---|---|---|---|---|---|---|---|---|---|
| **DT** | **JJ** | **NN** | **VB** | **DT** | **JJ** | **NN** | **IN** | **NN** | **NN** |

Table 9. Common part-of-speech tags in an example sentence.

[20] http://www.clips.ua.ac.be/pages/penn-treebank-tagset

## 6.1 Natural language processing

Natural language processing or NLP (Jurafsky & Martin, 2000) is a subfield of AI and linguistics that overlaps with related fields like text mining (Feldman & Sanger, 2007), text categorization (Sebastiani, 2002) and information retrieval (Manning, Raghavan & Schütze, 2009). Natural language processing is concerned with the structure of text and algorithms that extract meaningful information from text. Meaningful information includes objective facts, subjective opinions and writing style.

### Objective facts

"Who did what to whom where and when?" An important NLP task is part-of-speech tagging: identifying the parts of speech of words in a sentence. The task is complicated by ambiguous words that fall into multiple categories depending on their role in the sentence, e.g., "dashing" as a verb or an adjective. Consecutive words can be chunked into phrases based on their parts of speech. For example, "the stuffed bear" (who?) is a noun phrase: **DT** + **JJ** + **NN** = **NP**. The verbs "is napping" (did what?) constitute a verb phrase (**VP**). Phrases convey a single idea. They can consist of other phrases. "The stuffed bear in the box" breaks down into a noun phrase "the stuffed bear" and a prepositional phrase (**PP**) "in the box" (where?), which contains a noun phrase "the box". Identifying the part-of-speech tags is the first step in determining relations between the different parts of a sentence and extracting objective facts (who, what, where).

### Subjective opinions

Modal verbs (e.g., can, may, must, will) convey mood and modality. Together with adverbs and negation words (e.g., not, never) they indicate uncertainty (Morante & Sporleder, 2012). For example, the sentence "I <u>wish</u> the cat <u>would</u> stop pawing the box" uses the subjunctive mood. It expresses an opinion or a wish rather than a fact. The sentence "<u>Stop</u> pawing the box<u>!</u>" uses the imperative mood. It is a command. "The cat <u>didn't</u>" is a negation, and a fact. For an example NLP system used to predict modality, see Morante, Van Asch & Daelemans (2010). Adverbs and adjectives are often used to convey positive or negative sentiment (e.g., nice, irritating). Nouns and verbs do not convey sentiment but they can be associated with a positive or negative tone (e.g., party, punishment). This is discussed further in chapter 7.

### Writing style

Writing style pertains to how individuals write. It is studied as a branch of computational linguistics called stylometry. For example, research by Pennebaker (2011) shows how function words capture the author's personality and gender. Men use more determiners (a, the, that) whereas women use more pronouns (I, you, she). A person with a higher status uses less first-person singular pronouns (I, me, my). A person who is lying uses more plural pronouns (we), more generalizations (every, all) and more negative sentiment words (Newman, Pennebaker, Berry & Richards, 2003). For an example of how stylometry can be used to attribute authorship to anonymous documents, see Luyckx & Daelemans (2008).

## 6.2 MBSP for Python

MBSP FOR PYTHON (see De Smedt, Van Asch & Daelemans, 2010) is a memory-based shallow parser for Python, based on the MBT and TIMBL memory-based learning toolkits (Daelemans, Zavrel, van der Sloot & van den Bosch, 2004). It provides functionality for tokenization, sentence splitting, part-of-speech tagging, chunking, relation finding, prepositional phrase attachment and lemmatization for English. A short definition for each of these tasks is offered further below.

A parser transforms sentences into a representation called a parse tree. A parse tree describes how words are grouped into phrases, along with their relations (e.g. subject and object). Different approaches exist to building a parser. One option is to construct a grammar, a set of structural rules that describes the syntax expected to occur in the sentences of a language. For example, simple declarative sentences often have a subject-verb-object structure: "the cat inspects the box". Constituent-based grammars focus on the hierarchical phrase structure. Dependency-based grammars focus on grammatical relations between words. A grammar can be constructed by hand by a linguist or it can be induced automatically from a treebank. A treebank is a large collection of texts where each sentence has been manually annotated with its syntax structure. Treebank annotation is time-consuming but the effort pays off, because the induced grammar is based on actual language use instead of the linguist's intuition. Also, the induced grammar is probabilistic. It contains useful statistical information, such as how many times a sentence structure occurs in the treebank or how many times a word occurs in **NP** or **VP** phrases.

### Shallow parser

MBSP FOR PYTHON is a shallow parser. Shallow parsing (Abney, 1991) describes a set of tasks used to retrieve some syntactic-semantic information from text in an efficient, robust way. For example, by parsing the parts of speech and phrases in a sentence, at the expense of ignoring detailed configurational syntactic information. Our parser uses a supervised machine learning approach that handles tokenization, sentence splitting, part-of-speech tagging, chunking, relation finding, prepositional phrase attachment and lemmatization. This is illustrated in figure 26. Together, these steps produce a syntactic analysis detailed enough to drive practical applications such as information extraction, information retrieval, question answering and summarization, where large volumes of text (often with errors) have to be analyzed in an efficient way.

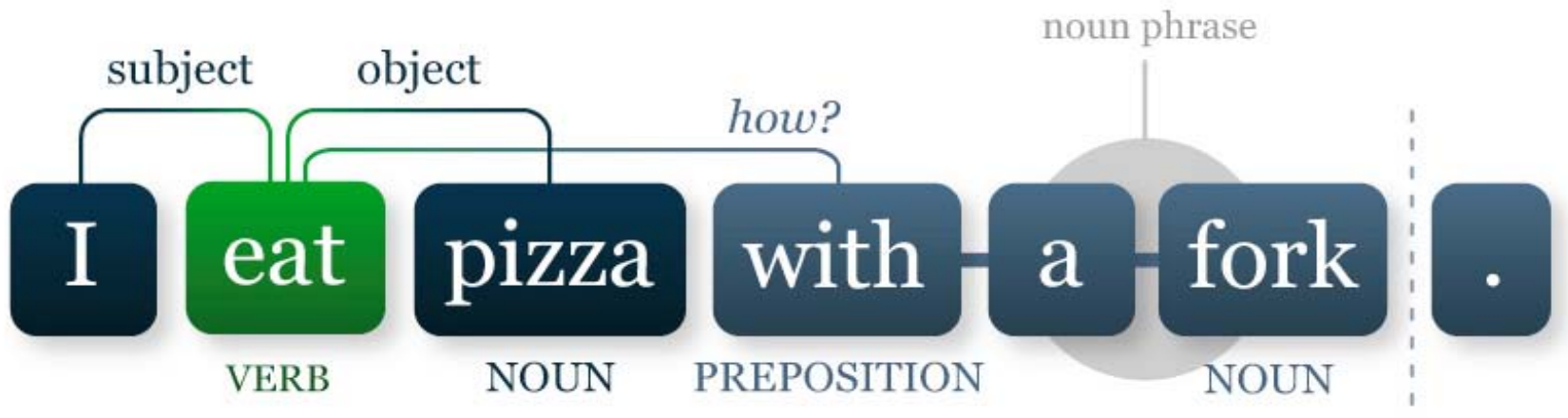

Figure 26. Shallow parsing. The verb "eat" relates sentence subject "I" to object "pizza". The verb is also related to the prepositional phrase "with a fork".

The different steps in the process are described as follows:

**TOKENIZATION**

First, sentence periods are determined and punctuation marks (e.g., ( )"$%) are split from words, which are separated by spaces. Regular expressions (see Friedl, 2002) are used at this stage, since the tokenizer has no idea what a word (token) is right now. Our tokenizer uses a non-trivial approach that deals with word abbreviations, unit suffixes, biomedical expressions, contractions, citations, hyphenation, parenthesis and common errors involving missing spaces and periods.

**PART-OF-SPEECH TAGGING + CHUNKING**

After tokenization and sentence splitting, each word in each sentence is assigned a part-of-speech tag. At the same time constituents like noun phrases (**NP**), verb phrases (**VP**) and prepositional noun phrases (**PNP**) are detected. The tagger/chunker is based on TIMBL (Daelemans et al., 2004). TIMBL implements a supervised, memory-based learning approach (Daelemans, Buchholz & Veenstra, 1999) using *k*-NN with information gain and the fast IGTREE algorithm (Daelemans, van den Bosch, Weijters, 1997). We will discuss *k*-NN and information gain in a little more detail shortly. For the Wall Street Journal corpus (WSJ), accuracy (F-score) is around 96.5% for part-of-speech tagging, 92.5% for **NP** chunking and 92% for **VP** chunking as reported in Daelemans & van den Bosch (2005). Across different experiments and data sets, accuracy is in the high nineties for part-of-speech tagging and in the low nineties for chunking.

**RELATION FINDING**

Grammatical relations (e.g. subject, object) are then predicted between the constituents. For example, in "the cat attacked the stuffed bear", the subject of the sentence (who attacked what?) is "the cat". The object (who or what was attacked?) is "the stuffed bear". Accuracy (F-score) is around 77% for **SBJ** detection and 79% for **OBJ** detection as reported in Buchholz (2002).

**PREPOSITIONAL PHRASE ATTACHMENT**

Prepositional phrases are related to the constituent to which they belong. For example, in "the cat attacked the bear with fervor", the "with fervor" **PNP** is related to the "attacked" **VP** and not to the "the bear" **NP**, since the expression "to attack with fervor" is statistically more probable. Accuracy (F-score) is around 83% as reported in Van Asch & Daelemans (2009).

**LEMMATIZATION**

Lemmatization computes the base form of each word, for example "attacked" → "attack". The lemmatizer in MBSP FOR PYTHON is based on MBLEM, see van den Bosch & Daelemans (1999) for further information.

MBSP FOR PYTHON is a so-called lazy learner. It keeps the initial training data available, including exceptions which may be productive. This technique has been shown to achieve higher accuracy than eager (or greedy) methods for many language processing tasks (Daelemans, Buchholz & Veenstra, 1999). However, the machine learning approach is only as good as the treebank that is used for training. This means that a shallow parser trained on a collection of texts containing descriptions of toys will not perform very well on texts about physics, and vice versa. MBSP is bundled with two sets of training data: newspaper language (WSJ) and biomedical language.

## Python example of use

MBSP FOR PYTHON uses a client-server architecture. This way, the training data is loaded once when the servers start. By default this happens automatically when the module is imported in Python. Otherwise, the `start()` function explicitly starts the four servers (`CHUNK`, `LEMMA`, `RELATION` and `PREPOSITION`) The module includes precompiled Mac OS X binaries for MBT and TIMBL. Instructions on how to compile binaries from source are included in the appendix. Once the servers have started, tagging jobs can be executed using the module's `parse()` function.

The `parse()` function takes a string and a number of optional parameters:

```
MBSP.parse(string,
     tokenize = True, # Tokenize input string?
         tags = True, # Parse part-of-speech tags?
       chunks = True, # Parse chunk tags?
    relations = True, # Parse chunk relations?
      anchors = True, # Parse PP-attachments?
      lemmata = True, # Parse word lemmata?
     encoding = 'utf-8')
```

The following example parses a given string and prints the output. The output is slash-formatted. Slashes that are part of words are encoded as `&slash;`.

```
from MBSP import parse

s = 'The cat studied the toy bear with disinterest.'
s = parse(s)
print s # The/DT/NP-SBJ-1/O/O/the cat/NN/NP-SBj-1/O/O/cat studied/VBD/VP-1/O/A1/study ...
```

See: http://www.clips.ua.ac.be/pages/MBSP#parser

To examine the output more closely, `pprint()` can be used. Its output is shown in table 10.

```
from MBSP import parse, pprint

s = 'The cat studied the toy bear with disinterest.'
s = parse(s)
pprint(s) # pretty-print
```

| WORD | TAG | CHUNK | ROLE | ID | PNP | ANCHOR | LEMMA |
|---|---|---|---|---|---|---|---|
| The | DT | NP | SBJ | 1 | - | - | the |
| cat | NN | NP | SBJ | 1 | - | - | cat |
| studied | VBD | VP | - | 1 | - | A1 | study |
| the | DT | NP | OBJ | 1 | - | - | the |
| toy | JJ | NP | OBJ | 1 | - | - | toy |
| bear | NN | NP | OBJ | 1 | - | - | bear |
| with | IN | PP | - | - | PNP | P1 | with |
| disinterest | NN | NP | - | - | PNP | P1 | disinterest |
| . | . | - | - | - | - | - | . |

Table 10. Example output of `parse()` and `pprint()` in MBSP.

The `parsetree()` function takes the same arguments as `parse()`. It returns the parse tree as a `Text` object that contains a list of sentences. Each `Sentence` in this list contains a list of words. A `Word` object has properties `string`, `type`, `chunk` and `lemma`:

```
from MBSP import parsetree

s = 'The cat studied the toy bear with disinterest.'
t = parsetree(s)

for sentence in t.sentences:
    for word in sentence.words:
        print word.string   # u'The'
        print word.type     # u'DT'
        print word.chunk    # Chunk('The cat/NP-SBJ-1')
        print word.lemma    # u'the'
        print
```

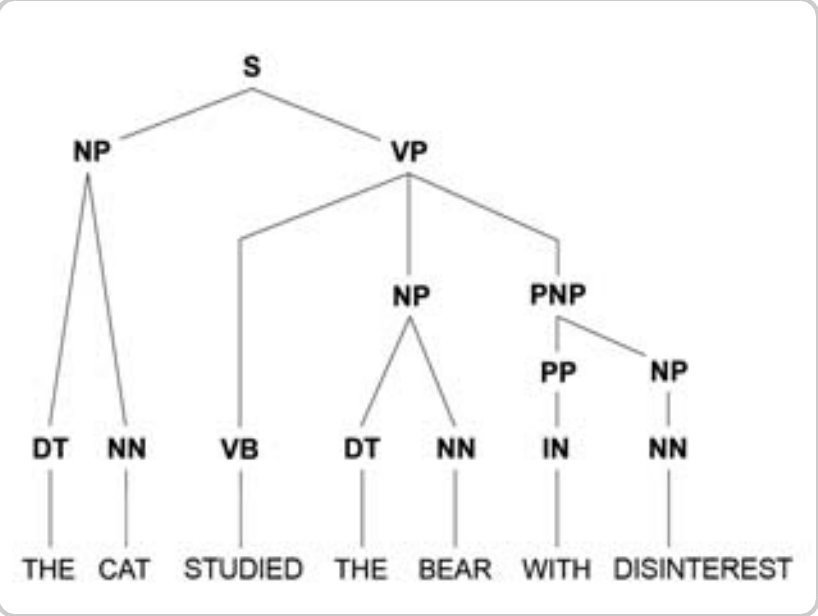

See: http://www.clips.ua.ac.be/pages/MBSP#tree

Figure 27. Parse tree.

A sentence contains a list of phrases (chunks). A `Chunk` object has properties `string`, `type`, `role`, `words`, `lemmata` and `related`:

```
for sentence in t.sentences:
    for chunk in sentence.chunks:
        print chunk.string  # u'The cat'
        print chunk.type    # u'NP'
        print chunk.role    # u'SBJ'
        print chunk.words   # [Word(u'The/DT'), Word(u'cat/NN')]
        print chunk.lemmata # [u'the', u'cat']
        print chunk.related # [Chunk('studied/VP-1'), Chunk('the toy bear/NP-OBJ-1')]
        print
```

A sentence contains a list of prepositional noun phrases. A `PNPChunk` object has properties `string`, `anchor` and `chunks`:

```
for sentence in t.sentences:
    for pnp in sentence.pnp:
        print pnp.string    # u'with disinterest'
        print pnp.anchor    # Chunk('studied/VP-1')
        print pnp.chunks    # [Chunk('with/PP'), Chunk('disinterest/NP')]
        print
```

The parsed `Text` can be exported as an XML-formatted string:

```
from MBSP import Text

f = open('output.xml', 'w')
f.write(t.xml)
f.close()
s = open('output.xml').read()
t = Text.from_xml(s)
```

Because language is ambiguous and because a probabilistic approach is used, the system is not always 100% correct. Arguably, if it was a 100% correct it would actually outperform humans.

## 6.3 Machine learning

Machine learning (ML) is a branch of AI concerned with the development of algorithms that learn from experience and by similarity. Popular applications include junk email detection, the Google search engine (Brin & Page, 1998), Google's translation service, Amazon.com product recommendations and iTunes Genius music playlists. A widely quoted definition of machine learning is offered by Mitchell (1997):

> A computer program is said to learn from experience E with respect to some class of tasks T and performance measure P, if its performance at tasks in T, as measured by P, improves with experience E.

Machine learning algorithms can be supervised, unsupervised or based on reinforcement:

**SUPERVISED LEARNING**

Supervised learning or classification algorithms are used to recognize patterns or make predictions by inferring from training examples. By analogy, this is not unlike pointing out the names of various animals in a picture book to a child. The child can infer from the pictures when it encounters a real animal.

**UNSUPERVISED LEARNING**

Unsupervised learning or clustering algorithms are used to recognize patterns without training examples. By analogy, this can be compared to sorting out a heap of toy blocks based on shared commonalities, for example block color or theme (e.g., separating space Lego from castle Lego).

**REINFORCEMENT LEARNING**

Reinforcement learning algorithms are based on reward. See Sutton & Barto (1998).

### Distance in n-dimensional space

Teddy bears, rag dolls and sock monkeys are similar. They are all stuffed toys. A stuffed bear and a pair of scissors on the other hand are clearly quite different: one is a toy, the other a tool. This is clear because our mind has a deep understanding of teddy bears and scissors ever since childhood. A machine does not. A teddy bear is a meaningless string of characters: t-e-d-d-y b-e-a-r. But we can train machines, for example using a heuristic approach like PERCEPTION. This approach is interesting but also ambiguous and difficult to evaluate. To clarify this, in chapter 5 we have seen how PERCEPTION attributes similarity between Brussels and a stag or a toad. This is ambiguous, since a stag can be seen as pretty while a toad can be seen as ugly, two ends of a continuum. This leads to interpretation bias. In the artistic sense this is not problematic. Both are viable artistic solutions; PERCEPTION's audience is allowed to fill in the blanks with their own interpretation. But suppose we have a program that needs to assess whether PERCEPTION's output is pretty or ugly. It responds with "both!" or "whatever!" In the context of creativity assessment this is not a very useful program. To *evaluate* the output, we need a program that yields a well-founded, unambiguous answer.

#### VECTOR SPACE MODEL

Modern machine learning algorithms favor statistical approaches. A well-known and effective technique is the vector space model that represents documents (i.e., descriptions of concepts) as a matrix of n × m dimensions (Salton, Wong & Yang, 1975). A distance metric can then be used as a function of the matrix to compute similarity between documents, as a value between `0.0`–`1.0`. The vector space model is fundamental to many tasks in natural language processing and machine learning, from search queries to classification and clustering (Manning et al., 2009). Consider the following documents, describing teddy bears, rag dolls and scissors:

**BEAR** A teddy bear is a stuffed toy bear used to entertain children.
**RAGDOLL** A rag doll is a stuffed child's toy made from cloth.
**SCISSORS** Scissors have sharp steel blades to cut paper or cloth.

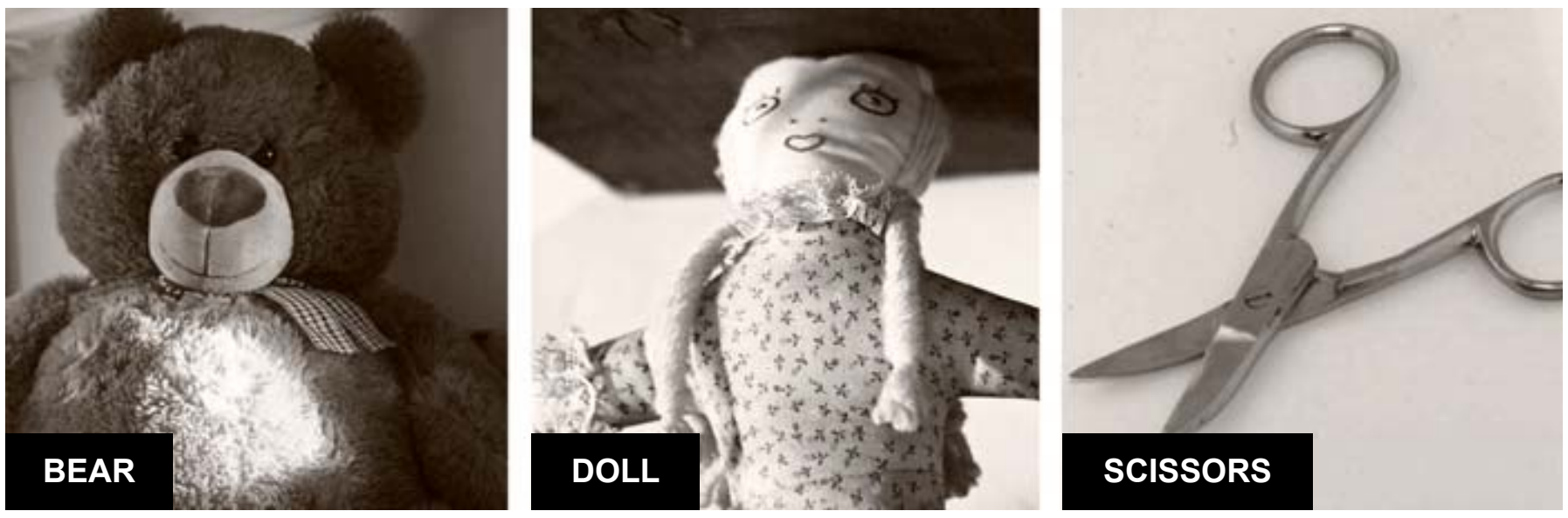

Figure 28. Visual representation of each document. Which is the odd one out?
Photography © Andy Norbo and © Xavi Temporal. Used with permission.

A visual representation of each document is shown in figure 28; we can easily see how one is very different from the other two. Now notice how the **BEAR** and **RAGDOLL** documents both have the words “stuffed” and “toy”, while the **SCISSORS** document does not. We can use this to our advantage. The idea is to count words in each document. We can also normalize each document by collapsing variant word forms (e.g., children → child) and by eliminating words that carry little meaning (e.g., a, or, to). This makes the words that matter stand out.

The normalized **BEAR** document then becomes: “~~a~~ teddy bear ~~be~~ ~~a~~ stuffed toy bear ~~use~~ ~~to~~ entertain child”. The *document vector* is the set of distinct words (called features) and their relative frequency, shown in table 11. Relative frequencies are used so that shorter documents are not at a disadvantage:

| bear (2x) | child | entertain | stuffed | teddy | toy |
|---|---|---|---|---|---|
| `0.28` | `0.14` | `0.14` | `0.14` | `0.14` | `0.14` |

Table 11. Document vector for “A teddy bear is a stuffed toy bear used to entertain children.”

Notice how the word order is discarded. This is a simplified representation called bag-of-words. For our purposes it is relevant that **BEAR** and **RAGDOLL** both have the word “toy”, not so much where this word occurs in each document.

The vector space is the n × m matrix with $n$ features in $m$ documents:

| FEATURES | DOCUMENTS | | |
|---|---|---|---|
| | **BEAR** | **RAGDOLL** | **SCISSORS** |
| bear | 0.28 | 0 | 0 |
| blade | 0 | 0 | 0.14 |
| child | 0.14 | 0.17 | 0 |
| cloth | 0 | 0.17 | 0.14 |
| cut | 0 | 0 | 0.14 |
| doll | 0 | 0.17 | 0 |
| entertain | 0.14 | 0 | 0 |
| paper | 0 | 0 | 0.14 |
| rag | 0 | 0.17 | 0 |
| scissors | 0 | 0 | 0.14 |
| sharp | 0 | 0 | 0.14 |
| steel | 0 | 0 | 0.14 |
| stuffed | 0.14 | 0.17 | 0 |
| teddy | 0.14 | 0 | 0 |
| toy | 0.14 | 0.17 | 0 |

Table 12. Vector space with features as rows and the feature weights for each document as columns.

We can use the matrix to calculate which documents are nearer to each other. As a useful metaphor, imagine the documents as dots on a piece of paper. In other words, as a number of points with ($x$, $y$) coordinates (horizontal and vertical position) in a 2D space. Some points will be further apart than others. We can calculate the distance between each two points using Pythagoras' theorem (Euclidean distance):

$$d = \sqrt{(x_2 - x_1)^2 + (y_2 - y_1)^2} \tag{1}$$

The formula can easily be scaled to points with ($x$, $y$, $z$) coordinates in a 3D space (and so on):

$$d = \sqrt{(x_2 - x_1)^2 + (y_2 - y_1)^2 + (z_2 - z_1)^2} \tag{2}$$

**COSINE SIMILARITY**

Each of our document vectors has $n$ features: ($x$, $y$, $z$, . . . $n$). In other words, they are points in an $n$-D space. We can then calculate the Euclidean distance in $n$-D space. Documents that are nearer to each other are more similar. Another distance metric that is often used for text data is called cosine similarity. It is the Euclidean dot product divided by the product of the Euclidean norm of each vector. The cosine similarity for **BEAR–RAGDOLL** is: (0.28 × 0) + (0 × 0) + . . . + (0.14 × 0.17) = 0.07 / (0.42 × 0.42) = 0.4. The cosine similarity for **BEAR–SCISSORS** is 0.0; there is no overlap for features in **BEAR** and **SCISSORS**. In this way we have learned that teddy bears and rag dolls are more similar than teddy bears and scissors.

In general, for two vectors $a$ and $b$ with $n$ features in the vector space:

$$\cos(\theta) = \frac{\sum_{i=1}^{n} a_i \times b_i}{\sqrt{\sum_{i=1}^{n} (a_i)^2} \times \sqrt{\sum_{i=1}^{n} (b_i)^2}} \quad (3)$$

In Python code:

```
class Vector(dict):
    # Feature => weight map.
    pass

def norm(v):
    return sum(w * w for w in v.values()) ** 0.5

def cos(v1, v2):
    s = sum(v1.get(f, 0) * w for f, w in v2.items())
    s = s / (norm(v1) * norm(v2) or 1)
    return s # 0.0-1.0
```

For an overview of different distance metrics, consult Van Asch (2012).

## Word frequency and relevance

Buying your toddler a dozen identical teddy bears serves no purpose. Which one to hug? It is more engaging if one bear stands out with a red coat and another stands out with a dashing cravat. In a vector space, we also want each document to stand out, so that we can more easily compare them. Imagine we have a vector space with many documents (e.g., a thousand different toys) and many features (e.g., the Wikipedia article for each toy). Some words will recur often. For the Wikipedia "teddy bear" article, frequent words include "bear" (100×), "teddy" (80×), "toy" (10×) and "stuffed" (5×). Frequent words have a higher impact on the distance metric. But it is likely that the word "toy" is prevalent in *every* document. It is true of course that all documents in this vector space are somewhat similar, since they are all about toys. But we are interested in what sets them apart; the words that are frequent in one document and rare in others.

#### TF × IDF

The tf × idf weight or term frequency–inverse document frequency (Salton & Buckley, 1988) is a statistic that reflects the relevance of a word in *one* document. It is the term frequency, that is, the number of times a word appears in the document, divided by the document frequency, the number of times a word appears in all documents. The tf × idf weight is often used instead of the simple relative word count, which we used earlier to compare teddy bears and scissors.

For a word $w$ and a collection of documents $D$:

$$\mathrm{idf}(w, D) = \log \frac{|D|}{1 + |\{d \in D : w \in d\}|} \tag{4}$$

$$\mathrm{tf} * \mathrm{idf}(w, d, D) = \mathrm{tf}(w, d) \times \mathrm{idf}(w, D) \tag{5}$$

In Python code:

```
from math import log

def tfidf(vectors=[]):
    df = {}
    for v in vectors:
        for f in v:
            df[f] = df.get(f, 0.0) + 1
    for v in vectors:
        for f in v: # Modified in-place.
            v[f] *= log(len(vectors) / df[f]) or 1
```

#### INFORMATION GAIN

Another, more elaborate measure for feature relevance is information gain (Kullback & Leibler, 1951). It is based on the Shannon entropy (Shannon, 1948). In information theory, high entropy means that a feature is more evenly distributed. Low entropy means that it is unevenly distributed: it stands out. For a given feature, information gain is the entropy of document probabilities (i.e., how many documents labeled **TOY** or **TOOL**), minus the probability of the feature across all documents multiplied by the *conditional* entropy of the feature's probabilities (e.g., how many times it occurs in a **TOY** document, in a **TOOL** document, etc.) In other words, information gain is a measure of the predictive probability of the given feature for certain document labels in the collection. For a detailed overview, consult Mitchell (1997).

For word $w$ in documents $D$, with entropy $H$ and $d_w$ = weight of word $w$ (`0` or `1`) in document $d$:

$$IG(D, w) = H(D) - \sum_{v \in values(w)} \frac{|\{d \in D | d_w = v\}|}{|D|} \cdot H(\{d \in D | d_w = v\}) \tag{6}$$

In Python code, for a given list of probabilities `P` (sum to one):

```
def entropy(P=[]):
    s = float(sum(P)) or 1
    return -sum(p / s * log(p / s, len(P)) for p in P if p != 0)
```

## Classification

Classification (supervised machine learning) can be used to predict the label of an unlabeled document based on manually labeled training examples. Well-known algorithms include Naive Bayes, *k*-NN and SVM. Naive Bayes is a classic example based on Bayes' theorem of posterior probability (Lewis, 1998). SVM or Support Vector Machine uses hyperplanes to separate a high-dimensional vector space (Cortes & Vapnik, 1995). Some methods such as Random Forests use an ensemble of multiple classifiers (Breiman, 2001).

#### KNN

The *k*-nearest neighbor algorithm (Fix & Hodges, 1951) is a lazy learning algorithm that classifies unlabeled documents based on the nearest training examples in the vector space. It computes the cosine similarity for a given unlabeled document to all of the labeled documents in the vector space. It then yields the label from the *k* nearest documents using a majority vote.

Consider the following training documents:

| | |
|---|---|
| **TOY** | A teddy bear is a stuffed toy bear used to entertain children. |
| **TOY** | A rag doll is a stuffed child's toy made from cloth. |
| **TOOL** | A bread knife has a jagged steel blade for cutting bread. |
| **TOOL** | Scissors have sharp steel blades to cut paper or cloth. |

Suppose we now want to learn the label of the unlabeled document: "A needle is steel pin for sewing cloth." Is it a toy or a tool? The *k*-NN algorithm ($k$=`3`) will classify it as a tool. It yields the following cosine similarity values for the normalized training examples: **TOY** `0.0`, **TOY** `0.18`, **TOOL** `0.15`, **TOOL** `0.34`. This means that the nearest neighbor is the fourth document about scissors (highest cosine similarity). This document also contains the words "steel" and "cloth". The $k$=`3` nearest neighbors include two tools and one toy, so by majority vote the unlabeled document is predicted to be tool.

In Python code:

```
class KNN:

    def __init__(self, train=[]):
        self.model = train # List of (label, vector)-tuples.

    def project(self, v1):
        return [(cos(v1, v2), label) for label, v2 in self.model]

    def classify(self, v, k=3):
        nn = {}
        for x, label in sorted(self.project(v))[-k:]:
            nn[label] = nn.get(label, 0) + x
        nn = [(x, label) for label, x in nn.items() if x > 0]
        if len(nn) > 0:
            return max(nn)[1]
```

#### K-FOLD CROSS VALIDATION

How accurate is our **TOY–TOOL** classifier? One straightforward way to know for sure is to test it with a set of hand-labeled documents not used for training (called a gold standard) and evaluate to what extent the predictions correspond to the annotated labels. The gold standard is comprised of documents of which we are certain that they are labeled correctly. This implies annotating or reviewing them by hand, which can be time-consuming, but the technique is vacuous without a reliable performance test. For example, if we have a set of a 1,000 correctly labeled documents (so far we have 4), we could use 750 for training and 250 for testing and see how many of these 250 are correctly predicted. We could also run 10 separate tests, each using a different 90%–10% train and test set, where each document is used one time and one time only as test data, and average the results. This is called 10-fold cross validation.

#### PRECISION & RECALL

How accurate is our **TOY–TOOL** classifier, *really*? Suppose we test it with 90 **TOY** documents and 10 **TOOL** documents. It predicts 100× **TOY** so accuracy is 90% = 90/100 correct predictions. But the estimate can be misleading because the classifier might *always* predict **TOY**, in which case its actual recognition rate for tools is 0%. Your toddler is in the kitchen happily playing with scissors. A better way is to calculate precision and recall. For two labels, called binary classification, there will be a number of true positives (e.g., toys classified as toys), true negatives (tools classified as tools), false positives (tools classified as toys) and false negatives (toys classified as tools). The distinction between *tp*, *tn*, *fp* and *fn* is called the confusion matrix.

Precision, recall and F-measure are then defined as:

$$P = \frac{tp}{tp + fp} \qquad R = \frac{tp}{tp + fn} \qquad F = \frac{2PR}{P + R} \tag{7}$$

Recall is the percentage of toys identified by a classifier. Precision is the percentage of documents classified as **TOY** that really are toys. F-score is the harmonic mean of precision and recall.

#### FEATURE SELECTION + EXTRACTION

In general, the performance of a classifier relies on the amount of training documents and the relevance of document features. For example, a vector space with too many features can degrade in performance because assumptions are made based on noisy or irrelevant features. This is called overfitting (see Schaffer, 1993). In this case we want to select a subset of the most predictive features. Techniques for automatic feature selection include information gain and genetic algorithms (see Guyon & Elisseeff, 2003).

Another technique used to reduce the vector space is feature extraction. A well-known example is latent semantic analysis (Landauer, McNamara, Dennis & Kintsch, 2007). LSA assumes that related words occur in similar documents. It is based on a matrix factorization called singular value decomposition (Golub & Van Loan, 1996), used to group related features into concepts. The result is a new, reduced vector space with these concepts as features, illustrated in figure 29.

Feature selection and feature extraction can also make the classifier faster, since less time is spent computing distance functions in the vector space.

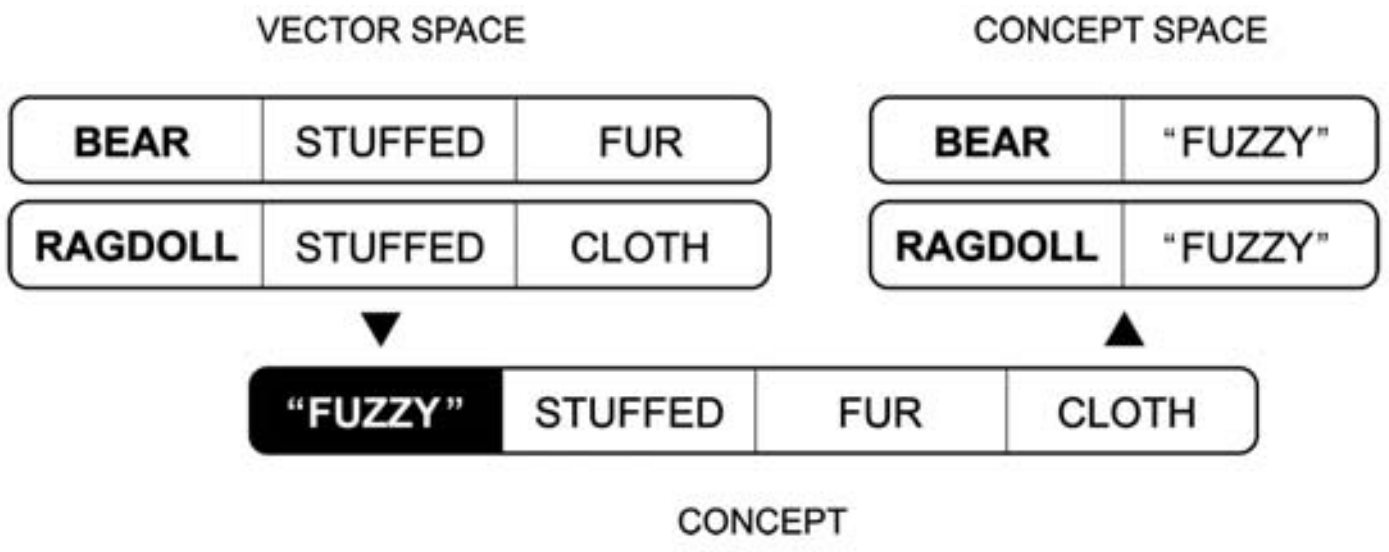

Figure 29. Concise illustration of latent semantic analysis.
Features are grouped into concepts, resulting in a reduced concept space.

## Clustering

Clustering (unsupervised machine learning) is used to partition a set of unlabeled documents into groups, each containing documents that are similar to each other. In the context of creativity, clustering can be employed for novelty detection by examining the smallest groups (Markou & Singh, 2003). Well-known clustering algorithms include *k*-means and hierarchical clustering.

#### *k*-MEANS

The *k*-means algorithm (Lloyd, 1982) uses a fast heuristic approach to partition a vector space into *k* clusters, so that each document belongs to the nearest cluster. The nearest cluster for a given document has the shortest distance between the document and the centroid of the cluster. The centroid is the mean vector for all documents in a cluster. Once each document is assigned to a cluster, clusters are then iteratively refined by updating their centroids and by re-assigning documents to nearer clusters. Since the initial *k* centroids are chosen randomly there is no guarantee that the algorithm will converge to a global optimum.

#### HIERARCHICAL CLUSTERING

The hierarchical clustering algorithm builds a hierarchy of clusters. Each document is assigned to its own cluster. Pairs of nearest clusters are then iteratively merged using a chosen distance metric. The result is a top-level cluster consisting of recursively nested clusters called a dendrogram. Distance between two clusters is measured as the distance between either the two nearest documents in the two clusters (Sibson, 1973) or the two most distant (Defays, 1977). The latter approach avoids chaining, where documents in a cluster are connected by a chain of other documents but very distant from each other.

With a basic understanding of NLP and ML, we can now turn our attention to PATTERN, a Python package containing tools for natural language processing (similar to MBSP) and the machine learning techniques discussed above: the cosine similarity distance metric, tf × idf, information gain, *k*-NN, *k*-means and hierarchical clustering.

## 6.4 Pattern for Python

The World Wide Web is an immense collection of linguistic information that has in the last decade gathered attention as a valuable resource for tasks such as information extraction, machine translation, opinion mining and trend detection, that is, Web as Corpus (Kilgarriff & Grefenstette, 2003). This use of the WWW poses a challenge, since the Web is interspersed with code (HTML markup) and lacks metadata (e.g., language identification, part-of-speech tags, semantic labels).

PATTERN (De Smedt & Daelemans, 2012) is a Python package for web data mining, natural language processing, machine learning and network analysis, with a focus on ease-of-use. It offers a mash-up of tools often used when harnessing the Web as a corpus. This usually requires several independent toolkits chained together. Several such toolkits with a user interface exist in the scientific community, for example ORANGE (Demšar, Zupan, Leban & Curk, 2004) for machine learning and GEPHI (Bastian, Heymann & Jacomy, 2009) for graph visualization. By contrast, PATTERN is more related to toolkits such as SCRAPY[21], NLTK (Bird, Klein & Loper, 2009), PYBRAIN (Schaul, Bayer, Wierstra, Sun, Felder et al., 2010) and NETWORKX (Hagberg, Schult & Swart, 2008) in that it is geared towards integration in the user's own programs.

PATTERN aims to be useful to both a scientific and a non-scientific audience. The syntax is straightforward. Function names and parameters were chosen to make the commands self-explanatory. The documentation[22] assumes no prior knowledge. The package is bundled with in-depth examples and unit tests, including a set of corpora for testing accuracy such as POLARITY DATASET 2.0 (Pang & Lee, 2004). The source code is hosted online on GitHub[23] and released under a BSD license. It can be incorporated into proprietary products or used in combination with other open source packages. We believe that the package is valuable as a learning environment for students, as a rapid development framework for web developers, and in research projects with a short development cycle.

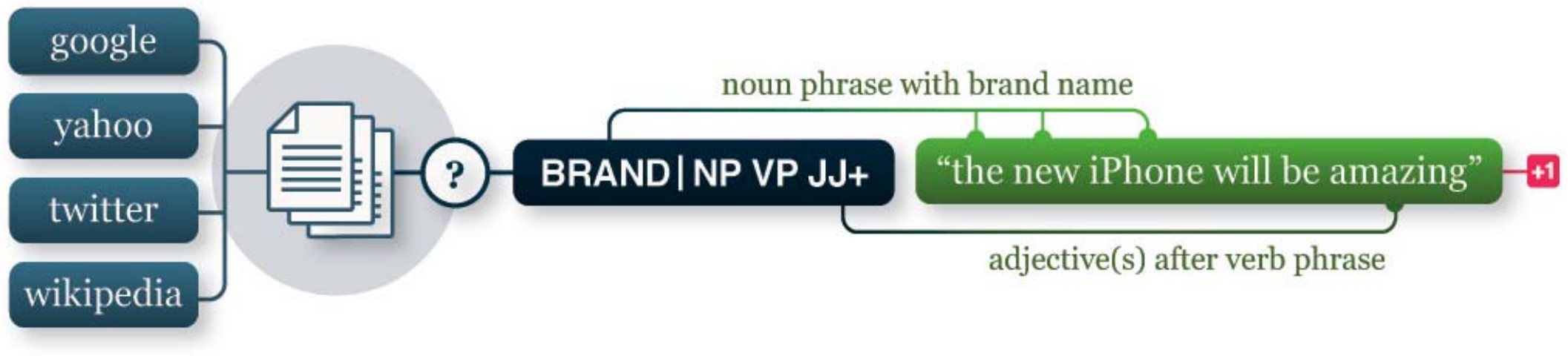

Figure 30. Example workflow. Text is mined from the web and searched by syntax and semantics. Sentiment analysis (positive/negative) is performed on matching phrases.

[21] http://scrapy.org/
[22] http://www.clips.ua.ac.be/pages/pattern
[23] https://github.com/clips/pattern

PATTERN is organized in separate modules that can be combined. For example, text mined from Wikipedia (module pattern.web) can be parsed for part-of-speech tags (pattern.en) and queried for specific patterns (pattern.search) that can be used as training examples for a classifier (pattern.vector). This is illustrated in figure 30.

## pattern.web

The pattern.web module contains tools for web mining, using a download mechanism that supports caching, proxy servers, asynchronous requests and redirection. It has a `SearchEngine` class that provides a uniform API to multiple web services such as Google, Bing, Yahoo!, Twitter, Facebook, Wikipedia, Flickr, and news feeds using FEEDPARSER[24]. The module includes a HTML parser based on BEAUTIFULSOUP[25], a PDF parser based on PDFMINER[26], a web crawler and a webmail interface. The following example executes a Google search query. The `Google` class is a subclass of `SearchEngine`. The `SearchEngine.search()` method returns a list of `Result` objects for the given query. Note the `license` parameter in the constructor. Some web services (i.e., Google, Bing, Yahoo) use a paid model requiring a license key.

```
from pattern.web import Google, plaintext

q = 'wind-up mouse'
G = Google(license=None)
for page in range(1, 3):
    for result in G.search(q, start=page, cached=False):
        print result.url
        print result.title
        print plaintext(result.description)
        print
```

See: http://www.clips.ua.ac.be/pages/pattern-web#services

The `plaintext()` function decodes entities (é → é), collapses superfluous whitespace and removes HTML tags from the given string. The approach is non-trivial using an SGML parser.

The following example retrieves an article from Wikipedia. Here the `search()` method returns a single `WikipediaArticle` instead of a list of `Result` objects. With the `cached=True` parameter the results will be cached locally so they can be retrieved faster next time.

```
q = 'rubber chicken'
W = Wikipedia(language='en')
article = W.search(q, cached=True)
for section in article.sections:
    print section.title.upper()
    print section.content
    print

print article.links
```

See: http://www.clips.ua.ac.be/pages/pattern-web#wikipedia

[24] http://packages.python.org/feedparser/
[25] http://www.crummy.com/software/BeautifulSoup/
[26] http://unixuser.org/~euske/python/pdfminer/

The module includes an HTML parser that represents HTML source code as a parse tree. The DOM class (Document Object Model) offers an efficient and robust way to traverse the parse tree and retrieve HTML elements by tag name, id and class. Suppose that there is a web page located at http://www.store.com/pets/toys/3 that displays a product with customer reviews:

```
<html lang="en">
<head>
    <meta charset="utf-8">
    <title>Wind-up mouse toy</title>
</head>
<body>
    [...]
    <div class="review">
        <h3>Pet loves this toy!</h3>
        <p>She loves the eeking sound it makes.</p>
        <span class="rating">****</span>
    </div>
    [...]
</body>
</html>
```

We want to retrieve customer reviews together with their star rating. In the code below, we download the HTML source of the web page, feed it to the DOM parser and then search the parse tree for <div class="review"> elements. This is accomplished with the DOM.by_class() method, which returns a list of Element objects whose class attribute is "review".

```
from pattern.web import URL, DOM, plaintext

url = URL('http://www.store.com/pets/toys/3')
src = url.download(unicode=True)

for e in DOM(src).by_class('review'):
    title, description, rating = (
        plaintext(e.by_tag('h3')[0].content),
        plaintext(e.by_tag('p')[0].content),
        plaintext(e.by_class('rating')[0].content).count('*')
    )
    print title, rating # u'Pet loves this toy', 4
```

See: http://www.clips.ua.ac.be/pages/pattern-web#DOM

We can use the reviews to construct a training corpus for sentiment analysis, for example.

## pattern.db

The pattern.db module facilitates file storage of corpora. A corpus is a collection of text documents or other data, used for statistical analysis and hypothesis testing. A text corpus can be annotated with part-of-speech tags. For example, Penn Treebank corpus (Marcus et al., 1993) has over 4.5 million tagged words in English text. Some corpora may contain spam email messages, others customer product reviews or newspaper articles. Some corpora are multilingual. Since there is no standard format for storing a corpus, many different formats are used: XML, CSV, plain text, and so on. Typical problems pertain to reading the file format of a particular corpus, dealing with Unicode special characters (e.g., ø or ñ) and storing a corpus, i.e., if a comma is used to separate different items, how do we store a comma as part of an item?

The pattern.db module has wrappers for Unicode CSV files and SQLITE and MYSQL databases. For example, the `Datasheet` class is an n × m table that imports and exports CSV files. It has functionality for selecting and sorting columns or grouping rows comparable to a MS Excel sheet.

## pattern.en

The pattern.en module implements a fast, regular expressions-based tagger/chunker for English based on Brill's finite state part-of-speech tagger (Brill, 1992). Brill's algorithm automatically acquires a lexicon of known words and a set of rules for tagging unknown words from a training corpus. The tagger is extended with a tokenizer, lemmatizer and chunker. The `parse()` and `parsetree()` functions are identical to those in MBSP FOR PYTHON, so it is trivial to switch to the MBSP parser in a PATTERN workflow. The module contains functions for noun singularization and pluralization (Conway, 1998), verb conjugation, linguistic mood and modality, and sentiment analysis (De Smedt & Daelemans, 2012). It comes bundled with WORDNET and PYWORDNET[27].

Accuracy for Brill's tagger has been shown to be 95%. Although taggers such as MBSP perform better, Brill's tagger is faster and smaller in storage size. We have no accuracy scores for **SBJ** and **OBJ** detection but performance is poorer than that of MBSP. Accuracy (F-score) is 96% for noun singularization and pluralization tested on CELEX (Baayen, Piepenbrock & van Rijn, 1993), 95% for verb conjugation on CELEX, 67% for mood and modality on the CONLL2010 SHARED TASK 1 Wikipedia corpus (Farkas, Vincze, Móra, Csirik & Szarvas, 2010), and 74% for sentiment analysis (discussed in chapter 7).

The following example demonstrates the `parse()` function. The output is shown in table 13.

```
from pattern.en import parse, pprint

s = 'The black cat sat on the mat.'
s = parse(s, relations=True, lemmata=True)
pprint(s)
```

See: http://www.clips.ua.ac.be/pages/pattern-en#parser

| WORD | TAG | CHUNK | ROLE | ID | PNP | LEMMA |
|---|---|---|---|---|---|---|
| The | **DT** | **NP** | **SBJ** | 1 | - | the |
| black | **JJ** | **NP** | **SBJ** | 1 | - | black |
| cat | **NN** | **NP** | **SBJ** | 1 | - | cat |
| sat | **VBD** | **VP** | - | 1 | - | sit |
| on | **IN** | **PP** | - | - | **PNP** | on |
| the | **DT** | **NP** | - | - | **PNP** | the |
| mat | **NN** | **NP** | - | - | **PNP** | mat |
| . | **.** | - | - | - | - | . |

Table 13. Example output of `parse()` and `pprint()` in PATTERN.

[27] http://osteele.com/projects/pywordnet

The `parsetree()` function returns a `Text` object consisting of `Sentence` objects. These provide an easy way to iterate over chunks and words in the parsed text:

```
from pattern.en import parsetree

s = 'The black cat sat on the mat.'
t = parsetree(s, relations=True, lemmata=True)
for sentence in t.sentences:
    for chunk in sentence.chunks:
        for word in chunk.words:
            print chunk.type, word.string, word.type, word.lemma
```

See: http://www.clips.ua.ac.be/pages/pattern-en#tree

When `parse()` and `parsetree()` are used with `lemmata=True`, the lemmatizer will compute the singular form of plural nouns (cats → cat) and the infinitive of conjugated verbs (sat → sit). Lemmatization can also be invoked with separate functions. This is useful to bring different word forms together when training a classifier, for example.

```
from pattern.en import singularize, pluralize
from pattern.en import conjugate

print singularize('cats')     # 'cat'
print pluralize('cat')        # 'cats'
print conjugate('sat', 'inf') # 'sit'
print conjugate('be', '1sg')  # 'am'
```

See: http://www.clips.ua.ac.be/pages/pattern-en#pluralization

The `Text` and `Sentence` objects can serve as input for other functions, such as the `search()` function in the pattern.search module, or `modality()` and `sentiment()`.

The `modality()` function returns a value between `-1.0` and `+1.0`, expressing the degree of certainty based on modal verbs and adverbs in the sentence. For example, "I wish it would stop raining" scores `-0.75` while "It will surely stop raining soon" scores `+0.75`. In Wikipedia terms, modality is sometimes referred to as *weaseling* when the impression is raised that something important is said, but what is really vague and misleading (Farkas et al., 2010). For example: "some people claim that" or "common sense dictates that".

```
from pattern.en import parsetree
from pattern.en import modality

print modality(parsetree('some people claim that')) # 0.125
```

The `sentiment()` function returns a `(polarity, subjectivity)`-tuple, where `polarity` is a value between `-1.0` and `+1.0` expressing negative vs. positive sentiment based on adjectives (e.g., nice) and adverbs (e.g., very). For example, "The weather is very nice" scores `+0.64`.

```
from pattern.en import parsetree
from pattern.en import sentiment

print sentiment(parsetree('What a wonderful day!')) # (1.0, 1.0)
```

The pattern.en.wordnet submodule is an interface to the WORDNET lexical database. It is organized in `Synset` objects (synonym sets) with relations to other synsets. Example relations are **hyponymy** (bird → hen), **holonymy** (bird → flock) and **meronymy** (bird → feather). Following are two functions taken from the FLOWEREWOLF program discussed in chapter 5. The `shift()` function returns a random hyponym for a given noun. The `alliterate()` function yields a list of alliterative adjectives for a given noun.

```
from pattern.en import wordnet
from random import choice

def shift(noun):
    s = wordnet.synsets(noun)
    s = s[0]
    h = choice(s.hyponyms(recursive=True) or [s])
    return (h.synonyms[0], h.gloss)

def alliterate(noun, head=2, tail=1):
    for a in wordnet.ADJECTIVES.keys():
        if noun[:head] == a[:head] and noun[-tail:] == a[-tail:]:
            yield a

print shift('soul')
print list(alliterate('specter', head=3))

> ('poltergeist', 'a ghost that announces its presence with rapping')
> ['spectacular', 'specular']
```

See: http://www.clips.ua.ac.be/pages/pattern-en#wordnet

## pattern.de

The pattern.de module contains a tagger/chunker for German (Schneider & Volk, 1998) and functions for German noun singularization and pluralization, predicative and attributive adjectives (e.g., neugierig → die neugierige Katze) and verb conjugation. Schneider & Volk report the accuracy of the tagger around 95% for 15% unknown words. Accuracy (F-score) is 84% for noun singularization, 72% for pluralization, 98% for predicative adjectives, 75% for attributive adjectives, and 87% for verb conjugation of unknown verbs. The results were tested on CELEX.

## pattern.nl

The pattern.nl module contains a tagger/chunker for Dutch (Geertzen, 2010) and functions for Dutch noun singularization and pluralization, predicative and attributive adjectives (e.g., nieuwsgierig → de nieuwsgierige kat), verb conjugation and sentiment analysis. Geertzen reports the accuracy of the tagger around 92%. Accuracy (F-score) is 91% for noun singularization, 80% for pluralization, 99% for predicative and attributive adjectives, and 80% for verb conjugation of unknown verbs. The results were tested on CELEX. Accuracy for sentiment analysis is 82%.

## pattern.search

The pattern.search module is useful for information extraction. Information extraction (IE) is the task of obtaining structured information from unstructured data, for example to recognize the names of a person or organization in text (named entity recognition), or to identify relations such as **PERSON works-for ORGANIZATION**. Interesting examples include TEXTRUNNER (Banko, Cafarella, Soderland, Broadhead & Etzioni, 2007), which mines relations from Wikipedia, and CONCEPTNET (Liu & Singh, 2004), an automatically acquired semantic network of common sense.

The pattern.search module implements a pattern matching algorithm for *n*-grams (*n* consecutive words) using an approach similar to regular expressions. Search queries can include a mixture of words, phrases, part-of-speech-tags, taxonomy terms (e.g., **PET** = dog + cat + goldfish) and control characters such as wildcards (`*`), operators (`|`), quantifiers (`?` and `+`) and groups (`{}`) to extract information from a string or a parsed `Text`. For example, a simple "**DT NN** is an animal" pattern can be used to learn animal names. It means: a determiner (a, an, the) followed by any noun, followed by the words "is an animal".

```
from pattern.en import parsetree
from pattern.search import search

s = 'A crocodile is an animal called a reptile.'
t = parsetree(s)
for match in search('DT {NN} is an animal', t):
    print match.group(1)
```

See: http://www.clips.ua.ac.be/pages/pattern-search

This yields a list of words matching the `{NN}` group in the pattern: `[Word('crocodile/NN')]`. Now that we know that a crocodile is an animal we can learn more about it, for example using a "**JJ+** crocodile" pattern. It means: one or more adjectives followed by the word "crocodile". It matches phrases such as "prehistoric crocodile" and "green crocodile".

A scalable approach is to group known animals in a taxonomy:

```
from pattern.en import parsetree
from pattern.search import taxonomy, search

for animal in ('crocodile', 'snail', 'toad'):
    taxonomy.append(animal, type='animal')

s = 'Picture of a green crocodile covered in pond lily.'
t = parsetree(s)
for match in search('{JJ+} {ANIMAL}', t):
    print match.group(1), match.group(2)
```

See: http://www.clips.ua.ac.be/pages/pattern-search#taxonomy

Since `search()` is case-insensitive, uppercase words in the pattern are assumed to be taxonomy terms: "ANIMAL" represents the **ANIMAL** category in the taxonomy, with instances "crocodile", "snail" and "toad". The pattern matches phrases like "green crocodile" but also "squishable snail".

To deploy this learner on the Web we can combine the search patterns with a custom `Spider` from the pattern.web module. In the code below, the `Spider.visit()` method is implemented in a subclass. It is called for each web page the spider visits, by following the links in each page.

```
from pattern.web import Spider, plaintext
from pattern.en import parsetree
from pattern.search import search

class Learner(Spider):
    def visit(self, link, source=None):
        t = parsetree(plaintext(source))
        for match in search('{NN} is an animal', t):
            print match.group(1)

learner = Learner(links=['http://en.wikipedia.org/wiki/Animal'])
while True: # Endless learning.
    learner.crawl(throttle=10)
```

See: http://www.clips.ua.ac.be/pages/pattern-web#spider

## pattern.vector

The pattern.vector module implements a vector space model using a `Document` and a `Corpus` class. Documents are lemmatized bag-of-words that can be grouped in a sparse corpus to compute tf × idf, distance metrics (cosine, Euclidean, Manhattan, Hamming) and dimension reduction (latent semantic analysis). The module includes a hierarchical and a *k*-means clustering algorithm, optimized with a *k*-means++ initialization algorithm (Arthur and Vassilvitskii, 2007) and a fast triangle inequality exploit (Elkan, 2003). A Naive Bayes, a *k*-NN and a SVM classifier using LIBSVM (Chang and Lin, 2011) are included, along with tools for feature selection (information gain) and K-fold cross validation.

The following example replicates our previous **BEAR–RAGDOLL–SCISSORS** example. Each `Document` object is initialized with a string. With `stemmer=LEMMA` lemmatization is enabled. With `stopwords=False` (default) words that carry little meaning (e.g., a, or, to) are eliminated. The `type` parameter defines the document label. The documents are bundled in a `Corpus` that computes tf (default is tf × idf) for each `Document.vector`. The `Corpus.similarity()` method returns the cosine similarity for two given documents.

```
from pattern.vector import Document, Corpus, LEMMA, TF, TFIDF

d1 = Document('A teddy bear is a stuffed toy bear used to entertain children.',
      stemmer=LEMMA, stopwords=False, type='BEAR')

d2 = Document('A rag doll is a stuffed child\'s toy made from cloth.',
      stemmer=LEMMA, stopwords=False, type='RAGDOLL')

d3 = Document('Scissors have sharp steel blades to cut paper or cloth.',
      stemmer=LEMMA, stopwords=False, type='SCISSORS')

corpus = Corpus([d1, d2, d3], weight=TF)

print corpus.similarity(d1, d2) # 0.27 bear-doll
print corpus.similarity(d1, d3) # 0.00 bear-scissors
print corpus.similarity(d2, d3) # 0.15 doll-scissors
```

See: http://www.clips.ua.ac.be/pages/pattern-vector#document

Documents can also be passed to a classifier. The `Classifier` base class has three subclasses: `Bayes`, `kNN` and `SVM`. The following example replicates our previous **TOY–TOOL** *k*-NN classifier:

```
from pattern.vector import Document, Corpus
from pattern.vector import LEMMA, TFIDF
from pattern.vector import kNN

train = (
    ('TOY',  'A teddy bear is a stuffed toy bear used to entertain children.'),
    ('TOY',  'A rag doll is a stuffed child\'s toy made from cloth.'),
    ('TOOL', 'Scissors have sharp steel blades to cut paper or cloth.'),
    ('TOOL', 'A bread knife has a jagged steel blade for cutting bread.')
)

test = 'A bread knife has a jagged steel blade for cutting bread.'

corpus = Corpus(weight=TFIDF)
for label, example in train:
    d = Document(example, stemmer=LEMMA, type=label)
    corpus.append(d)

knn = kNN()
for document in corpus:
    knn.train(document, type=document.type)

print knn.classify(test) # 'TOOL'
```

See: http://www.clips.ua.ac.be/pages/pattern-vector#classification

Finally, we chain together four PATTERN modules to train a *k*-NN classifier on adjectives mined from Twitter. We mine a 1,000 tweets with the hashtag #win or #fail (our labels), for example: "$20 tip off a sweet little old lady today #win". We parse the part-of-speech tags for each tweet, keeping adjectives. We group adjective vectors in a corpus and use it to train the classifier. It predicts "sweet" as `WIN` and "stupid" as `FAIL`. The results may vary depending on what is currently buzzing on Twitter. A real-world classifier will also need more training data and more rigorous feature selection.

```
from pattern.web    import Twitter
from pattern.en     import parsetree
from pattern.search import search
from pattern.vector import Document, Corpus, kNN

corpus = Corpus()

for i in range(1, 10):
    for tweet in Twitter().search('#win OR #fail', start=i, count=100):
        p = '#win' in tweet.description.lower() and 'WIN' or 'FAIL'
        s = tweet.description.lower()
        t = parsetree(s)
        m = search('JJ', t)
        m = [match[0].string for match in m]
        if len(m) > 0:
            corpus.append(Document(m, type=p))

classifier = kNN()
for document in corpus:
    classifier.train(document)

print classifier.classify('sweet')  # 'WIN'
print classifier.classify('stupid') # 'FAIL'
```

To apply LSA dimension reduction to a corpus before training a classifier we can do:

```
corpus = corpus.reduce(dimensions=k) # reduce k dimensions
```

To apply feature selection to a corpus before training a classifier we can do:

```
subset = corpus.feature_selection(top=100, method='infogain')
corpus = corpus.filter(features=subset)
```

## pattern.graph

The pattern.graph module implements a graph data structure using `Node`, `Edge` and `Graph` classes. It contains algorithms for shortest path finding, subgraph partitioning, eigenvector centrality and betweenness centrality (Brandes, 2001). Centrality algorithms were ported from NETWORKX. For an example of use, see chapter 5. The module has a force-based layout algorithm that positions nodes in 2D space. Visualizations can be exported to HTML and manipulated in a web browser using the `canvas.js` helper module.

## pattern.metrics

The pattern.metrics module contains statistical functions for data analysis, including sample mean, variance and standard deviation, histogram, quantiles, boxplot, Fisher's exact test and Pearson's chi-squared test. Other evaluation metrics include a Python code profiler, functions for accuracy, precision, recall and F-score, confusion matrix, inter-rater agreement (Fleiss' kappa), string similarity (Levenshtein, Dice), string readability (Flesch) and intertextuality.

## canvas.js

The canvas.js module is a simple and robust JavaScript API for the HTML5 `<canvas>` element, which can be used to generate interactive 2D graphics in a web browser, using lines, shapes, paths, images, image filters and text. The functionality in the module closely resembles NODEBOX FOR OPENGL. It is useful to support quantitative data analysis with a graphical representation of the data (e.g., line chart, histogram, scatter plot, box plot). More recent representation techniques include network visualization and custom, interactive and/or online visualizations, in effect any representation that reduces complexity while capturing important information (Fayyad, Wierse & Grinstein, 2002). This is well within the capabilities of canvas.js.

Below is a basic example that draws a rotating red square in the web browser. The source code imports the canvas.js module. Note the `<script type="text/canvas">` that defines the animation. It has a `setup()` function that will be executed once when the animation starts, and a `draw()` function that will be executed each animation frame.

```
<!doctype html>
<html>
<head>
    <script type="text/javascript" src="canvas.js"></script>
</head>
<body>
```

```
▶   <script type="text/canvas">
      function setup(canvas) {
          canvas.size(500, 500);
      }
      function draw(canvas) {
          canvas.clear();
          translate(250, 250);
          rotate(canvas.frame);
          rect(-150, -150, 300, 300, {fill: color(1,0,0,1)});
      }
    </script>
</body>
</html>
```

See: http://www.clips.ua.ac.be/pages/pattern-canvas

Figure 31 shows a screenshot of the online editor[28]. In live-editing mode, any modifications to the source code are instantly reflected in the animation.

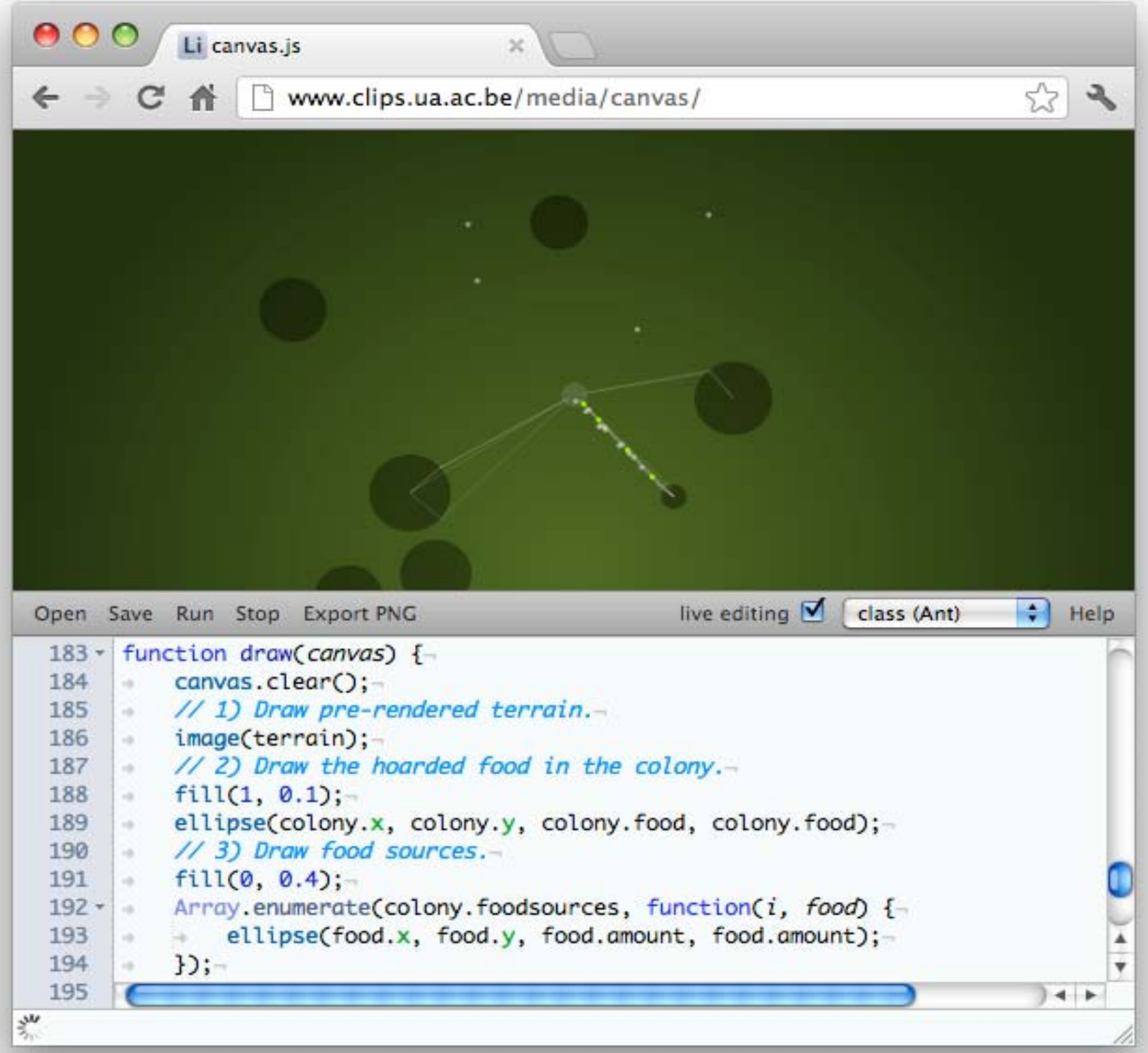

Figure 31. canvas.js editor. With live editing enabled, changes to the JavaScript code are instantly reflected in the animation that is currently running.

[28] http://www.clips.ua.ac.be/media/canvas

## 6.5 A Mad Tea-Party (creative thinking test revisited)

To conclude this chapter, we return to Guilford's Divergent Thinking test (DT) discussed in chapter 4. In a DT test, subjects are asked to come up with as many answers as possible to an open-ended problem solving task. In chapter 4 (section 4.6) we explained how we subjected 187 participants to a DT test ("How do you open a locked door if you don't have the key?"). Since the scoring process is essentially a language evaluation task, we should be able to replicate it using NLP and machine learning techniques. In other words, can we implement a system that is able to evaluate linguistic creativity? And subsequently, can we implement a system that generates answers that beat all answers given by the participants?

Creativity in the DT test is assessed as follows:

| | |
|---|---|
| **ORIGINALITY** | Unusual answers across all answers score `1`, unique answers score `2`. |
| **FLUENCY** | The number of answers. |
| **FLEXIBILITY** | The number of different categories of answers. |
| **ELABORATION** | The amount of detail.<br>For example, "break door" = `0` and "break door with a crowbar" = `1`. |

In our setup from chapter 4, we took each answer with a score greater than average + standard deviation as creative, and smaller as not creative. In the same way, we could implement a `creativity()` function that returns a score for each answer, then partition the scores into creative vs. not creative, and compare this outcome to the original outcome using precision and recall evaluation metrics. We obtain the best results if we slightly relax the threshold for creative answers = average + standard deviation × `0.95`.

Figure 32 shows the correlation between manual and automatic scores, where the jagged line shows the automatic scores. Following is a discussion of the automatic scoring implementation.

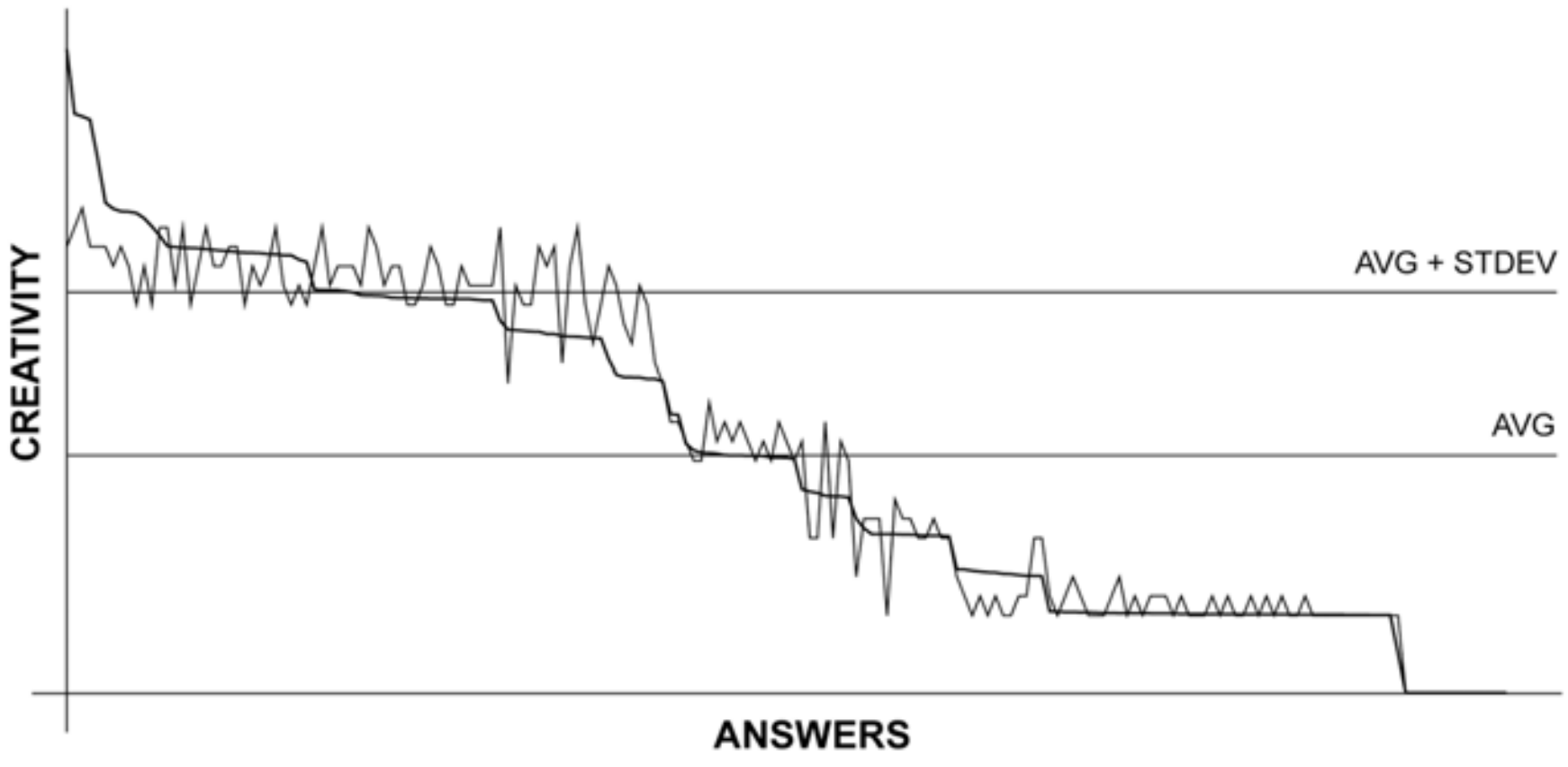

Figure 32. Correlation between the manual and automatic scores for each answer.
The jagged line shows the scores for the machine learner.

The **FLUENCY** score is simply the number of answers for a participant. The **ELABORATION** score corresponds to the number of prepositional phrases in each answer (e.g., “with a crowbar”). We can measure this by constructing a parse tree with MBSP or PATTERN:

```
from pattern.en import parsetree

def fluency(answers):
   return len(answers)

def elaboration(answers):
   return sum(min(len(parsetree(a)[0].pnp), 2) for a in answers)
```

If we only use these two scores to predict creative vs. not creative, we obtain a precision of `0.74` and a recall of `0.67` (combined F1-score of `0.70`). This is our performance baseline.

The **FLEXIBILITY** and **ORIGINALITY** scores rely on a categorization of the answer set. In chapter 4, the categorization was performed manually by two annotators. We can model this process using an unsupervised clustering analysis, where more closely related answers are assigned to the same cluster based on the words in their description. We use the hierarchical clustering algorithm in PATTERN. Assuming we have all the answers as a Python list of strings, we can transform the set into a sparse vector space where each vector is the word count of an answer:

```
from pattern.vector import Vector, count, words

answers = open('locked-door.txt').readlines()
vectors = []
for a in answers:
   v = count(words(a), stemmer='lemma')
   v = Vector(v)
   vectors.append(v)
```

Using the vector space we can then compute clusters of vectors, i.e., categories of similar answers. Since the number of clusters $k$ is arbitrary, we need to perform several iterations to see what value of $k$ yields the best results. In our setup, higher $k$ yields better results: more and smaller clusters have a smaller chance of covering multiple topics, perhaps including an original answer that should really belong to *no* cluster. Similarly, we are interested in “tight” clusters in which the vectors are near to each other. Loose clusters may be covering outliers. We define a “loose” cluster as a cluster with a high average distance of vectors to the center of the cluster. On our data, the best results are obtained with $k$=`250`, in combination with removing the top 50 loose clusters.

```
from pattern.vector import hierarchical, distance, centroid
from pattern.metrics import avg

def variance(cluster):
   return avg([distance(centroid(cluster), v) for v in cluster])

clusters = hierarchical(vectors, k=250, distance='cosine')
clusters = [isinstance(v, Vector) and [v] or v.flatten() for v in clusters]
clusters = sorted(clusters, key=variance)[:-50]

categories = {}
for i, cluster in enumerate(clusters):
   for v in cluster:
       categories[answers[vectors.index(v)]] = i
```

The **FLEXIBILITY** score is then defined as:

```
def flexibility(answers):
   return len(set(categories.get(a, a) for a in answers))
```

The **ORIGINALITY** score discerns between common, unusual and unique answers. We can define common answers as being part of a larger cluster, whereas unusual answers (top 5%) and unique answers (top 1%) correspond to smaller (or no) clusters. We need to know the relative probability of each cluster:

```
p = {}
for c in categories.values():
   p.setdefault(c, 0.0)
   p[c] += 1
s = sum(p.values())
for c in p:
   p[c] /= s
```

The **ORIGINALITY** score is then defined as:

```
def originality(answers):
   originality = 0
   for a in answers:
       if p.get(categories.get(a, a), 0) < 0.01:
           originality += 1
       if p.get(categories.get(a, a), 0) < 0.05:
           originality += 1
   return originality / (float(fluency(answers)) or 1)
```

Using the total score of `originality()`, `fluency()`, `flexibility()` and `elaboration()` for prediction, we obtain a precision of `0.72` and a recall of `0.83` (combined F1-score of `0.77`). Overall, the learning step has improved the accuracy by 7% over the baseline.

Since we can now automatically (approximately) predict creative answers, we can also generate new answers and evaluate how well they do. We can adopt many strategies to try and beat the answers given by the participants. For example, we can blend existing answers into new answers, employ Web search queries, WORDNET, and so on. But the simplest approach is to exploit a weakness in the automatic scoring implementation. This weakness has two angles.

First, as suggested by the high baseline, the scores emphasize the number of answers per participant and sentence length (i.e., the number of prepositional phrases) over originality. One can argue that the originality score is diminished by dividing it by the number of answers, but this is actually done to correct a known contamination problem in the DT test (Mouchiroud & Lubart, 2001). Second, the scores emphasize originality over appropriateness, since our system has no way to evaluate appropriateness (see chapter 4) as a factor.

This implies that longer answers are more creative even if they are nonsensical.

We can therefore beat the automatic scoring process by generating long, nonsensical sentences. We will use random nouns, verbs and adjectives linked to the *door* concept retrieved from the PERCEPTION semantic network. This is done to promote an air of meaningfulness: the retrieved words should in some way be semantically related to a door. However, as it turns out things quickly spiral out of control.

```
from pattern.graph.commonsense import Commonsense
from pattern.en import tag, wordnet, pluralize

from random import choice

concept = 'door'
g = Commonsense()
g = g[concept].flatten(4)

tagged = [tag(node.id)[0] for node in g]

JJ = [w for w, pos in tagged if pos == 'JJ']
NN = [w for w, pos in tagged if pos == 'NN']
VB = [w for w, pos in tagged if w.encode('utf-8') in wordnet.VERBS]

for i in range(5):
   a = '%sly %s the %s with a %s %s of %s %s.' % (
        choice(JJ).capitalize(),
        choice(VB),
        concept,
        choice(JJ),
        choice(NN),
        choice(JJ),
        pluralize(choice(NN)))
   print a
```

We used some additional post-processing to polish the output: the first word is capitalized, and the last noun is pluralized using the `pluralize()` function in PATTERN. This yields the following set of nonsensical answers, for example:

**Question**: How do you open a locked door?

"Stubbornly club the door with a graceful albatross of pungent swamps."
"Unpleasantly court the door with a massive exhibition of busy colors."
"Passionately pet the door with a weak obstacle of skeptical physicians."
"Rawly storm the door with a dry business lunch of primitive crustaceans."
"Sensationally duck the door with a fast locomotion of large grandmothers."

Marvelous, all we need now is an albatross. When we include the generator in the cluster analysis on our data and calculate its score, it beats all other scores by at least 15% since it has the maximum number of answers, each with two prepositions, and flavored with adjectives for the reader's enjoyment. As discussed above, the high score is the result of the system's inability to discern between useful and nonsensical answers. This also suggests that the DT test may benefit from a more explicit directive for scoring appropriateness.

## 6.6 Discussion

In this chapter we have discussed a statistical approach used in natural language processing and machine learning called the vector space model. In this approach, natural language is transformed into a matrix of numeric values. Statistical functions can then be calculated on the matrix to predict the similarity of vectors in the matrix, which represent sentences, paragraphs or text (or pixels, genes, EEG brain activity, and so on). We have presented two Python software packages developed in the course of our work that implement this approach: MBSP FOR PYTHON and PATTERN. Both packages are open source.

MBSP FOR PYTHON is a Python implementation of the MBSP memory-based shallow parser for English, which detects word types (part-of-speech tagging), word chunks and relations between chunks in text. In terms of modeling creativity, part-of-speech tags can offer valuable linguistic information. For example, to evaluate what adjectives are commonly associated with what nouns or what words constitute *x* and *y* in "*x* is a new *y*". Such information can be used (for example) to populate a semantic network of common sense or to acquire a model for sentiment analysis.

PATTERN is a Python package for web mining, natural language processing, machine learning and graph analysis. With a combination of PATTERN and MBSP it would not be difficult to construct an endless learning PERCEPTION (chapter 5) with information extracted from actual language use instead of relying on the intuition of an annotator. Using PATTERN's classification tools it is possible to replace PERCEPTION entirely with a statistical approach. Using its clustering tools it is possible to predict novelty, i.e., the instances left unclustered after the analysis. In recent work, Shamir & Tarakhovsky (2012) for example have used unsupervised learning to detect the artistic style of approximately 1,000 paintings, in agreement with the opinion of art historians and outperforming humans not trained in fine art assessment.

Does statistical AI model how the human mind works? It is hard to say for sure. But the state-of-the-art offers a robust approach to many practical tasks in AI and creativity. Whenever such applications yield results, the threshold of what contributes to human intelligence and creativity is raised. For example, when a chess computer beats a human we are quick to remark that it plays a boring chess or that it can't write poetry or translate Chinese. When a fine art learner reaches the level of an art historian, we respond with: "I'll wager I can think of a painting it can't handle... See! It doesn't work."

Many AI problems such as natural language understanding and machine translation are considered to be AI-complete. They may require that all other AI problems are solved as well. Many AI researchers hope that bundling the individual applications will one day lead to emergent strong AI. Futurists like Kurzweil (2005) argue that this would require artificial consciousness. Kurzweil believes that once machines reach this stage they will be powerful enough to solve (among other) all known medical problems, but such conjectures have also been criticized.

# 7 Sentiment analysis

Textual information can be broadly categorized into three types: objective facts and subjective opinions (Liu, 2010) and writing style. Opinions carry people's sentiments, appraisals and feelings toward the world. Before the World Wide Web, opinions were acquired by asking friends and families, or by polls and surveys. Since then, the online availability of opinionated text has grown substantially. Sentiment analysis (or opinion mining) is a field that in its more mature work focuses on two main approaches. The first approach is based on subjectivity lexicons (Taboada, Brooks, Tofiloski, Voll & Stede, 2011), dictionaries of words associated with a positive or negative sentiment score (called polarity or valence). Such lexicons can be used to classify sentences or phrases as subjective or objective, positive or negative. The second approach is by using supervised machine learning methods (Pang & Vaithyanathan, 2002).

Resources for sentiment analysis are interesting for marketing or sociological research. For example, to evaluate customer product reviews (Pang & Vaithyanathan, 2002), public mood (Mishne & de Rijke, 2006), electronic word-of-mouth (Jansen, Zhang, Sobel & Chowdhury, 2009) and informal political discourse (Tumasjan, Sprenger, Sandner & Welpe, 2010).

## 7.1 A subjectivity lexicon for Dutch adjectives

In De Smedt & Daelemans (2012) we describe a subjectivity lexicon for Dutch adjectives integrated with PATTERN and compatible with CORNETTO, an extension of the Dutch WordNet (Vossen, Hofmann, de Rijke, Tjong Kim Sang & Deschacht, 2007). Esuli & Sebastiani (2006) note that adverbs and adjectives are classified more frequently as subjective (40% and 36%) than verbs (11%). In our approach we focus on adjectives. We used a crawler to extract adjectives from online Dutch book reviews and manually annotated them for polarity, subjectivity and intensity strength. We then experimented with two machine learning methods for expanding the initial lexicon, one semi-supervised and one supervised. Each of the book reviews has an accompanying, user-given "star rating" (1–5), which we used to evaluate the lexicon.

### Manual annotation

Since adjectives with high subjectivity will occur more frequently in text that expresses an opinion, we collected 14,000 online Dutch book reviews in which approximately 4,200 Dutch adjective forms occurred. The texts were mined with PATTERN and part-of-speech tagged with FROG (van den Bosch, Busser, Canisius & Daelemans, 2007). We did not apply lemmatization at this stage and therefore some adjectives occur both in citation and inflected form, e.g., "goed<u>e</u>" vs. "goed" (good). The adjective frequency approximates an exponential Zipf distribution, with "goed" being the most frequent (6000+ occurrences), followed by "echt" (real, 4500+) and "heel" (very, 3500+). The top 10% constitutes roughly 90% of all occurrences. We took the top 1,100 most frequent adjectives, or all adjectives that occurred more than four times.

Seven human annotators were asked to classify each adjective in terms of positive–negative polarity and subjectivity. In Esuli & Sebastiani (2006) adjectives are not only classified in terms of polarity and subjectivity but also per word sense, since different senses of the same term may have different opinion-related properties. For example, crazy ≈ insane (negative) vs. crazy ≈ enamored (positive). We adopt a similar approach where annotators assessed word senses using a triangle representation (figure 33). The triangle representation implies that more positive or more negative adjectives are also more subjective. But not all subjective adjectives are necessarily positive or negative, e.g., "blijkbaar" (apparently). We used CORNETTO to retrieve the different senses of each adjective. This is to our knowledge the first subjectivity lexicon for Dutch that applies sense discrimination.

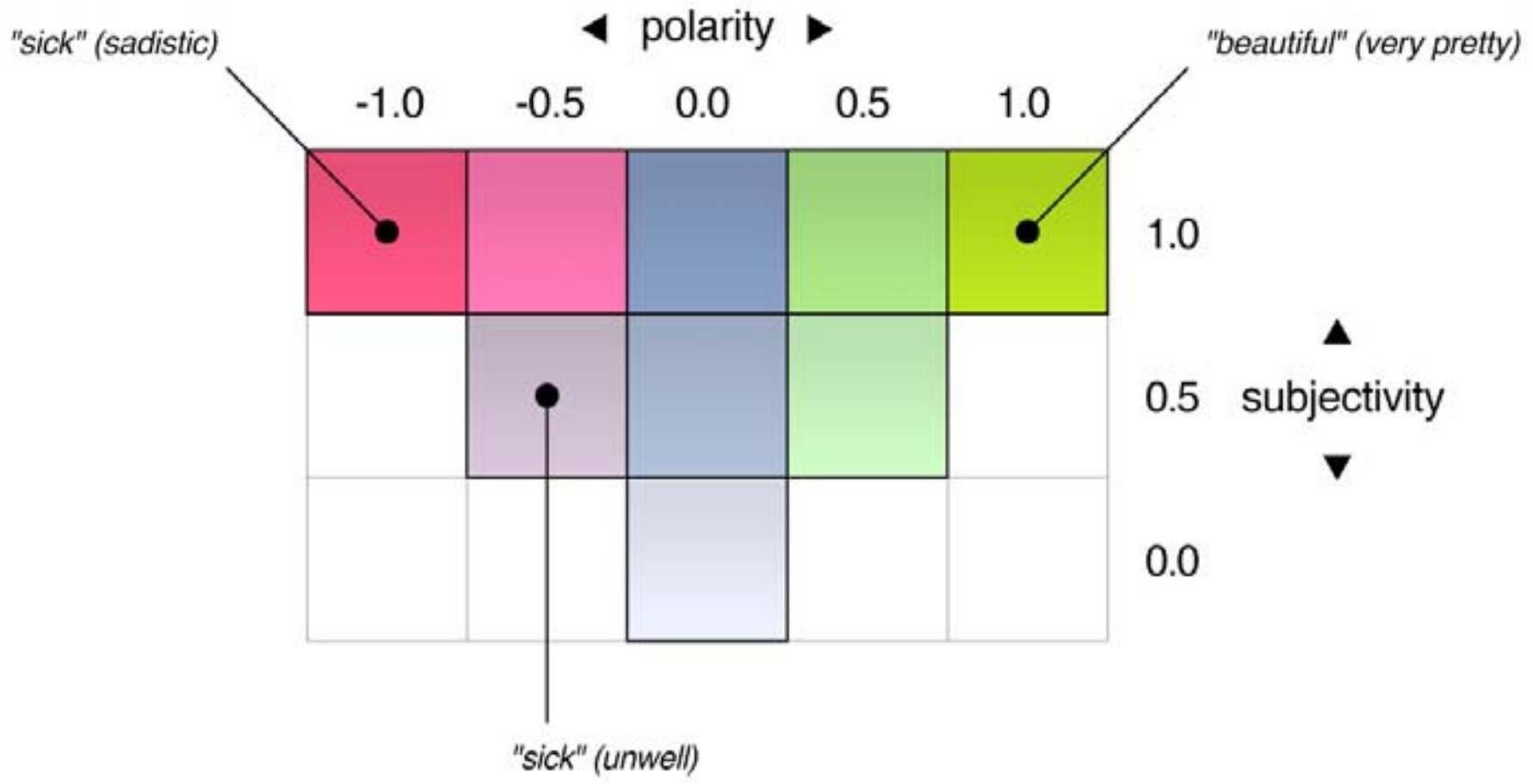

Figure 33. Triangle representation with polarity and subjectivity axes.

Dutch adjectives can be used as adverbs, where in English the ending `-ly` is usually required. For example: "ongelooflijk goed" is understood as "incredibly good" and not as "unbelievable" + "good". Annotators were asked to provide an additional intensity value, which can be used as a multiplier for the successive adjective's polarity.

#### AGREEMENT

We removed a number of inflected adjectives, spelling errors and adverbs, bringing the final GOLD1000 lexicon to about 1,000 adjectives (1,500 word senses) with the average scores of the annotators. The lexicon contains 48% positive, 36% negative and 16% neutral assessments. Figure 34 shows a breakdown of the distribution. Table 14 shows the inter-annotator agreement (Fleiss' kappa). We attain the best agreement for positive–neutral–negative polarity without considering polarity strength ($\varkappa$=`0.63`). Assessment of subjectivity strength is shown to be a much harder task ($\varkappa$=`0.34`), perhaps because it is more vague than classifying positive versus negative.

The DUOMAN subjectivity lexicon for Dutch (Jijkoun & Hofmann, 2009) contains 5,000+ words with polarity assessments by two annotators. About 50% of the adjectives in GOLD1000 also occur in DUOMAN. We compared the positive–neutral–negative polarity (without strength) of the adjectives in GOLD1000, using the average of word senses, to those that also occur in DUOMAN.

Agreement is $\varkappa$=`0.82`. Some adjectives are positive in GOLD1000 but negative in DUOMAN, or vice-versa. One explanation is the different semantics between Dutch in the Netherlands and Dutch in Flanders for some adjectives (e.g., “maf”). Agreement increases to $\varkappa$=`0.93` when these aberrant 27 adjectives are omitted from the measurement.

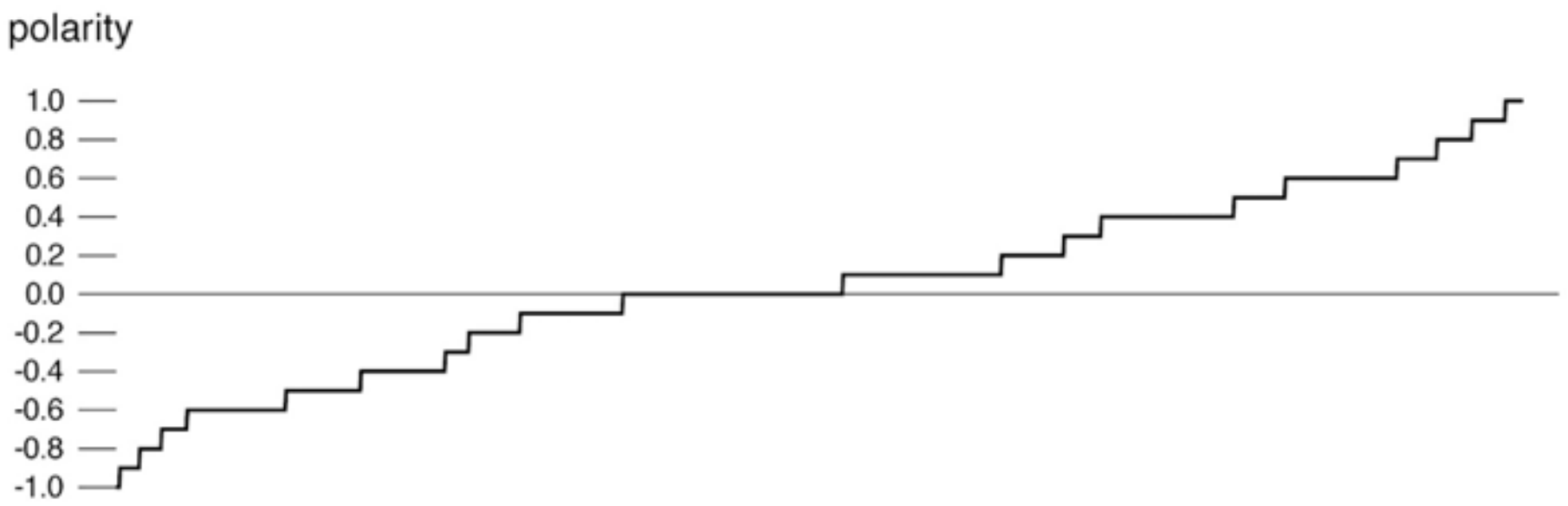

Figure 34. Distribution of positive-negative polarity strength in GOLD1000.

| RATING | K |
|---|---|
| polarity (`-1` or `0` or `+1`) | `+0.63` |
| polarity (`-1.0` to `+1.0`) | `+0.47` |
| polarity + subjectivity | `+0.30` |
| subjectivity | `+0.34` |
| intensity | `+0.32` |

Table 14. Agreement for seven annotators.

## Automatic expansion

#### DISTRIBUTIONAL EXTRACTION

There is a well-known approach in computational linguistics in which semantic relatedness between words is extracted from distributional information (Schütze & Pedersen, 1997). For an example for Dutch, see Van de Cruys (2010). Van de Cruys uses a vector space with adjectives as document labels and nouns as vector features. The value for each feature represents the frequency an adjective precedes a noun. Classification or dimension reduction is then used to create groups of semantically related words. We applied the distributional approach to annotate new adjectives. From the Dutch TWNC newspaper corpus (Ordelman et al., 2002), we analyzed 3,000,000 words and selected the top 2,500 most frequent nouns. For each adjective in TWNC that is also in CORNETTO, we counted the number of times it directly precedes one or more of the top nouns. This results in 5,500+ adjective vectors with 2,500 vector features. For each GOLD1000 adjective we then applied *k*-NN using cosine similarity to retrieve the top 20 most similar nearest neighbors from the set. For example, for “fantastisch” (fantastic) the top five nearest neighbors are: “geweldig” (great, 70%), “prachtig” (beautiful, 51%), “uitstekend” (excellent, 50%), “prima” (fine, 50%), “mooi” (nice, 49%) and “goed” (good, 47%). The best nearest neighbors were handpicked by 2 annotators. These inherit polarity, subjectivity and intensity from their GOLD1000 anchor. The AUTO3000 lexicon then contains about 3,200 adjectives (3,700 word senses).

**SPREADIING ACTIVATION**

We then iteratively extend the lexicon by traversing relations between CORNETTO synsets. In CORNETTO or in WORDNET, word senses with a similar meaning are grouped in synsets. Different synsets are related to each other by synonymy (**is-same-as**), antonymy (**is-opposite-of**), hyponymy (**is-a**), and so on. For each adjective word sense we retrieve the CORNETTO synset and inherit polarity, subjectivity and intensity to related word senses. In three iterations we can spread out to over 2,200 new adjectives (3,000 word senses). The AUTO5500 lexicon contains about 5,400 adjectives (6,675 word senses).

## Evaluation

For evaluation we tested with a set of 2,000 Dutch book reviews, which were not used in the GOLD1000 lexicon and evenly distributed over negative opinion (star rating 1 and 2) and positive opinion (4 and 5). For each review, we then scored polarity for adjectives that occur in the lexicon and compared the average strength to the original star rating. We took polarity `>= +0.1` as a positive observation and polarity `< +0.1` as a negative observation. This is a form of binary classification in which there will be a number of correctly classified words (true positives and negatives) and incorrectly classified words (false positives and negatives) by which we can calculate precision and recall. The polarity threshold can be lowered or raised, but `+0.1` yields the best results. Overall we attain a precision of 72% and a recall of 78%.

We then used the intensity strength. Instead of scoring "echt teleurgesteld" (truly disappointed) as "echt" (true) + "teleurgesteld" (disappointed) = `0.2` + `-0.4` = `-0.2`, we used "echt" (intensity `1.6`) × teleurgesteld = `1.6` × `-0.4` = `-0.64`. For book reviews, this increases recall to 82% without affecting precision. To provide a cross-domain measurement we tested with 2,000 Dutch music reviews evenly distributed by star rating. We attain a precision of 70% and a recall of 77%.

| **positive >= 0.1** | | | **BOOKS2000** | | |
|---|---|---|---|---|---|
| | # adjectives | A | P | R | F1 |
| GOLD1000 | 794 | 0.75 | 0.72 | 0.82 | 0.77 |
| AUTO3000 | 1,085 | 0.75 | **0.72** | **0.82** | 0.77 |
| AUTO5500 | 1,286 | 0.74 | 0.72 | 0.79 | 0.75 |

| **positive >= 0.1** | | | **MUSIC2000** | | |
|---|---|---|---|---|---|
| | # adjectives | A | P | R | F1 |
| GOLD1000 | 794 | 0.75 | 0.72 | 0.82 | 0.77 |
| AUTO3000 | 1,085 | 0.75 | **0.72** | **0.82** | 0.77 |

Table 15. Number of unique adjectives rated, accuracy, precision, recall and F1-scores for opinion prediction.

The results for the GOLD1000 and AUTO3000 lexicons are nearly identical. The reason that precision and recall do not increase by adding more adjectives is that 94.8% of top frequent adjectives is already covered in GOLD1000. As noted, the distribution of adjectives over book reviews is exponential: the most frequent "goed" (good) occurs 6,380 times whereas for example "aapachtig" (apish) occurs only one time. Adding words such as “aapachtig” has a minimal coverage effect. Nevertheless, distributional extraction is a useful expansion technique with no drawback as indicated by the stable results for AUTO3000. For AUTO5500, F-score is less (-2%). The reason is that the adjectives “een” (united), “in” (hip) and “op” (exhausted) are part of the expanded lexicon. In Dutch, these words can also function as a common determiner (een = a/an) and common prepositions (in = in, op = on). Without them the scores for AUTO5500 come close to the results for GOLD1000 and AUTO3000. This suggests that the CORNETTO expansion technique can benefit from manual correction and part-of-speech tagging. Combined with part-of-speech tagging, this technique could be eligible for on-the-fly classification of unknown adjectives, since it can be implemented as a fast PATTERN tagger + CORNETTO traversal algorithm.

Using regular expressions for lemmatization, negation (reverse score for “niet”, “nooit” and “geen”) and exclamation marks, precision and recall for books increases to 77% and 84% respectively.

## Analysis

In overall, positive predictions (57%) were more frequent than negative predictions (43%). By examining the test data, we see three reasons why it may be harder to identify negative opinions:

**COMPARISON** Some negative opinions make their point by referring to other instances, for example: “dat boek was grappig, origineel, pakkend, maar dit boek vond ik op al die punten tegenvallen” (that book was funny, inspiring, moving, but this book fails on all those points). The adjectives rate as positive but the review is negative.

**FEATURAL OPINION** In “de eerste tien pagina’s zijn sterk, maar dan zakt het als een pudding in elkaar” (the first ten pages are quite good, but it falls apart from thereon) the positive opinion accounts for a specific feature of the book (first ten pages), while the general negative opinion is carried by a figure of speech (falling apart).

**IRONY** It is not possible to detect irony using a subjectivity lexicon. For example, “zou niet weten wat ik met mijn leven moest als ik dit geweldige boek gemist had” (wouldn't know what to do with my life if I had missed this awesome book).

## English translation

Many word senses in CORNETTO have inter-language relations to WORDNET. We took advantage of this to map the polarity and subjectivity scores in the Dutch lexicon to an English lexicon. We then tested our English lexicon against POLARITY DATASET 2.0 (Pang & Lee, 2004) containing a 1,000 positive and a 1,000 negative IMDb movie reviews (imdb.com). Initial test results were poor: 66% precision and 54% recall. If we look at the 1,000 top frequent adjectives in 3,500 random English IMDb movie reviews, only 32% overlaps with the Dutch most frequent adjectives. We proceeded to manually annotate about 500 frequent English adjectives (1,500 word senses) to expand the English lexicon. This was done by a single annotator, but the effect is apparent: precision increases to 72% and recall to 71%. Currently, the F-score in PATTERN is 74%.

In the next section, we present a case study using the Dutch lexicon on online political discourse.

## 7.2 Sentiment analysis for political discourse

In 2011, Belgium experienced the longest government formation period known to modern-day democratic systems (BBC Europe, 2011). At the basis lies the dual federalist structure in Belgium, the two major regions being the Dutch-speaking Flemish Community and the French-speaking Walloon Community. In 2010, the Dutch-speaking right-wing N-VA (Nieuw-Vlaamse Alliantie) emerged both as newcomer and largest party of the Belgian federal elections. The second largest party was the French-speaking left-wing PS (Parti Socialiste). While the N-VA ultimately seeks secession of Flanders from Belgium, the PS is inclined towards state interventionism. Over the following year they were unable to form a government coalition. It eventually took a record-breaking 541 days of negotiations to form a government. The media has played an important and sometimes controversial role in the course taken by the political parties, and passionate assertions have been expressed towards the media and its favoritism (Niven, 2003). However, the question whether there truly exists a media bias or not should be answered by systematic, statistical analysis.

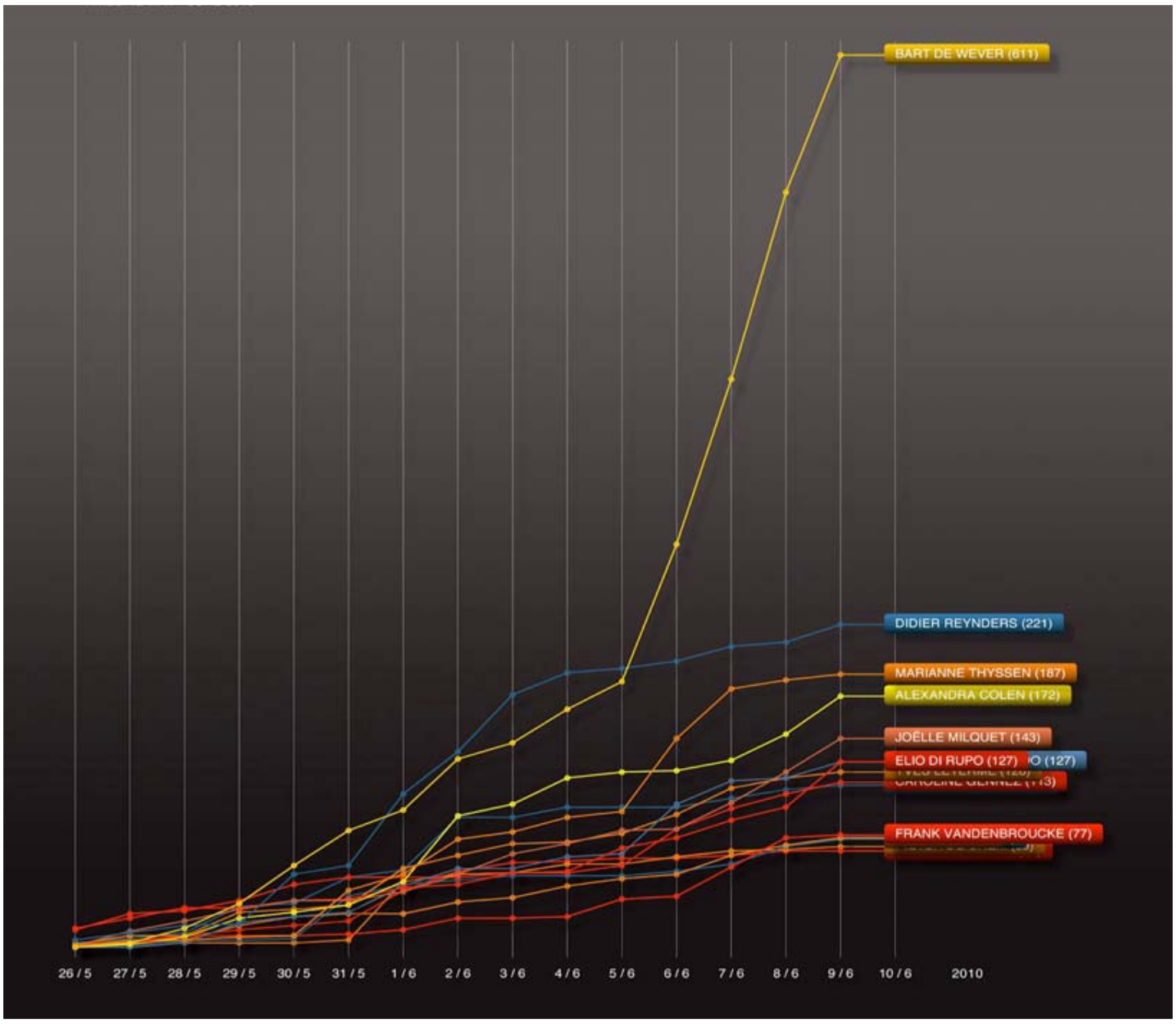

Figure 35. Timeline of Belgian political tweets, federal elections, May 26 to June 9 2010.
Highest ranking is "Bart De Wever" of the N-VA.

Nowadays, newspapers and other news providers are updating their online news feeds in near real-time, allowing those interested to follow the news in near real-time. The Web has thereby become an important broadcast medium for politicians. In Junqué de Fortuny, De Smedt, Martens & Daelemans (2012) we describe a text mining system that acquires online news feeds for Belgian political articles. We have used this system together with NODEBOX to visualize how the tone of reporting evolved throughout 2011, on party, politician and newspaper level.

In a first test[29] we were able to predict key figures in the government formation (e.g., Bart De Wever and Elio Di Rupo) by mining 7,500 Twitter messages containing the name of Belgian politicians prior to the 2010 elections. This is shown in figure 35.

## Media bias

A study by Benewick, Birch, Blumler & Ewbank (1969) shows that exposure to a political party's broadcasts is positively related to a more favorable attitude towards the party for those with medium or weak motivation to follow the campaign. Knowing this, McCombs & Shaw (1972) raise the question whether mass media sources actually reproduce the political world perfectly. In an imperfect setting, media bias could propagate to the public opinion and influence favoritism towards one or another party or politician, thus shaping political reality. Most of this bias is introduced by the selection and editing process. McCombs & Shaw conclude that mass media may well set the agenda of political campaigns. D'Alessio & Allen (2000) found three main bias metrics used to study partisan media bias:

#### GATEKEEPING

Selection, editing and rejection of news articles to be published introduces a bias. This so-called gatekeeping causes some topics to never surface in the media landscape. It is therefore a sampling bias introduced by editors and journalists (Dexter & White, 1964). However, measuring the gatekeeping bias turns out to be infeasible, since information about rejection is not available.

#### COVERAGE

Media coverage is the amount of attention that each side of the issue receives. We can measure coverage as the amount of online articles for each entity (i.e., a party or politician). We argue that fair coverage is determined by an *a priori* distribution. The *a priori* distribution represents all entities in the population by their relative importance, as measured by electoral votes in the latest elections. Deviations from the fair distribution show bias towards one or another entity.

#### STATEMENT

Media coverage can be favorable or unfavorable. We can use the subjectivity lexicon in PATTERN to measure statement bias in terms of positive–negative coverage. This is an associative measure: a party is associated with a negative image when most of its coverage contains negative content. However, when an online article is classified as negative, this does not necessarily imply favoritism from a news source or a journalist. The entity may be purposely interjecting criticism or the article may contain citations from rival entities.

[29] http://www.clips.ua.ac.be/pages/pattern-examples-elections

## Web crawler + sentiment analysis

We used the 2010 Chamber election results[30] as a gold standard against which we measured media coverage bias. The rationale is that the media is aware of these results and therefore knows the distribution of political parties beforehand. Similarly, we used the 2010 Senate election results to compare politicians. Politicians are voted for directly in the Senate elections whereas Chamber elections concern party votes. We then implemented a crawler to obtain a corpus of over 68,000 unique online articles from all major Flemish newspapers, spanning a ten-month period (January 1, 2011 to October 31, 2011). We can see how this is not hard to accomplish by following the steps in the pattern.web examples in chapter 6.4. Each article in the corpus contains the name of at least one Flemish political party or leading figure. The criterion for being a political party of interest is based on the votes for that party in the 2010 Chamber elections, that is, we only include parties voted into the Chamber. A leading figure of interest is a politician with a top ten ranking amount of preference votes in the 2010 Senate elections.

We analyzed the occurrences of each entity in each newspaper article. The sentiment associated with an entity in an article is the polarity of adjectives, in a window of two sentences before and two sentences after. The window is important, because an article can mention several entities or switch tone. It ensures a more reliable correlation between the entity and the polarity score, contrary to using all adjectives across the article. A similar approach with a 10-word window is used in Balahur et al. (2010). They report improved accuracy compared to measuring all words. We also excluded adjectives that score between `-0.1` and `+0.1` to reduce noise. This results in a set of about 365,000 assessments, where one assessment corresponds to one adjective linked to a party or politician.

For example:

> Bart De Wever (N-VA) verwijt de Franstalige krant Le Soir dat ze aanzet tot haat, omdat zij een opiniestuk publiceerde over de Vlaamse wooncode met daarbij een foto van een Nigeriaans massagraf. De hoofdredactrice legt uit waarom ze De Wever in zijn segregatiezucht hard zal blijven bestrijden. “Verontwaardigd? Ja, we zijn eerst en boven alles verontwaardigd.”
>
> Bart De Wever (N-VA) accuses the French newspaper Le Soir of inciting hate after they published an opinion piece on the Flemish housing regulations together with a picture of a Nigerian mass grave. The editor explains why they will continue to fight De Wever’s love of segregation firmly. “Outraged? Yes, we are first and above all outraged.”

The adjectives “Franstalig”, “Vlaams” and “Nigeriaans” are neutral, as are all adjectives in our subjectivity lexicon that indicate a region or a language. The adjective “hard” (firm) scores `-0.03` and is excluded. The adjective “verontwaardigd” (outraged) scores `-0.4`. In overall the item on the political party N-VA is assessed as negative.

[30] Federal Public Services Home Affairs: http://polling2010.belgium.be/
All percentages in this study were normalized over the Flemish parties and the selection.

## Evaluation

The *coverage* of an entity by a newspaper source is the number of articles published by the source on that entity, divided by the total amount of articles by that source in the corpus. The *popularity* of an entity is the relative number of votes (Chamber/Senate). This is the fair distribution. The *coverage bias* of a newspaper is the difference between the real distribution and the fair distribution. Table 16 shows the newspaper sources used for the analysis with their respective number of articles, number of readers and the coverage bias:

| SOURCE | REGIONAL | # ARTICLES | # READERS | BIAS |
|---|---|---|---|---|
| De Morgen | NO | 11,303 | 256,800 | 21.53% |
| De Redactie | NO | 5,767 | 146,250 | 16.16% |
| De Standaard | NO | 9,154 | 314,000 | 23.32% |
| De Tijd | NO | 10,061 | 123,300 | 22.71% |
| GVA | YES | 9,154 | 395,700 | **27.30**% |
| HBVL | YES | 3,511 | 423,700 | **37.43**% |
| HLN | NO | 11,380 | 1,125,600 | 21.62% |
| Het Nieuwsblad | NO | 7,320 | 1,002,200 | 20.24% |

Table 16. Newspaper source, number of articles, number of readers and coverage bias.

The highest coverage bias is found for the regional newspapers Gazet van Antwerpen (GVA) and Het Belang van Limburg (HBVL). Both newspaper are owned by the media group Concentra. De Morgen, De Tijd and Het Laatste Nieuws (HLN) are owned by De Persgroep. De Standaard and Het Nieuwsblad are owned by Corelio. De Redactie is an online news channel owned by VRT. It has the lowest coverage bias, but it has a high statement bias, as we will see.

Figure 36 shows how a coverage bias is found for some parties, with a maximal positive bias towards CD&V and a maximal negative bias towards the far-right Vlaams Belang (VB). The first result can be justified by the fact that CD&V was running the interim government while the new government formations took place. The latter result gives supportive evidence that the party is being quarantined by the media (Yperman, 2004). These tendencies propagate to the local level, for example Vlaams Belang is under-represented in all newspapers. Some local differences exist as well: the regional newspaper HBVL has a large negative bias towards N-VA. Regional newspapers in general (GVA, HBVL) have a larger coverage bias.

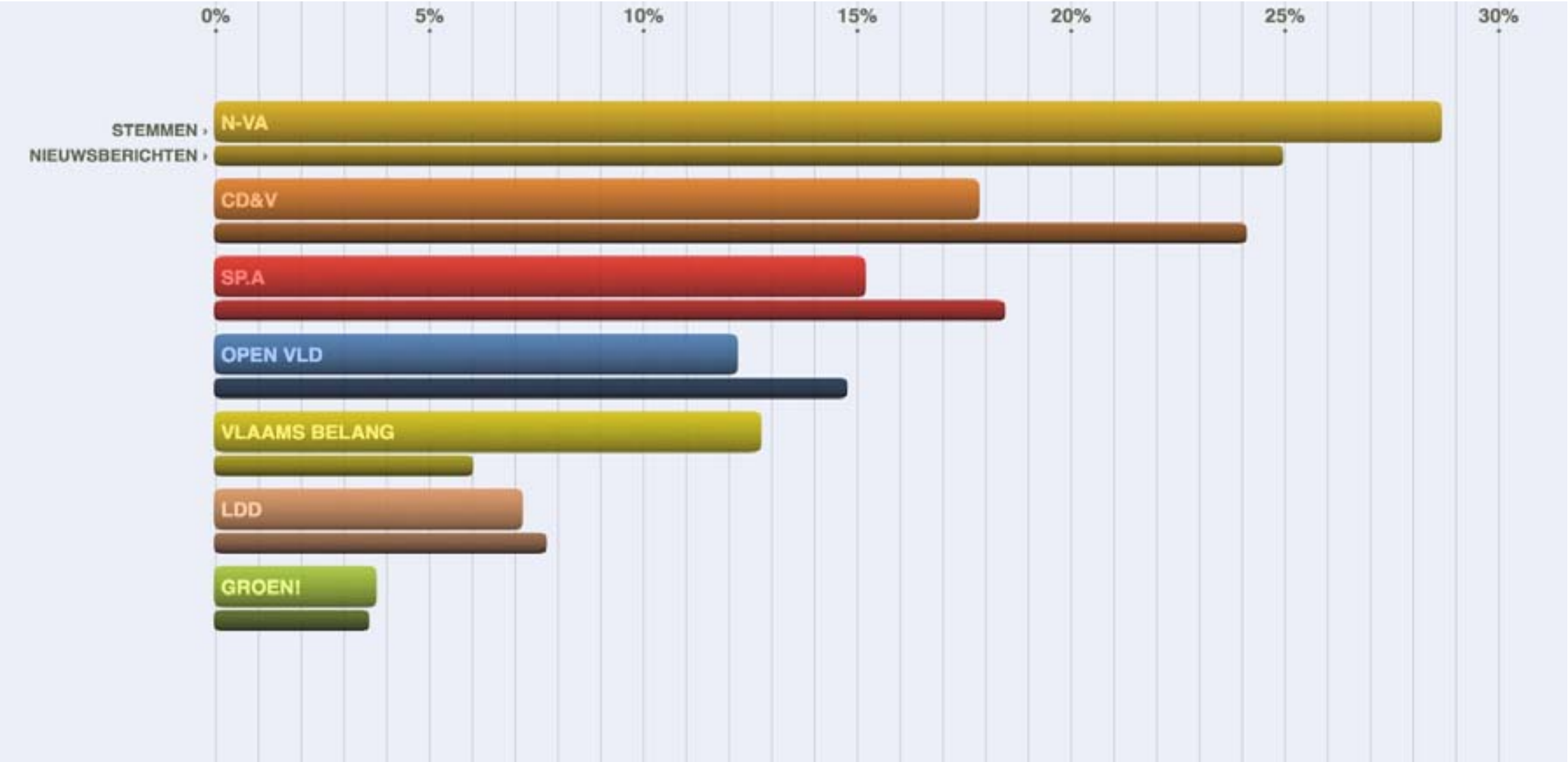

Figure 36. Discrepancy between media coverage and popularity for political parties.
The wide bar represents popularity, the narrow bar media coverage.

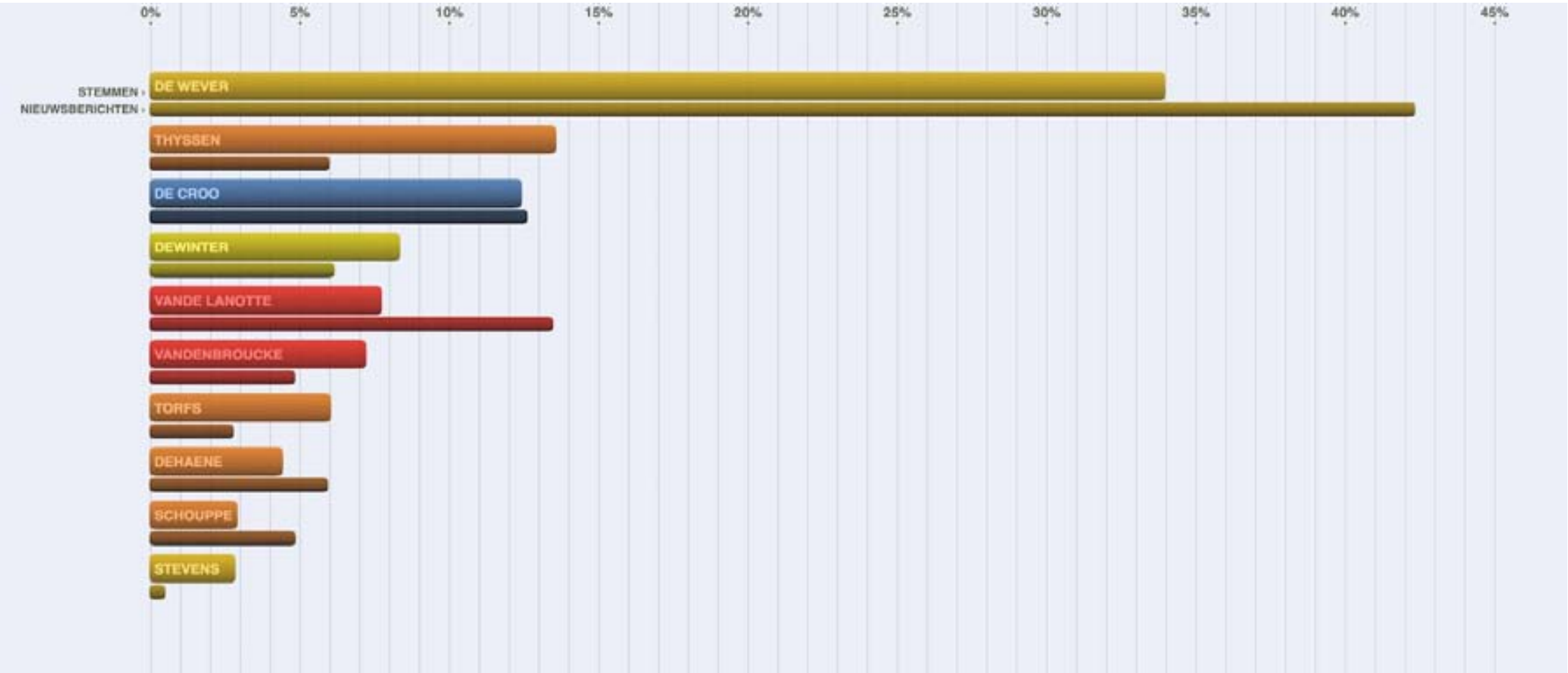

Figure 37. Discrepancy between media coverage and popularity for politicians.
The wide bar represents popularity, the narrow bar media coverage.

Figure 37 shows that the coverage bias for individual politicians varies irrespective of the party they represent. For example, a positive bias for Bart De Wever is not reflected in the negative bias for his party (N-VA).

Figure 38 shows the distribution of positive–negative sentiment for each political party. Overall, 20–30% of newspaper coverage is assessed as negative. Some statement bias is present under the assumption of uniformity of the sentiment distribution. Highest negative scores are measured for the far-right Vlaams Belang (–30.3%) and for the N-VA (–28.7%).

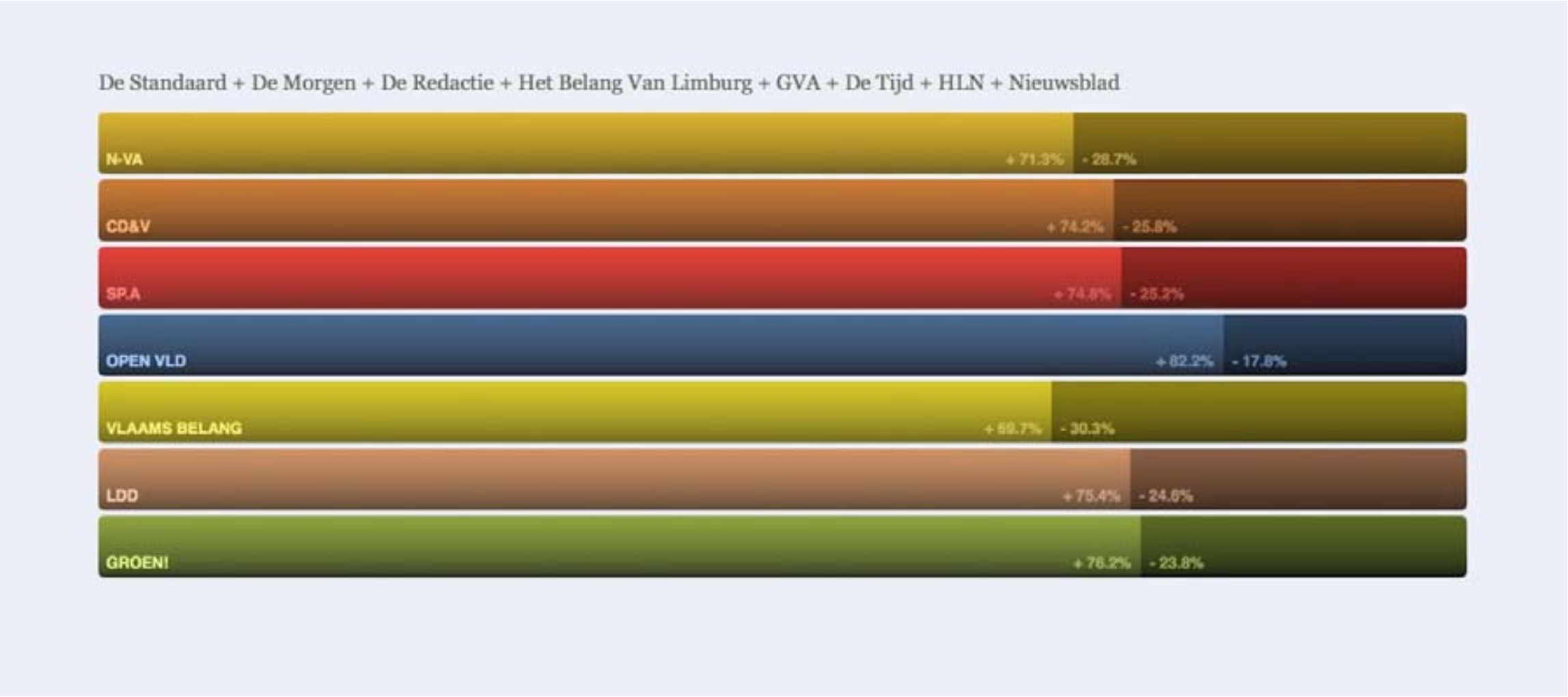

Figure 38. Sentiment for each political party, with the percentage of positive coverage on the left and negative on the right.

For each given political party, we then grouped sentiment assessments in subsets of one week and constructed a timeline of the consecutive weeks. We calculated a simple moving average (SMA) across all weeks to smoothen fluctuation in individual parties and emphasize the differences across parties.

Figure 39 shows the SMA for each political party across all newspapers. It is interesting to note the peak with all parties (except Vlaams Belang) in July-August. At this time the negotiating parties involved in the government formation were on a three-week leave. Once the negotiations resumed around August 15th, the peak drops. In the Belgian political crisis, it seems, no news equals good news.

Figure 40 shows the SMA for each newspaper across all political parties. The curves with the highest fluctuation are those for Het Belang Van Limburg and De Redactie. Het Belang Van Limburg also has the highest average sentiment: `+0.15` against [`0.13`, `0.14`] for all other newspapers. De Standaard appears to deliver the most neutral political articles.

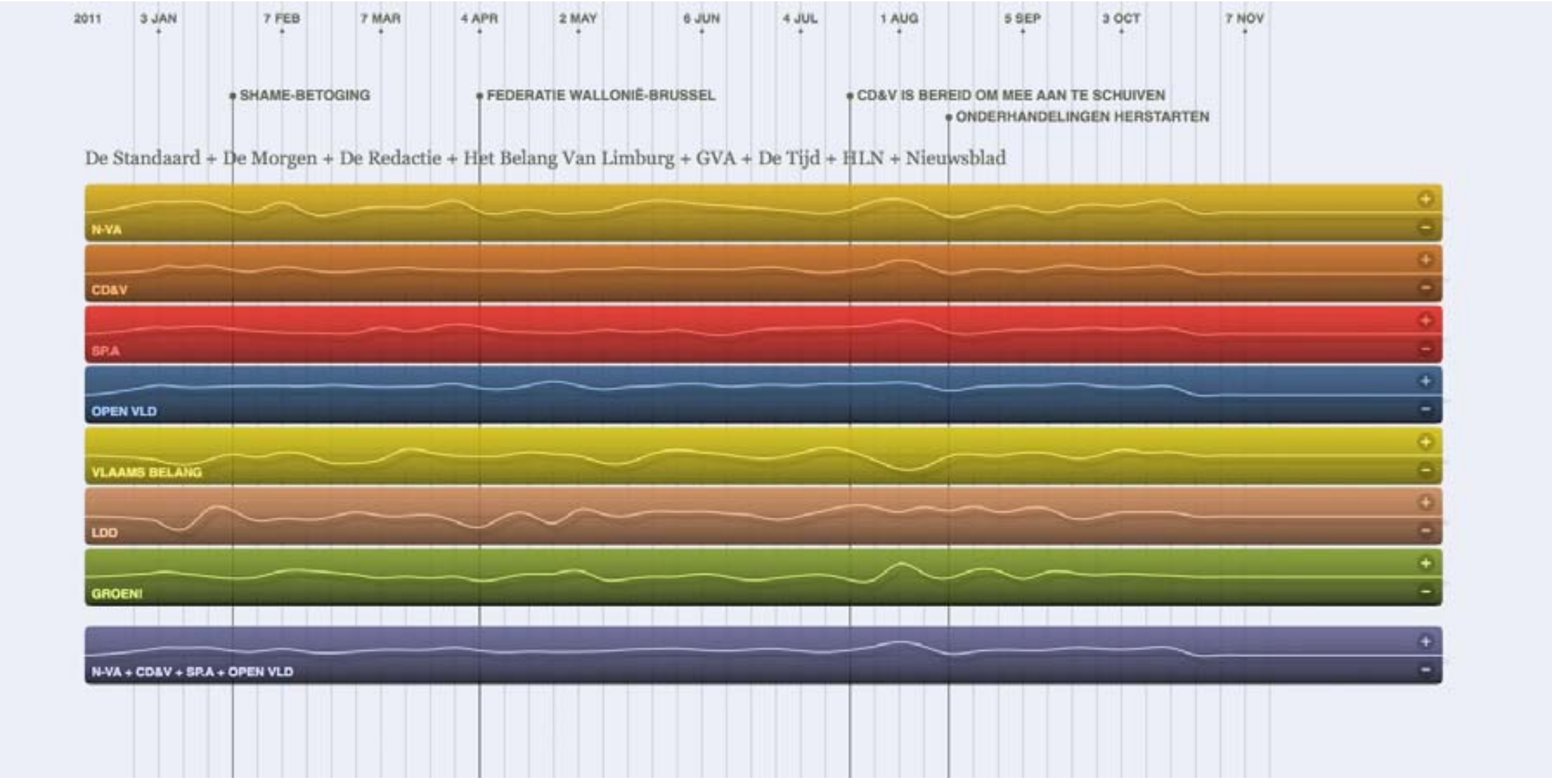

Figure 39. Sentiment of news items in 2011 for each party.

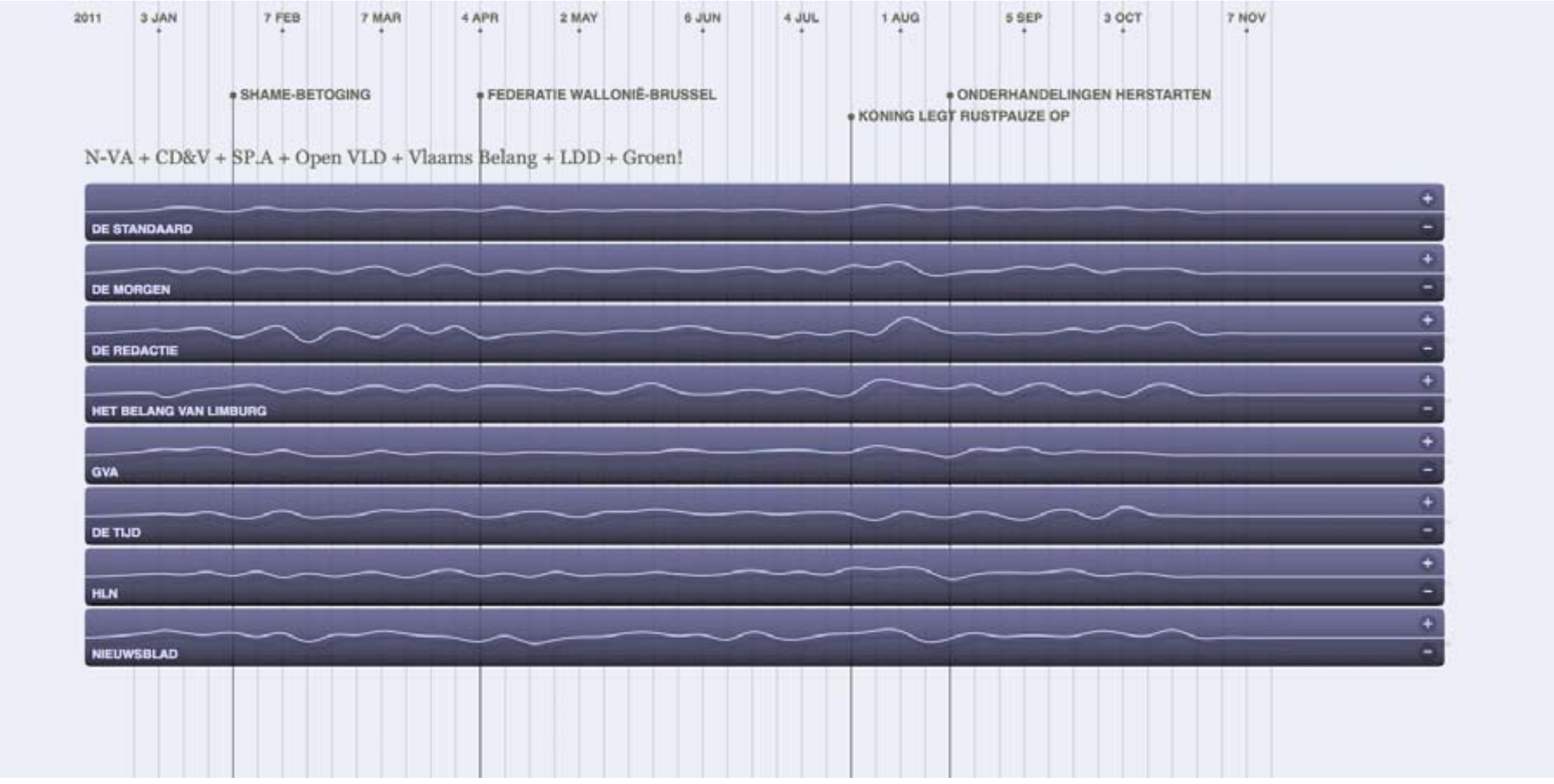

Figure 40. Sentiment of news items in 2011 for each newspaper.

In summary, quantitative analysis shows that a media coverage bias does indeed exist, most notably in regional newspapers. The figures provide support for the claim that some statement bias exists in the mass media. Statement bias is most notably found in the online news channel of De Redactie and the regional newspaper Het Belang van Limburg. Methodologically, we have shown that sentiment expert systems are a viable strategy for research in political text corpora.

## 7.3 Discussion

In this chapter we demonstrated how the PATTERN package discussed in chapter 6 can be used to build a subjectivity lexicon for sentiment analysis. In summary, we mined the Web for product reviews and annotated the most frequent adjectives with a positive–negative score. We applied two techniques (distributional extraction and spreading activation) to expand the lexicon automatically and tested its accuracy using a different set of product reviews + star rating. We then discussed a case study of sentiment analysis for online political discourse. Adjectives retain their polarity regardless of what words they qualify: "good book", "good dog" and "good speech" are all positive. It follows that the approach should be scalable to other domains besides product reviews, such as political discourse. In support of this, the U.S. Colorado Department of Transportation use the English subjectivity lexicon in PATTERN to enhance their asset condition forecasting abilities. Most of the asset condition records are expressed in natural language. An evaluation (figure 41, Cole, R., personal communication, 2012) indeed suggests a correlation between the polarity of the inspector's comment and the age of the asset (e.g., steel girder).

In the same way the approach can also be used for fine art assessment (e.g., "vibrant colors", "inspiring work", "dull exhibition"). This ties in with Csikszentmihalyi's systems view of creativity (chapter 4), which attributes creativity to the recognition and adoption by peers. We can use sentiment analysis as a means of evaluating the performance of an artistic model of creativity. For example as proposed in chapter 2, using a setup of PERCOLATOR that publishes its works to a web page where visitors can comment on it. A practical case study is left up to future work.

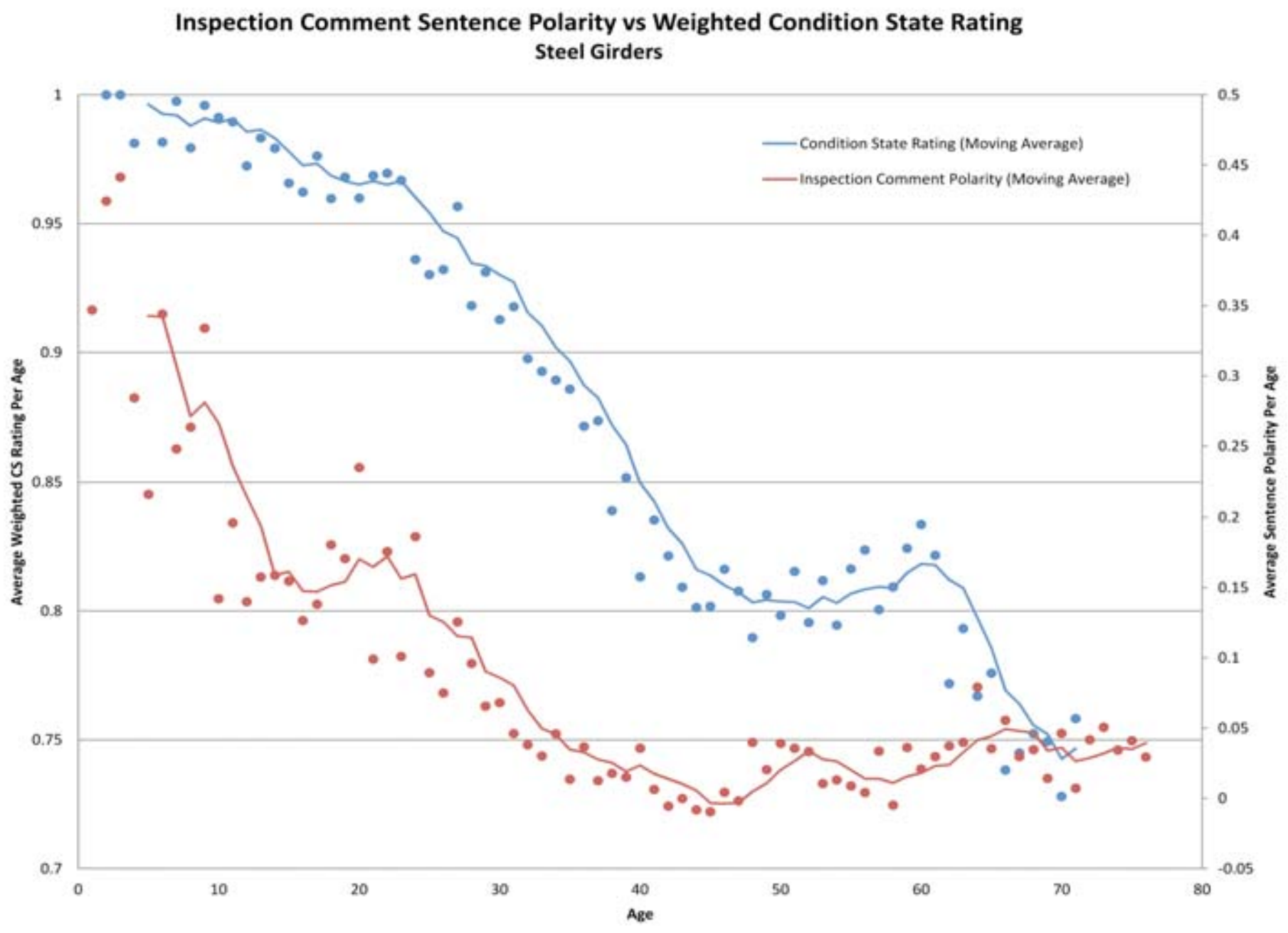

Figure 41. Inspection comment polarity vs. weighted condition state rating.
© U.S. Colorado Department of Transportation. Used with permission.

# Conclusions

Throughout this work we have examined creativity in an artistic context and from a number of different angles: from its origins in nature to humans and machines, from a generating perspective to an evaluating + learning perspective. In the introduction, we presented a diagram of how a model for computational creativity could work (figure 42), along with the following hypothesis:

> A "good idea" is selected or combined from a pool of many possible ideas (*a*). A work (of art) that captures this idea is then created (*b*). The work is evaluated by the author or a community of knowledgeable peers (*c*). Following a negative evaluation, the idea is subsequently adapted (*d*).

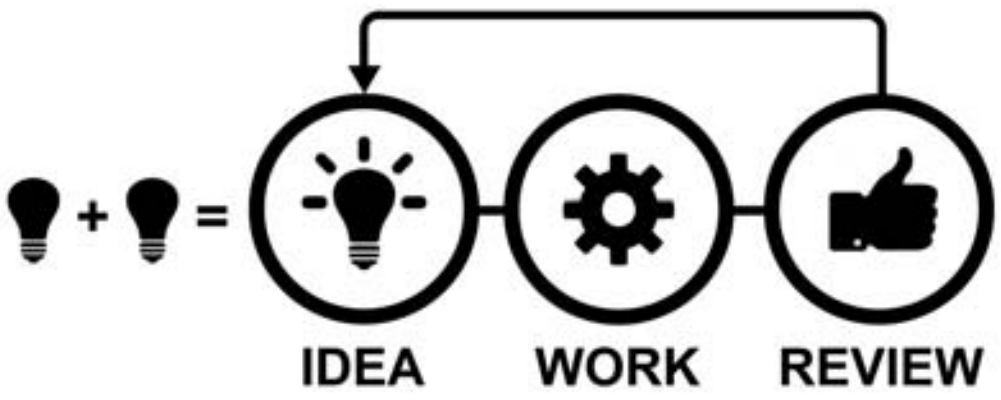

Figure 42. Illustrative example of a model for artistic creativity.

#### EVOLUTION & COMPLEXITY

In chapter 1, we looked at evolution by natural selection and emergence in complexity. In nature, evolution is the result of (slow) diversification by means of reproduction and genetic mutation, and the propagation of competitive variants that offer a better chance of survival. This process is essentially blind, since there is no predefined goal (1.1). To demonstrate this, we presented a case study called EVOLUTION that evolves virtual creatures using a genetic algorithm (1.2). Evolution can also be seen as the result of a complex interaction between a multitude of processes, such as competition within a species, interaction with other species, the environment, the climate (1.3). Unexpected creative behavior seemingly emerges out of nothing when such complex systems are regarded as a whole. To demonstrate this, we presented a case study called ANTAGONISM, which simulates competing ant colonies and the arising cooperative and competitive behavior (1.4). This case study showed how the emergent behavior is the result of simpler, underlying rules. Everything comes out of something. We have used this argument in defense of our hypothesis (*a*): *a new idea* = selected or combined from existing ideas.

#### RULES, CONSTRAINTS & AUTHORSHIP

In chapter 2, we provided an overview of generative art, a rule-based form of art inspired by nature and complexity, making use of techniques borrowed from AI and artificial life (1.5). We have seen how generative art relies on cooperation in the form of open source communities (2.1) and free software (2.2) to advance its own field. We presented our own software for generative art, NODEBOX and NODEBOX FOR OPENGL (2.3), and discussed four case studies: NEBULA, SUPERFOLIA, CREATURE and NANOPHYSICAL (2.4). Related to authorship we highlighted two additional case studies: PRISM and PERCOLATOR. These rely on the selection of an existing idea (e.g., color by majority vote, online news) to generate a new artwork. In this sense they are the author of their own artworks, since the decision is not made by the user or the author of the program. This ties in with our hypothesis (*b*): in order for there to be an *innovation* (e.g., a new artwork, a new technology) there must first be an original idea. Plagiarizing an existing idea does not count; it has to be manipulated in some way. We then discussed how PRISM and PERCOLATOR are not very creative systems, because they perform random manipulations instead of employing a creative thought process (2.5). This leads us to narrow our initial hypothesis (*a*): a new idea = a ~~selection or~~ combination of existing ideas, where the combination is driven by a *thought process*.

#### BLENDING

In chapter 3 we discussed VALENCE, a generative art installation created with NODEBOX FOR OPENGL that responds to brain activity (3.1–3.2). We used VALENCE as a fitting example of how (artistic) creativity works. First we selected two distinct techniques: agent-based modeling (artificial life) and electroencephalography (biomedical research). Then we blended them together in the context of generative art, motivated by curiosity, imagination, because we can, and the possibility of acclaim (e.g., media coverage, published papers). We based the idea on forerunners that experimented with similar setups, and in turn introduced a new element in the form of the valence hypothesis. This again refines our hypothesis (*a*): a new idea = a combination of existing ideas building on *previously acquired knowledge*, where the combination is driven by a thought process that satisfies a *problem* or a *motivation*.

#### NOVELTY & APPROPRIATENESS, LITTLE-C vs BIG-C, CONCEPT SEARCH SPACE

In chapter 4, we provided an overview of the psychological study of creativity. We discussed the psychological definition of human creativity, with novelty (it has to be new) and appropriateness (it has to be useful) as its determinants (4.1). We discerned between little-c creativity and Big-C creativity. Little-c refers to everyday creativity grounded in unconscious heuristics such as intuition and common sense. Arguably, it can be found in all humans and animals. Big-C refers to eminent creativity grounded in consciousness and hard work. Consciousness is a slow, self-evaluating feedback loop that (sometimes) emerges from the complexity of the mind. Hard work pertains to the modern-day definition of talent, which breaks down into a number of factors such as environment, personality, intelligence, experience, knowledge and motivation (4.2). We provided an example in the form of a DT test, which suggested that individuals with more experience in and knowledge of language do better at linguistic creativity (4.6). Finally, we examined a computational model of creativity: the concept search space (4.3–4.5). It models creative thought as a network of related concepts that can be traversed by different paths and thinking styles. Thinking styles can be unconscious heuristics (error-prone) or conscious thought (lazy). They can be combinatorial, explorative or transformational. The concept search space offers a way to model creative thought processes.

#### SEMANTIC NETWORK, COMMON SENSE, INTERPRETATION BIAS

In chapter 5, we discussed two case studies of computational creativity. We demonstrated FLOWEREWOLF, a program that generates poems by manipulating a network of lexical relations (5.1). We then demonstrated PERCEPTION, an implementation of the concept search space using a semantic network of common sense (5.2). The ideas generated by PERCEPTION are novel as well as witty. We can evaluate the degree of novelty using a web-based majority vote, for example. However, it is much harder to evaluate whether they are also appropriate (useful). We showed how both FLOWEREWOLF and PERCEPTION are prone to interpretation bias. The human mind has a tendency to fill in the blanks, for example by attributing meaning to random verses generated by FLOWEREWOLF. This foreshadows a problem in our hypothesis (*c*): novelty + appropriateness = *evaluated by the author* (5.3). How can a computer model self-evaluate the usefulness of its creations? This seems to imply machine consciousness, the pursuit of which has been largely abandoned by modern approaches. Currently, one more feasible approach is to rely on the evaluation of a third-party, for example using sentiment analysis.

#### LEARNING & EVALUATION

In chapter 6, we provided an introduction to the field of computational linguistics. In particular, we have looked at the vector space model, a technique for similarity-based learning and prediction used in natural language processing (6.1) and machine learning (6.3). We then demonstrated MBSP FOR PYTHON, a robust software package for text analysis (6.2). We also demonstrated PATTERN, a Python software package for web mining, natural language processing and machine learning (6.4). Finally, we discussed an example of how a combination of text analysis and unsupervised learning can be used to evaluate creativity (6.5–6.6). In short, we created an algorithm for scoring Guilford's Divergent Thinking test (DT), boosted it with a learning step and subsequently attempted to beat it with a simple approach for generating new responses. This ties in with our hypothesis (*d*): the *evaluation* and *adaptation* of an idea. Unfortunately, the success of this case study was partly based on a shortcoming in our algorithm, which did not clearly factor appropriateness.

#### OPINION

In chapter 7, we demonstrated our work on sentiment analysis for Dutch using the PATTERN package (7.1). We then discussed a case study of sentiment analysis for online political discourse (7.2). Finally, we argued how the technique can be applied to different domains (7.3), for example to assess the evaluation of creativity by peers. By assessing the opinions of humans, a computational model could learn to discern good ideas from bad ideas. This ties in with our hypothesis (*c*).

In conclusion, we argue that to call a computer model *creative*, it means that the model must be able to generate new and useful ideas by combining previously acquired knowledge (acquiring new knowledge if necessary). A generated idea should address a given problem or be driven by motivation. The model should be able to evaluate the novelty and usefulness of its ideas, or at least take into account the opinion of a third party. As we have shown, it is possible to an extent to address these tasks using AI approaches such as genetic algorithms, natural language processing and machine learning.

One particular problem is self-evaluation. How can we construct a computational model of creativity that is able to judge how well it is doing? Without this step, the model is of course unable to adapt and improve. It will continue to spawn random art, witty comparisons (e.g., cities + amphibians) and nonsensical answers (e.g., albatrosses to open locked doors); which may or may not be useful.

In the context of the arts, the problem is somewhat less perplexing since it is generally accepted that there is no good or bad art. Usefulness is split into the motivation of the author (e.g., a fascination for nature or decay) and the response of the audience. In future work we can explore this further along the lines of our discussion in chapter 2 (i.e., using genetic algorithms) and chapter 7 (i.e., using sentiment analysis as the GA's fitness function). However, to cite Hoare's Law of Large Problems in relation to complexity: inside every large problem there is a small problem struggling to get out. What if the robot artist is motivated by a fancy such as the perfect tessellation of a circle? It is one thing to have it formulate interesting ideas about this, but quite another issue to get it to achieve a visual result, which would require a very specific algorithm that is hard to come up with. In future work – for now – we may want to explore such problems separately.

In the context of practical little-c challenges (e.g., a locked door) we could resort to adding a massive amount of common sense knowledge, or restrict the model to a well-defined and bounded problem (i.e., an expert system). The rationale in line with chapter 4 is that more knowledge may make self-evaluation easier. An example of such a knowledge database is the OPEN MIND COMMON SENSE project (Havasi, Pustejovsky, Speer, Lieberman, 2009), a never-ending learner that has been learning for over a decade. It will be interesting to explore such approaches in future work.

# Appendix

## A.1 Genetic algorithm

Following is a concise, standard implementation of a genetic algorithm in Python:

```
from random import random, choice

class GA:

    def __init__(self, population=[]):
        self.population = population
        self.generation = 0

    def fitness(self, candidate):
        # Must be defined in a subclass, returns 0.0-1.0.
        return 1.0

    def crossover(self, candidate1, candidate2, mix=0.5):
        # Must be defined in a subclass.
        return None

    def mutate(self, candidate):
        # Must be defined in a subclass.
        return None or candidate

    def evolve(self, crossover=0.5, mutation=0.1, elitism=0.2):
        # 1) SELECTION.
        p = sorted((self.fitness(g), g) for g in self.population) # Weakest-first.
        n = sum(fitness for fitness, g in p) or 1
        for i, (fitness, g) in enumerate(p):
            p[i] = (fitness / n + (i > 0 and p[i-1][0] or 0), g)
        if elitism > 0:
            s = p[-int(elitism * len(p)):]
        else:
            s = []
        while len(s) < len(self.population):
            # Fitness proportionate selection.
            x = random()
            for fitness, g in p:
                if fitness > x:
                    s.append((fitness, g))
                    break
        # 2) REPRODUCTION.
        p = []
        while len(p) < len(self.population):
            # Recombine two random parents.
            i = int(random() * (len(s)-1))
            j = int(random() * (len(s)-1))
            p.append(self.crossover(s[i][1], s[j][1], mix=crossover))
            # Mutation avoids local optima by maintaining genetic diversity.
            if random() <= mutation:
                p[-1] = self.mutate(p[-1])
        self.population = p
        self.generation += 1

    @property
    def average_fitness(self):
        # Average fitness should increase each generation.
        return sum(map(self.fitness, self.population)) / len(self.population)
```

## A.2 NodeBox quick overview

NodeBox is a Mac OS X application that generates 2D visuals from Python code and exports them as a PDF or a QuickTime movie. The basic drawing functions are `rect()`, `oval()`, `line()`, `text()` and `image()`. The basic transformation functions are `translate()`, `rotate()` and `scale()`. The basic color functions are `fill()`, `stroke()` and `strokewidth()`. The application installer and the documentation are available from: http://nodebox.net/code/

| DRAWING | DESCRIPTION |
|---|---|
| `rect(x, y, width, height)` | Draws a rectangle at `(x, y)` with the given `width` and `height`. |
| `oval(x, y, width, height)` | Draws an ellipse at `(x, y)` with the given `width` and `height`. |
| `line(x1, y1, x2, y2)` | Draws a line from `(x1, y1)` to `(x2, y2)`. |
| `text(txt, x, y, width=None)` | Draws a string of `txt` at `(x, y)` wrapped at `width`. |
| `image(path, x, y, alpha=1.0)` | Draws the image file stored at `path` at position `(x, y)`. |

| TRANSFORMATION | DESCRIPTION |
|---|---|
| `translate(x, y)` | Translate the origin to `(x, y)`, by default the top-left `(0, 0)`. |
| `rotate(degrees)` | Rotate subsequent shapes like `rect()` by the given `degrees`. |
| `scale(x, y=None)` | Scale subsequent shapes by a relative factor, e.g., `0.5` = 50%. |

| COLOR | DESCRIPTION |
|---|---|
| `fill(r, g, b, a)` | Sets the current fill color as R,G,B,A values between `0.0`-`1.0`. |
| `stroke(r, g, b, a)` | Sets the current stroke outline color. |
| `strokewidth(width)` | Sets the `width` in pixels of the outline for subsequent shapes. |

### Example of use

The following example produces a 1,000 randomly rotated rectangles in shades of red:

```
for i in range(1000):
    rotate(random(8) * 45)
    fill(random(), 0, 0, random())
    x = random(WIDTH)
    y = random(HEIGHT)
    w = random(400)
    h = random(40)
    rect(x, y, w, h)
```

## A.3 NodeBox for OpenGL

NodeBox for OpenGL is a cross-platform toolkit that generates 2D animations with Python code. The documentation is available from http://www.cityinabottle.org/nodebox/. The toolkit requires Python and Pyglet. Python can be downloaded from http://www.python.org/ and Pyglet can be downloaded from http://www.pyglet.org/. On Linux, users can use their package management system to install Pyglet. On Ubuntu you can do `sudo apt-get install python-pyglet`. On Fedora you can do `sudo yum install pyglet`. On Mac OS X, Python is already installed. On Mac OS X 10.6+ this is a 64-bit Python. If you use Pyglet 1.1.4 you need to install a 32-bit Python or run scripts in 32-bit mode by executing this command from the terminal:

```
> arch -i386
> python flocking.py
```

### Installation

To install NodeBox, execute `setup.py` in the `nodebox/` folder:

```
> cd nodebox
> python setup.py install
```

If this doesn't work, place the `nodebox` folder in the download in the correct location manually. To be able to import NodeBox in a script, Python needs to know where the module is located. There are three options: 1) put the `nodebox` folder in the same folder as your script, 2) put it in the standard location[31] for modules so it is available to all scripts or 3) add the location of the folder to `sys.path` in your script, before importing the package. To modify `sys.path` do:

```
NODEBOX = '/users/tom/desktop/nodebox'
import sys
if NODEBOX not in sys.path: sys.path.append(NODEBOX)
from nodebox.graphics import *
```

### Hardware acceleration + C extensions

Your video hardware needs to support OpenGL 2.0. NodeBox may not work on older video hardware. Some indicators are: 1) using image filters only works with non-transparent images, 2) using image filters produces no effect and `nodebox.graphics.shader.SUPPORTED` is `False`, and 3) using the `render()` or `filter()` command throws an `OffscreenBufferError`.

NodeBox is bundled with a number of C extensions to increase performance. It will run fine without the extensions, but you can compile them to make it faster. In the `nodebox/ext/` folder, execute `setup.py` from the command line:

```
> cd nodebox/ext/
> python setup.py
```

[31] On Mac OS X, the standard location is `/Library/Python/2.5/site-packages/`
On Unix, it is `/usr/lib/python2.5/site-packages/`
On Windows, it is `c:\python25\Lib\site-packages\`

## Example of use

Try out if the following example works. It should produce a flock of BOIDS:

```
from nodebox.graphics import *
from nodebox.graphics.physics import Flock

flock = Flock(40, 0, 0, 500, 500)
flock.sight = 300

def draw(canvas):
    background(1, 0.5)
    fill(0, 0.75)
    flock.update(cohesion=0.15)
    for boid in flock:
        push()
        translate(boid.x, boid.y)
        scale(0.5 + 1.5 * boid.depth)
        rotate(boid.heading)
        arrow(0, 0, 15)
        pop()

canvas.fps = 30
canvas.size = 600, 400
canvas.run(draw)
```

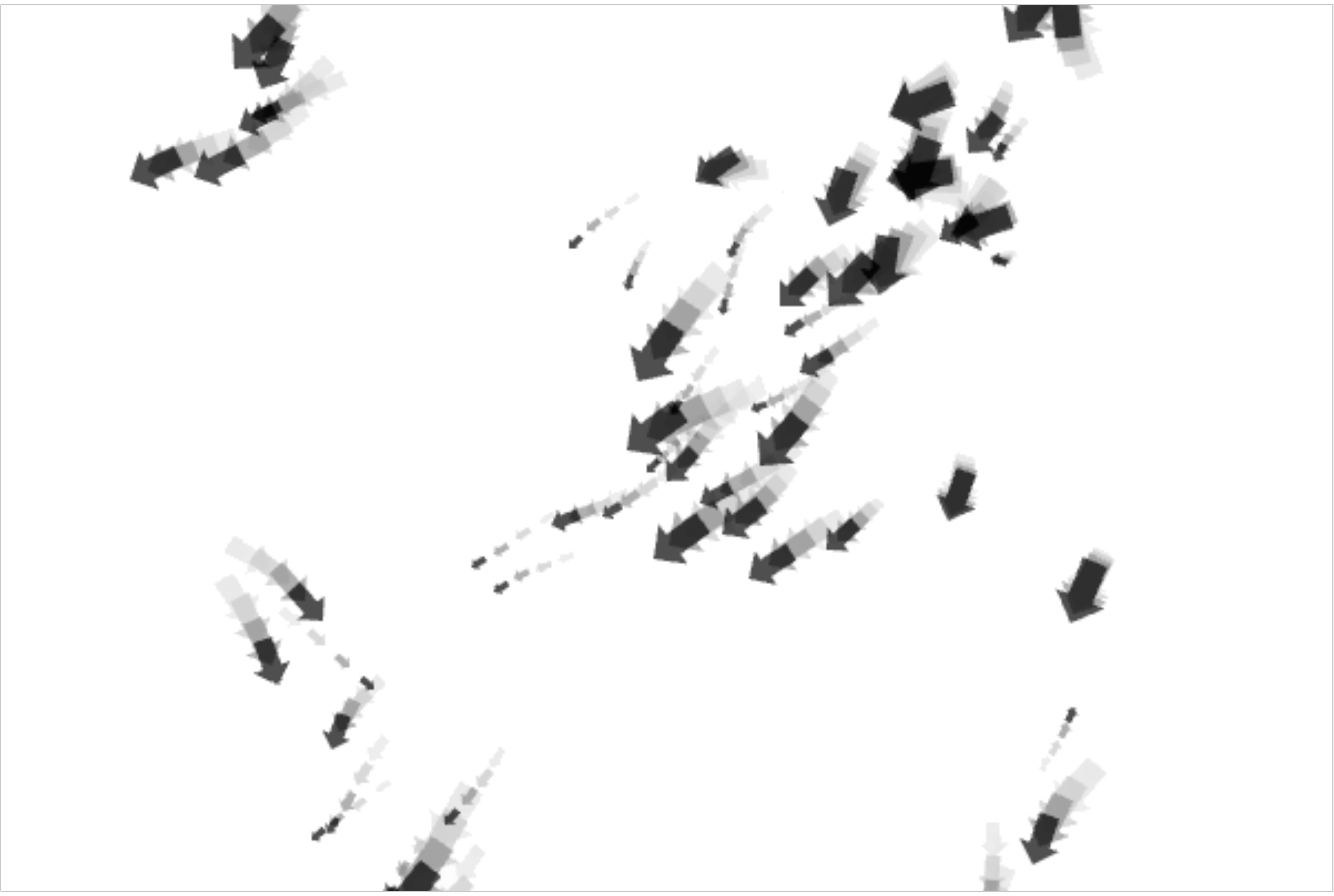

Figure 43. Screenshot of the output of the example.

## A.4 MBSP for Python

MBSP for Python is a text analysis system with tools for tokenization and sentence splitting, part-of-speech tagging, chunking, lemmatization, relation finding and prepositional phrase attachment. The documentation is available from http://www.clips.ua.ac.be/pages/mbsp.

### Installation

The package is bundled with three dependencies written in C/C++ (TIMBL learner, MBT tagger and MBLEM lemmatizer). Binaries have been compiled for Mac OS X. If these work no installation is required. Put the MBSP folder in the download in `/Library/Python/2.6/site-packages/` (on Mac OS X) or `/usr/lib/python2.6/site-packages/` (on Unix). If the binaries do not work, execute `setup.py`, which compiles and installs the toolkit in the right location. If `setup.py` does not work you need to compile binaries manually.

```
> cd MBSP
> python setup.py install
```

### Compiling TIMBL + MBT + MBLEM

Go to the `MBSP/timbl` folder. Uncompress the source code from the `timbl-6.x.tar` archive. From the command line, execute `configure` and `make` in the `MBSP/timbl/timbl-6.x` folder. The given `[FOLDER]` is an absolute path to the folder where TIMBL will be built. The executable will be in `[FOLDER]/bin`. Copy it to `MBSP/timbl`. Then build MBT in the same `[FOLDER]`.

```
> cd MBSP/timbl/timbl-6.1.5
> ./configure --enable-shared=no --enable-static=no --prefix=[FOLDER]
> make install
```

Go to the `MBSP/mbt` folder. Uncompress the source code from the `mbt-3.x.tar` archive. From the command line, execute `configure` and `make` in the `MBSP/mbt/mbt-3.x` folder. The executable will be in `[FOLDER]/bin`. Copy it to `MBSP/mbt`. Delete the build `[FOLDER]`.

```
> cd MBSP/mbt/mbt-3.1.3
> ./configure --enable-shared=no --enable-static=no --prefix=[FOLDER]
> make install
```

Go to the `MBSP/mblem` folder. Delete all the files with a `.o` extension and the current executable binary `mblem_english_bmt`. From the command line, execute `make`:

```
> cd MBSP/mblem
> make
```

### Example

Try out if the following example works. It should start the servers and print the tagged output. It will start four data servers that take up memory: `CHUNK` 80MB, `LEMMA` 10MB, `RELATION` 160MB and `PREPOSITION` 210MB. Only the `CHUNK` server is mandatory. The optional servers can be disabled in `config.py`.

```
from MBSP import parse
print parse('The black cat sat on the mat.')
```

## A.5 Pattern for Python

Pattern is a web mining module for the Python programming language. Pattern is written for Python 2.5+. The module has no external dependencies except when using LSA in the vector module, which requires the NumPy package (installed by default on Mac OS X). To install it so that the package is available in all your scripts, open the terminal and do:

```
> cd pattern-2.5
> python setup.py install
```

If you have pip, you can automatically download and install from the PyPi repository:

```
> pip install pattern
```

If none of the above works, you can make Python aware of the package in three ways: 1) put the `pattern` folder in the download in the same folder as your script, 2) put it in the standard location for modules or 3) add the location of the folder to `sys.path`. To modify `sys.path` do:

```
PATTERN = '/users/tom/desktop/pattern'
import sys
if PATTERN not in sys.path: sys.path.append(PATTERN)
import pattern
```

### Example

Try out if the following example works. It should print the tagged output.

```
from pattern.en import parse
print parse('The black cat sat on the mat.')
```

# Bibliography

Abney, S. (1991). Parsing by chunks. In R. Berwick, S. Abney, & C. Tenny (Eds.) *Principle-Based Parsing*. Dordrecht: Kluwer Academic Publishers.

Albert, R. S., & Runco, M. A. (1999). A history of research on creativity. In R. Sternberg (Ed.) *Handbook of Creativity*, 16–31. New York: Cambridge University Press.

Amabile, T. M., Hill, K. G., Hennessey, B. A., & Tighe, E. M. (1994). The work preference inventory: assessing intrinsic and extrinsic motivational orientations. *Journal of Personality and Social Psychology, 66*(5), 950–967.

Arthur, D., & Vassilvitskii, S. (2007). k-means++: the advantages of careful seeding. In *Proceedings of the Eighteenth Annual ACM-SIAM Symposium on Discrete Algorithms*, 1027–1035. New Orleans, Louisiana, USA.

Axelrod, R. (1984). *The Evolution of Cooperation*. New York: Basic Books.

BBC Europe (2011). Belgium swears in new government headed by Elio Di Rupo. Retrieved from: http://www.bbc.co.uk/news/world-europe-16042750

Baars, B. J. (1997). In the theatre of consciousness: global workspace theory, a rigorous scientific theory of consciousness. *Journal of Consciousness Studies, 4*(4), 292–309.

Baayen, R. H., Piepenbrock, R., & van Rijn, H. (1993). The CELEX lexical database. *Experimental Psychology, 56*(A), 445– 467.

Balahur, A., Steinberger, R., Kabadjov, M., Zavarella, V., Van Der Goot, E., Halkia, M., Pouliquen, B., & Belyaeva, J. (2010). Sentiment analysis in the news. In *Proceedings of the Seventh International Conference on Language Resources and Evaluation (LREC '10)*. Valletta, Malta.

Balota, D. A., & Lorch, R. F. (1986). Depth of automatic spreading activation: mediated priming effects in pronunciation but not in lexical decision. *Journal of Experimental Psychology: Learning, Memory, and Cognition, 12*(3): 336–345.

Banko, M., Cafarella, M. J., Soderland, S., Broadhead, M., & Etzioni, O. (2007). Open information extraction from the web. In *Proceedings of the 20th International Joint Conference on Artificial Intelligence (IJCAI '07)*, 2670–2676. Hyderabad, India.

Bastian, M., Heymann, S., & Jacomy, M. (2009). Gephi: An open source software for exploring and manipulating networks. In *Proceedings of the Third International AAAI Conference on Weblogs and Social Media (ICWSM09)*. San Jose, California, USA.

Batey, M., & Furnham, A. (2006). Creativity, intelligence and personality. A critical review of the scattered literature. *Genetic, Social, and General Psychology Monographs, 132*(4), 355–429.

Bedny, M., Pascual-Leone, A., & Saxe, R. R. (2009). Growing up blind does not change the neural bases of theory of mind. *Proceedings of the National Academy of Sciences of the United States of America (PNAS), 106*(27), 11312–11317.

Benamara, F., Cesarano, C., Picariello, A., Reforgiato, D., & Subrahmanian, V. S. (2007). Sentiment analysis: adjectives and adverbs are better than adjectives alone. In *Proceedings of the International Conference on Weblogs and Social Media (ICWSM '07)*, 203–206. Seattle, Washington, USA.

Benewick, R. J., Birch, A. H., Blumler, J. G., & Ewbank, A. (1969). The floating voter and the liberal view of representation. *Political Studies, 17*, 177–195.

Bersak, D., McDarby, G., Augenblick, N., McDarby, P., McDonnell, D., McDonald, B., & Karkun, R. (2001). Biofeedback using an immersive competitive environment. In *Online Proceedings of the Ubicomp 2001 Workshop on Designing Ubiquitous Computer Games*. Atlanta, Georgia, USA.

Bird, S., Klein, E., & Loper, E. (2009). *Natural Language Processing with Python*. O'Reilly Media.

Boden, M. (2003). *The Creative Mind: Myths and Mechanisms*. New York: Routledge.

Boden, M., & Edmonds, E. A. (2009). What is generative art? *Digital Creativity, 20*(1–2), 21–46.

Bowers, K. S., Regehr,G., Balthazard, C., & Parker, K. (1990). Intuition in the context of discovery. *Cognitive Psychology, 22*(1), 72–110.

Braitenberg, V. (1984). *Vehicles: Experiments in Synthetic Psychology*. Cambridge: MIT Press.

Brandes, U. (2001). A faster algorithm for betweenness centrality. *Journal of Mathematical Sociology, 25*(2), 163–177.

Breiman, L. (2001). Random forests. *Machine Learning, 45*, 5–32.

Brill, E. (1992). A simple rule-based part of speech tagger. In *Proceedings of the Third Conference on Applied Natural Language Processing*, 152–155. Trento, Italy.

Brin, S., & Page, L. (1998). The anatomy of a large-scale hypertextual web search engine. *Computer Networks and ISDN Systems, 30*, 107–117.

Brinton, L. (2000). *The structure of modern English: a linguistic introduction*. Amsterdam: John Benjamins.

Bruce, R. F., & Wiebe, J. M. (1999). Recognizing subjectivity: A case study in manual tagging. *Natural Language Engineering, 5*(2), 187–205.

Buchanan, R. (1992). Wicked problems in design thinking. *Design Issues, 8*(2), 5–21.

Buchholz, S. (2002). *Memory-Based Grammatical Relation Finding*. Tilburg: ILK.

Burch, G. S., Pavelis, C., Hemsley, D. R., & Corr, P. J. (2006). Schizotypy and creativity in visual artists. *British Journal of Psychology, 97*(2), 177–190.

Campbell, D.T. (1960). Blind variation and selective retention in creative thought as in other knowledge processes. *Psychological Review, 67*, 380–400.

Cardoso, A., Veale, T., & Wiggins, G. A. (2009). Converging on the divergent: The history (and future) of the international joint workshops in computational creativity. *AI Magazine, 30*(3), 15–22.

Chalmers, D., French, R., & Hofstadter, D. (1991). High-level perception, representation, and analogy: a critique of artificial intelligence methodology. *Journal of Experimental & Theoretical Artificial Intelligence, 4*, 185–211.

CHAMBERLAIN, W. (1984). *The Policeman's Beard is Half-Constructed: Computer Prose and Poetry.* New York: Warner Software/Books.

CHANG, C.-C., & LIN, C.-J. (2011). LIBSVM: a library for support vector machines. *ACM Transactions on Intelligent Systems and Technology, 2*(3).

CHOMSKY, N. (1957). *Syntactic Structures.* The Hague: Mouton.

CLEVELAND, P. (2004). Bound to technology - the telltale signs in print. *Design Studies, 25*(2), 113–153.

COHEN, H. (1995). The further exploits of Aaron, painter. *Stanford Humanities Review, 4*(2), 141–158.

COLERIDGE, S. T. (1884). *The Friend.* Harper.

COLLINS, A., & LOFTUS, E. (1975). A spreading-activation theory of semantic processing. *Psychological Review, 82*(6), 407–428.

COLLINS, C., & STEPHENSON, K. (2002). A circle packing algorithm. *Computational Geometry, 25*(3), 233–256.

COLTON, S. (2008). Creativity versus the perception of creativity in computational systems. In *Proceedings of the AAAI Spring Symposium on Creative Systems 2008.* Palo Alto, California, USA.

COMINGS, D. E., WU, S., ROSTAMKHANI, M., MCGUE, M., IACONO, W. G., CHENG, L. S-C., & MACMURRAY, J. P. (2003). Role of the cholinergic muscarinic 2 receptor (CHRM2) gene in cognition. *Molecular Psychiatry, 8*, 10–11.

CONWAY, D. (1998). An algorithmic approach to English pluralization. In *Proceedings of the Second Annual Perl Conference.* San Jose, California, USA.

COPE, D. (2001). *Virtual Music.* Cambridge: MIT Press.

CORTES, C., & VAPNIK, V. (1995). Support-vector networks. *Machine Learning, 20*(3), 273–297.

COXETER, H. S. M. (1968). The problem of Apollonius. *The American Mathematical Monthly, 75*(1), 5–15.

COXETER, H. S. M. (1979). The non-euclidean symmetry of Escher's picture 'Circle Limit III'. *Leonardo, 12*, 19–25.

CRAIG, C. L. (1987). The ecological and evolutionary interdependence between web architecture and web silk spun by orb web weaving spiders. *Biological Journal of the Linnean Society, 30*(2), 135–162.

CRAIG, C. L. (1997). Evolution of arthropod silks. *Annual Review of Entomology, 42*, 231–267.

CSIKSZENTMIHALYI, M. (1990). *Flow: The psychology of optimal experience.* New York: Harper & Row.

CSIKSZENTMIHALYI, M. (1999). Implications of a systems perspective for the study of creativity. In R. Sternberg (Ed.) *Handbook of Creativity*, 16–31. New York: Cambridge University Press.

DAELEMANS, W., & VAN DEN BOSCH, A. (2005). *Memory-based language processing.* Cambridge: Cambridge University Press.

DAELEMANS, W., BUCHHOLZ, S., & VEENSTRA, J. (1999). Memory-based shallow parsing. In *Proceedings of CoNLL*, Bergen, Norway.

DAELEMANS, W., ZAVREL, J., VAN DER SLOOT, K. & VAN DEN BOSCH, A. (2004). *TiMBL : Tilburg Memory-Based Learner (version 5.1). Reference guide.* Tilburg: ILK.

DAELEMANS, W., VAN DEN BOSCH, A., & WEIJTERS, T. (1997). IGTree: using trees for compression and classification in lazy learning algorithms. *Journal of Artificial Intelligence Review, 11*, 407–423.

DARWIN, C. (1859). *On the origin of species by means of natural selection.* London: John Murray.

DAWKINS, R. (1976). *The Selfish Gene.* Oxford: Oxford University Press.

DE BLESER, F., DE SMEDT, T., & NIJS L. (2003). NodeBox 1.9.5 for Mac OS X. Retrieved May 2012, from: http://nodebox.net/code

DE BLESER, F., & GABRIËLS, S. (2009). NodeBox 2 version 2.2.1. Retrieved May 2012, from: http://beta.nodebox.net/

DE SMEDT T., LECHAT L., & DAELEMANS W. (2011). Generative art inspired by nature, in NodeBox. In *Applications of Evolutionary Computation*, LNCS 6625(2), 264–272.

DE SMEDT T., VAN ASCH V. & DAELEMANS, W. (2010). Memory-based shallow parser for Python. CLiPS Technical Report Series 2.

DE SMEDT, T. (2005). Flowerewolf. Retrieved June 2012, from: http://nodebox.net/code/index.php/Flowerewolf

DE SMEDT, T. (2005). Percolator. Retrieved June 2012, from: http://nodebox.net/code/index.php/Percolator

DE SMEDT, T. (2010). Nebula. In *Creativity World Biennale*, exhibition catalogue, 15 November–8 January 2010, Untitled Gallery, Oklahoma, US.

DE SMEDT, T. (2011). NodeBox for OpenGL version 1.7, retrieved May 2012, from: http://cityinabottle.org/nodebox/

DE SMEDT, T., & DAELEMANS, W. (2012). "Vreselijk mooi!" (terribly beautiful): a subjectivity lexicon for Dutch adjectives. In *Proceedings of the 8th Language Resources and Evaluation Conference (LREC'12)*, 3568–3572. Istanbul, Turkey.

DE SMEDT, T., & DAELEMANS, W. (2012). Pattern for Python. *Journal of Machine Learning Research, 13*, 2031–2035.

DE SMEDT, T., & DE BLESER, F. (2005). Prism. Retrieved June 2012, from: http://nodebox.net/code/index.php/Prism

DE SMEDT, T., DE BLESER, F., VAN ASCH, V., NIJS, L., & DAELEMANS, W. (forthcoming 2013). Gravital: natural language processing for computer graphics. In T. Veale, K. Feyaerts, & C. Forceville (Eds.) *Creativity and the Agile Mind: A Multidisciplinary Approach to a Multifaceted Phenomenon.* Berlin: Mouton De Gruyter.

DE SMEDT, T., & LECHAT, L. (2010). Self-replication. In *Travelling Letters*, exhibition catalogue, 18 June–5 September 2010, Lahti Art Museum, Finland.

DE SMEDT, T., & LECHAT, L. (2010). Superfolia. In *Creativity World Biennale*, exhibition catalogue, 15 November–8 January 2010, Untitled Gallery, Oklahoma, US.

DE SMEDT, T., LECHAT, L., & COLS, J. (2010). Creature. In *Creativity World Biennale*, exhibition catalogue, 15 November–8 January 2010, Untitled Gallery, Oklahoma, US.

DE SMEDT, T., & MENSCHAERT, L. (2012). Affective visualization using EEG. *Digital Creativity.* DOI:10.1080/14626268.2012.719240

DEFAYS, D. (1977). An efficient algorithm for a complete link method. *The Computer Journal, 20*(4), 364–366.

Delph, L., & Lively, C. (1989). The evolution of floral color change: pollinator attraction versus physiological constraints in Fuchsia excorticata. *Evolution*, *43*(6), 1252–1262.

Demšar, J., Zupan, B., Leban, G., & Curk, T. (2004). Orange: from experimental machine learning to interactive data mining. *Knowledge Discovery in Databases*, 3202, 537–539.

Dennett, D. C. (1991). *Consciousness Explained.* Boston: Little Brown.

Dennett, D. C. (2003). *Freedom evolves*. Harmondsworth: Penguin.

Dennett, D. C. (2004). Could there be a Darwinian account of human creativity? In A. Moya, & E. Font (Eds.) *Evolution: From Molecules to Ecosystems*, 272–279. Oxford University Press.

Dexter, L., & White, D. (1964). People, society, and mass communications. In *Free Press of Glencoe*. London: Collier-MacMillan.

Dijkstra, E. (1959). A note on two problems in connexion with graphs. *Numerische Mathematik*, *1*, 269–271.

D'Alessio, D., & Allen, M. (2000). Media bias in presidential elections: a meta-analysis. *Journal of communication*, *50*, 133–156.

Ebert, D., Musgrave F. K., Peachey, D., Perlin, K., & Worley, S. (2003). *Texturing & Modeling. A Procedural Approach.* Morgan Kaufmann.

Eisenstadt, J. M. (1978). Parental loss and genius. *American Psychologist*, *33*(3), 211–223.

Elia, M. (1992). Organ and tissue contribution to metabolic rate. In H. Tucker, & J. Kinney (Eds.) *Energy Metabolism: Tissue Determinants and Cellular Corollaries*, 61–77. New York: Raven.

Elkan, C. (2003). Using the triangle inequality to accelerate k-means. In *Proceedings of the Twentieth International Conference on Machine Learning*, 147–153. Washington, USA.

Ericsson, K. A. (1998). The scientific study of expert levels of performance: general implications for optimal learning and creativity. *High Ability Studies*, *9*(1), 75–100.

Escher, M. C. (1960). *The graphic work of M. C. Escher.* New York: Hawthorn Books.

Esuli, A., & Sebastiani, F. (2006). Sentiwordnet: A publicly available lexical resource for opinion mining. In *Proceedings of LREC 2006*, 417–422.

Eysenck, H. J. (1995). *Genius: The natural history of creativity.* Cambridge: Cambridge University Press.

Farkas R., Vincze, V. , Móra, G., Csirik, J., & Szarvas, G. (2010). The CoNLL-2010 shared task: learning to detect hedges and their scope in natural language text. In *Fourteenth Conference on Computational Natural Language Learning, Proceedings of the Shared Task*, 1–12. Uppsala, Sweden.

Fauconnier, G., & Turner, M. (2003). *The Way We Think: Conceptual Blending And The Mind's Hidden Complexities.* New York: Basic Books.

Fayyad, U. M., Wierse, A., & Grinstein, G., G. (2002). *Information visualization in data mining and knowledge discovery.* Morgan Kaufmann.

Feist, G. J. (1999). Influence of personality on artistic and scientific creativity. In R. Sternberg (Ed.) *Handbook of Creativity*, 273–296. New York: Cambridge University Press.

Feldman, R., & Sanger, J. (2007). *The Text Mining Handbook: Advanced Approaches in Analyzing Unstructured Data.* Cambridge: Cambridge University Press.

Fellbaum, C. (1998). *WordNet: An Electronic Lexical Database.* Cambridge: MIT Press.

Fisher, R. A. (1930). *The Genetical Theory of Natural Selection.* Oxford: Clarendon Press.

Fisher, R. S., van Emde Boas, W., Blume, W., Elger, C., Genton, P., Lee, P., & Engel, J. (2005). Epileptic seizures and epilepsy: definitions proposed by the International League Against Epilepsy (ILAE) and the International Bureau for Epilepsy (IBE). *Epilepsy*, *46*(4), 470–472.

Fix, E., & Hodges, J.L. (1951). Discriminatory analysis, nonparametric discrimination: Consistency properties. Technical Report 4, USAF School of Aviation Medicine, Randolph Field, Texas.

Fodor, J. (2000). *The Mind Doesn't Work That Way: The Scope and Limits of Computational Psychology.* Cambridge: MIT Press.

Freud, S. (1913). *The Interpretation of Dreams*. Macmillan.

Friedl, J. (2002). *Mastering Regular Expressions: A Modern Approach.* O'Reilly.

Fry, B. (2008). *Visualizing Data.* O'Reilly.

Galanter, P. (2003). What is generative art? Complexity theory as a context for art theory. In *Proceedings of the 6th International Conference on Generative Art*, 216–236. Milan, Italy.

Galanter, P. (2008). Complexism and the role of evolutionary art. In J. Romero, & P. Machado (Eds.) *The Art of Artificial Evolution: A Handbook on Evolutionary Art and Music.* Berlin: Springer-Verlag.

Galanter, P. (2012). Message 15 in: There must be no generative, procedural or computational art. Retrieved May 2012, from: http://yaxu.org/there-must-be-no-generative-procedural-or-computational-art

Gallai, N., Salles, J. M., Settele, J., & Vaissière, B. (2009). Economic valuation of the vulnerability of world agriculture confronted with pollinator decline. *Ecological Economics*, *68*, 810–821.

Gallo, L., De Pietro, G., & Marra I. (2008). 3D interaction with volumetric medical data: experiencing the Wiimote. In *Proceedings of the 1st International Conference on Ambient Media and Systems (Ambi-Sys '08).* Quebec City, Canada.

Gallup, G. G. (1970). *Chimpanzees: Self-recognition.* Science, *167*(3914), 86–87.

Gallup, G. G. (1982). Self-Awareness and the emergence of mind in primates. *American Journal of Primatology*, *2*, 237–248.

Gallup, G. G., Anderson, J. R., & Shillito, D. J. (2002). The mirror test. In M. Bekoff, C. Allen, & G. Burghardt (Eds.) *The cognitive animal*, 325–334. Cambridge: MIT Press.

Gardner, H. (1993). Seven creators of the modern era. In J. Brockman (Ed.) *Creativity*, 28–47. New York: Simon & Schuster.

Gardner, H. (1999). *Intelligence reframed: Multiple intelligences for the 21st century.* New York: Basic Books.

Gardner, M. (1970). Mathematical games: the fantastic combinations of John Conway's new solitaire game "life". *Scientific American*, *223*: 120–123.

Geertzen, J. (2010). Jeroen geertzen :: Brill-nl, Retrieved June 2012, from: http://cosmion.net/jeroen/software/brill_pos/

Geschwind, N. (1970). The organization of language and the brain. *Science*, *27*, 940–944.

Gielis, J. (2003). A generic geometric transformation that unifies a wide range of natural and abstract shapes. *American Journal of Botany*, *90*, 333–338.

Gilleade, K., Dix, A., & Allanson, J. (2005). Affective videogames and modes of affective gaming: assist me, challenge me, emote me. In *Proceedings of DiGRA 2005*. Vancouver, Canada.

Gillies, D. (1996). *Artificial Intelligence and Scientific Method*. New York: Oxford University Press.

Gilovich, T., Griffin, D., & Kahneman, D. (2002). *Heuristics and Biases: The Psychology of Intuitive Judgment*. New York: Cambridge University Press.

Goldstein, J. (1999). Emergence as a construct: history and issues. *Emergence*, *11*, 49–72.

Golub, G. H., & Van Loan, C. F. (1996). *Matrix Computations*. Baltimore: Johns Hopkins.

Gombrich, E. H. (1960). *Art and Illusion: A Study in the Psychology of Pictorial Representation*. New York: Pantheon Books.

Gruszka, A., & Necka, E. (2002). Priming and acceptance of close and remote associations by creative and less creative people. *Creativity Research Journal*, *14*(2), 193–205.

Guilford, J. P. (1956). The structure of intellect. *Psychological Bulletin*, *53*(4), 267–293.

Guilford, J. P. (1967). Creativity: yesterday, today and tomorrow. *The Journal of Creative Behavior*, *1*(1), 3–14.

Guilford, J. P. (1977). *Way Beyond The IQ*. New York: Creative Education Foundation.

Guyon, I., & Elisseeff, A. (2003). An introduction to variable and feature selection. *Journal of Machine Learning Research*, *3*, 1157–1182.

Gyselinckx, B., Van Hoof, C., Ryckaert, J., Yazicioglu, R. F., Fiorini, P., & Leonov, V. (2005). Human++: autonomous wireless sensors for body area networks. In *Proceedings of the Custom Integrated Circuits Conference, IEEE 2005*, 13–19. San Jose, California, USA.

Hagberg A., Schult, D., & Swart, P. (2008). Exploring network structure, dynamics, and function using networkx. In *Proceedings of the 7th Python in Science Conference (SCIPY 2008)*, 11–15. Pasadena, California, USA.

Haill, L. (2010). Tunes on the brain: Luciana Haill's EEG art. Retrieved August 2012, from: http://www.wired.co.uk/magazine/archive/2010/09/play/tunes-brain-luciana-haill-eeg-art

Harnad, S. (2001). What's wrong and right about Searle's Chinese Room argument? In M. Bishop, & J. Preston (Eds). *Essays on Searle's Chinese Room Argument*. Oxford: Oxford University Press.

Hars, A., & Ou, S. (2002). Working for free? Motivations of participating in open source projects. *International Journal of Electronic Commerce*, *6*, 25–39.

Havasi, C., Pustejovsky, J., Speer, R., & Lieberman, H. (2009). Digital intuition: applying common sense using dimensionality reduction. *Intelligent Systems*, *IEEE 24*(4): 24-35.

Hellesoy, A., & Hoover, D. (2006). Graph JavaScript framework version 0.0.1, retrieved June 2012, from: http://snipplr.com/view/1950/graph-javascript-framework-version-001/

Hjelm, S. (2003). The making of Brainball. *Interactions*, *10*(1), 26–34.

Hobson, A., & McCarley, R. W. (1977). The brain as a dream state generator: an activation-synthesis hypothesis of the dream process. *American Journal of Psychiatry*, *134*(12), 1335–1348.

Hofstadter, D. (2007). *I Am a Strange Loop*. New York: Basic Books.

Hofstadter, D. R. (1979). *Gödel, Escher, Bach: an Eternal Golden Braid*. New York: Basic Books.

Hofstadter, D. R. (1985). *Metamagical Themas: Questing for the Essence of Mind and Pattern*. New York: Basic Books.

Hofstadter, D. R. (1995). *Fluid Concepts and Creative Analogies: Computer Models of the Fundamental Mechanisms of Thought*. New York: Basic Books.

Hofstadter, D. R. (2002). Staring Emmy straight in the eye – and doing my best not to flinch. In T. Dartnall (Ed.) *Creativity, Cognition and Knowledge*, 67–100. Westport, CT: Praeger.

Holkner, A. (2008). Pyglet version 1.1.4. Retrieved May 2012, from: http://pyglet.org/

Holland, J. H. (1992). Genetic Algorithms. *Scientific American*, *267*, 66–72.

Hu, J., & Goodman, E. (2002). The hierarchical fair competition (HFC) model for parallel evolutionary algorithms. In *Proceedings of the Congress on Evolutionary Computation*, 49–54. Brisbane, Australia.

Hölldobler, B., & Wilson E. O. (1990). *The Ants*. Cambridge: Harvard University Press.

Hölldobler, B., & Wilson, E. O. (2009). *The Superorganism: The Beauty, Elegance, And Strangeness of Insect Societies*. New York: W. W. Norton.

Igoe, T. (2007). *Making Things Talk*. O'Reilly.

Jackson, E., & Ratnieks, F. (2007). Communication in ants. *Current Biology*, *16*(15), 570–574.

Jansen, B. J., Zhang, M., Sobel, K., & Chowdhury, A. (2009). Twitter power: tweets as electronic word of mouth. *Journal of the American Society for Information Science and Technology*, *60*, 2169–2188.

Jijkoun, V., & Hofmann, K. (2009). Generating a non-English subjectivity lexicon: relations that matter. In *Proceedings of the 12th Conference of the European Chapter of the ACL*, 398--405. Athens, Greece.

Jordà S., Geiger G., Alonso M., & Kaltenbrunner, M. (2007). The reacTable: exploring the synergy between live music performance and tabletop tangible interfaces. In *Proceedings of the 1st International Conference on Tangible and Embedded Interaction (TEI '07)*. New York, USA.

Jung-Beeman M., Bowden E. M., Haberman J., Frymiare J. L., Arambel-Liu S., et al. (2004). Neural activity when people solve verbal problems with insight. *PLoS Biol, 2*(4), e97.

Junqué de Fortuny, E., De Smedt, T., Martens, D., & Daelemans, W. (2012). Media coverage in times of political crisis: a text mining approach. *Expert Systems with Applications*, *39*(14), 11616–11622.

Jurafsky, D., & Martin, J. H. (2000). *Speech and Language Processing: Introduction to Natural Language Processing, Computational Linguistics and Speech Recognition*. New Jersey: Prentice Hall.

Kahneman, D. (2011). Thinking, fast and slow. New York: Farrar, Straus, and Giroux.

Kauffman, S. (1995). *At Home in the Universe: The Search for the Laws of Self-Organization and Complexity*. Oxford: Oxford University Press.

Kaufman, J. C., & Beghetto, R. A. (2009). Beyond big and little: the four c model of creativity. *Review of General Psychology*, *13*(1), 1–12.

Kilgarriff, A., & Grefenstette, G. (2003). Introduction to the special issue on the web as corpus. *Computational Linguistics*, *29*(3), 333–347.

Kilgour, M. (2007). Big C versus little c creative findings: domain-specific knowledge combination effects on the eminence of creative contributions. In *Call for Creative Futures Conference Proceedings*, 15–35. Pori, Finland.

Kirton, M. J. (1976). Adaptors and innovators: a description and measure. *Journal of Applied Psychology*, *61*(5), 622–629.

Klein A., Vaissière B., Cane, J., Steffan-Dewenter, I., Cunningham, S., Kremen, C., & Teja Tscharntke, T. (2007). Importance of pollinators in changing landscapes for world crops. *Proceedings of the Royal Society B*, *274*(1608), 303–313.

Koestler, A. (1964). *The Act of Creation*. New York: Penguin Books.

Kriegel, U . (2003). Consciousness as intransitive self-consciousness: two views and an argument. *Canadian Journal of Philosophy*, *33*(1), 103–132.

Kullback, S., & Leibler, R. A. (1951). On information and sufficiency. *Annals of Mathematical Statistics*, *22*(1), 79–86.

Kurzweil, R. (2005). *The Singularity Is Near: When Humans Transcend Biology*. New York: Viking.

Lakoff, G., & Johnson, M. (1980). *Metaphors We Live By*. Chicago: University of Chicago Press.

Landauer, T. K., McNamara, D. S., Dennis, S., & Kintsch, W. (2007). *Handbook of Latent Semantic Analysis*. London: Lawrence Erlbaum.

LeWitt, S. (1967). Paragraphs on conceptual art. *Artforum*, *5*(10), 79–83.

Lechat, L., & De Smedt, T., (2010). Nanophysical: an artistic visualization of a biochip. Imec Leuven, permanent exhibition.

Lechat, L., & De Smedt, T. (2010). Scanning, Parsing, Understanding. CLiPS, University of Antwerp. Permanent exhibition.

Lee, E. A., & Parks. T. (1995). Dataflow process networks. *Proceedings of the IEEE*, *83*(5), 773–799.

Lerner, J., & Tirole, J. (2002). Some simple economics of open source. *The Journal of Industrial Economics, 50*(2), 197–234.

Lewis, D. (1998). Naive (Bayes) at forty: the independence assumption in information retrieval. In *Machine Learning: ECML-98*, LNCS 1398, 4–15.

Libet, B. C., Gleason, C. A., Wright, E. W., & Pearl, D. K. (1983). Time of conscious intention to act in relation to onset of cerebral activities (readiness potential); the unconscious initiation of a freely voluntary act. *Brain*, *106*, 623–42.

Lieberman, J. N. (1965). Playfulness and divergent thinking: an investigation of their relationship at the kindergarten level. *The Journal of Genetic Psychology: Research and Theory on Human Development*, *107*(2), 219–224.

Liu, B. (2010). Sentiment analysis and subjectivity. In N. Indurkhya, & F. Damerau (Eds.) *Handbook of Natural Language Processing, Second Edition*. CRC Press.

Liu, H., & Singh, P. (2004). ConceptNet – a practical commonsense reasoning tool-kit. *BT Technology Journal*, *22*(4), 211–226.

Livio, M. (2003). *The Golden Ratio: The Story of Phi, the World's Most Astonishing Number*. New York: Broadway Books.

Lloyd, S. P. (1987). Least squares quantization in PCM. *IEEE Transactions on Information Theory*, *28*(2), 129–137.

Luyckx, K., & Daelemans, W. (2008). Authorship attribution and verification with many authors and limited data. In *Proceedings of the 22nd International Conference on Computational Linguistics (COLING '08)*, *1*, 513–520. Manchester, UK.

Lynch, A. (1996). *Thought Contagion: : How Belief Spreads through Society*. New York: Basic Books

Maeda, J. (2001). *Design By Numbers*. Cambridge: MIT Press.

Maeda, J. (2006). *The Laws of Simplicity*. Cambridge: MIT Press.

Mandelbrot, B. (1982). *The Fractal Geometry of Nature*. New York: W. H. Freeman.

Manning, C., Raghavan, P., & Schütze, H. (2009). *Introduction to Information Retrieval*. Cambridge: Cambridge University Press.

Marcus, M., Santorini, B., & Marcinkiewicz, M. (1993). Building a large annotated corpus of English: The Penn Treebank. *Computational Linguistics*, *19*(2), 313–330.

Mareschal, D., Quinn, P. C., & Lea, S. E. G. (2010). *The Making of Human Concepts*. New York: Oxford University Press.

Marinus, N., Lechat, L., Vets, T., De Bleser, F., & De Smedt, T. (2009). City in a Bottle. Retrieved June 2012, from: http://www.cityinabottle.org/

Markou, M., & Singh, S. (2003). Novelty detection: a review—part 1: statistical approaches. *Signal Processing*, *83*, 2481–2497.

Marsella, A. J. (2005). Culture and conflict: understanding, negotiating, and reconciling conflicting constructions of reality. *International Journal of Intercultural Relations*, *29*, 651–673.

Martindale, C. (1989). Personality, situation, and creativity. In R. Sternberg (Ed.) *Handbook of Creativity*, 211–232. New York: Plenum Press.

Martindale, C. (1999). Biological Bases of Creativity. In R. Sternberg (Ed.). *Handbook of Creativity*, 16–31. New York: Cambridge University Press.

Maynard Smith, J., & Price, G. R. (1973). The logic of animal conflict. *Nature*, *246*, 15–18.

McCombs, M. E., & Shaw, D. L. (1972). The agenda-setting function of mass media. *Public Opinion Quarterly*, *36*, 176.

McCormack, J. (2005). Open problems in evolutionary music and art. In *Applications of Evolutionary Computing*, LNCS 3449, 428–436.

McCormack, J. (2007). Artificial ecosystems for creative discovery. In *Proceedings of the 9th annual conference on Genetic and evolutionary computation (GECCO '07)*, 301–307. London, UK.

McCormack, J., & Dorin, A. (2001). Art, emergence and the computational sublime. In *A Conference on Generative Systems in the Electronic Arts (CEMA)*, 67-81. Melbourne, Australia.

McCrae, R. R. (1987). Creativity, divergent thinking, and openness to experience. *Journal of Personality and Social Psychology*, *52*(6), 1258–1265.

McGlone, M. S., & Tofighbakhsh, J. (1999). The Keats heuristic: rhyme as reason in aphorism interpretation. *Poetics*, *26*, 235–244.

Mednick, M. T., Mednick, S. A., & Mednick, E. V. (1964). Incubation of creative performance and specific associative priming. *The Journal of Abnormal and Social Psychology*, *69*(1), 84–88.

Mednick, S. A. (1962). The associative basis of the creative process. *Psychological Review*, *69*(3), 220–232.

Menschaert, L., De Bleser, F., & De Smedt, T. (2005). BLOTTER. In *Time Canvas 2005*, exhibition catalogue. December 3–December 4, Museum of Modern Art Antwerp, Belgium.

Meyer, D. E., & Kieras, D. E. (1997). A computational theory of executive cognitive processes and human multiple-task performance: part 2. Accounts of psychological refractory-period phenomena. *Psychological Review*, *104*(4), 749–791.

Minsky, M. (1991). Logical versus analogical or symbolic versus connectionist or neat versus scruffy. *AI Magazine*, *12*(2), 34–51.

Mishne, G., & de Rijke, M. (2006). Capturing global mood levels using blog posts. In *Proceedings of the Spring Symposia on Computational Approaches to Analyzing Weblogs (AAAI-CAAW-06).* Stanford, USA.

Mitchell, M., Crutchfield, J. P., & Hraber, P. T., (1994). Dynamics, computation, and the "edge of chaos": a re-examination. In *Complexity: Metaphors, Models, and Reality, 19*, 497–513.

Mitchell, T. (1997). *Machine Learning.* McGraw Hill.

Morante, R., & Sporleder, C. (2012). Modality and negation: an introduction to the special issue. *Computational Linguistics*, *38*(2), 223–260.

Morante, R., Van Asch, V., & Daelemans, W. (2010). Memory-based resolution of in-sentence scopes of hedge cues. In *Shared Task Proceedings of the Fourteenth Conference on Computational Natural Language Learning (CoNLL'10)*, 40–47. Uppsala, Sweden.

Mouchiroud, C., & Lubart, T. (2001). Children's original thinking: an empirical examination of alternative measures derived from divergent thinking tasks. *Journal of Genetic Psychology*, *162*(4), 382–401.

Nake, F. (2010). Paragraphs on computer art, past and present. In *Proceedings of CAT 2010 London Conference*, 55–63. London, UK.

Nettle, D. (2006). Schizotypy and mental health amongst poets, visual artists, and mathematicians. *Journal of Research in Personality*, *40*(6), 876–890.

Newell, A., Shaw, J. C., & Simon H. A. (1962). The processes of creative thinking. In H. Gruber, G. Terrell, M. Wertheimer (Eds.) *Contemporary Approaches to Creative Thinking*, 63–119. New York: Atherton Press.

Newell, A., & Simon, H. A. (1976). Computer science as empirical inquiry: symbols and Search. *Communications of the ACM*, *19*(3): 113–126.

Newman, M. L., Pennebaker, J. W., Berry, D. S., & Richards, J. M. (2003). Lying words: predicting deception from linguistic styles. *Personality and Social Psychology Bulletin*, *29*, 665–675.

Nickerson, R. S. (1999). Enhancing creativity. In R. Sternberg (Ed.) *Handbook of Creativity*, 392–430. New York: Cambridge University Press.

Niedermeyer, N., & Lopes da Silva, F. (2004). *Electroencephalography: Basic Principles, Clinical Applications, and Related Fields.* Lippincott Williams & Wilkins.

Nijholt, A., Plass-Oude Bos, D., Reuderink, B. (2008). Brain-computer interfaces for hci and games. *Entertainment Computing*, *1*(2), 85–94.

Niles, E., & Gould, S. J. (1972). Punctuated equilibria: an alternative to phyletic gradualism. *Models in Paleobiology*, 82–115.

Niu, W., & Sternberg, R. J. (2006). The philosophical roots of Western and Eastern conceptions of creativity. *Journal of Theoretical and Philosophical Psychology*, *26*, 18–38.

Niven, D. (2003). Objective evidence on media bias: newspaper coverage of congressional party switchers. *Journalism and Mass Communication Quarterly*, *80*, 311–326.

Norris, P., & Epstein, S. (2011). An experiential thinking style: its facets and relations with objective and subjective criterion measures. *Journal of Personality*, *79*(5), 1043–1080.

Norvig, P. (1987). Inference in text understanding. In *National Conference on Artificial Intelligence (AAAI-87)*, 561–565. Seattle, Washington, USA.

Nyffeler, M., & Sunderland, K. D. (2003). Composition, abundance and pest control potential of spider communities in agroecosystems: a comparison of European and US studies. *Agriculture, Ecosystems and Environment*, *95*, 579–612.

Oakley, K. P. (1961). On man's use of fire, with comments on tool-making and hunting. In S. Washburn (Ed.) *Social Life of Early Man*, 176–193. Chicago: Aldine Publishing.

Ordelman, R., de Jong, F., van Hessen, A., & Hondorp, H. (2007). TwNC: a multifaceted Dutch news corpus. *ELRA Newsletter*, *12*(3–4).

Osborn, A. F. (1963). *Applied Imagination.* New York: Scribner's.

Pakkenberg, B., & Gundersen, H. J. (1997). Neocortical neuron number in humans: effect of sex and age. *Journal of Comparative Neurology*, *384*(2), 312–320.

Pang, B., & Lee, L. (2004). A sentimental education: sentiment analysis using subjectivity summarization based on minimum cuts. In *Proceedings of the Association for Computational Linguistics (ACL 2004)*, 271–278. Barcelona, Spain.

Pang, B., Lee, L., & Vaithyanathan, S. (2002). Thumbs up? Sentiment classification using machine learning techniques. In *Proceedings of the Conference on Empirical Methods in Natural Language Processing*, 79–86. Philadelphia, USA.

Parish, Y. I. H., & Müller, P. (2001). Procedural modeling of cities. In *Proceedings of the 28th Annual Conference on Computer Graphics and Interactive Techniques* (*SIGGRAPH '01).* Los Angeles, California, USA.

Patki, S., Grundlehner, B., Nakada, T., & Penders, J. (2011). Low power wireless EEG headset for BCI applications. In *14th International Conference on Human-Computer Interaction (HCI 2011)*, LCNS 6762, 481–490. Orlando, Florida, USA.

Penders, J., Grundlehner, B., Vullers, R., & Gyselinckx, B. (2009). Potential and challenges of body area networks for affective human computer interaction. In *Proceedings of the 5th Foundations of Augmented Cognition conference (FAC 2009)*, LCNS 5638, 202–211. San Diego, California, USA.

Pennebaker, J. W. (2011). *The Secret Life of Pronouns: What Our Words Say About Us.* Bloomsbury Press.

Pereira, F. C. (2007). *Creativity and Artificial Intelligence: A Conceptual Blending Approach.* Berlin: Mouton de Gruyter.

Perkins, D. N. (1988). The possibility of invention. In R. Sternberg (Ed.) *The Nature of Creativity.* Cambridge: Cambridge University Press.

Perlin, K. (1985). An image synthesizer. *Computer Graphics*, *19*, 287–296.

Piantadosi, S. T., Tily, H., Gibson, E. (2012). The communicative function of ambiguity in language. *Cognition*, 122: 280–291.

Pinker, S. (1999). How The Mind Works. *Annals of the New York Academy of Sciences*, *882*, 119–127.

Pinker, S. (2007). *The Stuff of Thought: Language as a Window into Human Nature.* New York: Viking.

Pinker. S. (2005). So How Does The Mind Work?. *Mind & Language*, *20*(1): 1–24.

Policastro, E., & Gardner, H. (1999). From case studies to robust generalizations: an approach to the study of creativity. In R. Sternberg (Ed.) *Handbook of Creativity*, 213–225. New York: Cambridge University Press.

Prusinkiewicz, P., & Lindenmayer, A. (1990). *The Algorithmic Beauty of Plants.* New York: Springer-Verlag.

Pyers, J. E., & Senghas, A. (2009). Language promotes false-belief understanding. *Psychological Science*, *20*(70), 805–812.

Quillian, M. (1968). Semantic memory. In M. Minsky (Ed.) *Semantic Information Processing*, 227–270. Cambridge: MIT Press.

Reas, C., Fry, B. (2007). *Processing: a Programming Handbook for Visual Designers and Artists.* Cambridge: MIT Press.

Reeves, W. T. (1983). Particle systems - a technique for modelling a class of fuzzy objects. *ACM Transactions on Graphics*, *2*(2), 91–108.

Rennard, J. P. (2002). Implementation of logical functions in the Game of Life. *Collision-Based Computing*, 1321–1346.

Rethemeyer, K. R. (2007). The Empires Strike Back: is the internet corporatizing rather than democratizing policy processes? *Public Administration Review*, 199–215.

Reynolds, C. (1987). Flocks, herds, and schools: a distributed behavioral model. *Computer Graphics*, *21*(4).

Rittel, H. W. J., & Webber, M. M. (1973). Dilemmas in a general theory of planning. *Policy Sciences, 4*(2), 155–169.

Roth, G., & Dicke, U (2005). Evolution of the brain and intelligence. *Trends in Cognitive Sciences*, *9*(5), 250–257.

Runco, M. A., & Charles, R. E. (1992). Judgments of originality and appropriateness as predictors of creativity. *Personality and Individual Differences*, *15*(5), 537–546.

Runco, M. A., & Sakamoto, S. O. (1999). Experimental studies of creativity. In R. Sternberg (Ed.) *Handbook of Creativity*, 62–92. New York: Cambridge University Press.

Russell, S. J., & Norvig, P. (2003). *Artificial Intelligence: A Modern Approach.* New Jersey: Prentice Hall.

Salton, G., & Buckley, C. (1988). Term-weighting approaches in automatic text retrieval. *Information Processing & Management*, *24*(5), 513–523.

Salton, G., Wong, A., & Yang, C. S. (1975). A vector space model for automatic indexing. *Communications of the ACM*, *18*(11), 613–620.

Sassenberg, K., & Moskowitz, G. B. (2005). Don't stereotype, think different! Overcoming automatic stereotype activation by mindset priming. *Journal of Experimental Social Psychology*, *41*(5), 506–514.

Saunders, R., & Gero, J. S. (2001). Artificial creativity: a synthetic approach to the study of creative behaviour. In *Proceedings of the Fifth Conference on Computational and Cognitive Models of Creative Design*, 113–139. Queensland, Australia.

Schaffer, C. (1993). Overfitting avoidance as bias. *Machine Learning*, *10*(2), 153–178.

Schank, R., & Abelson, R. (1977). *Scripts, Plans, Goals, and Understanding: An Inquiry into Human Knowledge Structures.* New Jersey: Erlbaum.

Schaul, T., Bayer, J., Wierstra, D., Sun, Y., Felder, M., Sehnke, F., Rückstiess, T., & Schmidhuber, J. (2010). Pybrain. *Journal of Machine Learning Research*, 743–746.

Schilling, M.A., (2005). A "small world" network model of cognitive insight. *Creativity Research Journal*, *17*(2–3), 131–154.

Schneider, G., & Volk, M. (1998). Adding manual constraints and lexical look-up to a Brill-Tagger for German. In *Proceedings of the ESSLLI Workshop on Recent Advances in Corpus Annotation.* Saarbrücken, Germany.

Schütze, H., & Pedersen, J.O. (1997). A cooccurrence-based thesaurus and two applications to information retrieval. *Information Processing & Management*, *33*(3), 307–318.

Searle, J. (1999). The Chinese Room. In R. Wilson, & F. Keil (Eds.) *The MIT Encyclopedia of the Cognitive Sciences.* Cambridge: MIT Press.

Sebastiani, F. (2002). Machine learning in automated text categorization. *ACM Computing Surveys*, *34*(1), 1–47.

Shamir, L., & Tarakhovsky, J. A. (2012). Computer analysis of art. *Journal on Computing and Cultural Heritage*, *5*(2), 7:1–7:11.

Shannon, C. E. (1948). A mathematical theory of communication. *Bell System Technical Journal*, *27*(3), 379–423.

Shermer, M. (2008). Patternicity: finding meaningful patterns in meaningless noise. *Scientific American*, *299*(6), 48.

Sibson, R. (1973). SLINK: an optimally efficient algorithm for the single-link cluster method. *The Computer Journal*, *16*(1), 30–34.

Simonton, D. K. (1999). Creativity and genius. In L. Pervin, & O. John (Ed.) *Handbook of Personality Theory and Research*, 629–652. New York: Guilford Press.

Simonton, D. K. (1999). Creativity as blind variation and selective retention: is the creative process Darwinian? *Psychological Inquiry*, *10*, 309–328.

Simonton, D. K. (2000). Creativity: cognitive, personal, developmental, and social aspects. *American Psychologist*, *55*(1), 151–158.

Simonton, D. K. (2004). *Creativity in Science: Chance, Logic, Genius, and Zeitgeist.* Cambridge: Cambridge University Press.

Sims, K. (1994). Evolving virtual creatures. *Computer Graphics (SIGGRAPH '94 Proceedings)*, *28*(3), 15–22.

Snow, R. E., Corno, L., Cronbach, L. J., Kupermintz, H., Lohman, D. F., Mandinach, E. B., Porteus, A. W., & Talbert, J. E. (2002). *Remaking the Concept of Aptitude: Extending the Legacy of Richard E. Snow.* New Jersey: Erlbaum.

Sokal, A., & Bricmont, A. (1998). *Fashionable Nonsense: Postmodern Intellectuals' Abuse of Science.* New York: St. Martin's Press.

Sowa, J. (1987). Semantic networks. *Encyclopedia of Artificial Intelligence.* New York: Wiley.

Steels, L. (1997). The synthetic modeling of language origins. *Evolution of Communication*, *1*(1), 1–34.

Sternberg, R. J. (1999). Darwinian creativity as a conventional religious faith. *Psychological Inquiry*, *10*, 357–359.

Sternberg, R. J., & Lubart, T. I. (1999). The concept of creativity: prospects and paradigms. In R. Sternberg (Ed.) *Handbook of Creativity*, 3–15. New York: Cambridge University Press.

Sulzer, D. (2012). David Sulzer turns brain waves into music. Retrieved August 2012, from: http://medicalxpress.com/news/2012-08-neuroscientist-david-sulzer-brain-music.html

Sung, K. (2011). Recent videogame console technologies. *Computer*, February 2011, 91–93.

Sutherland, S. (1989). "Consciousness". *Macmillan Dictionary of Psychology.* Macmillan.

SUTHERLAND, S. (1992). *Irrationality: The Enemy Within. London*: Constable.

SUTTON, R. S., & BARTO, A. G. (1998). *Reinforcement Learning: An Introduction.* Cambridge: MIT Press.

TABOADA, M., BROOKS, J., TOFILOSKI, M., VOLL, K., & STEDE, M. (2011). Lexicon-based methods for sentiment analysis. *Computational Linguistics*, *37*(2), 267–307.

TAKEUCHI, H., TAKI, Y., HASHIZUME, H., SASSA, Y., NAGASE, T., NOUCHI, R., & KAWASHIMA R. (2011). Failing to deactivate: the association between brain activity during a working memory task and creativity. *Neuroimage*, *55*(2), 681–687.

TORRANCE, E. P. (1988). The nature of creativity as manifest in its testing. In R. Sternberg (Ed.) *The Nature of Creativity: Contemporary Psychological Perspectives*, 43–75. New York: Cambridge University Press.

TUMASJAN, A., SPRENGER, T. O., SANDNER, P. G., & WELPE, I. M. (2010). Predicting elections with Twitter: what 140 characters reveal about political sentiment. In *Proceedings of the Fourth International AAAI Conference on Weblogs and Social Media.* Washington, USA.

TURING, A. M. (1950). Computing machinery and intelligence. *Mind*, *59*(236), 433–460.

TVERSKY, A. (1977). Features of similarity. *Psychological Review*, *84*(4), 327–352.

VAN ASCH, V. (2012). *Domain Similarity Measures: On the Use of Distance Metrics in Natural Language Processing.* PhD dissertation.

VAN ASCH, V., & DAELEMANS., W. (2009). Prepositional phrase attachment in shallow parsing. In *7th International Conference on Recent Advances in Natural Language Processing (RANLP)*, 14–16. Borovets, Bulgaria.

VAN DE CRUYS, T. (2010). Mining for meaning: the extraction of lexico-semantic knowledge from text. *Groningen Dissertations in Linguistics*, *82*.

VAN DEN BOSCH, A., BUSSER, B., CANISIUS, S., & DAELEMANS, W. (2007). An efficient memory-based morphosyntactic tagger and parser for Dutch. In *Selected Papers of the 17th Computational Linguistics in the Netherlands Meeting*, 99–114. Leuven, Belgium.

VEALE, T., & HAO, Y. (2007). Comprehending and generating apt metaphors: a web-driven, case-based approach to figurative language. In *Proceedings of AAAI 2007*, 1471–1476. Vancouver, Canada.

VEALE, T., & HAO, Y. (2007). Learning to understand figurative language: from similes to metaphors to irony. In *Proceedings of the 29th Annual Meeting of the Cognitive Science Society (CogSci 2007).* Nashville, Tennessee, USA.

VEALE, T., O'DONOGHUE, D., & KEANE, M. T. (2000). Computation and blending. *Cognitive Linguistics*, *11*(3–4), 253–281.

VEALE, T., SECO, N., & HAYES, J. (2004). Creative discovery in lexical ontologies. In *Proceedings of Coling 2004*, 1333–1338. Switzerland: Geneva.

VEALE, T., & YAO, H. (2008). A fluid knowledge representation for understanding and generating creative metaphors. In *Proceedings of the 22nd International Conference on Computational Linguistics (COLING '08)*, *1*, 945-952. Manchester, UK.

VERLE, L. (2007). MIND VJ: An interface to manipulate moving images in real-time using brain waves. The First Summit Meeting of the Planetary Collegium.

VOLLRATH, F., & SELDEN, P. (2007). The role of behavior in the evolution of spiders, silks, and webs. *Annual Review of Ecology, Evolution, and Systematics*, *38*, 819–846.

VOSSEN, P., HOFMANN, K., DE RIJKE, M., TJONG KIM SANG, E., & DESCHACHT, K. (2007). The cornetto database: architecture and user-scenarios. In *Proceedings of 7th Dutch-Belgian Information Retrieval Workshop DIR2007.* Leuven, Belgium.

WALDROP, M. M. (1992). *Complexity: The Emerging Science and the Edge of Order and Chaos.* New York: Touchstone Press.

WATZ, M. (2005). Generative art now. An interview with Marius Watz. Retrieved April 2012, from: http://www.artificial.dk/articles/watz.htm.

WEISBERG, R. W. (1999). Creativity and knowledge: A challenge to theories. In R. Sternberg (Ed.) *Handbook of Creativity*, 227–250. New York: Cambridge University Press.

WEISS, R. (1995). Floral color change: a widespread functional convergence. *American Journal of Botany*, *82*(2), 167–185.

WIGGINS, G. A. (2006). A preliminary framework for description, analysis and comparison of creative systems. *Knowledge-Based Systems*, *19*, 449–458.

WINOGRAD, T. (1972). *Understanding Natural Language.* New York: Academic Press.

YPERMAN, T. (2004). *Van Vlaams Blok naar Vlaams Belang.* PhD dissertation, University of Ghent.

ZIMMER, M. (2008). Critical perspectives on Web 2.0. *First Monday*, *13*(3).

VAN DOORN, W. (1997). Effects of pollination on floral attraction and longevity. *Journal of Experimental Botany*, *48*(314), 1615–1622.

VAN DEN BOSCH, A., & DAELEMANS, W. (1999). Memory-based morphological analysis. In *Proceedings of the 37th Annual Meeting of the Association for Computational Linguistics on Computational Linguistics*, 285–292. Maryland, USA.

# Index